\documentclass[acmtog,]{acmart}

\AtBeginDocument{%
  \providecommand\BibTeX{{%
    \normalfont B\kern-0.5em{\scshape i\kern-0.25em b}\kern-0.8em\TeX}}}

\setcopyright{rightsretained}
\usepackage[export]{adjustbox}
\usepackage[ruled]{algorithm2e} 
\usepackage{amsmath}
\usepackage{newtxmath}
\usepackage{booktabs} 
\usepackage{epsfig}
\usepackage[toc]{glossaries}
\usepackage{graphicx}
\usepackage{times}
\usepackage{xcolor} 
\usepackage{tikz}
\usepackage{multirow}
\usepackage{subcaption}

\newcommand{\omicron}{o}

\SetKwComment{comment}{// }{} 
\DontPrintSemicolon
\LinesNumbered
\let\oldnl\nl
\newcommand{\nonl}{\renewcommand{\nl}{\let\nl\oldnl}}

\newcommand{\modifiedOrder}[1]{}
\newcommand{\modifiedNote}[1]{}

\newcommand{\deleted}[1]{}

\newcommand{\etal}{et~al.\ }
\newcommand{\cf}{cf.\ }
\newcommand{\ours}{ours}

\newcommand{\numberedThingRef}[2]{\mbox{#1.~{#2}}}
\newcommand{\Alg}[1]{\numberedThingRef{Alg}{#1}}
\newcommand{\Eq}[1]{\numberedThingRef{Eq}{#1}}
\newcommand{\Fig}[1]{\numberedThingRef{Fig}{#1}}

\newcommand{\Tab}[1]{\numberedThingRef{Tab}{#1}}

\def\authornote#1#2#3{{\textcolor{#2}{{\small#1:[*#3*]}}}}
\definecolor{darkgreen}{RGB}{0,128,0}
\definecolor{darkcyan}{RGB}{0,192,96}

\definecolor{Gray}{gray}{0.95}

  {\end{list}}

\renewcommand{\d}[1]{\ensuremath{\operatorname{d}{#1}}}

\newcommand{\real}{\vmathbb{R}} 

\newcommand{\dimension}{D}
\newcommand{\modelParametersSymbol}{\Theta}
\newcommand{\allParams}{{\modelParametersSymbol}}

\newcommand{\neighborsSymbol}{\mathcal{N}}

\newcommand{\neighborsSix}[1]{\neighborsSymbol_{6}(#1)}

\newcommand{\neighborsSixteen}[1]{\neighborsSymbol_{16}(#1)}

\newcommand{\pos}{\mathbf{x}}

\newcommand{\querySymbol}{q}
\newcommand{\posQuery}{\mathbf{\querySymbol}}
\newcommand{\pQuery}{\posQuery}

\newcommand{\direction}{\mathbf{v}}

\newcommand{\normalSymbol}{\mathbf{n}}

\newcommand{\normal}[1]{\normalSymbol(#1)}

\newcommand{\raw}[1]{\widehat{#1}}

\newcommand{\far}{\infty}
\newcommand{\near}{0}

\newcommand{\camIdx}{c}
\newcommand{\camPos}{\pos_\camIdx}
\newcommand{\camPosIndexed}[1]{\pos_{\camIdx,#1}}

\newcommand{\raySymbol}{\mathbf{r}}
\newcommand{\rayIndexed}[1]{\raySymbol_{#1}}
\newcommand{\rayDir}{\mathbf{d}}
\newcommand{\rayDirIndexed}[1]{\rayDir_{#1}}
\newcommand{\rayT}{t}
\newcommand{\rayTIndexed}[1]{\rayT_{#1}}
\newcommand{\rayTIndexedij}[2]{\rayT_{#1,#2}}
\newcommand{\rayTFar}{\rayTIndexed{\far}}
\newcommand{\rayTNear}{\rayTIndexed{\near}}
\newcommand{\rayTInner}{\tilde{\rayT}}

\newcommand{\rayShortIndexed}[1]{\rayIndexed{#1}\left(\rayT\right)}

\newcommand{\rayFullIndexed}[1]{
    \rayShortIndexed{#1} = \camPosIndexed{#1} + \rayT \cdot \rayDirIndexed{#1}
}

\newcommand{\rayBatch}[1]{\left\{\rayShortIndexed{#1}\right\}}
\newcommand{\raySamplePoints}[2]{\left\{\rayTIndexedij{#1}{#2}\right\}}

\newcommand{\maxSamplesPerRayUninformed}{N_{\text{max}}}
\newcommand{\maxSamplesPerRayInformed}{N_{\text{max}, \opacity}}

\newcommand{\weightSymbol}{w}

\newcommand{\spatialSymbol}{\text{3D}}
\newcommand{\wTLerpSymbol}{\weightSymbol_\spatialSymbol}
\newcommand{\weightTrilinearInterpolation}[2]{\wTLerpSymbol(#1,#2)}
\newcommand{\wTLerp}[2]{\weightTrilinearInterpolation{#1}{#2}}

\newcommand{\levelOfDetailSymbol}{\text{LoD}}
\newcommand{\LoDSymbol}{\levelOfDetailSymbol}
\newcommand{\wLoDSymbol}{\weightSymbol_{\LoDSymbol}}
\newcommand{\weightLevelOfDetail}[2]{\wLoDSymbol(#1, #2)}
\newcommand{\weightLoD}[2]{\weightLevelOfDetail{#1}{#2}}
\newcommand{\wLoD}[2]{\weightLoD{#1}{#2}}

\newcommand{\weightQuadrilinearInterpolation}[2]{\weightSymbol_{\text{4D}}(#1,#2)}

\newcommand{\wQLerp}[2]{\weightQuadrilinearInterpolation{#1}{#2}}

\newcommand{\fieldSymbol}{f}
\newcommand{\fieldGradientSymbol}{\nabla \fieldSymbol}
\newcommand{\field}[1]{\fieldSymbol(#1)}
\newcommand{\fieldGradient}[1]{\fieldGradientSymbol(#1)}
\newcommand{\fieldIndexed}[2]{\fieldSymbol_{#1}(#2)}

\newcommand{\density}{\rho}

\newcommand{\opacitySymbol}{\omicron}
\newcommand{\opacity}{\opacitySymbol}

\newcommand{\opacitySampleCoordsij}[2]{\rayTIndexedij{#1}{#2}}
\newcommand{\opacitySampleij}[2]{\opacity(\opacitySampleCoordsij{#1}{#2})}
\newcommand{\opacitySamplesij}[2]{\left\{\opacitySampleij{#1}{#2}\right\}}

\newcommand{\radiance}{\mathbf{L}}

\newcommand{\SHBandCountSymbol}{b}
\newcommand{\SHBasisSymbol}{Y}

\newcommand{\SHBasisIndexed}[2]{\SHBasisSymbol_{#1,#2}}
\newcommand{\SHBasisFunctionIndexedlm}[3]{\SHBasisSymbol_{#1,#2}(#3)}
\newcommand{\SHBasisFunctionslm}[2]{\{\SHBasisSymbol_{#1,#2}\}}
\newcommand{\SHCoefficientSymbol}{\mathbf{c}}

\newcommand{\SHCoefficientIndexedlm}[2]{\SHCoefficientSymbol_{#1,#2}}
\newcommand{\SHCoefficientsIndexedlm}[2]{\{\SHCoefficientIndexedlm{l}{m}\}}

\newcommand{\SHSoftServeSLFBandCount}{3}
\newcommand{\SHSoftServeSLFParamsPerNode}{3 \times \SHBandCountSymbol \times \SHBandCountSymbol}

\newcommand{\transparencyCumSymbol}{T}
\newcommand{\transparencyCum}[1]{\transparencyCumSymbol(#1)}

\newcommand{\radianceOutgoingSymbol}{\radiance_o}
\newcommand{\radianceOutgoing}[2]{\radiance_o(#1, #2)}
\newcommand{\radianceOutgoingRaySample}[1]{\radianceOutgoing{\rayTIndexed{#1}}{-\rayDir}}
\newcommand{\radianceSampleCoords}[2]
{
    \left(
        \rayTIndexedij{#1}{#2}, -\rayDirIndexed{#1}
    \right)
}
\newcommand{\radianceSample}[2]
{
    \radianceOutgoingSymbol\radianceSampleCoords{#1}{#2}
}

\newcommand{\radianceSamplesij}[2]
{
    \left\{\radianceSample{#1}{#2}\right\}
}

\newcommand{\shadingSymbol}{\radianceOutgoingSymbol}

\newcommand{\shading}[2]{\shadingSymbol(#1, #2)}
\newcommand{\shadingFull}{\shading{\rayT}{-\rayDir}}

\newcommand{\pixelRadiance}{\radiance_{\pixelSymbol}}

\newcommand{\pixelRadianceCamPosRayDir}{\pixelRadiance(\camPos,-\rayDir)}
\newcommand{\distantRadianceSymbol}{\radiance_{\far}}
\newcommand{\distantRadiance}[1]{\distantRadianceSymbol(#1)}

\newcommand{\pixelRadiancePixelCoordsIndexed}[1]{\pixelRadiance(\pixelCoordsVecIndexed{#1})}
\newcommand{\pixelLPCVIndexed}[1]{\pixelRadiancePixelCoordsIndexed{#1}}
\newcommand{\radianceSamples}[1]{\left\{\pixelRadiancePixelCoordsIndexed{i}\right\}}
\newcommand{\opacitySample}[1]{\opacity_\pixelSymbol(\pixelCoordsVecIndexed{#1})}
\newcommand{\opacitySamples}[1]{\left\{\opacitySample{#1}\right\}}

\newcommand{\groundTruth}{\text{gt}}

\newcommand{\pixelSymbol}{p}
\newcommand{\pixelIndexed}[1]{\pixelSymbol_{#1}}
\newcommand{\pixelChannelSymbol}{c}
\newcommand{\channelSymbol}{\pixelChannelSymbol}
\newcommand{\intensitySymbol}{\mathcal{I}}

\newcommand{\pixelCoordsVecIndexed}[1]{\mathbf{\pixelSymbol}_{#1}}

\newcommand{\pixelBatch}[1]{\left\{\pixelCoordsVecIndexed{#1}\right\}}
\newcommand{\batchSizeSoftServe}{4096}

\newcommand{\pixelIntensity}{\intensitySymbol_{\pixelSymbol}}

\newcommand{\pixelIntensityGT}{\pixelIntensity^{\groundTruth}}
\newcommand{\pixelIntensityIndexed}[1]{\pixelIntensity({#1})}
\newcommand{\pixelIntensityGTIndexed}[1]{\pixelIntensityGT({#1})}
\newcommand{\pixelIntensityPixelIndexed}[1]{\pixelIntensity(\pixelCoordsVecIndexed{#1})}
\newcommand{\pixelIntensityGTPixelIndexed}[1]{\pixelIntensityGT(\pixelCoordsVecIndexed{#1})}
\newcommand{\pixelIntensityBatch}[1]
{
    \left\{
        \pixelIntensityPixelIndexed{#1}
    \right\}
}
\newcommand{\pixelIntensityGTBatch}[1]
{
    \left\{
        \pixelIntensityGTPixelIndexed{#1}
    \right\}
}

\newcommand{\loss}{l}
\newcommand{\lossHuber}{l_1}

\newcommand{\lossModel}{\loss_{\allParams}}
\newcommand{\lossPixel}{\loss_{\pixelSymbol}}
\newcommand{\lossPixelIndexed}[1]{\lossPixel(\pixelCoordsVecIndexed{#1})}
\newcommand{\lossBatch}[1]{\left\{\lossPixelIndexed{#1}\right\}}
\newcommand{\lossCache}{\mathcal{C}}
\newcommand{\lossCacheUpdateRate}{5000}
\newcommand{\lossPixelIntensityChannel}[2]%
{
    \|
        \pixelIntensityIndexed{\pixelCoordsVecIndexed{#1}, #2}
        - \pixelIntensityGTIndexed{\pixelCoordsVecIndexed{#1}, #2}
    \|^2
}

\newcommand{\lossLocalSmoothness}{\loss_{\text{3D}}}
\newcommand{\lossLocalSmoothnessNormals}{\loss_{\text{n}}}
\newcommand{\lossLoDSmoothness}{\loss_{\text{LoD}}}
\newcommand{\lossLowFields}{\loss_{\text{0}}}

\newcommand{\lossLocalSmoothnessLambda}{\lambda_{\text{3D}}}

\newcommand{\lossLoDSmoothnessLambda}{\lambda_{\text{LoD}}}
\newcommand{\lossLowFieldsLambda}{\lambda_{\text{0}}}

\newcommand{\depth}{d}

\newcommand{\footprint}{\sigma}

\newcommand{\footprintBackProj}{\footprint(\rayT)}

\newcommand{\node}{n}
\newcommand{\nodeIndexed}[1]{\node_{#1}}

\newcommand{\footprintNode}{\footprint_\node}
\newcommand{\footprintNodeIndexed}[1]{\footprint_{#1}}
\newcommand{\extendNode}{\footprintNode}
\newcommand{\extendNodeIndexed}[1]{\footprintNodeIndexed{#1}}

\newcommand{\posNodeIndexed}[1]{\pos_{#1}}
\newcommand{\posNodeRandomIndexed}[1]{\tilde{\pos}_{#1}}
\newcommand{\pNodeIndexed}[1]{\posNodeIndexed{#1}}
\newcommand{\nodeBatch}[1]{\left\{\nodeIndexed{#1}\right\}}
\newcommand{\nodesRequired}[1]{\left\{\nodeIndexed{#1}\right\}}

\newcommand{\nodeScope}{\mathbf{\node}}
\newcommand{\nodeScopeIndexed}[1]{\nodeScope_{#1}}

\newcommand{\depthNode}{\depth_{\node}}
\newcommand{\depthNodeIndexed}[1]{\depth_{#1}}

\newcommand{\nodePlaneIndexed}[1]{\pi_{\node_{#1}}}

\newcommand{\borderIndexed}[1]{b}

\newcommand{\imageGap}{\hspace{2pt}}
\newcommand{\privateLabeledImageNodes}[3]%
{%
    \node(mainImage)%
        [inner sep=0pt,anchor=south west]%
        at (0,0)%
    {%
        \includegraphics[width=#1]{#2}%
    };%
    \node(label)%
        [inner sep=2pt,anchor=south east,black,fill=white,align=right]%
        at (node cs:name=mainImage,anchor=south east)%
    {%
        \small\textsf{#3}%
    };%
}

\newcommand{\labelImage}[3]%
{%
    \begin{tikzpicture}%
        \privateLabeledImageNodes{#1}{#2}{#3}
    \end{tikzpicture}%
}

\newcommand{\normalMap}[3]%
{%
    \begin{tikzpicture}%
        \privateLabeledImageNodes{#1}{#2}{#3}
        \node(colorMap)%
            [inner sep=0pt,anchor=north west,white,align=right]%
            at (node cs:name=mainImage,anchor=north west)%
        {%
            \includegraphics[width=20pt]{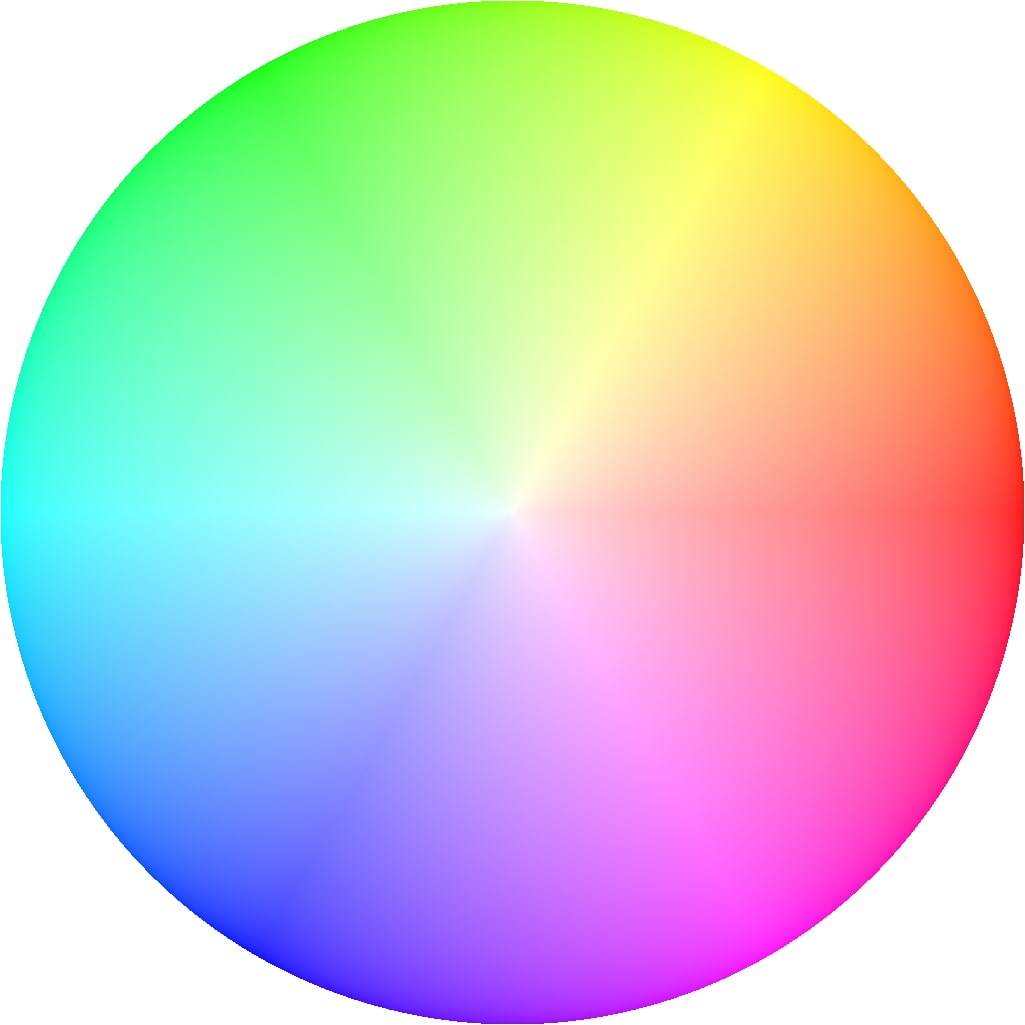}%
        };%
    \end{tikzpicture}%
}

\makeatletter 
\newcommand{\figlabel}[1]{\scriptsize\sffamily\bfseries\color{white}{#1}}
\newcommand{\figlabelblk}[1]{\scriptsize\sffamily\bfseries\color{black}{#1}}
\newlength{\sfp@hseplen}\newlength{\sfp@vseplen}
\define@cmdkey{subfigpos}[sfp@]{vsep}[0.6\baselineskip]{\setlength{\sfp@vseplen}{\sfp@vsep}}
\define@cmdkey{subfigpos}[sfp@]{hsep}[1.5pt]{\setlength{\sfp@hseplen}{\sfp@hsep}}
\newcommand{\subfigimg}[3][,]{%
    \setkeys{Gin,subfigpos}{vsep,hsep,#1}
    \setbox1=\hbox{\includegraphics{#3}}
    \leavevmode\rlap{\usebox1}
    \rlap{\hspace*{\sfp@hsep}\raisebox{\dimexpr\ht1-\sfp@vsep}{\figlabel{#2}}}
    \phantom{\usebox1}
}
\newcommand{\subfigimgblk}[3][,]{%
    \setkeys{Gin,subfigpos}{vsep,hsep,#1}
    \setbox1=\hbox{\includegraphics{#3}}
    \leavevmode\rlap{\usebox1}
    \rlap{\hspace*{\sfp@hsep}\raisebox{\dimexpr\ht1-\sfp@vsep}{\figlabelblk{#2}}}
    \phantom{\usebox1}
}
\makeatother 

\newacronym{CPU}{CPU}{Central Processing Unit}

\newacronym{GPU}{GPU}{Graphics Processing Unit}

\newcommand{\RGB}{RGB}
\newcommand{\RGBA}{RGBA}

\newcommand{\texel}{texel}

\newacronym{LoD}{{LoD}}{level of detail}
\newcommand{\LoD}{\gls{LoD}}

\newglossaryentry{Adam}
{
    name = {Adam},
    description =
    {
        \SGD{} optimizer tracking first and second momentum of gradients by Kingma and Ba \cite{kingma2014Adam}
    },
}
\newcommand{\Adam}{\gls{Adam}}

\newacronym{BFGS}{{BFGS}}{Broyden–Fletcher–Goldfarb–Shanno optimization algorithm}

\newacronym{LBFGS}{{LBFGS}}{limited-memory BFGS}
\newcommand{\LBFGS}{\gls{LBFGS}}

\newacronym{LM}{LM}{Levenberg-Marquardt optimization algorithm \cite{more1978LevenbergMarquardt}}
\newcommand{\LM}{\gls{LM}}

\newacronym{PCG}{PCG}{preconditioned conjugate gradients}
\newcommand{\PCG}{\gls{PCG}}

\newacronym{SGD}{SGD}{stochastic gradient descent}
\newcommand{\SGD}{\gls{SGD}}

\newacronym{CNN}{CNN}{convolutional neural network}
\newcommand{\CNN}{\gls{CNN}}

\newacronym{KDE}{KDE}{kernel density estimation}
\newcommand{\KDE}{\gls{KDE}} 

\newglossaryentry{leakyReLU}
{
    name = {leaky ReLU},
    description =
    {
        leaky rectified linear unit \ReLU{} allowing invalid constraint: $f(x) = x if x > 0 else a \cdot x$
    },
}

\newglossaryentry{LiLU}
{
    name = {LiLU},
    description =
    {
        Limited linear unit; activation function similar to \ReLU{} but without gradient loss at constraint border
    },
}
\newcommand{\LiLU}{\gls{LiLU}}

\newacronym{NCC}{NCC}{Normalized Cross Correlation}

\newacronym{ODKDE}{ODKDE}{observation-dependent \KDE from Aroudj \etal \cite{aroudj2017ThinSurface}}

\newacronym{RMSProp}{RMSProp}{root mean square propagation algorithm}

\newglossaryentry{ReLU}
{
    name = {ReLU},
    description =
    {
        Rectified linear unit; constraint or activation $f(x) = x if x > 0 else 0$
    },
}
\newcommand{\ReLU}{\gls{ReLU}}

\newglossaryentry{Softplus}
{
    name = {Softplus},
    description =
    {
        Constraint or activation function $f(x) = \ln(1 + e^x)$; also called SmoothReLU  
    },
}
\newcommand{\Softplus}{\gls{Softplus}}

\newacronym{SLAM}{SLAM}{simultaneous localization and mapping}
\newcommand{\SLAM}{\gls{SLAM}}

\newacronym{SfM}{SfM}{structure from motion}
\newcommand{\SfM}{\gls{SfM}}

\newacronym{LIDAR}{LIDAR}{light detection and ranging}

\newacronym{MVS}{MVS}{multi-view stereo}
\newcommand{\MVS}{\gls{MVS}}

\newacronym{AABB}{AABB}{axis-aligned bounding box}
\newcommand{\AABB}{\gls{AABB}}

\newacronym{BRDF}{BRDF}{Bidirectional reflectance distribution function}

\newglossaryentry{Lumisphere}
{
    name = {lumisphere},
    description =
    {
        2D spherical function modeling reflected light for all directions from a single surface point or 3D density field sample
        introduced by Wood \etal \cite{wood2000SurfaceLightFields}
    },
}

\newglossaryentry{Lumigraph}
{
    name = {lumigraph},
    description =
    {
        Abstract 4D function representing light reflected from surfaces
        introduced by Gortler \etal \cite{gortler1996Lumigraph}
    },
}

\newacronym{MLP}{MLP}{multi-layer perceptron}
\newcommand{\MLP}{\gls{MLP}}

\newacronym{MPI}{MPI}{multi-plane image}
\newcommand{\MPI}{\gls{MPI}}

\newacronym{MSI}{MSI}{multi-sphere image}
\newcommand{\MSI}{\gls{MSI}}

\newglossaryentry{NeRF}
{
    name = {NeRF},
    description = {Neural Radiance Field from Mildenhall \etal \cite{mildenhall2020NeRF}},
}
\newcommand{\NeRF}{\gls{NeRF}}

\newacronym{OBB}{OBB}{oriented bounding box}

\newacronym{PDF}{PDF}{probability distribution function}

\newacronym{SDF}{SDF}{signed distance function}
\newcommand{\SDF}{\gls{SDF}}

\newacronym{SG}{SG}{spherical Gaussian}

\newacronym{SH}{SH}{spherical harmonic}
\newcommand{\SH}{\gls{SH}}

\newacronym{SLF}{SLF}{surface light field}
\newcommand{\SLF}{\gls{SLF}}

\newacronym{SVBRDF}{SVBRDF}{spatially varying bidirectional reflectance distribution function}

\newacronym{SVO}{SVO}{sparse voxel octree}
\newcommand{\SVO}{\gls{SVO}}

\newacronym{TSDF}{TSDF}{truncated signed distance function}

\newacronym{IBR}{IBR}{image-based rendering}

\newacronym{IDR}{IDR}{inverse differentiable rendering}
\newcommand{\IDR}{\gls{IDR}}

\newacronym{PBIDR}{PBIDR}{physically-based inverse differentiable rendering}
\newacronym{PBIDRV}{PBVIDR}{physically-based volumetric inverse differentiable rendering}

\newacronym{PSNR}{PSNR}{Peak Signal-to-Noise Ratio}

\newacronym{SSIM}{SSIM}{Structural Similarity Index Measure by Wang \cite{wang2004SSIM}}

\newacronym{LPIPS}{LPIPS}{Learned Perceptual Image Patch Similarity \cite{zhang2018LPIPS}}
\newcommand{\LPIPS}{\gls{LPIPS}}

\newglossaryentry{Blender}{
    name=Blender,
    description={Open source and free 3D rendering and modeling software},
}
\newcommand{\Blender}{\gls{Blender}}

\newglossaryentry{Ceres}
{
    name=Ceres,
    description={Google's open source C++ optimization library},
}

\newglossaryentry{CMPMVS}
{
    name=CMPMVS,
    description={\MVS and surface reconstruction library from Jancosek and Pajdla \cite{jancosek2011CMPMVS}},
}
\newcommand{\CMPMVS}{\gls{CMPMVS}}

\newglossaryentry{CMPMVSOURS}
{
    name={CMPMVS*},
    description={Our implmementation of \CMPMVS.}
}

\newglossaryentry{Colmap}
{
    name=COLMAP,
    description={\SfM and \MVS-based 3D reconstruction pipeline library from Schoenberger \etal \cite{schoenberger2016SfM, schoenberger2016PixelwiseMVS}}
}
\newcommand{\Colmap}{\gls{Colmap}}

\newacronym{DRV}{DRV}{deep reflectance volumes from Bi \etal \cite{bi2020DeepReflectanceVolumes}}

\newglossaryentry{DualMarchingCubes}
{
    name={dual marching cubes},
    description={Dual marching cubes for surface extraction from an implicit representation \cite{schaefer2004DMC}},
}

\newglossaryentry{Embree}
{
    name=Embree,
    description={Intel's CPU-based ray tracing engine \cite{wald2014Embree}}
}

\newglossaryentry{Enoki}
{
    name=Enoki,
    description={Vectorization and auto differentiation framework from Wenzel Jakob \cite{jakob2019Enoki}},
}

\newglossaryentry{ERF}
{
    name=ERF,
    description={Reconstruction method with explicit radiance fields \cite{aroudj2021}},
}

\newacronym{FSSR}{FSSR}{floating scale surface reconstruction from Fuhrmann and Goesele \cite{fuhrmann2014FSSR}}

\newglossaryentry{Halide}
{
    name=Halide,
    description=
    {
        Auto differentiation and scheduling C++ library from Ragan-Kelley \etal \cite{ragan2014Halide,li2018DifferentiableProgrammingInHalide}
    },
}

\newglossaryentry{IDRYariv}
{
    name=IDR,
    description=
    {
        \SDF{} and \MLP{}-based reconstruction method from Yariv \etal \cite{yariv2020IDR}
    },
}

\newglossaryentry{JaxNeRF}
{
    name = {JaxNeRF},
    description = {Efficient and corrected implementation of \NeRF{} by Deng \etal \cite{deng2020JaxNeRF}},
}
\newcommand{\JaxNeRF}{\gls{JaxNeRF}}

\newglossaryentry{LLFF}
{
    name=LLFF,
    description={Implicit predecessor of \NeRF{} \cite{mildenhall2019LLFF}},
}
\newcommand{\LLFF}{\gls{LLFF}}

\newglossaryentry{KinectFusion}
{
    name=KinectFusion,
    description={Real-time \SLAM-based 3D reconstruction pipeline for RGBD-sensors from Newcombe \etal \cite{newcombe2011KinectFusion}},
}

\newglossaryentry{MarchingCubes}
{
    name={marching cubes},
    description={Marching cubes for surface extraction from an implicit representation \cite{lorensen1987marchingCubesJournal}},
}

\newglossaryentry{Maple}
{
    name=Maple,
    description={Mathematics software including symbolic differentiation support \cite{maple}},
}

\newglossaryentry{MipNeRF}
{
    name = {MipNeRF},
    description = {Anti-aliasing extension of \NeRF{} by Barron \etal \cite{barron2021MipNerf}},
}
\newcommand{\MipNeRF}{\gls{MipNeRF}}

\newglossaryentry{Mitsuba}
{
    name={Mitsuba},
    description={differentiable rendering framework from Nimier-David \etal \cite{nimier-David2019Mitsuba2}}
}

\newacronym{MVE}{MVE}{multi-view environment: 3D reconstruction pipeline tool from Fuhrmann \etal \cite{fuhrmann2014MVE}}

\newglossaryentry{NeRD}
{
    name = {NeRD},
    description = {\NeRF extension focused on physically-based scene decomposition by Boss \etal \cite{boss2020NeRD}},
}

\newglossaryentry{NeRV}
{
    name = {NeRV},
    description = {\NeRF variant with additional visibility \MLP for more efficient relighting by Srinivasan \etal \cite{srinivasan2020NeRV}},
}
\newcommand{\NeRV}{\gls{NeRV}}

\newglossaryentry{NeuralVolumes}
{
    name = {{Neural Volumes}},
    description = {
        Neural reconstruction method with encoder and decoder network
        which predicts a scene into a regular \RGBA{} voxel grid
        by Lombardi \etal \cite{lombardi2019NeuralVolumes}
    },
}

\newglossaryentry{PhySG}
{
    name = {PhySG},
    description = {\MLP and \SDF-based 3D reconstruction method modeling lights and materials using spherical Gaussians \cite{zhang2021PhySG}},
}

\newglossaryentry{PlenOctree}
{
    name = {PlenOctree},
    description = {\SVO storing \SH-based \glspl{Lumisphere} introduced by Yu \etal \cite{yu2021PlenOctrees}},
}
\newcommand{\PlenOctree}{\gls{PlenOctree}}

\newacronym{PSR}{{PSR}}
{
    poisson surface reconstruction from Kazhdan \etal \cite{kazhdan2005SolidModelsFromOrientedPointSets,kazhdan2006PSR,kazhdan2013ScreenedPSR}
}

\newglossaryentry{PyTorch}
{
    name={PyTorch},
    description={Facebook's auto differentiation and machine learning framework \cite{pytorch2019}}
}
\newcommand{\PyTorch}{\gls{PyTorch}}

\newacronym{SSD}{{SSD}}{smooth signed distance surface reconstruction from Calackli and Taubin \cite{calakli2011SSD,calakli2012SSD-C}}

\newacronym{SURFMRS}{{SURFMRS}}{surface reconstruction from multi-resolution sample points from M\"ucke \etal \cite{mucke2011Surface}}

\newglossaryentry{RealityCapture}
{
    name={RealityCapture},
    description={State-of-the-art commercial 3D reconstruction software for image data sets \cite{realityCapture}}
}

\newglossaryentry{Redner}
{
    name={Redner},
    description={differentiable rendering framework from Li \etal \cite{li2018Redner}}
}

\newglossaryentry{Soft3D}
{
    name={Soft3D},
    description={\MPI-based 3D reconstruction and novel view synthesis method from Penner and Li \cite{penner2017Soft3DReconstruction}}
}

\newglossaryentry{TensorFlow}
{
    name={TensorFlow},
    description={Google's python and auto differentiation machine learning framework \cite{tensorflow2015whitepaper}}
}

\newacronym{TSR}{{TSR}}
{
    thin surface reconstruction method, Aroudj \etal \cite{aroudj2017ThinSurface}
}

\newacronym{UBG}{{UBG}}
{
    Ummenhofer and Brox's global dense multi-scale reconstruction \cite{ummenhofer2017GlobalDenseMultiscaleReconstruction}
}

\newacronym{UBT}{{UBT}}
{
    Ummenhofer and Brox's point-based 3D reconstruction of thin objects \cite{ummenhofer2013ThinObjects}
}

\newglossaryentry{VolSDF}
{
   name={{VolSDF}},
   description={\SDF{}-based and volumetric reconstrution method by Yariv \etal \cite{yariv2021VolSDF}},
}
\newcommand{\VolSDF}{\gls{VolSDF}}

\begin{document}

\title{ERF: Explicit Radiance Field Reconstruction From Scratch}

\author{Samir Aroudj}
\email{samaroud@fb.com}
\affiliation{%
  \institution{Meta, Reality Labs (RL)}
  \streetaddress{10 Brock Street Regents Place}
  \city{London}
  \country{United Kingdom}
  \postcode{NW1 3FG}
}

\author{Steven Lovegrove}
\email{stevenlovegrove@gmail.com}

\affiliation{%
  \institution{Unaffiliated}
  \city{London}
  \country{United Kingdom}
}

\author{Eddy Ilg}
\email{eddyilg@fb.com}

\author{Tanner Schmidt}
\email{tanner.schmidt@fb.com}

\author{Michael Goesele}
\email{goesele@fb.com}

\author{Richard Newcombe}
\email{newcombe@fb.com}

\affiliation{%
  \institution{RL}
  \streetaddress{9845 Willows Rd}
  \city{Redmond}
  \state{Washington}
  \country{USA}
  \postcode{98052}
}

\renewcommand{\shortauthors}{Aroudj et al.}

\begin{teaserfigure}%
    \centering%
    \resizebox{\columnwidth}{!}{%
        \begin{subfigure}[t]{0.25\textwidth}
            \centering%
            \includegraphics[width=\textwidth]{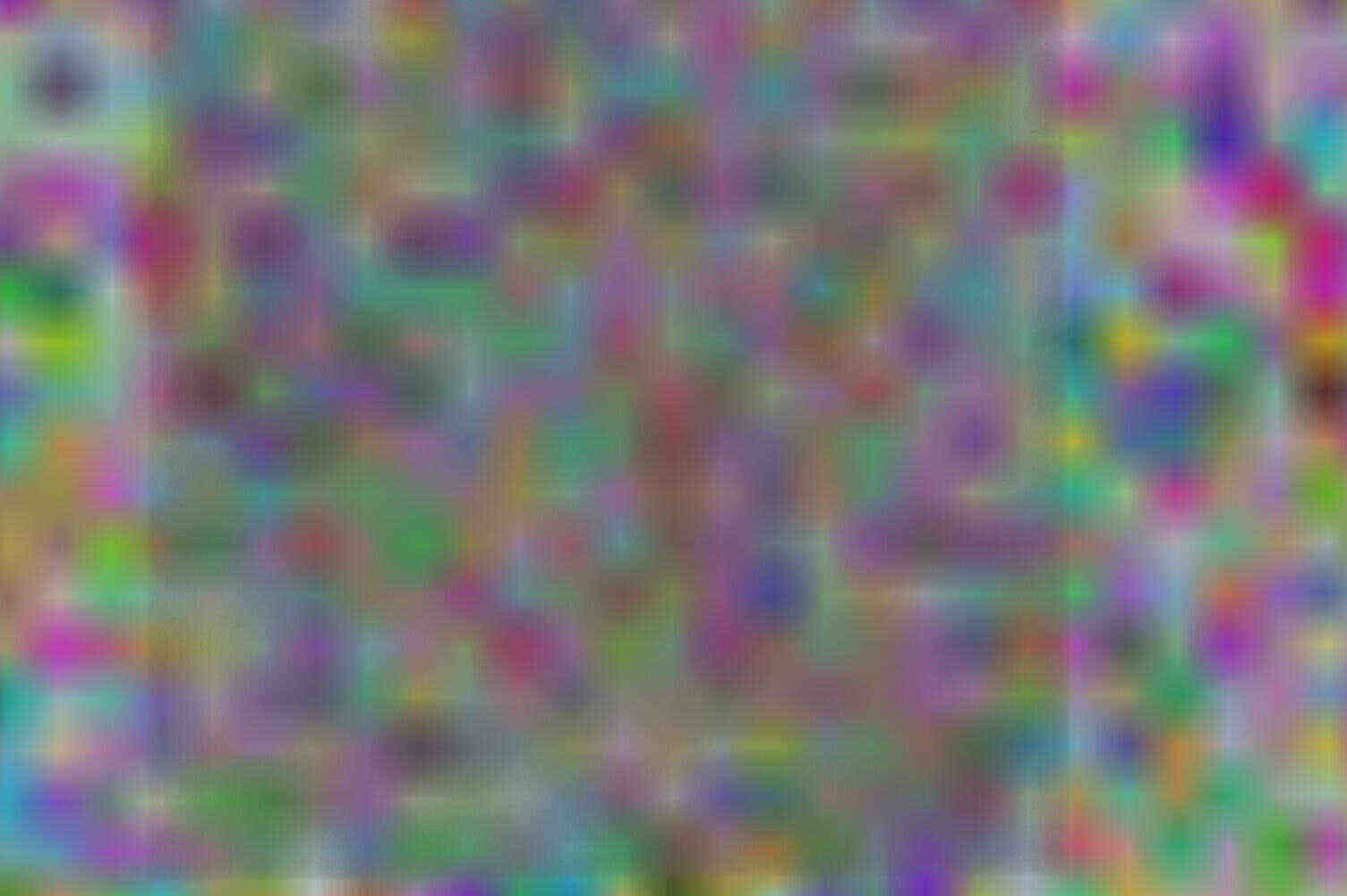}
            \subcaption{
                initialization
                \label{fig:optimizationProgress:initialization}
            }
        \end{subfigure}%
        \imageGap{}%
        \begin{subfigure}[t]{0.25\textwidth}
            \centering%
            \includegraphics[width=\textwidth]{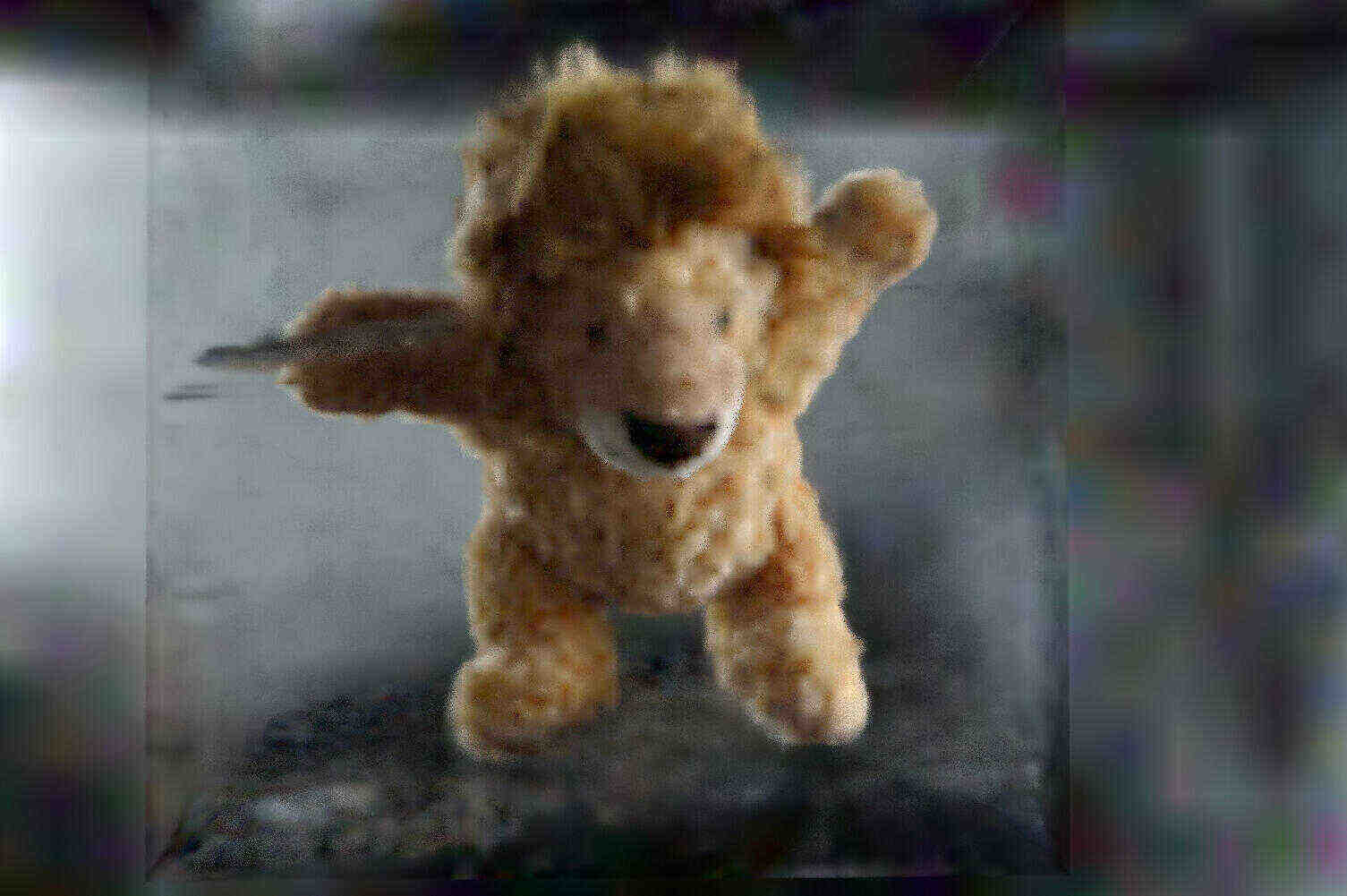}
            \subcaption{
                intermediate result
                \label{fig:optimizationProgress:coarseResult}
            }
        \end{subfigure}%
        \imageGap{}%
        \begin{subfigure}[t]{0.25\textwidth}
            \centering%
            \includegraphics[width=\textwidth]{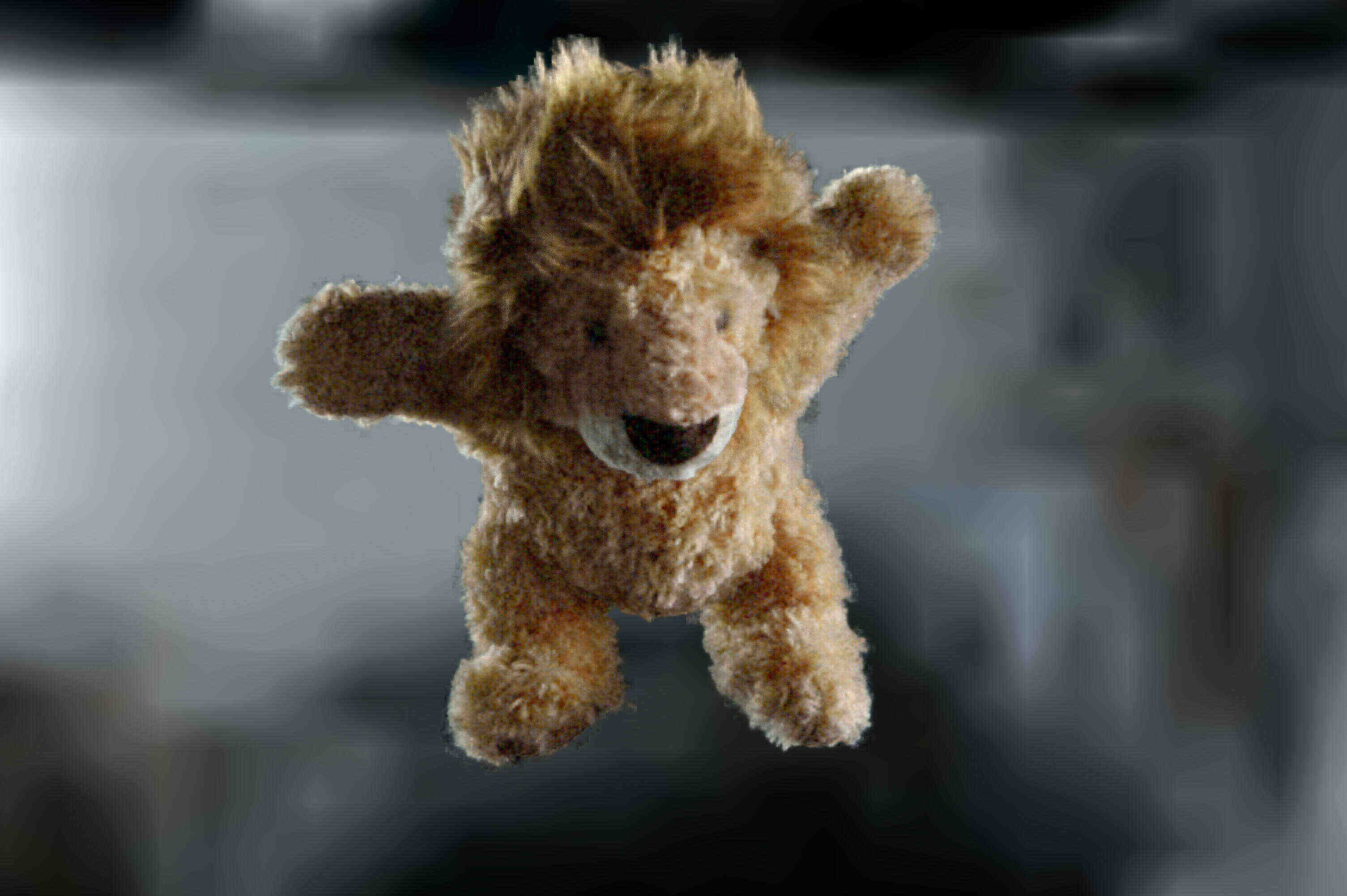}
            \subcaption{
                final result
                \label{fig:optimizationProgress:detailedResult}
            }
        \end{subfigure}%
        \imageGap{}%
        \begin{subfigure}[t]{0.25\textwidth}
            \centering%
            \includegraphics[width=\textwidth]{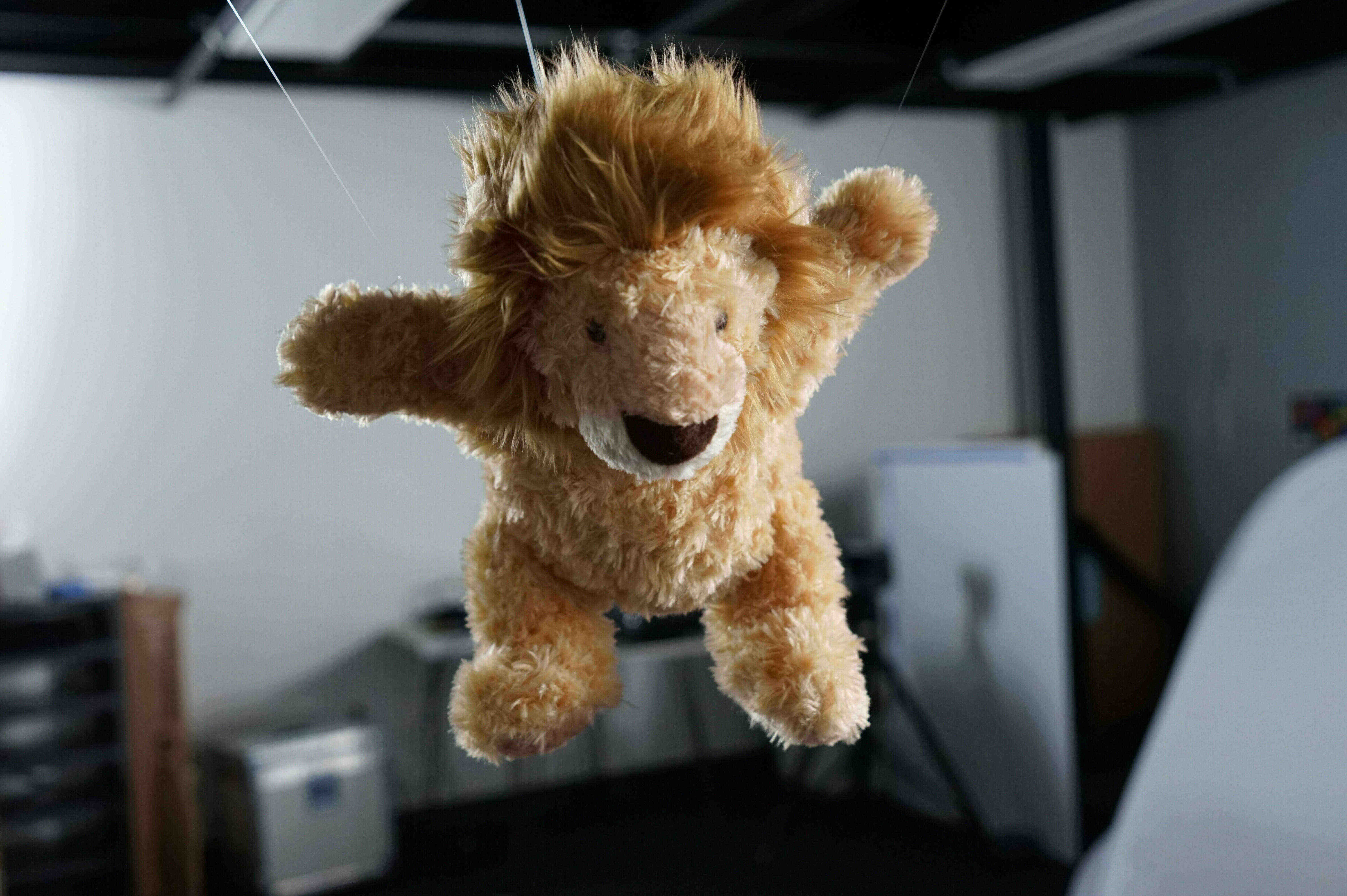}
            \subcaption{
                ground truth
                \label{fig:optimizationProgress:groundTruth}
            }
        \end{subfigure}%
    }
    \caption%
    {%
        Our from-scratch reconstruction of a plush lion
        demonstrated by
        three renderings (left)
        which gradually match an hold-out test photo (right).
    }%
    \label{fig:teaser}
    \label{fig:optimizationProgress}
    \label{fig:lion}
\end{teaserfigure}

\begin{abstract}

We propose a novel explicit dense 3D reconstruction approach
that processes a set of images of a scene with sensor poses and calibrations
and estimates a photo-real digital model.
One of the key innovations is that
the underlying volumetric representation is completely explicit
in contrast to neural network-based (implicit) alternatives.
We encode scenes explicitly
using clear and understandable mappings
of optimization variables to scene geometry and their outgoing surface radiance.
We represent them
using hierarchical volumetric  fields stored in a sparse voxel octree.
Robustly reconstructing such a volumetric scene model
with millions of unknown variables
from registered scene images only
is a highly non-convex and complex optimization problem.
To this end,
we employ stochastic gradient descent (Adam)
which is steered by an inverse differentiable renderer.

We demonstrate that our method can reconstruct models of high quality
that are comparable to state-of-the-art implicit methods.
Importantly,
we do not use a sequential reconstruction pipeline
where individual steps suffer from incomplete or unreliable information
from previous stages,
but start our optimizations from uniformed initial solutions
with scene geometry and radiance that is far off from the ground truth.
We show that our method is general and practical.
It does not require a highly controlled lab setup for capturing,
but allows for reconstructing scenes with a vast variety of objects,
including challenging ones,
such as outdoor plants or furry toys.
Finally,
our reconstructed scene models are versatile thanks to their explicit design.
They can be edited interactively
which is computationally too costly for implicit alternatives.
        
\end{abstract}



\maketitle

\section{Introduction}
\label{sec:introduction}


The vast field of 3D reconstruction has been researched actively for decades.
Diverse pioneering works were already published in the 1990s
\cite{%
hoppe1992UnorganizedPoints,%
turk1994ZipperedPolygonMeshes,%
curless1996VRIP,%
gortler1996Lumigraph,%
szeliski1998MPIStackOfAcetates,%
seitz1999VoxelColoring%
}.
Yet, 
there has recently been a strong increase in interest in the field
due to the availability of powerful optimization techniques
such as \Adam{} \cite{kingma2014Adam}
combined with novel neural-network-based extensions
of such traditional earlier works
\cite{lombardi2019NeuralVolumes,mildenhall2019LLFF,mildenhall2020NeRF,yariv2020IDR,zhou2018StereoLearnedMPI}.
We also employ powerful optimization techniques,
but in contrast to the current research trend,
which creates the impression that 
state-of-the-art reconstructions are only possible
using neural-network-based models,
we reconstruct \emph{explicit high-quality 3D models from scratch};
i.e., only from multi-view images (with sensor poses and calibrations).
In the course of this,
we employ inverse differentiable rendering
paired with \Adam{} as a variant of \SGD{} 
and without any implicit components based on neural networks.
That way,
we can provide a practical reconstruction method for editable models.
Specifically,
it allows for capturing of static scenes
by simply taking photos from different view points.

In contrast,
the recently popular works \NeRF{}~\cite{mildenhall2020NeRF}
and IDR~\cite{yariv2020IDR}
as well as their many follow-up works
employ implicit scene representations (often \glspl{MLP}).
These network-based methods are able
to generate novel views with extremely high fidelity,
while only requiring very compact implicit scene models.
However, the fact that
the internals of these implicit models cannot be interpreted
poses significant challenges and 
leveraging the success of traditional graphics and vision techniques by
combining them with implicit models is an open and challenging research question.
Scaling purely implicit models for large scale scenes
is also challenging.
It is unclear
how to properly increase the capacity of purely implicit models,
i.e., how to extend the black box internals,
in a controlled manner without overfitting or oversmoothing artifacts.
Avoiding these limitations motivated recent hybrid extensions as 
mixes of im- and explicit models
\cite{hedman2021BakingNeRFs,liu2020NeRFSparseVoxelFields,martel2021ACORN,reiser2021KiloNeRF,yu2021PlenOctrees}.
Additionally,
implicit models come at the cost of reduced versatility.
In particular,
they are less suited for 3D content authoring,
e.g., using interactive tools like \Blender{}.
Editing operations on implicitly defined model parts
first have to go through a black box compression layer
inherent to their implicit definition
which entails costly optimization.
Although,
some  early works focus on editing of implicit models
\cite{liu2021EditCondRadFields,yang2021GeomProcNeuralFields},
they are rather a proof of concept targeting
small scale synthetic objects and
are too costly for practical use.

The goal of this work is to address these shortcomings.
We designed an explicit approach with the benefits of
being interpretable, scalable and editable.
Our method can handle complex scenes with tiny intricate scene details,
e.g., the fur of the plush lion in~\Fig{\ref{fig:lion}}. 
We further demonstrate that
our reconstructed models are
suitable for post processing,
such as interactive editing via tools like \Blender{},
and that they are comparable to state-of-the-art implicit models 
regarding photo consistency.

Our contributions are the following:
\begin{enumerate}
    \item a hierarchical, multi-resolution, sparse grid data structure 
        using \glspl{SVO} with 3D fields
        for opacity and outgoing radiance \glspl{SLF}
        to explicitly represent scene geometry and appearance,
    \item a storage and interpolation scheme using local planes 
        to efficiently represent 3D fields
        with little voxel artifacts,
    \item a simple, yet effective background model
        for distant scene radiance to handle unbounded scene volumes,
    \item an opacity compositing rendering algorithm
        that takes pixel footprints into account and
        thus avoids \LoD{} aliasing,
    \item a uniform, hierarchical (coarse-to-fine) optimization scheme
        to make the approach feasible and scalable,
    \item a practical reconstruction method
        for freely captured multi-view images of static scenes,
        neither requiring object masks, nor a sophisticated model initialization.
\end{enumerate}

Our novel scene representation has several benefits.
We represent scenes using complex (high-dimensional), continuous, volumetric and differentiable  3D fields
suitable for intricate geometry details
that leverage powerful optimization methods such as \Adam{} \cite{kingma2014Adam}.
Unlike ours, 
previous explicit and discrete representations,
such as \glspl{MSI} or general meshes,
make strong assumptions or are inherently difficult to optimize
due to their challenging objective function originating from their discrete model design.
Compared to network-based alternatives, while using more memory, 
our novel scene representation is explicit and better suited for interactive editing
as transformation operations do not go through an additional compression layer
inherent to implicit model definitions.
Further,
we argue that our explicit models facilitate research
to leverage the strengths of traditional graphics techniques.
E.g., our spatial scene partitioning can directly accelerate ray-based queries of scenes,
which are fundamental to implement complex shading,
instead of directly storing \glspl{SLF}.
We demonstrate a versatile and practical explicit approach that 
neither requires a restrictive laboratory setup 
(\cite{bi2020DeepReflectanceVolumes,bi2020NeRFForAppearanceAcquisition}),
nor does it need object masks (\cite{yariv2020IDR,boss2021NeRD,zhang2021PhySG})
which tend to be difficult to acquire or might be inaccurate
(hard or impossible for intricate geometry like the fur in \Fig{\ref{fig:lion}}).
Finally,
our 4D scene partitioning (3D space plus \LoD{}) is uniform and straightforward
to apply to captured scenes.
It neither requires an artificial \LoD{} separation 
(\cite{mildenhall2020NeRF,barron2021Mip-NeRF}),
nor does it require a special scene-dependent parameterization 
(\cite{mildenhall2020NeRF,zhang2020NerfPlusPlus}).


\section{Related Work}
\label{sec:relatedWork}

In the following,
we provide a short overview of closely related works,
whereas we focus on the im- or explicity of the underlying scene representations
and the corresponding practicality or usability implications.
First,
we discuss a group of methods
that model scenes using single-hypothesis surfaces.
Representations from this group tend to be
challenging during optimization
or they make strong assumptions limiting their use.
Second,
we look at methods trying to circumvent the latter disadvantages.
The second group models scenes using soft-relaxed surfaces.
I.e., these methods employ volumetric representations
that support multiple simultaneous hypotheses for the same surface.
Our approach falls into the second group.
We provide a soft-relaxed, but very explicit representation.
Our focus is on comprehensibility and versatility of the underlying representation 
thanks to its explicit design.

\subsection{Models with single surface hypothesis}
First,
we review representations with "strict" surfaces (without geometry soft relaxation).

\paragraph{Layered meshes}
Layered meshes are special cases
designed for novel view synthesis only.
That is why
they can be implemented using very regular and simple geometric structures.
I.e.,
they model a complete scene
using only a pyramid of rectangles (\MPI{}) or via concentric spheres (\MSI{}).
Due to their strong focus and simplicity,
they allow for efficient and high-quality novel view rendering at the same time.
One of the first such methods
reconstructs a scene as explicit \MPI{} with regular depths and
plane texels directly consisting of opacity and color values 
that are rendered using opacity compositing \cite{szeliski1998MPIStackOfAcetates}.
More recent works are rather hybrid methods with learned neural networks
that predict explicit \RGBA{} layers for
\glspl{MPI}~\cite{zhou2018StereoLearnedMPI}
or \glspl{MSI}~\cite{broxton2020ImmersiveLFVideo},
respectively assuming a novel viewer
in front of the rectangle pyramid or
at the center of the concentric spheres.
Even more on the implicit end,
\cite{wizadwongsa2021NeX} predict hybrid mesh layers
with neural basis functions for the appearance of scene surfaces,
since they better model view-dependent effects than simple \RGBA{} texels.
Such learned layered models interpolate the captured radiance fields of individual scenes well
within a limited range of views. 
However,
they fail to synthesize farther away views and
the reconstructed geometry can deviate significantly from the actual surfaces.
To maintain the quality,
faking view-dependent effects via "ghost" layers is required,
which prevents manual scene editing.

\paragraph{General, directly optimized meshes}
In contrast to \glspl{MPI} and \glspl{MSI},
general mesh-based methods aim at obtaining a complete and accurate surface reconstruction.
Therefore the results are much more versatile,
but also more difficult to reconstruct.
Direct full mesh optimization 
\cite{aroudj2017ThinSurface,luan2021UnifiedShapeSVBRDFDiffRendering} is difficult,
because the discrete representation leads to highly non-convex objective functions.
In particular,
such approaches are prone to miss necessary gradients during optimization and
they require an initialization
that is already close to the global optimum.
They can quickly degrade to invalid manifolds and
often cannot improve the topology during optimization.

\paragraph{Continuously and implicitly defined surfaces}
Owing to the drawbacks of directly optimized meshes,
the community also researched continuous representations
that implicitly define scene surfaces and
entail a better behaviour of the objective function.
For example,
the surfaces extracted from the \SDF{} representations~\cite{yariv2020IDR,zhang2021PhySG,kellnhofer2021NeuralLumigraphRendering} 
are manifolds by definition and can easily adapt topology during optimization.
However,
these methods still assume that
the scene can be well reconstructed using clearly defined single-hypothesis surfaces.
Depending on where the implicitly defined surface intersects or does not intersect view rays,
this leads to discontinuities in the objective function
that are hard to handle.
To avoid the local optima originating from these discontinuities,
additional constraints from object segmentation masks are required.
Since these masks by themselves are hard or impossible to obtain without human assistance,
this poses a significant limitation. 
\VolSDF{} is a follow up work
which circumvents this limitation.
In particular,
the authors again suggest modeling scene surfaces using \glspl{SDF}.
However,
the key difference is that,
to account for surface uncertainty and enforce geometric constraints (Eikonal loss) at the same time,
they infer scene density fields from the underlying \glspl{SDF}.
Then they can render the density fields using volume rendering
which facilitates optimization and removes the need for masks.
Note that,
thanks to its density-based uncertainty,
\VolSDF{} is a hybrid
that belongs to both related work groups discussed here.
Disadvantageously,
a single, scene-global parameter controls surface uncertainty
which makes it less suited for varying fine and intricate geometry.
Also only results for small scale scenes were shown by the authors.
In general, 
intricate geometry of for example plants or fur as in \Fig{\ref{fig:lion}}
is difficult to represent using \SDF{}- and mesh-based methods.
Representing such geometry accurately often requires a resolution
with prohibitively high costs. 
The limitations of methods with single-hypothesis surfaces
motivated not only \VolSDF{},
but also other recent, continuous representations.
As the next subsection explains,
these approaches represent geometry using volumetric fields
that inherently support multiple surface estimates at the same time
to facilitate optimization and 
also support fine and intricate geometry approximations.

\subsection{Models with multiple surface hypotheses}
This subsection reviews methods
that model scenes with soft-relaxed geometry.
We begin with models that are on the very implicit end and
continue going towards the very explicit end of scene representations.

\paragraph{\glspl{NeRF} with baked-in \glspl{SLF}}
Thanks to the pioneering implicit approach \NeRF{}~\cite{mildenhall2020NeRF},
\glspl{MLP} that encode the geometry and surface radiance (\SLF{}) of individual scenes volumetrically
have become the recently dominant representation.
They model individual scenes via 5D fields consisting of
continuous volumetric density for geometry
coupled with view-dependent surface radiance for appearance.
These compact \MLP{} models represent surfaces
continuously and in a soft-relaxed and statistical manner.
Meaning they consist of continuous fields that smoothly change during optimization.
They furthermore implement a soft relaxation by
allowing to model opaque surfaces as spread out or partially transparent.
The latter allows for multiple surface hypotheses during optimization
which improves convergence by reducing the issue of missing correct gradients.
To avoid novel view synthesis errors,
they can furthermore approximate fine and intricate surfaces statistically.
Nevertheless,
the original \NeRF{} method came with severe limitations owing to, for example,
its special scene parameterization,
capture setup requirements or
its static scene radiance assumption.
I.e.,
aliasing issues and multi-scale input data are focused by \cite{barron2021Mip-NeRF},
the follow-up work \cite{zhang2020NerfPlusPlus} allows for a more flexible capture setup
or \cite{martinbrualla2020NerfInTheWild} targeted the static radiance assumption
by disentangling transient occluders from the static parts of a scene and
by additional latent codes for per-image variations.

\paragraph{Decomposed \glspl{NeRF} with physically-based components}
Other follow-up works,
such as \cite{bi2020DeepReflectanceVolumes,boss2021NeRD,srinivasan2021NeRV,zhang2021NeRFactor}
focus on more explicit models by
decomposing the previously directly stored scene radiance
into more explicit components.
By jointly estimating
incoming illumination
as well as surface geometry and materials,
they aim at re-achieving
some of the versatility of traditional explicit representations.
However,
these approaches either require
a very restrictive laboratory capture setup,
object masks
or they only work for small scale scenes with centered objects.
Note that object masks also implicitly prevent intricate materials, e.g., fur or grass,
for which it is difficult to acquire accurate masks in practice.
Furthermore, in the case of \NeRV{}~\cite{srinivasan2021NeRV}),
only results for limited synthetic data
instead of captured real-world scenes are presented.
Like the original \NeRF{}, 
our scene models also directly store the outgoing surface radiance
of individual static radiance scenes.
However,
we store the outgoing radiance
using a sparse hierarchical grid with \glspl{SH}
instead of a black box \MLP{}.
We leave decomposing the convolution of incoming light with surface materials
as an extension for further research. 
Given our simplifying design choice
for directly storing and optimizing static \glspl{SLF},
our models allow for
direct geometry editing and
simple transformations of the surface appearance,
as we will demonstrate later.
Physically-based editing of
surface materials or scene light transport
is also left for future work at this time.

%
%
The earlier mentioned decomposition approach~\cite{bi2020DeepReflectanceVolumes}
is an exception regarding two aspects.
First,
it does not employ \glspl{MLP}, but a 3D \CNN{}
that decodes surface geometry and materials
into an explicit and dense voxel grid.
Second, similar to our work,
the authors suggest to implement volume rendering
using traditional opacity compositing.
Compared to our approach,
their method is however limited
to a small scale laboratory capture setup with black background.
It furthermore requires a single point light
that coincides with the capturing sensor.
Finally,
it is strongly limited by its
simple dense grid scene structure and
naive scene sampling.
In contrast,
while our scene models only have baked-in appearance,
we are able to optimize for more general scenes
with less controlled and unknown static radiance fields.
To support highly detailed reconstructions,
we present our coarse-to-fine optimization using sparse hierarchical grids
with our comparatively more efficient importance sampling scheme.

\paragraph{Spatially explicit hybrids}
The hybrid \emph{\gls{NeuralVolumes}} \cite{lombardi2019NeuralVolumes}
represents scenes captured with a light stage
using an encoder and decoder network.
It decodes a latent code into a regular \RGBA{} voxel grid
that is rendered using ray marching and alpha blending.
To allow for detailed reconstructions despite the dense regular grids,
the authors suggest to also learn warp fields
to unfold compressed learned models.
However,
ray marching through dense grids is still inefficient
and \RGBA{} grids cannot handle view-dependent effects
without ghost geometry.
%
%
More recent and more efficient hybrids
with implicit and explicit model parts 
\cite{liu2020NeRFSparseVoxelFields,hedman2021BakingNeRFs,reiser2021KiloNeRF}
also explicitly partition the 3D scene space into cells,
but using more efficient and view-dependent sparse voxel grids.
This allows for
overall higher model resolution,
more efficient scene sampling or
faster rendering.
These methods respectively 
cache computationally expensive volume rendering samples,
use a single feature-conditioned \MLP{} or
many simple and thus low-cost \glspl{MLP}
distributed over the sparsely allocated grid cells.
On the contrary,
we employ completely explicit scene models.
For multi-resolution rendering,
efficient sampling and
to limit memory consumption,
our models built on sparse hierarchical grids.
To keep our approach practical and
allow for optimizing freely captured and thus uncontrolled scenes,
we also directly cache the \SLF{} of scene surfaces using \glspl{SH}.

\paragraph{\SVO{} hybrids using \SH{}-based appearance}
%
%
The recently published \PlenOctree{} models \cite{yu2021PlenOctrees}
are also more explicit models
and most similar to our scene representations.
The authors likewise model individual scenes with static radiance
with continuous fields for geometry and appearance
that are stored in \glspl{SVO}.
Similar to us,
they handle view-dependent effects using \glspl{SH}.
However, in contrast to our method,
their method requires a multi-step reconstruction pipeline
starting from registered images.
The first step of this pipeline is
to obtain a coarse \NeRF{} scene model.
The \NeRF{} model then defines the space partitioning (and sparsity)
of their \SVO{} via its density field.
The authors suggest converting the \NeRF{} scene model
into the new \SVO{} by dense sampling
and finally optimize its per-node geometry and appearance parameters
to obtain an improved model.
Having the initial coarse \NeRF{} model as free space constraint,
reduces memory consumption and
helps avoid false clutter in free space.

In contrast,
we show that it is feasible to
reconstruct 3D scenes directly and uniformly from images with sensor poses and calibrations
using an explicit representation.
We achieve high model resolution
using \glspl{SVO}
that we gradually build and 
which we optimize with \Adam{},
but without any implicit, network-based model parts.
Since in our case free space and surfaces are initially completely unknown,
our coarse-to-fine optimization with dynamic voxel allocations is critical
to not run out of memory.
As important part of the coarse-to-fine optimization,
we present our local plane-based storage and interpolation schemes
for the volumetric fields attached to our \glspl{SVO}.
These schemes allow for approximating thin and fine geometric details,
even initially, when only a coarse \SVO{} is available.
The explicit coarse-to-fine reconstruction from registered images
furthermore requires efficient scene sampling.
We implemented an importance sampling scheme
that filters sampling points gradually and 
according to the current geometry estimate.
In that way,
our method is
not limited by the restrictions coming from an initialization
and the \SVO{} structure can dynamically adapt
to the scene content without external guide.
To avoid blurry transitions from free to occupied space and
to obtain clear surface boundaries, 
our volume rendering also differs
by implementing traditional opacity compositing
instead of an exponential transmittance model.
Note that
we model geometry not using a density field,
but via an opacity field
representing soft-relaxed surfaces only and not occupied space.
Our geometry representation is well suited for
inverse differentiable rendering,
opaque surfaces as well as
intricate geometry such as fur.
Finally,
we use our \SVO{} structure for \LoD{} interpolation
and provide a background model for more flexibility regarding capture setups.
In contrast to the \PlenOctree{} work,
we can also reconstruct scenes with an unbounded volume,
e.g., an outdoor scene with all sensor poses roughly facing the same direction.
\section{Overview}
\label{sec:overview}

In the following,
we present our scene representation and 
the corresponding reconstruction algorithm.

\subsection{Input and output}
\label{subsec:dataOverview}

\begin{figure}
    \centering%
        \resizebox{\columnwidth}{!}{%
            \begin{tikzpicture}%
                \draw (0, 0) node
                {
                    \includegraphics[width=\columnwidth]{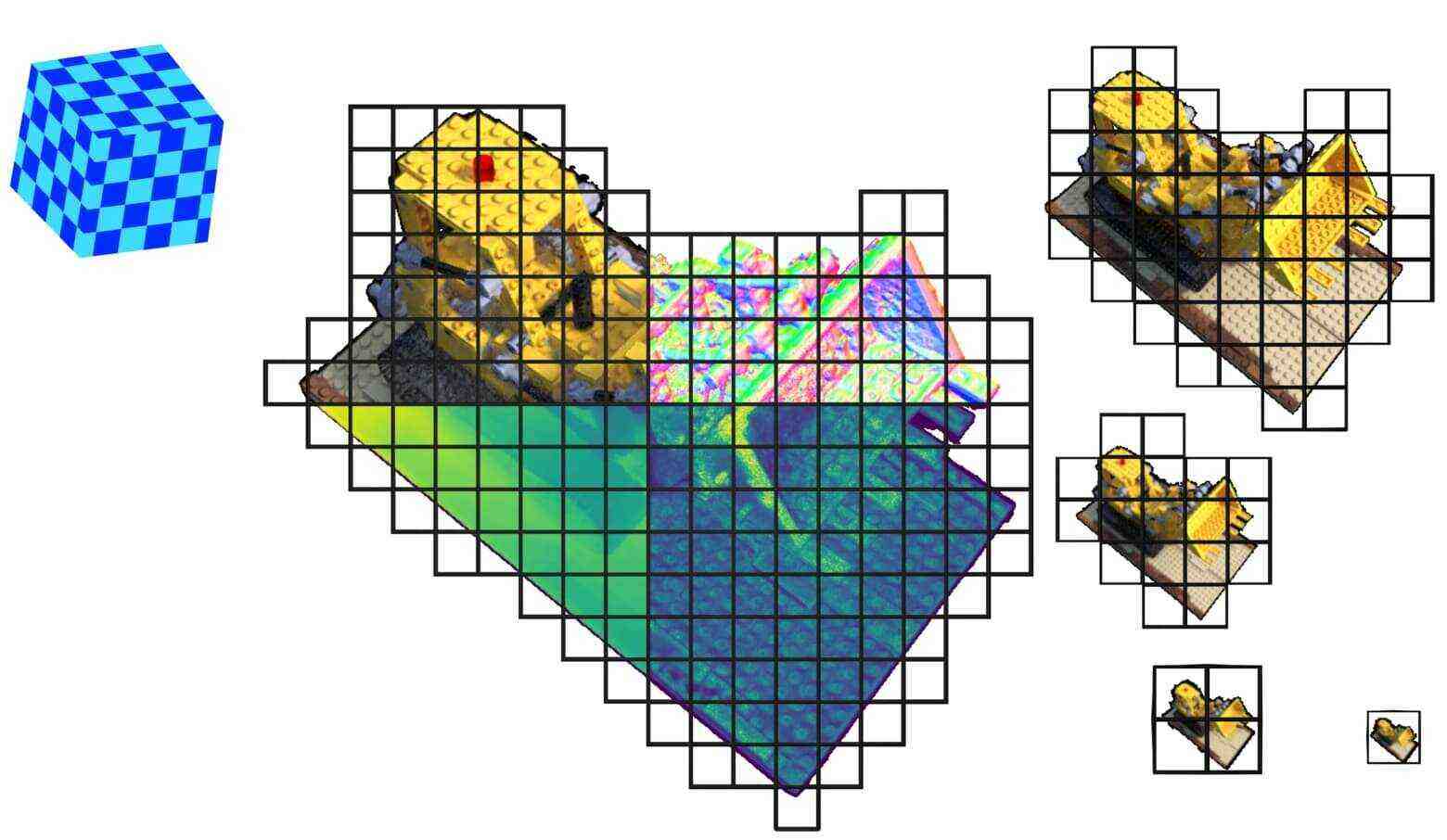}
                };
        	    \node[black] at ( -1.5,  2.2) {\small leaves};
        	    \node[black,rotate=50] at (   1.6, -1.75) {\small opacity};
        	    \node[black,rotate=-40] at ( -1.6, -1.35) {\small depth};
        	    \node[black,align=center] at ( -3.3, 0.5)
        	    {
        	        \small environment \\
        	        \small map
        	    };
        	    \node[black,align=center] at (3.75, -1.0)
        	    {
        	        \small coarse \\
        	        \small \SVO{} \\
        	        \small levels
        	    };
                \node (colorMapNormals) at (0.2, 2.0) []
                {
                    \includegraphics[width=0.05\textwidth]{Figures/NormalsWhiteness0.25}%
                };
        	    \node[black] at (node cs:name=colorMapNormals,anchor=south) {\small normals};
        	\end{tikzpicture}%
        }%
    \\
    \resizebox{\columnwidth}{!}{%
        \begin{tikzpicture}
    	    \node (colorMap) at (0, 0)
    	    {
    	        \includegraphics[width=\columnwidth,height=0.15cm]{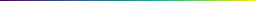}
            };
    	    \node (colorMapLowLabel) at (node cs:name=colorMap,anchor=south west)
    	    [xshift=1.7cm,yshift=-2pt,black]  {\small close to sensor/low opacity};
    	    \node (colorMapHighLabel) at (node cs:name=colorMap,anchor=south east)
    	    [xshift=-1.4cm,yshift=-2pt,black]  {\small far away/high opacity};
    	\end{tikzpicture}%
    }%
    \caption%
    {%
        Scene model sketch.
        The background model (cube map, left) complements the \SVO{}.
        The \SVO{} stores detailed surfaces in its leaves (center)
        and
        coarser approximations in its inner nodes (right).
        Each node has an opacity and multiple \SH{} parameters.
    }%
    \label{fig:modelOverview}%
\end{figure}

We reconstruct explicit 3D models from unordered multi-view input images.
In particular,
given an unstructured set of images,
we first run standard \SfM{} techniques in a preprocess.
I.e.,
we first run \Colmap{} for sensor poses and intrinsics as well as a sparse feature point cloud.
Second,
to bound the scene parts of interest to be reconstructed,
we manually estimate conservative min- and maximum corners of the scene \AABB{}
using the \SfM{} feature points.
Third,
we run our actual reconstruction algorithm with input data consisting of
\begin{enumerate}
    \item the multi-view images,
    \item their sensor poses and calibrations and
    \item the coarse, conservative \AABB{}.
\end{enumerate}
Our reconstruction algorithm outputs a scene model containing
an \SVO{} within the given scene \AABB{}.
The output scene model further consists of a background model,
an environment map that complements the \SVO{}.
It represents distant scene regions such as sky for example.
\Fig{\ref{fig:modelOverview}} sketches our scene representation.

The \SVO{} stores the "actual" scene.
First,
it stores a volumetric scalar field with opacity defining surface geometry.
Second,
it stores a volumetric vector field with \glspl{SH} defining the scene \SLF{}.
Note that
the opacity models soft-relaxed surfaces and not occupied space.
The scene \SLF{} contains the total outgoing radiance
for each surface point along each hemispherical direction.
Also note that
each tree level represents the scene at a specific \LoD{}.
I.e., to support varying scene detail level,
our \SVO{} represents the scene also using inner nodes
analogously to mipmap textures.

\paragraph{Use and limitations}
Our scene models are explicit, differentiable and statistical representations.
To facilitate robust reconstruction from scratch and editing,
the volumetric \SVO{} fields for opacity and outgoing radiance
statistically approximate surfaces,
allow for multiple surface hypothesis
during optimization contrary to "accurate" surface models and
have a clear meaning in contrast to network weights.
The model parameters only go through straightforward constraints
ensuring physically meaningful values
while an \SGD{} solver can still freely update the parameters.
Data transformations are simpler for explicit scene models
since the operations do not need to go through an additional compression layer
inherent to compact network-based models.
In case of networks,
these operations again require costly optimizations
when they target implicitly defined model parts.
Equally,
initializing our model with a specific state is simpler.
In this work,
we started reconstructions from scratch
with a mostly transparent and uninformed random fog
to demonstrate the flexibility and robustness of our approach,
see \Fig{\ref{fig:optimizationProgress}}.
Though,
initializing models
using results from prior steps such as
\SfM{} would also be straightforward.
Physically based scene transformations,
e.g., recalculating shadows when the opacity field changes,
are however out of scope of this work.
We made the simplifying decision
to directly store and optimize the radiance outgoing from the
scene surfaces using \glspl{SH}.

\subsection{Algorithm overview.}
\label{subsec:algorithmOverview}

Please see \Tab{\ref{tab:notation}} for the notation
we will use throughout the rest of the document.
\Alg{\ref{alg:overview}} describes our method on a high level and
details will follow in the remaining method sections.
Our algorithm first coarsely initializes the new scene model
and then gradually extends the \SVO{} (outer loop)
according to the repeated optimization of its fields (inner loop).

\begin{table}
    \centering
    \begin{tabular}{cl|cl}
         $\radianceOutgoingSymbol$              & \SVO{} \SLF{} &
         $\opacity$                             & \SVO{} opacity field \\
         $\distantRadianceSymbol$               & distant radiance &
         $\raw{o}$                              & unconstrained $\opacity$ \\
         $\normal{\pos}$                        & surf.\ normal at $\pos$ &
         $\lossCache$                           & loss cache \\
         $\pixelBatch{i}$                       & pixel batch; pixel $i$ &
         $\rayIndexed{i}$                       & ray of pixel $\pixelIndexed{i}$ \\
         $\camPosIndexed{i}$                    & sensor center of $\rayIndexed{i}$ &
         $\rayDirIndexed{i}$                    & direction of $\rayIndexed{i}$ \\
         $\opacitySampleij{i}{k}$               & opacity at $\rayIndexed{i}(\opacitySampleCoordsij{i}{k})$ &
         $\raySamplePoints{i}{j}$               & ray batch depths \\
         $\opacitySample{i}$                    & pixel opacity $i$ &
         $\raySamplePoints{i}{k}$               & depths subset \\
         $\radianceSample{i}{l}$                & \SLF{} at $\radianceSampleCoords{i}{l}$ &
         $\raySamplePoints{i}{l}$               & subset of subset \\
         $\pixelLPCVIndexed{i}$                 & pixel radiance $i$ & 
         $\pixelIntensityPixelIndexed{i}$       & rendered pixel $i$ \\
         $\lossPixelIndexed{i}$                 & photo loss $i$ &
         $\pixelIntensityGTPixelIndexed{i}$     & ground truth $i$ \\
         $\footprint(\pixelIndexed{i})$         & pixel footprint i & 
         $\SHBasisIndexed{l}{m}$                & \SH{} basis function \\
         $\density$                             & density &
         $\SHCoefficientIndexedlm{l}{m}$        & \SH{} coefficient \\
    \end{tabular}
    \caption
    {
        Notation overview,
        see \Alg{\ref{alg:overview}} for use.
        We abbreviate the notation of sampling points on
        rays when used within our equations. E.g.,
        we denote the opacity of
        the $j$-th sampling point
        on the ray $\rayIndexed{i}$
        at ray depth $\rayT_{i,j}$
        which is located at the 3D location $\pos = \rayIndexed{i}(\opacitySampleCoordsij{i}{j})$
        by $\opacitySampleij{i}{j}$.
        An indexed exemplar element $\pixelIndexed{i}$ surrounded
        by curly brackets denotes a set.
        E.g.,
        $\pixelBatch{i}$ is a pixel batch and
        the renderer samples
        each view ray $\rayIndexed{i}$
        for each optimization batch pixel $\pixelIndexed{i}$
        at depths $\raySamplePoints{i}{j}$.
    }
    \label{tab:notation}
\end{table}

\begin{algorithm}
    \SetAlgoLined
    \caption{Hierarchical optimization}
    \label{alg:overview}
    
    \comment{init model \& pixel errors cache}
    \SVO{} = createDenseGrid(\AABB{}) \comment{random $\opacity$, $\radianceOutgoingSymbol$}
    \label{alg:initSVO}
    $\distantRadianceSymbol$ = randomEnvMapRadiance()  \;
    \label{alg:initBackgroundRadiance}
    $\lossCache$ = highLossForAllInputPixels() \;
    \label{alg:initLossCache}
    \nonl\;
    
    \comment{Optimize: inv.~diff.~rendering \& \SGD{}}
    \For{$n = 0$ \KwTo $N$}
    {
        \label{alg:coreSGD}
        
        $\pixelBatch{i}$ = importanceSample($\lossCache$) \comment{error driven}
        \label{alg:importanceSampling}
        
        $\left\{\rayFullIndexed{i}\right\}$ = 
            castRays($\pixelBatch{i}$) \;
        \label{alg:castRays}
         
        $\raySamplePoints{i}{j}$ =
            stratifiedSampling(\SVO{}, $\rayBatch{i}$) \;
            \label{alg:stratifiedSampling}
        $\raySamplePoints{i}{k}$ =
            selectRandomly($\raySamplePoints{i}{j}$) \comment{uniform}
            \label{alg:stochasticLimitingUninformed}
            
        $\opacitySamplesij{i}{k}$ =
            getOpacity(\SVO{}, $\raySamplePoints{i}{k}$) \;
            \label{alg:getOpacity}
            
        $\raySamplePoints{i}{l}, \opacitySamplesij{i}{l}$ =
            selectRandomly($\raySamplePoints{i}{k}, \opacitySamplesij{i}{k}$) \;
            \label{alg:stochasticLimitingInformed}
            
        $\radianceSamplesij{i}{l}$ =
            getSLF(\SVO{}, $\rayBatch{i}, \raySamplePoints{i}{l}$) \;
            \label{alg:getSLF}
            
        $\radianceSamples{i}, \opacitySamples{i}$ =
            blend%
            (%
                $\opacitySamplesij{i}{l}$,%
                $\radianceSamplesij{i}{l}$%
            ) \;
            \label{alg:alphaCompasitingSVORadiance}
        $\radianceSamples{i}$ = 
            blend%
            (%
                $\radianceSamples{i}$,%
                $\opacitySamples{i}$,%
                $\left\{\distantRadiance{-\rayDir_i}\right\}$%
            ) \;
            \label{alg:alphaCompasitingBackgroundRadiance}
            
        $\pixelIntensityBatch{i}$ =
            sensorResponses($\radianceSamples{i}$) \;
        \label{alg:sensorResponses}
            
        $\lossBatch{i}$ =
            loss($\pixelIntensityBatch{i}$, $\pixelIntensityGTBatch{i}$) \;
            \label{alg:loss}
        \SVO{}, $\distantRadianceSymbol$ = 
            makeStep(\SVO{}, $\distantRadianceSymbol$, $\nabla(\lossBatch{i})$) \;
            \label{alg:step}
            
        $\lossCache$ = update($\lossCache$, $\lossBatch{i}$)  \comment{track errors}
            \label{alg:cacheUpdate}
    }
    \label{alg:SGDEnd}
    \nonl\;
    
    \comment{New \SVO{} via opacity $\opacity$ and footprints $\footprint$}
    mergeLeaves(\SVO{}) \comment*{compact free space}
    \If{subdivideLeaves(\SVO{}, $\left\{\footprint(\pixelIndexed{i})\right\}$)}%
    {
        resetOptimizer() \comment*{due to new unknowns}
        go to line \ref{alg:coreSGD} \;
    } 
\end{algorithm}

First,
our algorithm creates a coarse dense grid,
i.e., a full and shallow \SVO{},
within the given scene bounding box.
We randomly initialize the 3D fields of the \SVO{}
with grayish fog and random radiance
(line \ref{alg:initSVO}).
We similarly randomly initialize the environment map
(line \ref{alg:initBackgroundRadiance}),
as shown in \Fig{\ref{fig:optimizationProgress:initialization}}.

We then mainly optimize the parameters of the 3D fields without changing the tree structure
using multi-view volumetric \IDR{} and \SGD{}
(lines \ref{alg:coreSGD} - \ref{alg:SGDEnd}).
To this end,
we randomly select small batches of input image pixels using importance sampling 
(line \ref{alg:importanceSampling});
cast a ray for each selected pixel into the scene 
(line \ref{alg:castRays});
distribute scene sampling points along each ray using stratified importance sampling 
(line \ref{alg:stratifiedSampling});
query the scene \SVO{} for opacity and \SLF{} samples at these ray sampling points
(lines \ref{alg:getOpacity}, \ref{alg:getSLF});
accumulate the returned field samples along each ray 
and also add the visible background radiance
using classical opacity compositing
to estimate the totally received scene radiance for each selected pixel
(lines \ref{alg:alphaCompasitingSVORadiance} - \ref{alg:alphaCompasitingBackgroundRadiance});
map the received radiance to pixel intensities using the response curve of the sensor
(line \ref{alg:sensorResponses})
and, finally, compare the estimated against the input image pixel intensity
(line \ref{alg:loss})
for a model update step
(line \ref{alg:step}).

Using the gradients of our differentiable volumetric rendering,
we iteratively update the scene model using \SGD{}
to fit the scene model parameters to the input images
for a fixed model resolution (constant model parameter count).
Additionally,
we infrequently update the tree structure.
In particular,
we merge or subdivide tree nodes
to adapt the resolution based on the current surface geometry estimate.
We do so until the \SVO{} is sufficiently detailed
with respect to the input images.
The following subsections describe these algorithmic steps
in more detail.

\section{Explicit, sparse, hierarchical model}
\label{sec:representation}

Our explicit scene model consists of a sparse hierarchical grid,
i.e., an \SVO{}.
It stores an opacity and \RGB{} \SH{} parameters per node
to encode scene surfaces and the radiance leaving them
as a scalar and a vector field.
The \SVO{} stores both of these fields defined next
at each tree level and not only using the leaf nodes
to support multiple levels of detail for rendering and optimization.
We assume that everything outside of the \AABB{}
that bounds the scene \SVO{} is infinitely far away and
therefore represent all remaining scene parts
using an environment map implemented as a cube map.

\subsection{Surface geometry}
\label{subsec:opacity}
Our \SVO{} provides a continuous multi-resolution scalar field $\opacity$.
To implement it,
the \SVO{} stores a continuous, scalar, volumetric field
$\opacity: \real^3 \mapsto [0, 1]$ per tree level.
Each tree level with its individual field represents a single \LoD{}.
To this end,
each tree node,
including inner ones,
stores 1 floating-point opacity parameter (besides the \SLF{} parameters).
Note that
inner nodes hence approximate surfaces at a larger scale.
The continous opacity field $\opacity$ represents surfaces statistically.
Specifically,
the opacity $\opacity(\pos)$ represents the coverage of a planar slice
perpendicular to the radiance traveling through $\pos$ and
thus what percentage of it gets locally absorbed.
I.e., it is a surface property expressing
what relative percentage of photons statistically hits the surface at $\pos$,
e.g., $\opacity(\pos_\text{free}) = 0$ and $\opacity(\pos_\text{wall}) = 1$.
As detailed later,
the \SVO{} not only interpolates within 3D space,
but also blends between the individual \LoD{} fields
to serve scale-extended position queries.
Only regarding a single \LoD{} and given a query location $\pos$,
the \SVO{} interpolates the parameters of the tree nodes
surrounding the scene location $\pos$.
This results in a raw, unconstrained estimate $\raw{\opacity}(\pos)$
which needs to be constrained to be physically meaningful as explained next,
but which allows the optimizer to freely update the opacity parameters.

\begin{figure}
    \centering%
    \begin{subfigure}[t]{0.48\columnwidth}
        \centering%
        \includegraphics[width=\textwidth]{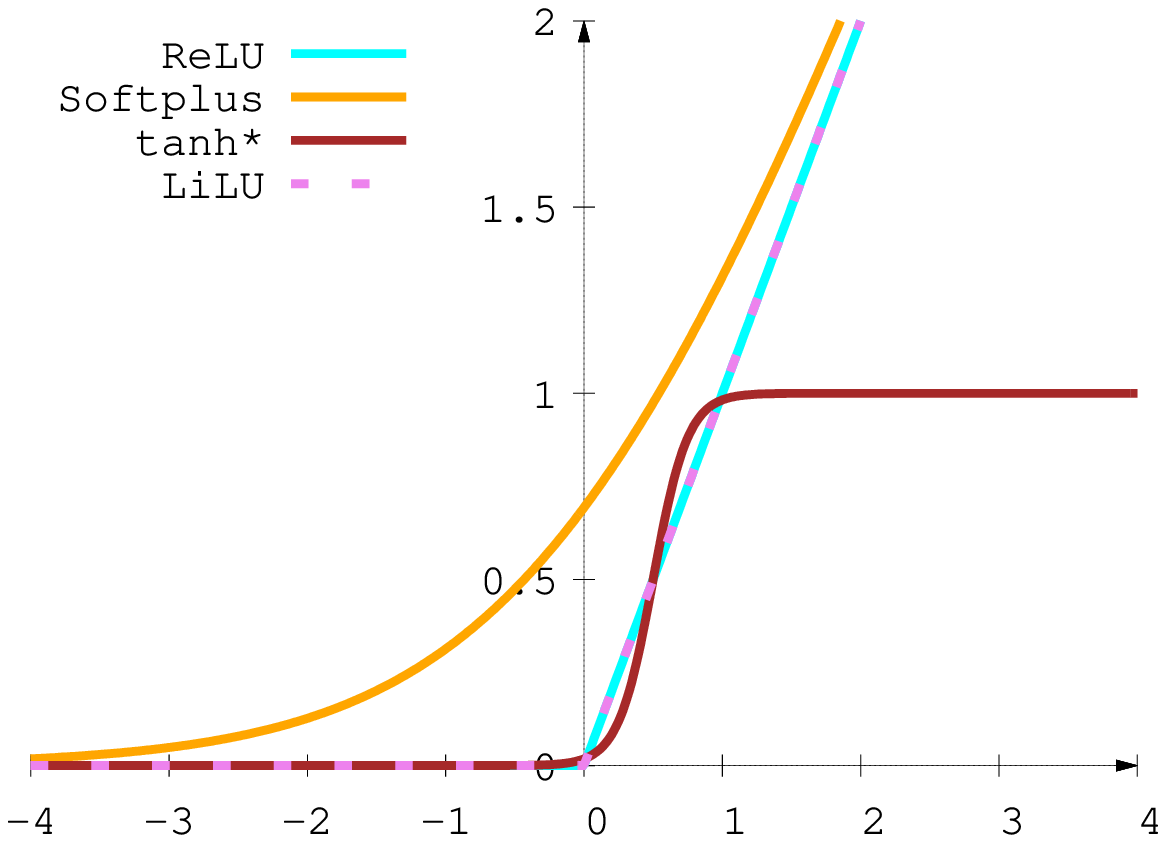}
        \subcaption{Model constraints.}
        \label{fig:activationFunctions}
        \imageGap{}%
    \end{subfigure}%
    \begin{subfigure}[t]{0.48\columnwidth}
        \centering%
        \includegraphics[width=\textwidth]{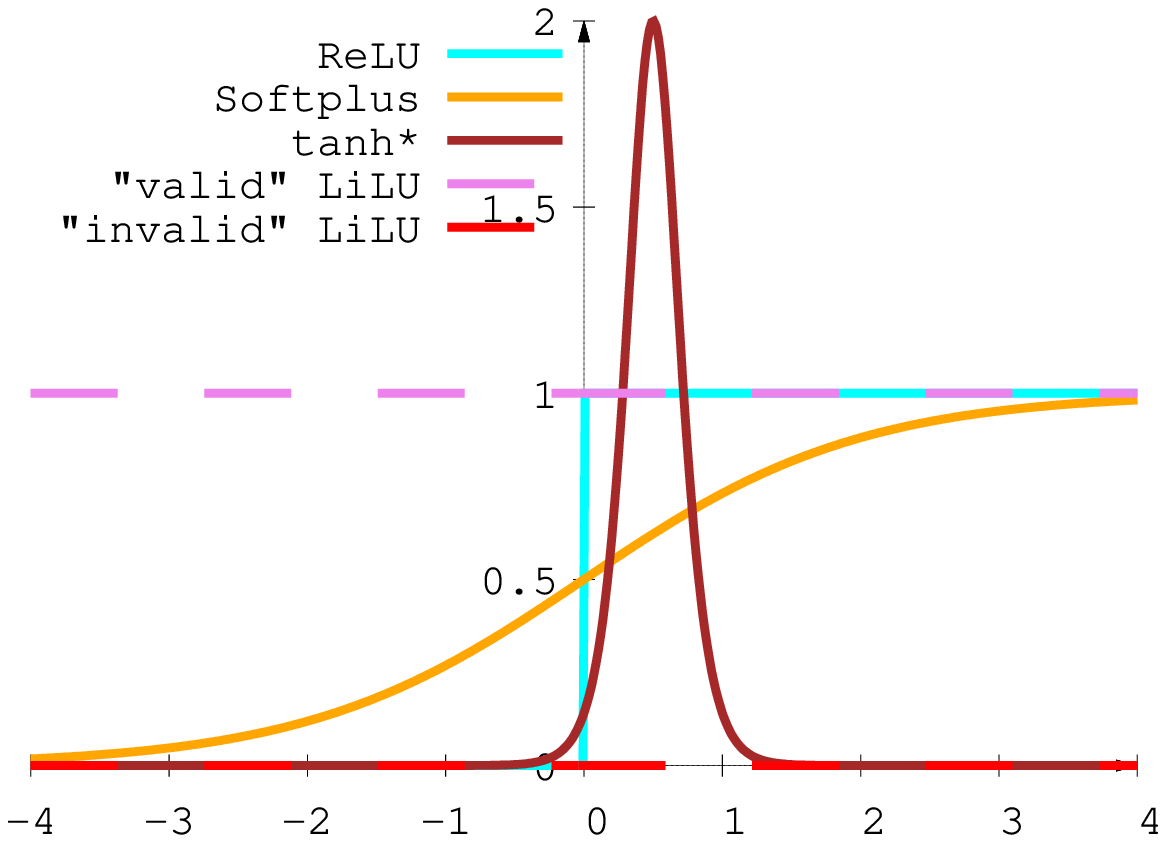}
        \subcaption{Model constraints' gradients.}
        \label{fig:activationFunctionGradients}
        \imageGap{}%
    \end{subfigure}%
    \caption%
    {%
        Model constraints.
        \subref{fig:activationFunctions}) shows multiple model constraints
        (activation functions)
        to map model unknowns to physically valid ranges.
        \Softplus{} only slowly approaches zero.
        \gls{ReLU} has no gradient for $x \leq 0$
	where optimizers can easily fail.
        Our \LiLU{} from \Eq{\ref{eq:LiLU}} avoids these problems.
	Note that
	\LiLU{} and \ReLU{} are equal except for their gradients.
        \subref{fig:activationFunctionGradients}) shows \LiLU{}'s twofold gradient
        which is constantly 1 for all valid optimization steps:
        $\LiLU{}(x_{i}) < \LiLU{}(x_{i+1})$.
        However, \ReLU{} and \LiLU{} are both not C1-continuous.
        This can cause the optimizer to oscillate around
	$\LiLU{}(x)=\ReLU{}(x)=0$.
        Our tanh variant is
        continuous, 
        mostly linear 
        and quickly approaches its borders 
        which facilitates opacity optimization.
    }%
    \label{fig:activationFunctionsAndGradients}
\end{figure}

\paragraph{Constraining the opacity field $\opacity$}
Unlike \NeRF{},
which uses the nonlinear \Softplus{} model constraint (activation function)
to limit density to the interval $[0, \infty)$,
we constrain opacity to $\left[0, 1\right]$
using a variant of $\tanh$:
\begin{equation}
    f(x) = 0.5 \cdot (\tanh(4 x - 2) + 1)
    \label{eq:clamp01TanH}
\end{equation}
which is mostly linear, but smoothly approaches its borders
as illustrated in \Fig{\ref{fig:activationFunctions}}.
This is necessary to prevent the optimizer from oscillating
when updating opacity parameters close to the interval borders.
Note that
it also approaches its borders much faster than \Softplus{} approaches zero.
These properties make it more suitable for free space reconstruction
(zero opacity border).

\paragraph{Surface normals}
As shown by previous works \cite{boss2021NeRD},
storing additional parameters for surface normals and optimizing
them independently of the surface geometry representation
does not work well in practice.
For this reason,
we do not store, but directly infer surface normals
from the raw opacity field gradient via:
\begin{equation}
    \normal{\pos} =
	- \frac{\nabla \raw{\opacity}(\pos)}{\|\nabla \raw{\opacity}(\pos)\|_2}.
    \label{eq:normals}
\end{equation}

\subsection{Surface appearance}
\label{par:SLF}
Our \SVO{} directly stores the "surface appearance".
Similar to related \NeRF{} variants,
we store and optimize the outgoing radiance,
i.e., the convolution of incoming light with surfaces
as a volumetric and view-dependent \SLF{}
denoted by $\radianceOutgoingSymbol$.
Analogous to the surface geometry,
the \SVO{} stores an \RGB{} radiance field per \LoD{} tree level:
$\radianceOutgoingSymbol: \real^5 \mapsto [0, \infty)^3$.
In particular,
each node stores low frequency \RGB{} \SH{} coefficients
$\SHCoefficientIndexedlm{l}{m} \in \real^3$
\cite{ramamoorthi2001EnvMaps}
besides the opacity parameter.
Given a 5D query $(\pos, \direction)$ for evaluating the \SLF{}
at the 3D scene location $\pos$ and
along the direction $\direction$,
we interpolate the \SH{} coefficients of the tree nodes surrounding $\pos$
resulting in a continuous vector field of \SH{} coefficients $\{\SHCoefficientIndexedlm{l}{m}(\pos)\}$.
Next,
we evaluate the \SH{} basis functions $\SHBasisFunctionslm{l}{m}$
with the interpolated coefficients at $\pos{}$
for the radiance traveling direction $\direction{}$
using their Cartesian form:
\begin{equation}
    \raw{\radianceOutgoingSymbol}(\pos, \direction) =
        \sum^{l=b}_{l=0} \sum^{m=l}_{m=-l}
            \SHCoefficientIndexedlm{l}{m}(\pos) \cdot \SHBasisFunctionIndexedlm{l}{m}{\direction}
\end{equation}
whereas $\raw{\radianceOutgoingSymbol} \in \real^3$ denotes the raw, unconstrained \RGB{} radiance,
which again allows the \SGD{} optimizer to freely update the per-node coefficients
$\SHCoefficientsIndexedlm{l}{m}$.
For memory reasons,
we only store the low frequency components of the \SLF{} in practice,
i.e., the first $\SHBandCountSymbol{} = \SHSoftServeSLFBandCount{}$ bands of each color channel
($\SHSoftServeSLFParamsPerNode{}$ coefficients per node in total).

\begin{figure}
    \centering%
    \begin{subfigure}[t]{0.48\columnwidth}
        \includegraphics[width=\columnwidth,trim={0 1cm 0 3.5cm},clip]%
        {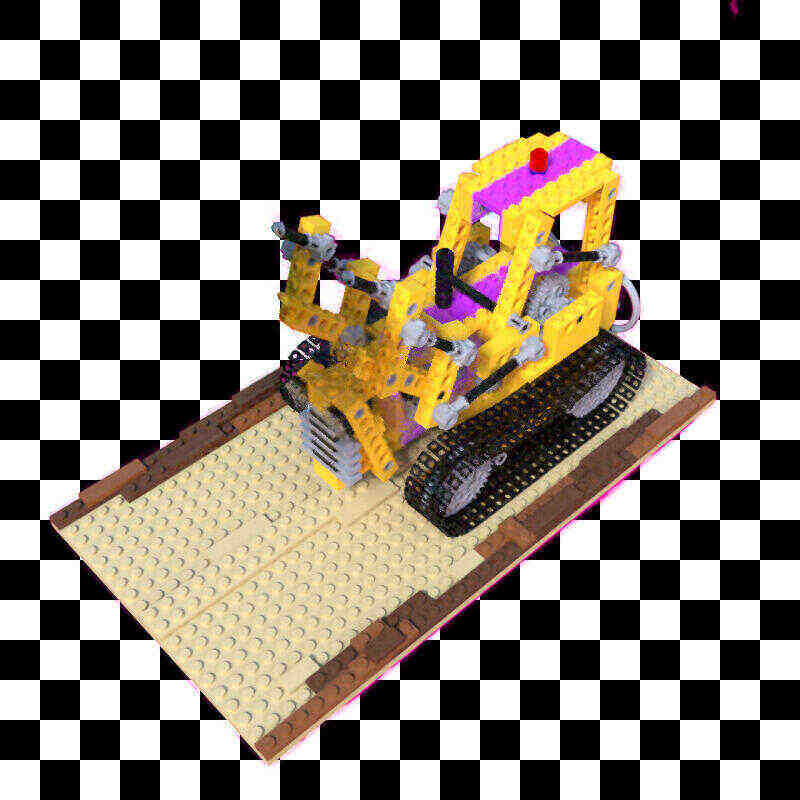}
        \subcaption{Explicit scene editing.}%
        \label{fig:editingExample}%
        \imageGap{}%
    \end{subfigure}
    \begin{subfigure}[t]{0.48\columnwidth}
        \includegraphics[width=\textwidth]{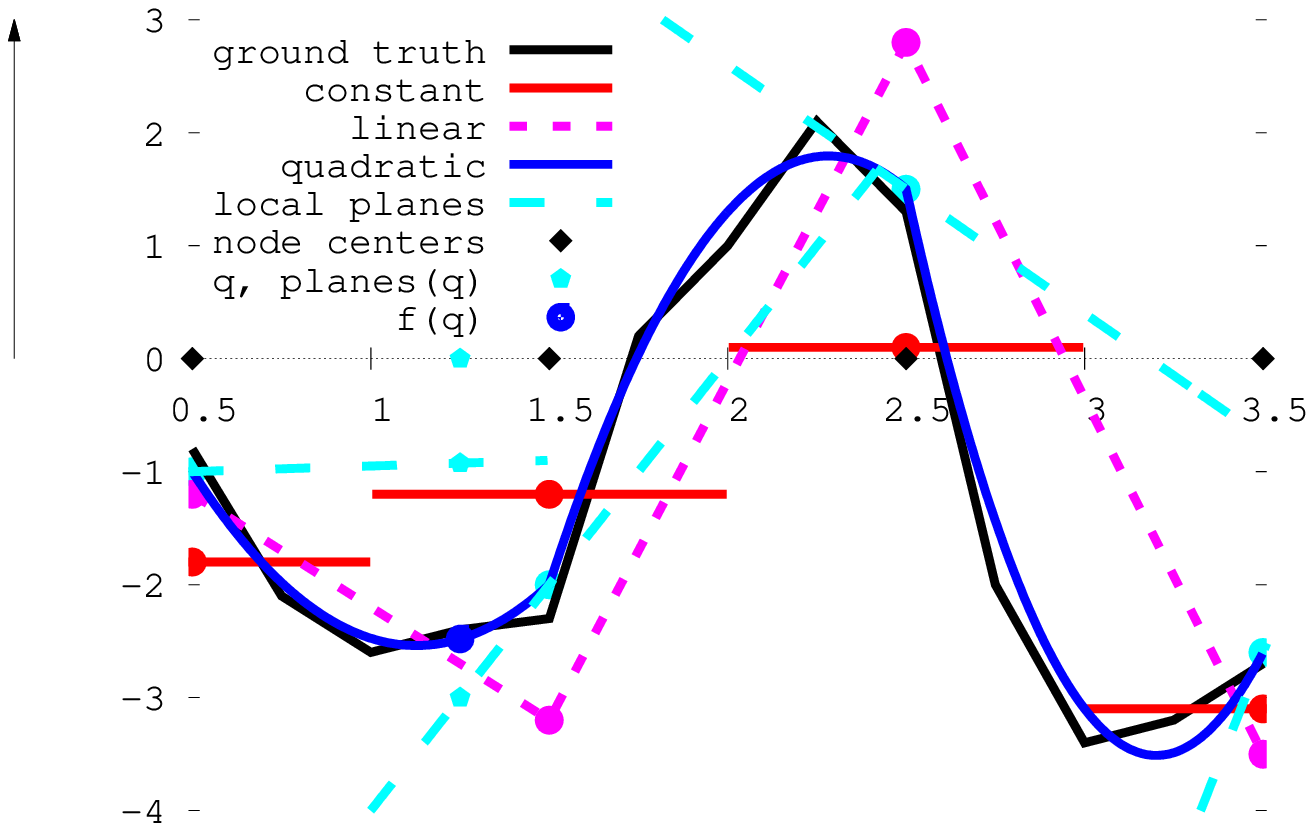}
        \subcaption{Interpolation comparison.}
        \label{fig:interpolation}%
    \end{subfigure}%
    \caption%
    {%
        (\subref{fig:editingExample})
        Our representation is suited for tools such as \Blender{}
        which we demonstrate using exemplary operations on the \NeRF{} bulldozer.
        We switched \SLF{} \RGB{} channels to purple
	    within a narrow \AABB{} at the vehicle center and
        cut out an \AABB{} volume enclosing the shovel.\newline
        (\subref{fig:interpolation})
        More complex interpolation of discrete samples stored by the scene \SVO{}
        to represent continuous fields is more expensive,
        but also  achieves a better model fit
        as shown by the example target 1D scalar field $f$ (black).
        Linear interpolation (magenta) achieves a much better model fit
        than simple nearest neighbor lookup (red).
        However, our quadratic interpolation (blue)
        using optimized local planes (cyan) provides the best fit.
        The extrema of the approximating function
        need not coincide with the tree node centers.
        Quadratic interpolation for the field value $\field{\pQuery}$ (blue dot)
        at the query $\pQuery$ (cyan dot)
        entails evaluating the local planes
        and blending the local plane results (cyan dots)
        based on the distance of $\pQuery$
        from its surrounding nodes (black dots).
    }%
    \label{fig:interpolationSchemesAndEditing}
\end{figure}

\paragraph{Constraining the \SLF{}}
To compute the physically meaningful non-negative radiance $\radianceOutgoingSymbol(\pos, \direction)$
after evaluating the \SH{} basis functions for a query $(\pos, \direction)$,
we map the unconstrained outgoing radiance $\raw{\radianceOutgoingSymbol}$ to $[0, \infty)$.
To this end,
we avoid any model constraint (activation function)
producing invalid negative radiance such as leaky \glspl{ReLU},
since they can introduce severe model overfitting.
Also, the frequently used \Softplus{} and \ReLU{}
both have severe disadvantages for this use case,
see \Fig{\ref{fig:activationFunctions}} and \subref{fig:activationFunctionGradients} for details.
For these reasons,
we introduce \glspl{LiLU} (Limited Linear Units) to constrain \SLF{} radiance
(\gls{LiLU} code in the supplemental).
\glspl{LiLU} are variants of \glspl{ReLU} with pseudo gradients.
I.e., their actual gradient depends on the state of the input unknown $x$
before ($x_i$) and after its update ($x_{i+1}$):
\begin{equation}
    \begin{split}
        \text{\LiLU{}}(x) &=
            \begin{cases}
                x  & \text{if } x \geq 0\\
                0  & \text{otherwise}
            \end{cases}
        \\
        \frac
        {
            \d{\text{\LiLU{}}}(x)
        }
        {
            \d{x_{i\mapsto i + 1}}
        } &=
            \begin{cases}
                1  & \text{if } x_{i+1} \geq 0  \\
                0  & \text{otherwise}
            \end{cases}
        \label{eq:LiLU}
    \end{split}
\end{equation}
which means we practically \emph{limit the function domain} to $[0, \infty)$.
The gradient is zero only for update steps that would result in an invalid state:
$x_{i+1} < 0$;
but the gradient is 1 for all valid updates, including the very border:
$x_i = 0 \land x_{i+1} \geq 0$.
Hence, the constrained variable is always in the physically valid function image:
$\text{\LiLU{}}(x) \geq 0$.
Our \glspl{LiLU} can be seen as \gls{ReLU} extentions
which do not suffer from complete gradient loss like
\glspl{ReLU}.
They linearly go to zero within the physically valid range
making them more suitable for optimizing
low radiance surfaces than \Softplus{} or other common constraints
that slowly approach the constraint border.
Note that
the stored model parameters in general only go through such easy to understand
constraints and not through black box compression layers, i.e., networks,
which simplifies transforming scenes, e.g., for editing,
see \Fig{\ref{fig:editingExample}}.

\subsection{Plane-based, quadratic 4D interpolation}
\label{subsec:4DInterpolation}
\label{subsec:quadraticInterpolation}
In order to support multi-resolution scene models
which adapt to the viewing distance,
we store scene data
using a tree hierarchy of discrete samples
to allow for 4D interpolation (spatial and \LoD{}).
In particular,
our \SVO{} stores all multi-resolution volumetric fields
using local plane-based samples (function value plus spatial gradient)
which we interpolate between.
Specifically,
an \SVO{} stores
the opacity $\opacity: \real^4 \mapsto \real$ and
the \SLF{} $\radianceOutgoingSymbol: \real^4 \mapsto \real^{\SHSoftServeSLFParamsPerNode{}}$
of a scene.
Each of these two multi-resolution fields is in turn composed of multiple single-resolution fields,
one per tree level.
Note that
this same scheme can be applied to other fields as well.
For example,
surface materials can be attached to the \SVO{} and interpolated in 4D analogously.
In the following,
we abstractly refer to such fields as
$\fieldSymbol: \real^4 \mapsto \real^\dimension$.

Our quadratic 4D field interpolation
for evaluating a field $\fieldSymbol$ as in
\Alg{\ref{alg:overview}}, lines \ref{alg:getOpacity} and \ref{alg:getSLF},
works as follows:
When processing an interpolation query $\field{\pQuery}$
for a scale-extended scene sampling point
$\pQuery = [\pos^t = \rayShortIndexed{i}, \footprintBackProj{}] \in \real^4$
on a view ray $\rayIndexed{i}$,
we first compute its footprint $\footprintBackProj{} \in \real$ (spatial extend)
via back projecting the diameter of the corresponding pixel $\pixelIndexed{i}$
along $\rayIndexed{i}$ to the depth $\rayT$.
Computing $\field{\pQuery}$ then entails
interpolating between the discrete local plane-based samples surrounding $\pQuery$.
Each tree node $j$ stores one such local plane
$\nodePlaneIndexed{j} =
 [
    \fieldIndexed{0}{\pNodeIndexed{j}, \depthNodeIndexed{j}},
    \fieldGradient{\pNodeIndexed{j}, \depthNodeIndexed{j}}^t
 ]
 \in \real^4$.
The local planes are addressed using their 4D coordinates
consisting of the node center and depth
$(\pNodeIndexed{j}, \depthNodeIndexed{j})$.
\Fig{\ref{fig:interpolation}} shows a 1D field example
with 3 such plane-based samples (cyan dashed lines)
and the result of blending them together (blue graph).
Based on the distance $\Delta\pos$ of $\pQuery$ to the surrounding nodes (black dots),
we evaluate each local plane individually (cyan dots) and
blend them together for $\field{\pQuery}$ (blue dot)
using weights $\weightSymbol{}$.
The weights $\weightSymbol{}$ are also based on the distance $\Delta\pos$
(simple linear \LoD{} and trilinear spatial interpolation)
making the overall interpolation quadratic.
In particular,
for a single 4D point query $\pQuery$,
the linear blending functions $\weightSymbol{}$ interpolate between the local planes
$\{\nodePlaneIndexed{j} | j \in \neighborsSixteen{\pQuery}\}$
of the 4D 16-node neighborhood $\neighborsSixteen{\pQuery}$ surrounding $\pQuery$ as follows
\begin{equation}
    \begin{split}
       \field{\pQuery} &=\!\!\!\!
        \sum_{j \in \neighborsSixteen{\pQuery}}\!\!\!\!
        {
            \wQLerp{\pQuery}{\nodeScopeIndexed{j}}
            \cdot
            \left(
                \fieldIndexed{0}{\pNodeIndexed{j}, \depthNodeIndexed{j}}
                + \Delta\pos \cdot \fieldGradient{\pNodeIndexed{j}, \depthNodeIndexed{j}}
            \right)
        }\\
        \wQLerp{\pQuery}{\nodeScopeIndexed{j}} &=
            \wLoD{\footprintBackProj{}}{\footprintNodeIndexed{j}}
            \cdot
            \wTLerp{\pos}{\pNodeIndexed{j}}
    \end{split}
    \label{eq:interpolation}
\end{equation}
wherein
$\nodeScopeIndexed{j} = [\pNodeIndexed{j}^t, \footprintNodeIndexed{j}] \in \real^4$
depicts the center position and diameter of the neighboring tree node $j$;
$\Delta_{\pos} = (\pos - \pNodeIndexed{j})$
is the distance vector from the node center to the 3D query position
and $\depthNodeIndexed{j}$ is the tree depth of node $j$.
The blending functions $\wTLerpSymbol{}$ and $\wLoDSymbol{}$ respectively provide
trilinear and linear weights
depending on the distance of the query within Euclidean 3D and within \LoD{} space.

\begin{figure}
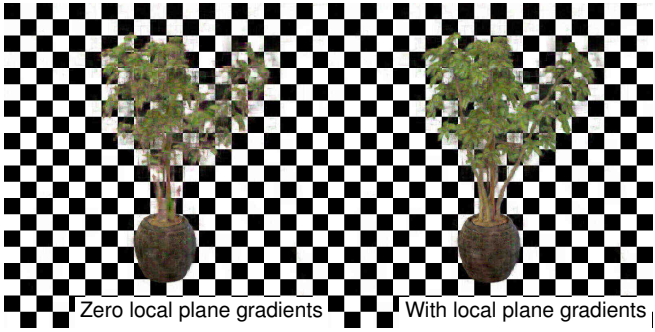

    \centering%
    \begin{subfigure}[t]{0.24\textwidth}
        \centering%
        \labelImage{\textwidth}{Results/AblationStudy/SpatialGradients/FicusView300It185K/2022_01_19_07_44_59_f21/PhotoWithZeroSpatialGradients} {Zero local plane gradients}%
        \imageGap{}%
    \end{subfigure}%
    \begin{subfigure}[t]{0.24\textwidth}
        \centering%
        \labelImage{\textwidth}{Results/AblationStudy/SpatialGradients/FicusView300It185K/2022_01_19_07_44_59_f21/PhotoWithSpatialGradients} {With local plane gradients}%
        \imageGap{}%
    \end{subfigure}%
    \caption%
    {%
        Plane-based storage and interpolation of \SVO{} fields
        using the ficus scene \cite{mildenhall2020NeRF}.
        Given only the coarse initial \SVO{},
        linear interpolation of only function samples results in extrema fixed to voxel centers
        and a worse model fit (left) compared to
        using the spatial gradients of our local plane samples
        as sketched by \Fig{\ref{fig:interpolation}} (right).
    }%
    \label{fig:gradientBasedInterpolation}
\end{figure}

Importantly,
this allows the optimizer to freely place the field extrema in 3D space
despite that it updates the pairs of function samples and their local gradients
($\fieldSymbol_0, \fieldGradientSymbol)$
which are only stored at the \SVO{} node centers.
This is in contrast to only optimizing direct function samples $\fieldSymbol_0$
which would only support function extrema at limited discrete node centers $\pNodeIndexed{j}$
as shown theoretically by the 1D scalar field example in \Fig{\ref{fig:interpolation}}.
Allowing the optimizer to continuously position the field extrema
(instead of fixing them to the discrete centers of the \SVO{} nodes)
is critical for fine geometry reconstruction
when the initial \SVO{} is only coarse
as demonstrated by \Fig{\ref{fig:gradientBasedInterpolation}}.
For regions where the \SVO{} is sparse,
a globally constant "border" plane $\nodePlaneIndexed{b}$ representing free space
substitutes the data of all missing neighbors.

A single point query entails two trilinear interpolations
using the blending function $\wTLerpSymbol$,
each one for the scene sampling point $\pos$ and
the corresponding 8 surrounding node centers $\pNodeIndexed{j}$
at the same tree level $\depthNodeIndexed{j}$.
We then linearly blend both 3D interpolation results
along the \LoD{} dimension of the tree
using the function $\wLoDSymbol$.
Our 4D interpolation algorithm determines the two depths
$\depthNode$ and $(\depthNode - 1)$
of the two surrounding 8-neighborhoods
(meaning $\depthNodeIndexed{j} = \depthNode$ or $\depthNodeIndexed{j} = \depthNode - 1$)
according to the Nyquist sampling theorem to avoid aliasing:
\begin{equation}
    \footprintNodeIndexed{\node}
	\leq \footprintBackProj{}
	< 0.5 \cdot \footprintNodeIndexed{\node}
    \label{eq:samplingTheorem}
\end{equation}
whereas a lower depth and thus a "blurry" query result is returned
if the tree is not deep enough for the query.
Note that
this \LoD{}-aware sampling scheme is similar to sampling mipmap textures.

\subsection{Background model and initial state.}
\label{par:backgroundModel}
Since our \SVO{} is limited to a given scene \AABB{},
we need to represent all captured radiance which emerged from outside the \SVO{}.
To this end,
we assume that all scene parts outside the \SVO{} are infinitely far away
and model the corresponding radiance using an environment map
$\distantRadianceSymbol: \real^2 \mapsto \real^3$
which only depends on the radiance traveling direction.
Specifically,
each model contains a background cube map that complements the \SVO{}.
Cube maps have the advantage of consisting of locally limited \texel{}s.
This prevents oscillations during optimization,
in contrast to, for example, an \SH{}-based background
for which each single frequency band parameter influences the whole background.
Exactly like for the outgoing radiance $\radianceOutgoingSymbol$ stored in our \SVO{},
we constrain the distant radiance $\distantRadianceSymbol$
using our \LiLU{}-constraint.
The background \LiLU{} processes the bilinear interpolation result
of the optimized cube map radiance \texel{}s.

\paragraph{Initial state}
\label{par:backgroundModelInitialState}
We initialize the background with random radiance and
the opacity and radiance fields of the \SVO{} with "grayish fog".
The initial opacity field is mainly transparent
to avoid false occlusions that would decrease convergence speed.
I.e,
opacity parameters are drawn from a uniform distribution,
such that a ray going from the min- to the maximum scene \AABB{} corner
accumulates only up to $0.05$ total opacity.
\SH{} coefficients are respectively drawn
from the uniform random distributions $[0.2475, 0.5025]$ and
$[-0.025, 0.025]$ for band 0 and all higher bands;
background radiance texels from $[0, 1]$.
See \Fig{\ref{fig:optimizationProgress:initialization}} for an example.

\section{Rendering ERFs}
\label{sec:rendering}

To render pixels
(\Alg{\ref{alg:overview}} lines \ref{alg:castRays} to \ref{alg:sensorResponses}),
we cast a ray $\rayFullIndexed{i}$ into the scene
for each pixel $\pixelCoordsVecIndexed{i}$
starting at the camera center $\camPosIndexed{i}$ and
going along the viewing direction $\rayDirIndexed{i}$.
Our renderer gathers all the visible scene radiance
along a ray from potentially multiple surfaces
to estimate the \RGB{} intensity of the corresponding pixel.
For this purpose, 
we distribute sample points along each ray within intersected \SVO{} nodes;
filter the resulting point set multiple times
to make later expensive gradient computations feasible;
query the \SVO{} fields and apply our 4D interpolation scheme,
see \Eq{\ref{eq:interpolation}} and \Fig{\ref{fig:interpolation}}
and accumulate the drawn samples along each ray as detailed next.

\subsection{Volume rendering algorithm}
\label{par:opacityCompositing}

The authors of \NeRF{} \cite{mildenhall2020NeRF} suggested
rendering scenes via an exponential transmittance function:
\begin{equation}
    \begin{split}
        \pixelRadianceCamPosRayDir &=
            \transparencyCum{\rayTFar} \cdot \distantRadiance{-\rayDir} +
            \int_{\rayTNear}^{\rayTFar}
            {
                \transparencyCum{\rayT} \cdot \density(\rayT) \cdot \shadingFull d\rayT
            }
        \\
        \transparencyCum{\rayT} &=
        \exp
        \left(
            -\int_{\rayTNear}^{\rayT}
            {
                \density(\rayTInner) d\rayTInner
            }
        \right).
    \end{split}
    \label{eq:NeRFVolumeRendering}
\end{equation}
This formulation is a twice adapted exponential transmittance model
for volumetric rendering of participating media only absorbing or emitting radiance
\cite{pharr16PhysicallyBasedRendering}.
The traditional parts include
the emitted light
via the outgoing radiance field $\radianceOutgoingRaySample{}$
and the occlusion term $\transparencyCum{\rayT}$
via the extinction coefficients $\density$.
The first adaptation is by the \NeRF{} authors
who called the extinction coefficients density
and they suggested to add the multiplication
by the scene density $\density$ within the outer integral
and interpreted the outer integral as the expected radiance.

Note that
the second adaptation of the volume rendering
of \Eq{\ref{eq:NeRFVolumeRendering}} is by us
to support a broader variety of capture setups than the \NeRF{} formulation
using $\distantRadianceSymbol$.
See for example the different setup of \Fig{\ref{fig:optimizationProgress}}.
The extension $\distantRadianceSymbol$ adds background radiance to the model,
see \ref{par:backgroundModel}.

However,
we find the aforementioned transmittance model unsuited
for our use cases for the following reasons:
First,
the latter exponential transmittance model assumes that
scene geometry consists of uncorrelated particles,
which is not true for opaque surfaces \cite{vicini2021NonExponential}.
Second,
our goal is modeling soft-relaxed surfaces
suited for optimization via inverse differential rendering and \SGD{}
and also suited for approximating intricate geometry such as grass.
Modeling uncorrelated particles of participating media is not our goal,
but we rather estimate coverage by approximated, but structured surfaces
as explained under \ref{subsec:opacity}.
Observed scenes usually contain mostly free space and opaque surfaces,
but not participating media.
Finally,
there is no scientific physical background
for the mentioned density multiplication of \Eq{\ref{eq:NeRFVolumeRendering}}.
Note that \Eq{\ref{eq:NeRFVolumeRendering}} is also too simplistic to model participating media.
For these reasons,
we implemented our forward rendering model
using traditional opacity compositing (alpha blending).

For each ray,
we draw outgoing \SLF{} radiance samples $\radianceOutgoingRaySample{i}$ as well as
opacity samples $\opacity(\rayT_{j})$ determining the blending weights for the totally received radiance along a ray:
\begin{equation}
    \begin{split}
        \pixelRadianceCamPosRayDir &=
            \transparencyCum{\rayTFar} \cdot \distantRadiance{-\rayDir} +
            \!\!\!\sum_{i=0}^{N}
            {
                (\transparencyCum{\rayT_{i}} \!\!-\!\! \transparencyCum{\rayT_{i+1}}) \cdot \radianceOutgoingRaySample{i}
            }
        \\
        \transparencyCum{\rayT_{i}} &= \prod_{j=0}^{i-1} (1 - \opacity(\rayT_{j}))
        \label{eq:opacityCompositing}
    \end{split}
\end{equation}
wherein $\transparencyCumSymbol$ models the leftover transparency.
Since view rays start close to the sensor and in free space,
we set $\transparencyCum{\rayT=0}$ to $1$.
For traditional opacity compositing and in contrast to \NeRF{},
$\transparencyCumSymbol$ directly depends on the relative surface opacity
$\opacity \in [0, 1]$, see \Eq{\ref{eq:clamp01TanH}}.
Similar to layered meshes with transparency,
the opacity $\opacity$ models soft-relaxed surfaces and not filled space.
It is a differentiable coverage term
in contrast to discrete opaque mesh surfaces and
thus more suited for optimization.
However,
in contrast to layered representations,
the underlying geometry is continuously defined over 3D space
which facilitates optimizing its exact location.
Thanks to these properties,
the opacity $\opacity$ can model opaque surfaces,
represent partially occluded pixels and
it can also approximate intricate fine geometry such as fur,
see for example \Fig{\ref{fig:lion}}.

\paragraph{Volume rendering comparison}
We compared opacity compositing
with the exponential transmittance model of \NeRF{}
in \Fig{\ref{fig:volumeRenderingAlgorithms}}.
Besides the fact that
opacity compositing is simpler and cheaper to evaluate,
it consistently helped the optimizer reconstruct actually opaque surfaces
instead of blurry and semi-transparent results
produced by the exponential transmittance model.
However,
the potentially more opaque opacity field $\opacity$ is harder to optimize
in case of false occlusions and thus missing gradients for occluded surfaces.
To alleviate the lack of gradients leading to a correct reconstruction (besides efficiency reasons),
we devised a custom scene sampling strategy for optimizing our models,
which we describe in the next subsection.

\newcommand{\imageFactor}{0.2475}
\begin{figure*}%
    \centering%
        \begin{subfigure}[t]{\imageFactor\textwidth}
            \centering%
            \labelImage{\textwidth}{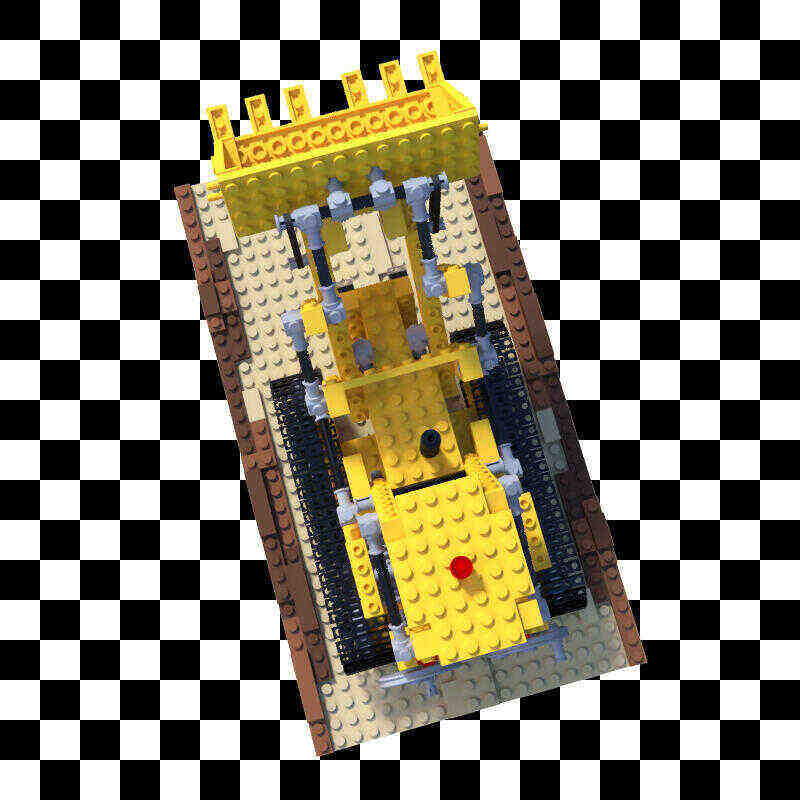}{\MipNeRF{}, 1m it.}
            \label{fig:NeRF:testViewRendering}
            \imageGap{}%
        \end{subfigure}%
        \begin{subfigure}[t]{\imageFactor\textwidth}
            \centering%
            \labelImage{\textwidth}{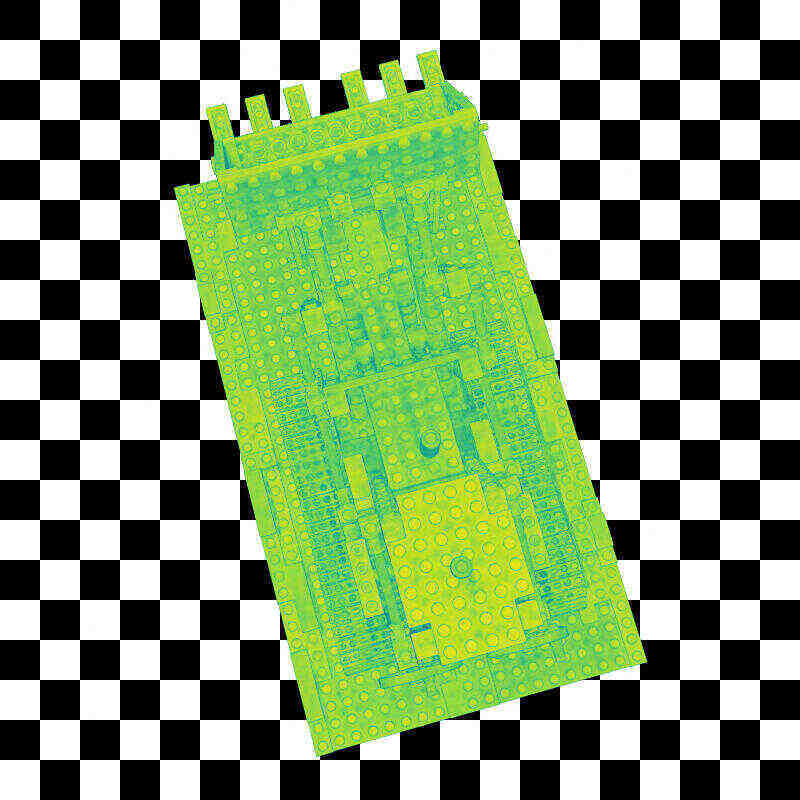}{\MipNeRF{}, 1m it.}
            \label{fig:NeRF:geometryVis}
            \imageGap{}%
        \end{subfigure}%
        \begin{subfigure}[t]{\imageFactor\textwidth}
            \centering%
            \normalMap{\textwidth}{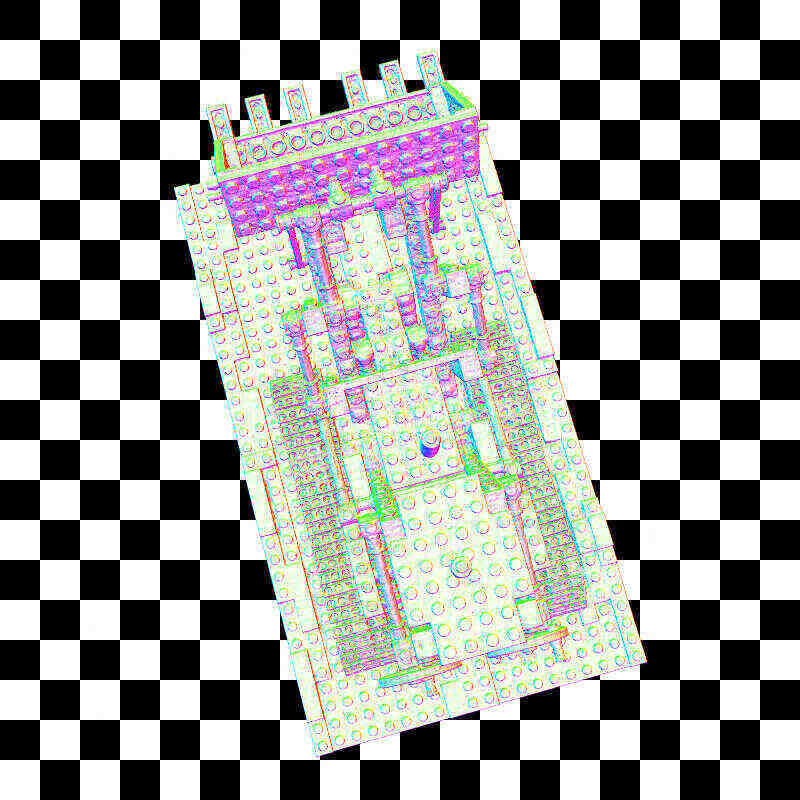}{\MipNeRF{}, 1m it.}
            \label{fig:NeRF:normals}
            \imageGap{}%
        \end{subfigure}%
        \begin{subfigure}[t]{\imageFactor\textwidth}
            \centering%
            \labelImage{\textwidth}{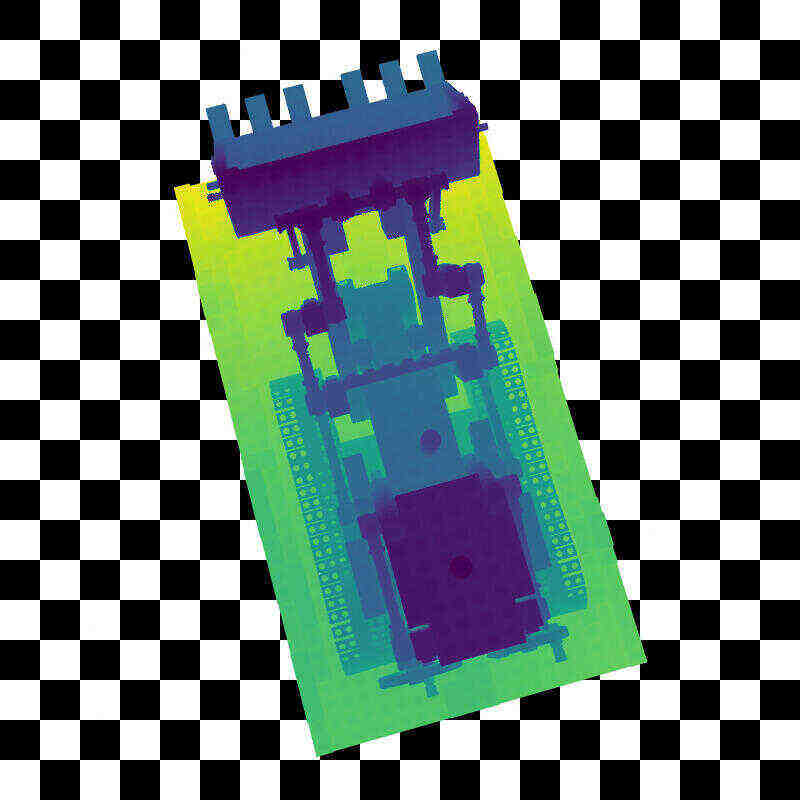}{\MipNeRF{}, 1m it.}
            \label{fig:NeRF:depth}
            \imageGap{}%
        \end{subfigure}%
        \\
        \begin{subfigure}[t]{\imageFactor\textwidth}
            \centering%
            \labelImage{\textwidth}{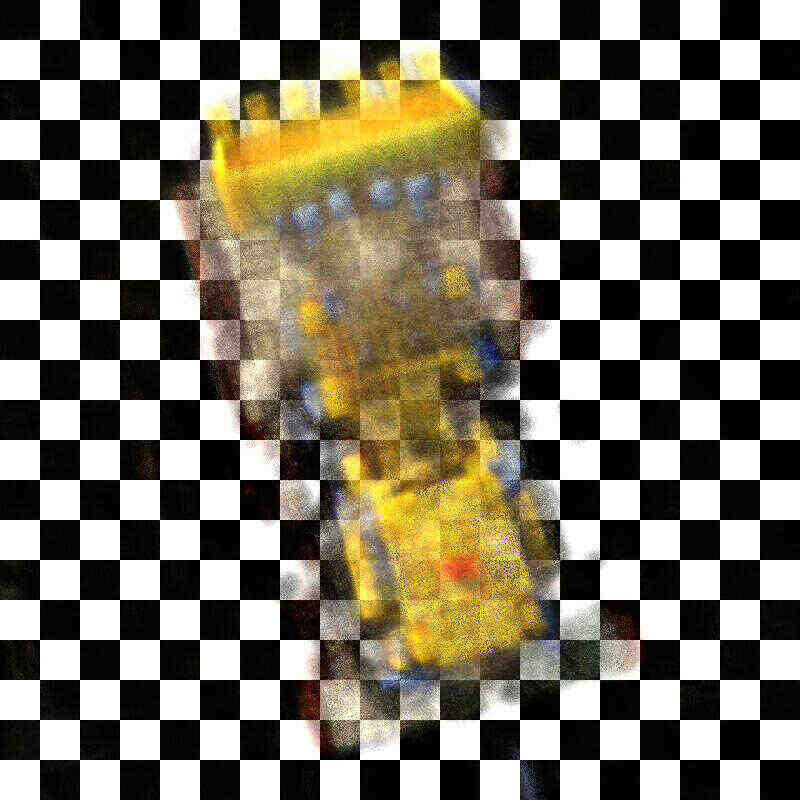}{ours (a), 70k it.}
            \label{fig:oursExpSoftPlus:testViewRendering}
            \imageGap{}%
        \end{subfigure}%
        \begin{subfigure}[t]{\imageFactor\textwidth}
            \centering%
            \labelImage{\textwidth}{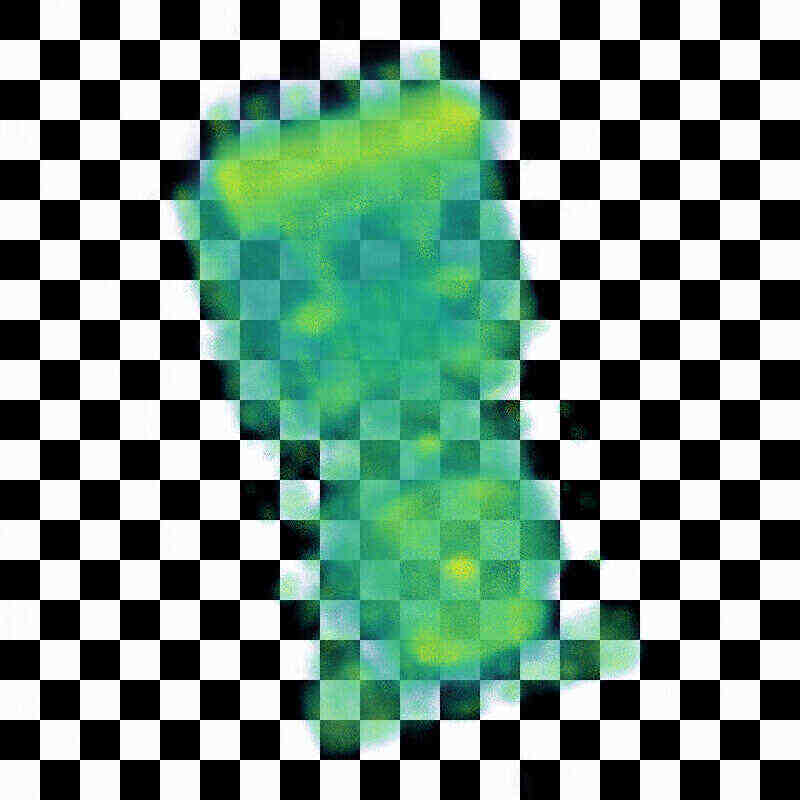}{ours (a), 70k it.}
            \label{fig:oursExpSoftPlus:geometryVis}
            \imageGap{}%
        \end{subfigure}%
        \begin{subfigure}[t]{\imageFactor\textwidth}
            \centering%
            \normalMap{\textwidth}{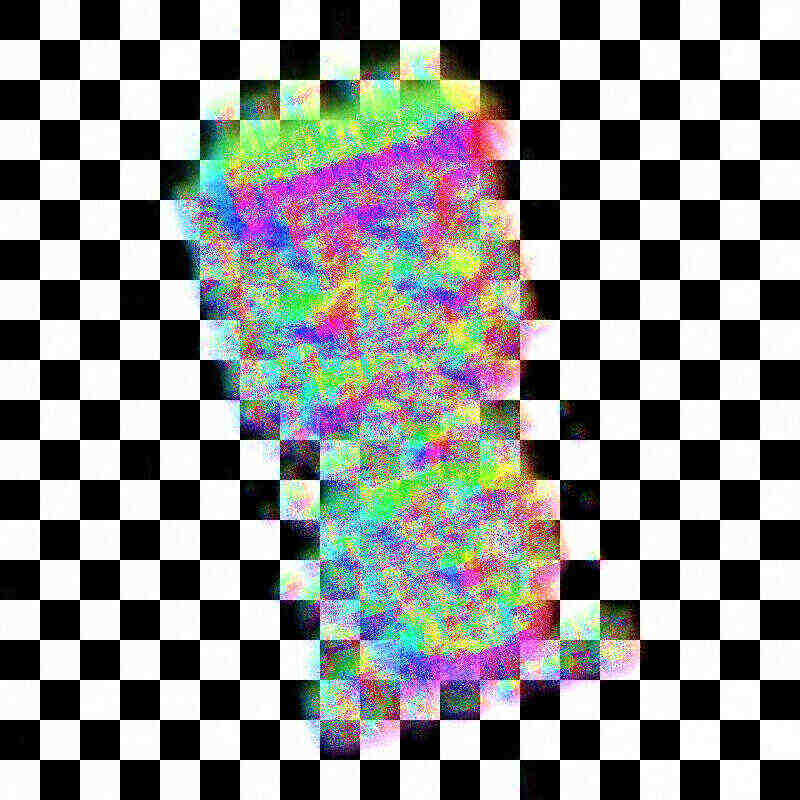}{ours (a), 70k it.}
            \label{fig:oursExpSoftPlus:normals}
            \imageGap{}%
        \end{subfigure}%
        \begin{subfigure}[t]{\imageFactor\textwidth}
            \centering%
            \labelImage{\textwidth}{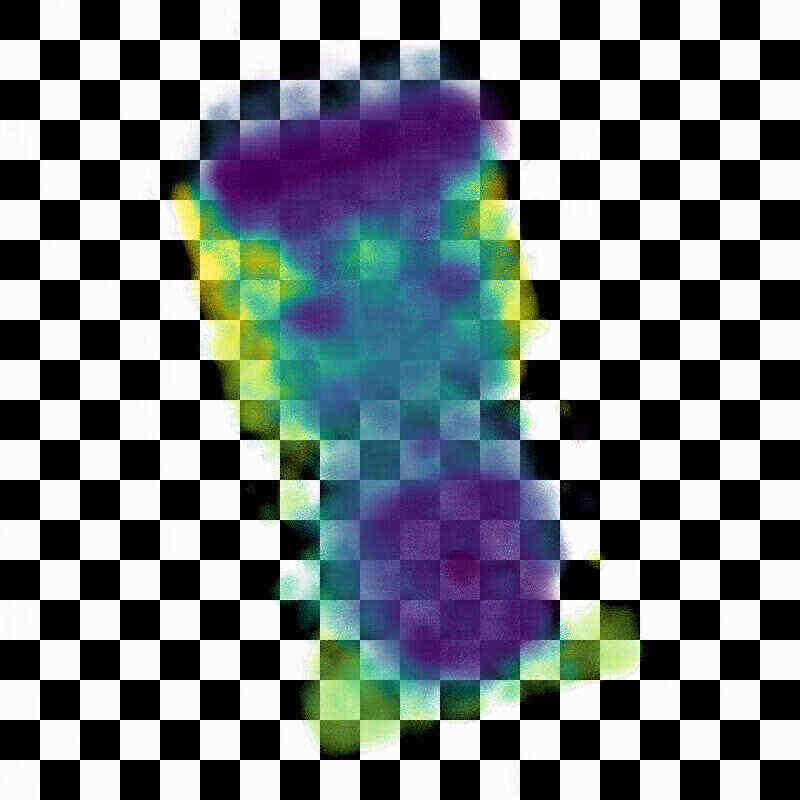}{ours (a), 70k it.}
            \label{fig:ours*:depth}
            \imageGap{}%
        \end{subfigure}%
        \\
        \begin{subfigure}[t]{\imageFactor\textwidth}
            \centering%
            \labelImage{\textwidth}{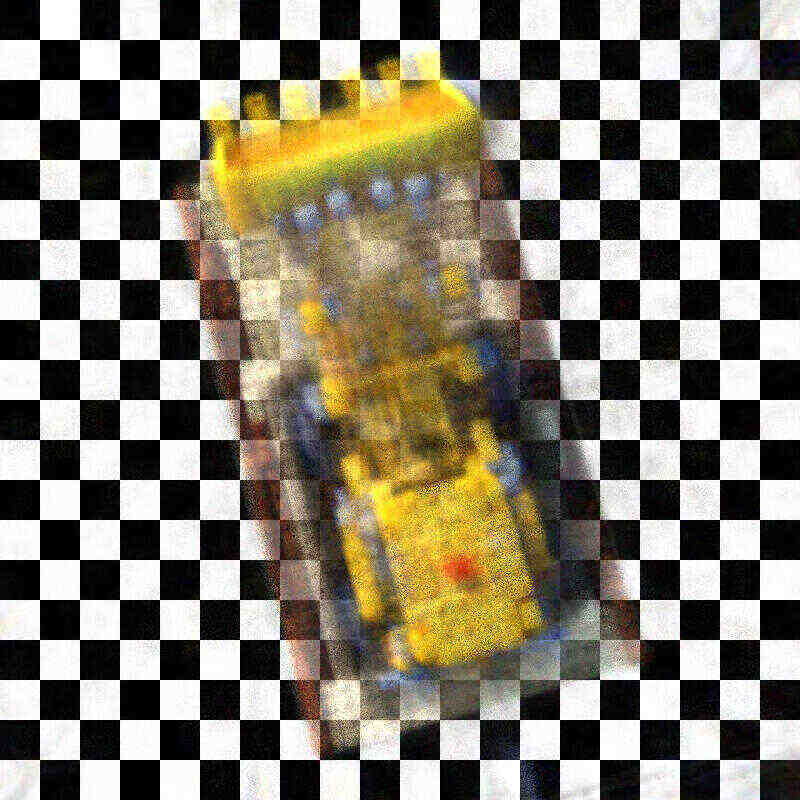}{ours (b), 70k it.}
            \label{fig:oursExpLiLU:testViewRendering}
            \imageGap{}%
        \end{subfigure}%
        \begin{subfigure}[t]{\imageFactor\textwidth}
            \centering%
            \labelImage{\textwidth}{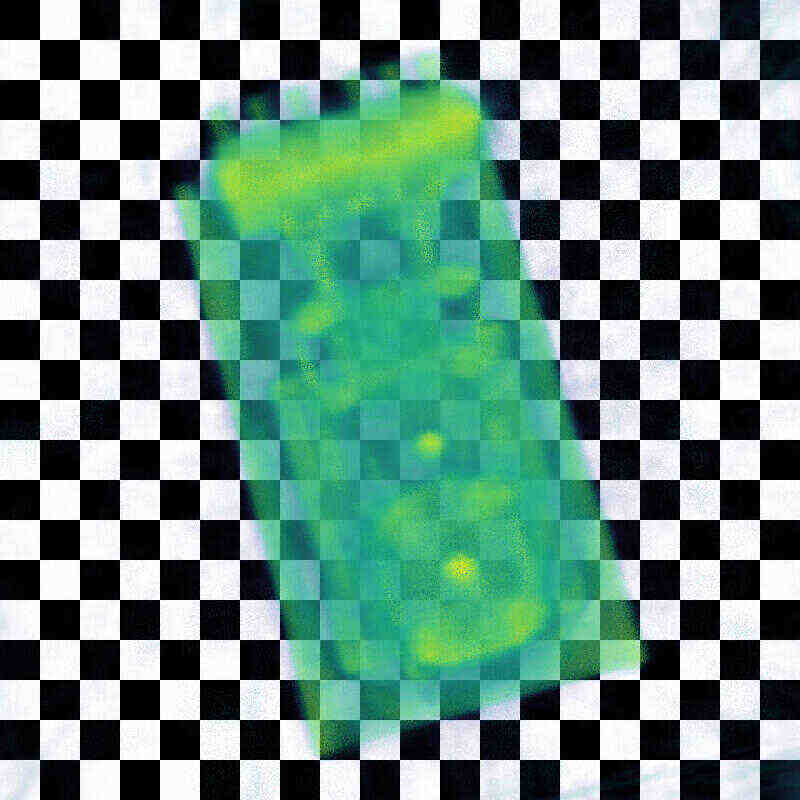}{ours (b), 70k it.}
            \label{fig:oursExpLiLU:geometryVis}
            \imageGap{}%
        \end{subfigure}%
        \begin{subfigure}[t]{\imageFactor\textwidth}
            \centering%
            \normalMap{\textwidth}{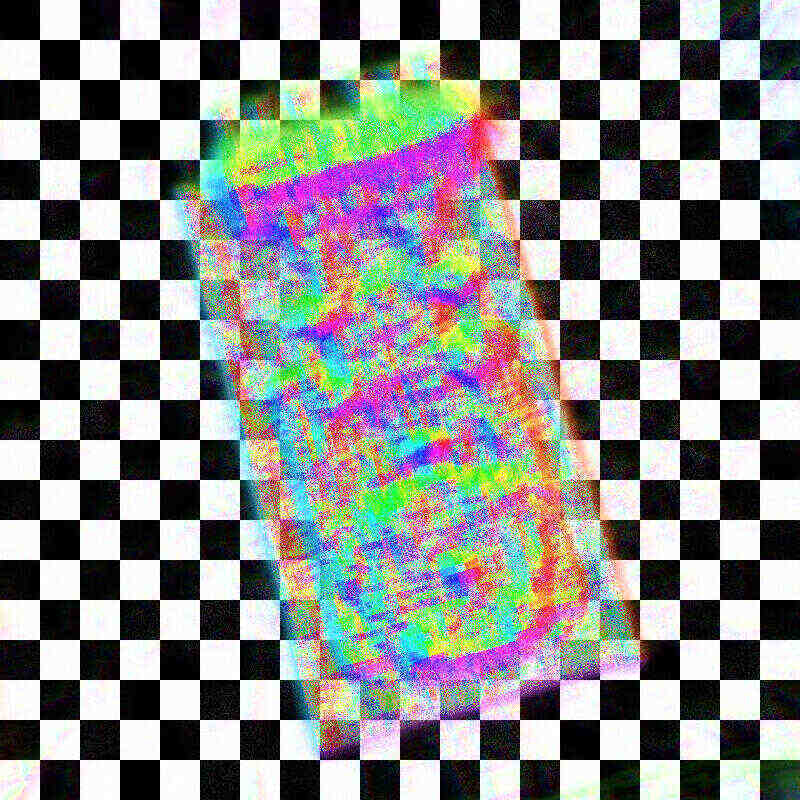}{ours (b), 70k it.}
            \label{fig:oursExpLiLU:normals}
            \imageGap{}%
        \end{subfigure}%
        \begin{subfigure}[t]{\imageFactor\textwidth}
            \centering%
            \labelImage{\textwidth}{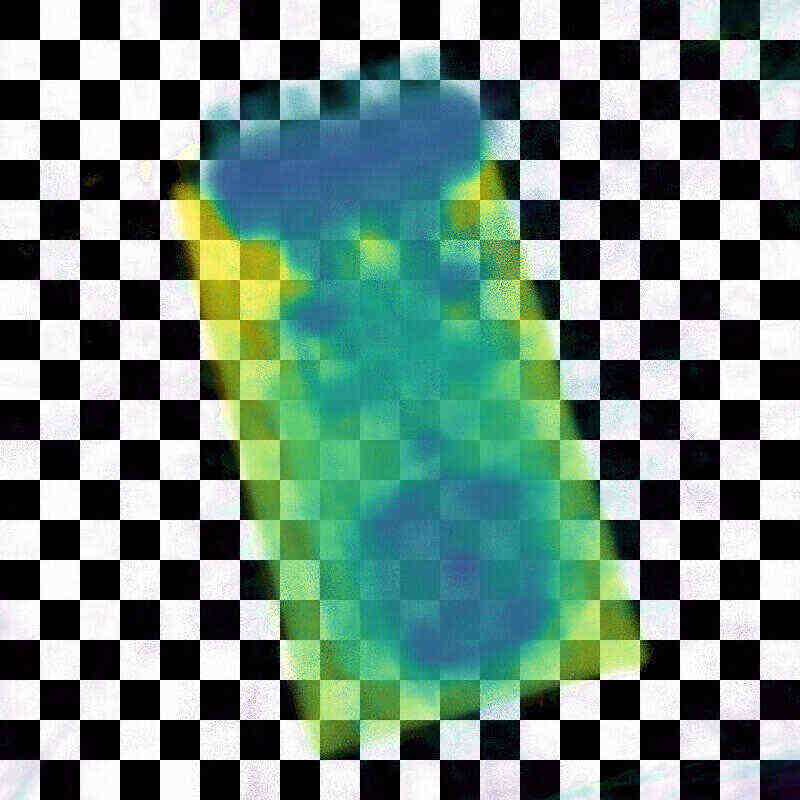}{ours (b), 70k it.}
            \label{fig:ours**:depth}
            \imageGap{}%
        \end{subfigure}%
        \\
        \begin{subfigure}[t]{\imageFactor\textwidth}
            \centering%
            \labelImage{\textwidth}{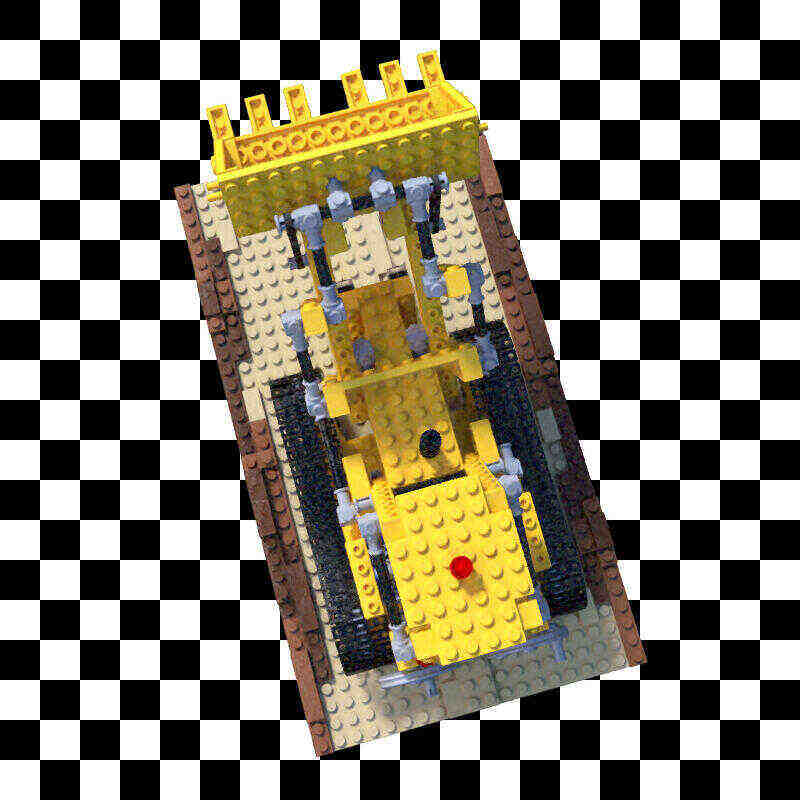}{ours (c), 115k it.}
            \label{fig:ours:testViewRendering}
            \imageGap{}%
        \end{subfigure}%
        \begin{subfigure}[t]{\imageFactor\textwidth}
            \centering%
            \labelImage{\textwidth}{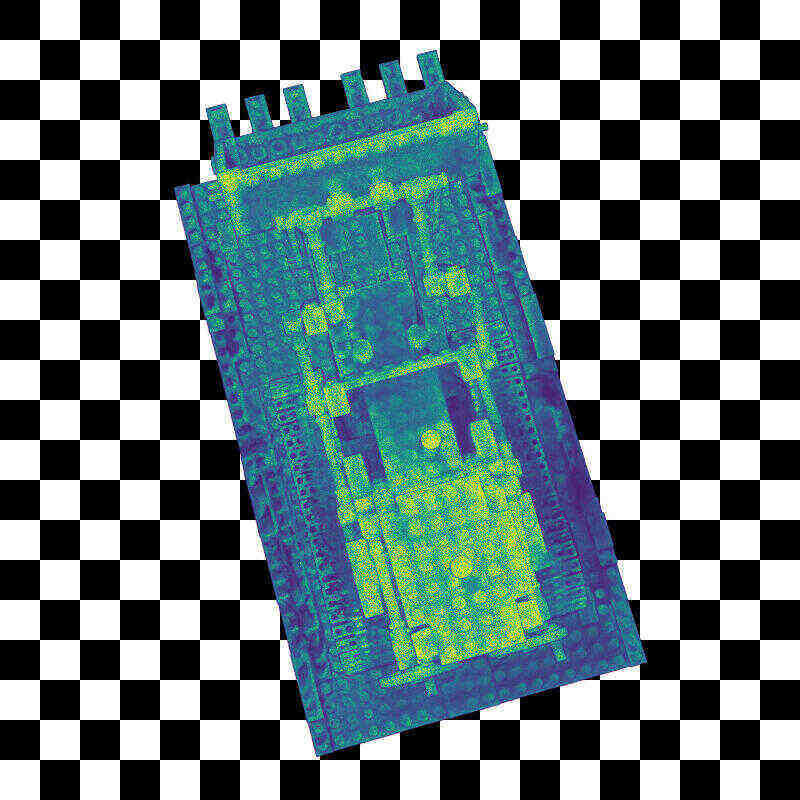}{ours (c), 115k it.}
            \label{fig:ours:geometryVis}
            \imageGap{}%
        \end{subfigure}%
        \begin{subfigure}[t]{\imageFactor\textwidth}
            \centering%
            \normalMap{\textwidth}{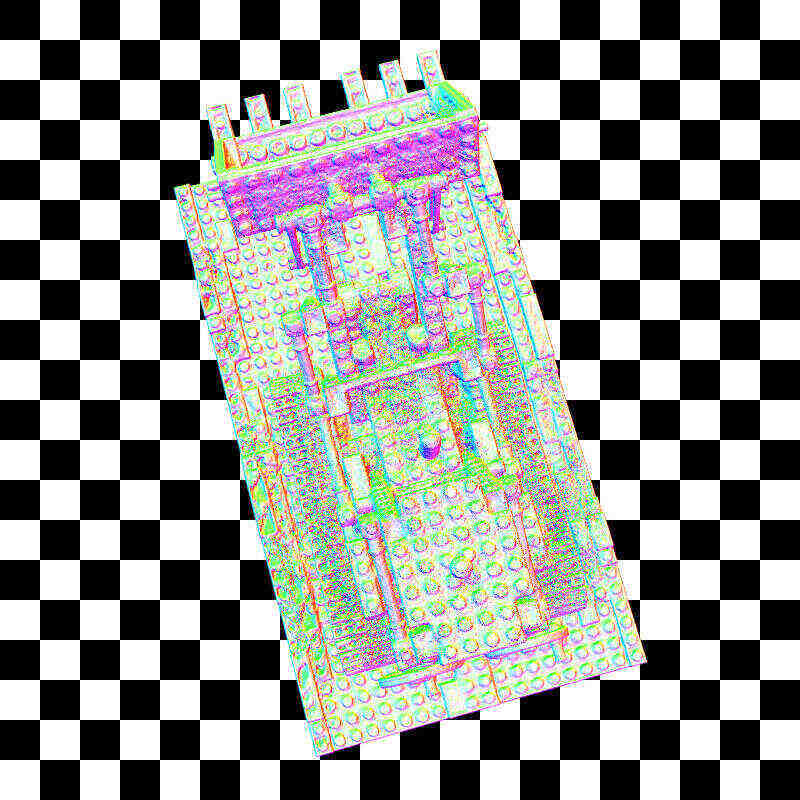}{ours (c), 115k it.}
            \imageGap{}%
        \end{subfigure}%
        \begin{subfigure}[t]{\imageFactor\textwidth}
            \centering%
            \labelImage{\textwidth}{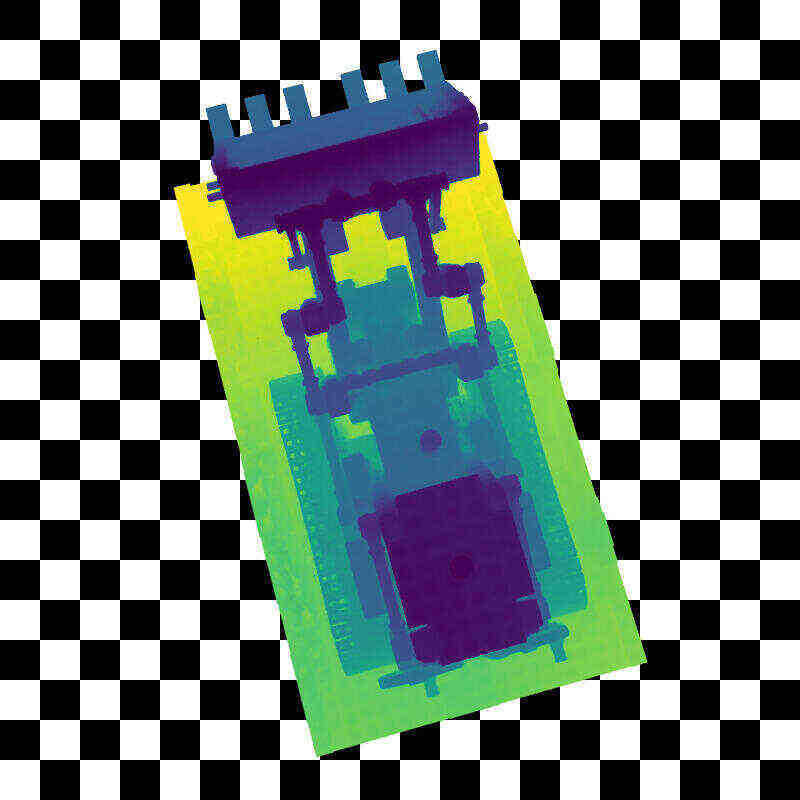}{ours (c), 115k it.}
            \label{fig:ours:depth}
            \imageGap{}%
        \end{subfigure}%
    \caption{Volume rendering comparison using the bulldozer scene \cite{mildenhall2020NeRF}.
        The image rows show the bulldozer scene from \cite{mildenhall2020NeRF} reconstructed with
        \MipNeRF{} after 1 million optimization iterations
        (\Eq{\ref{eq:NeRFVolumeRendering}});
        ours (a): with the same exponential transmittance and \Softplus{} activation function
        (\Eq{\ref{eq:NeRFVolumeRendering}});
        ours (b): with the exponential transmittance and \LiLU{} activation function
        (\Eq{\ref{eq:NeRFVolumeRendering}}, \Eq{\ref{eq:LiLU}});
        Ours (a) and (b) did not converge with density fields accurate enough for \SVO{} node subdivisions,
        even after 70k optimization iterations.
        Ours (c): with traditional opacity compositing after 85k iterations.
        (\Eq{\ref{eq:opacityCompositing}}).
    }
    \label{fig:volumeRenderingAlgorithms}
\end{figure*}


\subsection{Scene sampling}
\label{subsec:sceneSampling}
Efficient and robust scene sampling is critical
for explicit high resolution scene reconstruction
via inverse differentiable rendering.
It is especially important for explicit models
that are less compact than implicit neural network-based alternatives.
The 3D locations of scene surfaces must be sufficiently sampled,
which is difficult when their locations are initially completely unknown.
For example,
there need to be enough samples along each ray
to not miss surfaces intersected by a ray, especially thin structures,
False intermediate free space
where surfaces still have to emerge during optimization
also needs to be sampled densely enough.
However at the same time,
the number of drawn scene sampling points must be kept low
to limit the costs of following rendering and gradient computations.
We tackle this challenging problem using the following scene sampling scheme.

\begin{figure}
    \centering%
    \includegraphics[width=\columnwidth]{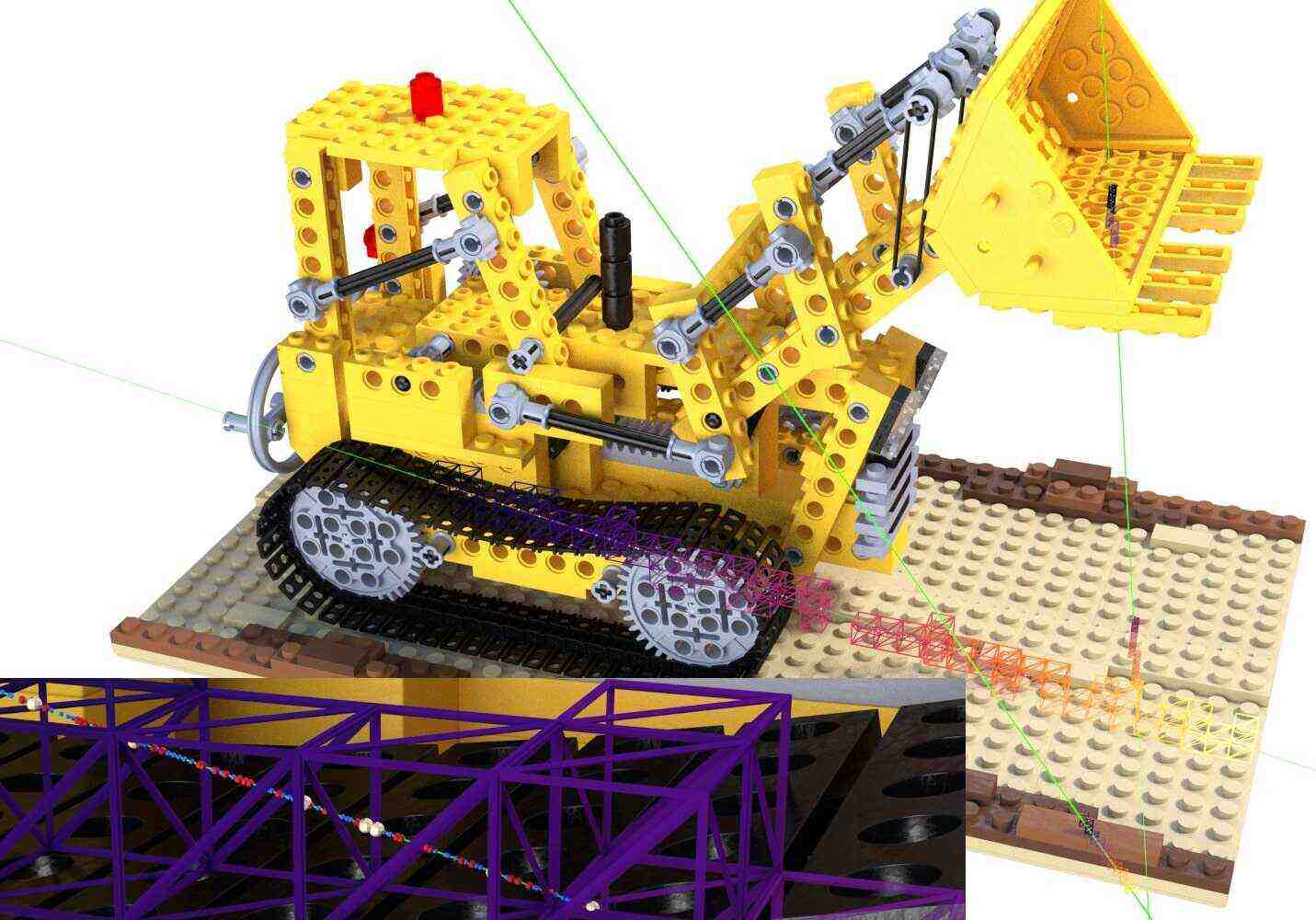}%
    \caption%
    {%
        \NeRF{} bulldozer scene rendering and sampling:
        the original \Blender{} model overlayed
        with relevant \SVO{} nodes for exemplar query rays and their sampling points.
        The renderer queries the scene \SVO{}
        for rays (green) through 3 example input pixels
        at different mipmap pyramid levels.
        Only the intersected \SVO{} nodes (wire frame)
        with side lengths fitting to the pixel back projections
        contain sampling points.
        For the ray hitting the treads,
        the close-up view shows
        the initial stratified sampling points $\raySamplePoints{i}{j}$ (blue),
        the subset $\raySamplePoints{i}{k}$ (red) from uniform random selection and
        finally the subset $\raySamplePoints{i}{l}$ (white) from importance sampling
        used for rendering.
    }%
    \label{fig:sceneSampling}%
\end{figure}

\paragraph{\LoD{}-aware stratified sampling of view rays}
\label{para:raySampling}
To tackle the challenging scene sampling requirements,
our renderer samples scene models via multiple steps,
which \Fig{\ref{fig:sceneSampling}} illustrates.
First,
the renderer draws uninformed samples
from a uniform distribution along each ray using stratified sampling,
see \Alg{\ref{alg:overview} line \ref{alg:stratifiedSampling}}.
Second,
it filters samples randomly and only keeps a subset,
see line \ref{alg:stochasticLimitingUninformed}.
We exploit the fact that \Adam{} keeps track of gradient histories.
This allows for deferring dense ray sampling over multiple optimization iterations
instead of densely sampling each ray within each single iteration.
Third,
the renderer filters the ray sampling points again after
it queried the \SVO{} for opacity samples
to  keep samples which are likely close to the true scene surfaces,
see line \ref{alg:stochasticLimitingInformed}.

%
%

\paragraph{1. Uninformed, stratified sampling}
For each ray $\rayIndexed{i}$,
the renderer creates sampling points $\raySamplePoints{i}{j}$ via stratified sampling,
see \Alg{\ref{alg:overview}} line \ref{alg:stratifiedSampling};
it randomly distributes these sampling points
within each \SVO{} node intersected by $\rayIndexed{i}$
while ensuring a sampling density
that depends on the side length of the intersected node
as detailed next.
To account for
for the projective nature of capturing devices
and the spatially varying \LoD{} within an \SVO{},
while marching along a ray $\rayIndexed{i}$,
the renderer goes down the \SVO{} to nodes of depth $\depthNode$
with side length $\extendNode$
that fit to the \SVO{} sampling rate $\footprintBackProj{}$ at the current ray sampling depth $\rayT$.
In this case,
the \SVO{} sampling rate $\footprint(\rayT)$ is the back projection of the diameter of the corresponding pixel.
The ideal tree depth $\depthNode$ for finding the node to be sampled
which has the highest \LoD{} and is still free of aliasing
is then inferred using the Nyquist sampling theorem
in the same manner as for general \SVO{} field queries via \Eq{\ref{eq:samplingTheorem}}.
If there is no tree node allocated at this depth,
the corresponding ray depth interval is treated as free space.
Though in the special case,
if the \SVO{} is still built up and
if the traversal reached the global maximum depth of the \SVO{},
a coarser higher level node is returned for sampling and not treated as free space.
Meaning we handle the ray query with coarse nodes if not possible otherwise and
with more detailed nodes at later optimization iterations.
The renderer samples each node in the intersected set
according to the node sizes $\left\{\footprintNodeIndexed{m}\right\}$
which may vary and increase with depth.
In particular,
for a node $\nodeIndexed{m}$ to be sampled,
the renderer uniformly draws a constant number of $N$ samples per side length $\extendNodeIndexed{m}$
within the intersection interval of $\rayIndexed{i}$ with $\nodeIndexed{m}$.
The constant relative sampling density $s(\nodeIndexed{m}) = N / \extendNodeIndexed{m}$
results in a varying spatial sampling density
that decreases with depth similar to inverse depth sampling.
It adapts to
the back projected pixel diameter $\footprintBackProj{}$ and
according to the spatially varying available \SVO{} \LoD{}.
The renderer filters the resulting sampling points $\raySamplePoints{i}{j}$ next.

\paragraph{2. Stochastic filtering of ray sampling points}
The renderer has a maximum budget for the number of sample points per ray
which it enforces via deferred stochastic filtering of the sampling points $\raySamplePoints{i}{j}$
from stratified sampling,
see \Alg{\ref{alg:overview}} line \ref{alg:stochasticLimitingUninformed}.
Varying sample counts per ray are more difficult to process in parallel and
storing their shading and gradients data in limited GPU memory is also more challenging
compared to a capped budget for scene sampling points.
Skipping sampling points randomly during optimization induces noise on the loss gradients
which \SGD{} is per design robust against however.
Also note that overall convergence is higher
since intermediate false occluders are skipped randomly.
They would otherwise potentially consistently occlude sampling points at the true surface locations
and result in missing gradients required for correct reconstruction.
Hence,
the renderer randomly (uniformly) selects a subset of sampling points
for each ray $\rayIndexed{i}$ with a sample count
exceeding a given budget $\maxSamplesPerRayUninformed$
and produces a limited set of ray sampling points
$\raySamplePoints{i}{k} \subset \raySamplePoints{i}{j}$.

\paragraph{3. Stochastic filtering of ray sampling points using opacity}
The renderer then filters the already limited sampling points $\raySamplePoints{i}{k}$ again
according to the current scene geometry estimate.
After querying the \SVO{} opacity via our 4D interpolation scheme of \Eq{\ref{eq:interpolation}}
with the limited ray sampling points $\raySamplePoints{i}{k}$,
see \Alg{\ref{alg:overview}} line \ref{alg:getOpacity},
the renderer reduces the number of samples per ray
using importance sampling to
$\maxSamplesPerRayInformed < \maxSamplesPerRayUninformed$.
It prefers samples
$\raySamplePoints{i}{l} \subset \raySamplePoints{i}{k}$
which are probably close to the true surfaces.
In particular,
it assigns a sampling weight
\begin{equation}
    w_{s, r}(\rayTIndexedij{i}{k}) = c + \opacity(\rayTIndexedij{i}{k})
    \label{eq:importanceSamplingRaySamplePoints}
\end{equation}
to each sampling point.
The user-defined constant $c = 0.05$ ensures
that the whole ray is sampled at least infrequently to
handle intermediate false free space regions.
Finally,
the renderer uses
the two times reduced samples $\raySamplePoints{i}{l}$
to retrieve outgoing \SLF{} radiance samples from the \SVO{},
see \Alg{\ref{alg:overview}} line \ref{alg:getSLF}.
We set the number of samples per node edge length to $N = 8$
and limit the per-ray sample sets to
$\maxSamplesPerRayUninformed = 256$ and $\maxSamplesPerRayInformed = 32$
during optimization,
see the supplemental for different sample budgets.
%
Note that this stochastic limiting is mainly required
to limit the costs of the gradients computations during optimization.
It is optional for only rendering once the \SVO{} is built;
when no loss gradients are computed and
when the \SVO{} nodes tightly bound the observed surfaces
and thus greatly limit the ray sampling intervals.

\paragraph{Sampling opacity instead of density fields}
Sampling along rays is simpler for opacity than for density fields.
The advantage of opacity fields is
they directly model the relative reduction of radiance.
Each opacity sample is independent of the distance to its neighboring sampling points.
This is similar to rendering fragments of layered mesh representations
\cite%
{%
    shade1998LayeredDepthImages,%
    flynn2019DeepView,%
    broxton2020ImmersiveLFVideo,%
    wizadwongsa2021NeX%
}.
As opposed to this,
sampling and optimizing density fields not only requires
estimating correct  extinction coefficients.
But also finding the right step sizes between samples is critical.
Nevertheless,
our opacity fields can theoretically be converted to equivalent density fields.

\section{Optimization}
\label{sec:optimization}

Our method iteratively reconstructs a scene model from scratch using \SGD{} and importance pixel sampling
in a coarse to fine manner
by comparing the given input images against differentiable renderings of that model from the same view
and updating it according to the resulting model loss and gradients.
Besides photo consistency loss from comparing renderings against the input images,
we employ light priors to improve convergence, see \Eq{\ref{eq:lossModel}}.
Our method starts reconstruction with a dense but coarse \SVO{} grid
that it attaches uninformed 3D fields to,
see \Fig{\ref{fig:optimizationProgress:initialization}}.
It then mainly optimizes these fields with the \SVO{} structure being fixed,
see \Fig{\ref{fig:optimizationProgress:coarseResult}}.
Further,
it infrequently updates the \SVO{} structure
to exploit free space and adapt the resolution given the current fields
and then restarts field optimization for a more detailed result.
see \Fig{\ref{fig:optimizationProgress:detailedResult}}.
Note that representing geometry using opacity fields is a soft relaxation
similar to layered mesh representations \cite{shade1998LayeredDepthImages}.
Both employ differentiable opacity parameters
defining local coverage and radiance reduction directly.
But otherwise,
our opacity fields are continuously defined over 3D space and
also provide fully differentiable surface locations like density fields \cite{mildenhall2020NeRF}.

\subsection{Objective function}
During \SGD{}, see \Alg{\ref{alg:overview}} line \ref{alg:loss},
we compute the model loss for small batches of image pixels and \SVO{} nodes.
The objective function of our optimization problem 
contains multiple priors besides a photo consistency term
to avoid convergence at solutions
that exhibit low photo consistency error,
but also physically implausible surface geometry.
Ambiguous reconstruction cases can cause such solutions
if only photo consistency is optimized for.
E.g., scenes might not have been captured sufficiently or
there can be surfaces with little texture
which do not sufficiently constrain their underlying geometry.
To avoid these local minima,
we suggest
\SVO{} priors defined on tree nodes.
They prefer smooth and physically meaningful results.
The priors also prevent that parameters derange
if they lack correct gradients intermediately or consistently,
\cf Gaussian prior on network parameters.

In particular,
for a random batch of pixels $\pixelBatch{i}$ and 
random bath of \SVO{} nodes $\nodeBatch{j}$,
we evaluate the objective function:
\begin{equation}
    \begin{split}
        \lossModel(\pixelBatch{i}, \nodeBatch{j}) =
            \frac{1}{|\pixelBatch{i}|}
            \sum_{i}
            \left[
                \lossPixel(\pixelCoordsVecIndexed{i})
            \right]
            \\
            + \frac{1}{|\nodeBatch{j}|} \cdot
            \left[
                \lossLocalSmoothness(\nodeBatch{j})
                + \lossLoDSmoothness(\nodeBatch{j})
                + \lossLowFields(\nodeBatch{j})
            \right]
    \end{split}
    \label{eq:lossModel}
\end{equation}
whereas the individual loss terms are as follows.
The squared pixel photo consistency loss $\lossPixel$ 
compares pixel intensity differences per color channel:
$\lossPixel(\pixelCoordsVecIndexed{i}) =
    \sum_\channelSymbol \lossPixelIntensityChannel{i}{\channelSymbol}$.
The \SVO{} priors
$\lossLocalSmoothness$, $\lossLoDSmoothness$, $\lossLocalSmoothnessNormals$ and $\lossLowFields$
are losses preferring local smoothness in 3D space,
local smoothness along the \LoD{} dimension,
as well as zero opacity and radiance for sparse models without clutter.
    
The photo consistency and prior losses are both normalized
by their individual batch size for comparability.
We set the batch size to $\batchSizeSoftServe$ for both batch types in our experiments (pixels and nodes).
Note that we also employed background priors enforcing local smoothness and zero radiance.
They are analogous to the their \SVO{} counterparts and
thus we omit them here for brevity.

\paragraph{\SVO{} priors}
\label{subsec:priors}

The objective function contains the following priors:
\begin{align}
    \lossLocalSmoothness(\nodeBatch{j}) &= 
        \lossLocalSmoothnessLambda \cdot \sum_{\nodeIndexed{j}}
            \sum_{\nodeIndexed{k} \in \neighborsSix{\nodeIndexed{j}}}
                \lossHuber
                (
                    \field{\posNodeIndexed{j}, \depthNodeIndexed{j}} - \field{\posNodeIndexed{k}, \depthNodeIndexed{j}}
                )
    \label{eq:priorSmoothNeighbors}
    \\
    \lossLoDSmoothness(\nodeBatch{j}) &=
        \lossLoDSmoothnessLambda  \cdot 
            \sum_{\nodeIndexed{j}}
            {
                \lossHuber
                (
                    \field{\posNodeRandomIndexed{j}, \depthNodeIndexed{j}} 
                    - \field{\posNodeRandomIndexed{j}, \depthNodeIndexed{j} + 1}
                )
            }
    \label{eq:priorSmoothLoD}
    \\
    \lossLowFields(\nodeBatch{j}) &=
        \lossLowFieldsLambda \cdot 
            \sum_{\nodeIndexed{j}}
            {
                \lossHuber(\field{\posNodeRandomIndexed{j}, \depthNodeIndexed{j}})
            }
    \label{eq:priorTargetValue}
\end{align}
which regularize the \SVO{} node parameters.
\Eq{\ref{eq:priorSmoothNeighbors}} prefers local smoothness.
\Eq{\ref{eq:priorSmoothLoD}} enforces smoothness between tree levels.
\Eq{\ref{eq:priorTargetValue}} punishes deranging parameters
by preferring zero density and radiance.
Note that
we applied the \SVO{} priors to
the opacity $\opacitySymbol$ and
the outgoing radiance field $\radianceOutgoingSymbol$.
We also analogously applied local smoothness and low radiance priors
to the background cube map texels
which we omit here for brevity.
Hereby, 
$\neighborsSix{\nodeIndexed{j}}$ are the six axis aligned neighbors
of node $\nodeIndexed{j}$;
$\lossHuber$ is the smooth Huber loss function;
$\posNodeIndexed{j}$ is the center of node $\nodeIndexed{j}$;
$\posNodeRandomIndexed{j}$ is a random 3D position within the scope of node $j$
and $\depthNodeIndexed{j}$ is its depth within the tree.
We choose the nodes $\left\{\nodeIndexed{j}\right\}$
to which the priors are applied as detailed next.
Further,
we uniformly set the strength of all priors via $\lambda = 1e-3$
for all the experiments shown here
(supplemental material with varying $\lambda$ experiment).

\paragraph{Stochastic priors}
Our stochastic priors improve convergence
via normalized random batches.
For the \SVO{} priors,
we simply randomly choose \SVO{} nodes (and neighborhoods).
We directly apply the rendering priors
to the rays of the random input pixel batches
(that are already available for reducing photo consistency errors).
The priors hence similarly work to the data term
that is based on the random input pixel batches.
Applying the priors every iteration
to every ray or voxel 
would result in very consistent and hence overly strong priors.
Contrary,
our random prior batches facilitate convergence,
but are also treated as noisy outliers in cases
where they do not fit,
since the stochastic priors are tracked like the data term
by \Adam{}'s gradient histories.
E.g., applying local smoothness to an edge infrequently
results in wrong, but inconsistent gradients
\Adam{} is robust against. 
Additionally,
the costs of our stochastic priors scale with
the batch and not the model size
making them more suitable for complex models with many parameters.

\paragraph{Suitable optimizer.}
We investigated different solvers
for fitting our scene models against registered images of a scene.
Higher order solvers such as \LM{}, \LBFGS{} or \PCG{} were either
too expensive given the high number of unknowns of our scene models or
they failed to reconstruct scenes from scratch.
That means they converged in a bad local optimum due to
our optimization problems being highly non-convex 
and due to starting far from the globally optimal solution
while making simplifying assumptions to approximate the inverse Hessian
which are not applicable in our case.
So we eventually decided for a relatively cheap \SGD{}-based optimizer
which is still powerful enough for the targeted non-convex and high-dimensional objective functions.
I.e., we employed \Adam{} for all our experiments and
ran it with the recommended settings \cite{kingma2014Adam}.

%
%
\begin{figure}
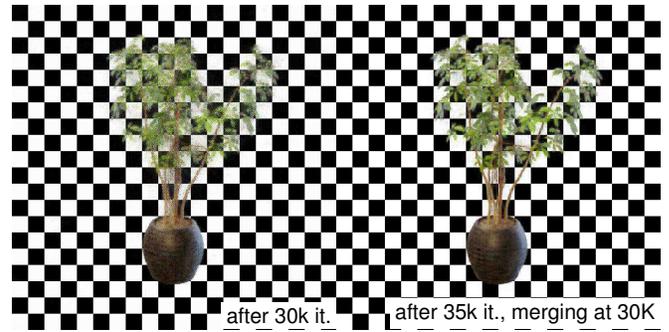

    \centering%
    \begin{subfigure}[t]{0.24\textwidth}
        \centering%
        \labelImage{\textwidth}{Method/Optimization/MergingNodes/MergingBefore}{after 30k it.}
        \label{fig:mergingBefore}
        \imageGap{}%
    \end{subfigure}%
    \begin{subfigure}[t]{0.24\textwidth}
        \centering%
        \labelImage{\textwidth}{Method/Optimization/MergingNodes/MergingAfter}{after 35k it., merging at 30K}
        \label{fig:mergingAfter}
    \end{subfigure}%
    \caption%
    {%
        Intermediate ficus scene \cite{mildenhall2020NeRF} models
        after optimizing the \SVO{} fields
        given the initial dense \SVO{} (left)
        and given the \SVO{} sparsified after 30k mini batch iterations (right).
    }%
    \label{fig:mergingSVONodes}%
\end{figure}

\subsection{Hierarchical optimization}
\label{subsec:hierarchicalOptimization}
Our method reconstructs scene models in a coarse to fine manner.
It starts with a dense, but coarse scene grid,
i.e.,
a full, shallow \SVO{}.
The tree then gradually becomes
sparser by merging of nodes or
more detailed thanks to new leaves.
The \SVO{} structure changes infrequently
depending on its attached fields
after optimizing them for a fixed \SVO{} structure
as shown by the outer loop of \Alg{\ref{alg:overview}}.

\paragraph{Merging \SVO{} nodes.}
We merge nodes in free space
to save memory and reduce rendering costs.
Merging nodes also considerably increases the quality
of the view ray sampling point distributions
as demonstrated by \Fig{\ref{fig:mergingSVONodes}}.
By means of the optimized opacity field $\opacity$,
we determine the set of nodes $\nodesRequired{r}$ required for rendering
using Dijkstra searches
that we run on each tree level's 27-neighborhood graph.
The Dijkstra search is hysteresis-based
similarly to the Canny edge detector \cite{canny1986CannyEdgeDetection}
for robustness.
First,
we sample each tree node using a regular pattern of $8^3$ sampling points
and determine its maximum opacity $\opacity_\text{max}(j)$.
Second,
we start a Dijkstra search on each unvisited node $j$ with $\opacity_\text{max}(j) \geq 0.75$
and only expand the search to nodes $\{k\}$ with $\opacity_\text{max}(k) \geq 0.075$.
Third,
a 27-neighborhood dilation augments
the set of all visited nodes
to ensure complete opacity function ramps from free to occupied space
The resulting set of $\nodesRequired{r}$
then defines nodes which must not be discarded:
We either keep all 8 children of an \SVO{} node
if any child is within $\nodesRequired{r}$ or discard all of them,
which is similar to the work \cite{laine2010EfficientSVOs}.

\paragraph{Subdividing \SVO{} nodes.}
To subdivide nodes,
we first find the set of required nodes $\nodesRequired{r}$ in the same manner as for merging.
However,
a leaf node within $\nodesRequired{r}$ is only eligible for subdivision
if the subdivision does not induces under sampling.
That means
there must be an input pixel with small enough 3D footprint
such that the Nyquist sampling theorem still holds
(similar to \Eq{\ref{eq:samplingTheorem}}).
The footprint $\footprintBackProj{}$ from back projecting the pixel
to the depth $\rayT$ of the node
defines the sampling rate and
the node edge length defines the signal rate.
If a leaf is within $\nodesRequired{r}$ and does not induce aliasing,
we allocate all of its 8 children.
The \SVO{} fills all new leaf nodes
with the 3D-interpolated data of their parents
resulting in a smooth initialization
with less block artifacts compared to copying parent values.
Our method afterwards optimizes
the new and smoothed, but also more detailed \SVO{}
using \SGD{} again as denoted in the inner loop of \Alg{\ref{alg:overview}}.
The refinement finally stops
if there are no new leaf nodes
according to the Nyquist sampling theorem.

\subsection{Input pixel sampling for \SGD{}}
To facilitate convergence,
our method employs Gaussian pyramids of the input images
and importance sampling
as described next.

For better optimization convergence and before the actual optimization,
we compute a complete Gaussian (mipmap) pyramid of each input image.
We then randomly select pixels from all input image pyramids
to simultaneously optimize the whole \SVO{}
at multiple levels of detail.
Coarser input pixels from higher mipmap levels have a larger footprint
and thus help optimize inner or higher level \SVO{} nodes as
in the sampling scheme of \Eq{\ref{eq:samplingTheorem}}.

\begin{figure}
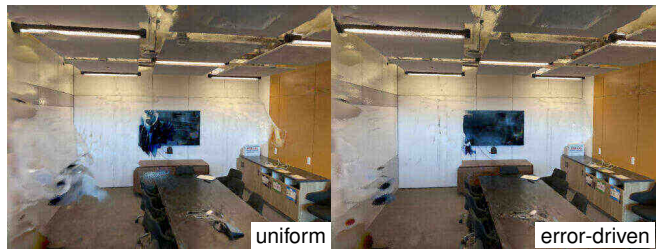
%
    \centering%
    \begin{subfigure}[t]{0.24\textwidth}
        \centering%
        \labelImage{\textwidth}{Results/AblationStudy/ImportanceSampling/RoomView0It175K/2021_09_29_17_04_43_f300366221}{uniform}
        \label{fig:withoutImportanceSampling}
        \imageGap{}%
    \end{subfigure}%
    \begin{subfigure}[t]{0.24\textwidth}
        \centering%
        \labelImage{\textwidth}{Results/AblationStudy/ImportanceSampling/RoomView0It175K/2021_10_12_08_59_04_f302526557}{error-driven}
        \label{fig:withImportanceSampling}
    \end{subfigure}%
    \caption%
    {%
        Pixel sampling comparison
        using the \NeRF{} room scene \cite{mildenhall2020NeRF}
        for a test view after 175k iterations.
    }%
    \label{fig:pixelImportanceSamplingComparison}
\end{figure}

\paragraph{Importance sampling}
For faster convergence where scene reconstruction has not finished yet,
we implemented importance sampling.
We prefer pixels with high photo consistency error
as depicted by
\Alg{\ref{alg:overview}} line \ref{alg:importanceSampling}
and demonstrated using
\Fig{\ref{fig:pixelImportanceSamplingComparison}}.
The sampling weights are basically
the per-pixel maxima of the color channel errors:
$w_{s, p}(\pixelCoordsVecIndexed{i}) =
    \max_c(\lossPixelIntensityChannel{i}{\channelSymbol}) + c$.
Similarly to the ray sampling of
\Eq{\ref{eq:importanceSamplingRaySamplePoints}},
we add a small constant $c = 0.05$
to also sample low error pixels infrequently
and prevent oscillating errors.
To implement this scheme,
a loss cache $\lossCache$ steers the importance sampler, see line \ref{alg:initLossCache}.
It caches running error averages of the input pixels
via the  photo consistency loss of the pixel batches $\lossBatch{i}$,
see line \ref{alg:cacheUpdate}.
The cache stores prefix sum offsets
of the sampling weights  $\left\{w_{s, p}(\pixelCoordsVecIndexed{i}) \right\}$
for fast random pixel selection
whereas it updates these offsets infrequently,
i.e., every $\lossCacheUpdateRate$ iterations.
Further,
we only store coarse error cache data,
meaning only at higher image mipmap pyramid levels for efficiency reasons
and to broaden the image sampling area of the importance sampler.
If a low mipmap level is selected
where multiple fine pixels fall into the same coarse pixel with a single running average error,
then the sampler uniformly selects between all fine pixels.

\section{Results}
\label{sec:results}

In the following,
we evaluate individual contributions of our approach
and show that our reconstructed explicit models are comparable
to state-of-the-art implicit alternatives.
Our models converge faster in general,
but their implicit competitors generally converge
at the best solutions after many more iterations.

\paragraph{Volume rendering algorithms.}
\label{par:volumeRenderingAlgorithms}
We compared
direct opacity compositing against the exponential transmittance formulation from \cite{mildenhall2020NeRF}.
When employing the exponential transmittance formulation (\Eq{\ref{eq:NeRFVolumeRendering}}),
our explicit model are less accurate compared to
their opacity compositing-based alternatives from (\Eq{\ref{eq:opacityCompositing}}).
The density field is smeared out and much blurrier than the opacity field.
The density fields are also less suited for our hierarchical refinement
as empty nodes are more difficult to distinguish from \SVO{} nodes in occupied space.
Interestingly,
the \MLP{} representation of \MipNeRF{} is not affected in the same way.
The \MipNeRF{} models of the synthetic scenes have low photo consistency error
and accurate underlying geometry despite the exponential transmittance model.
However,
the \MipNeRF{} models need many more mini batch optimization iterations.

\paragraph{Ablation study.}
%
%
First,
the comparison of \Fig{\ref{fig:gradientBasedInterpolation}}
demonstrates the importance of our quadratic 4D interpolation
when sampling scenes for scale-augmented 3D points.
Interpolating between local planes (field function samples plus local gradients)
allows the optimizer to select continuous instead of limited discrete 3D positions
for the extrema of the fields.
It hence better fits the scene model to the input images when voxels are only coarse.
This is especially beneficial for fine and intricate geometry like from the shown ficus.
%
%
Second,
the error-driven importance sampling improves convergence
as demonstrated by \Fig{\ref{fig:pixelImportanceSamplingComparison}}.

\paragraph{Qualitative evaluation}

\begin{figure*}%
    \centering%
    \begin{tabular}{ll}
        %
        %
        \raisebox{-0.5\height}[0pt][0pt]
        {
            \begin{subfigure}[t]{0.19\textwidth}
                \centering%
                \labelImage{\textwidth}{Results/QualitativeEvaluation/ChairView351/ChairView351L0Original}{gt}
                \label{fig:qualitativeEvaluation:chairOriginal}
                \imageGap{}%
            \end{subfigure}
        } &%
        \begin{subfigure}[t]{0.19\textwidth}
            \centering%
            \labelImage{\textwidth}{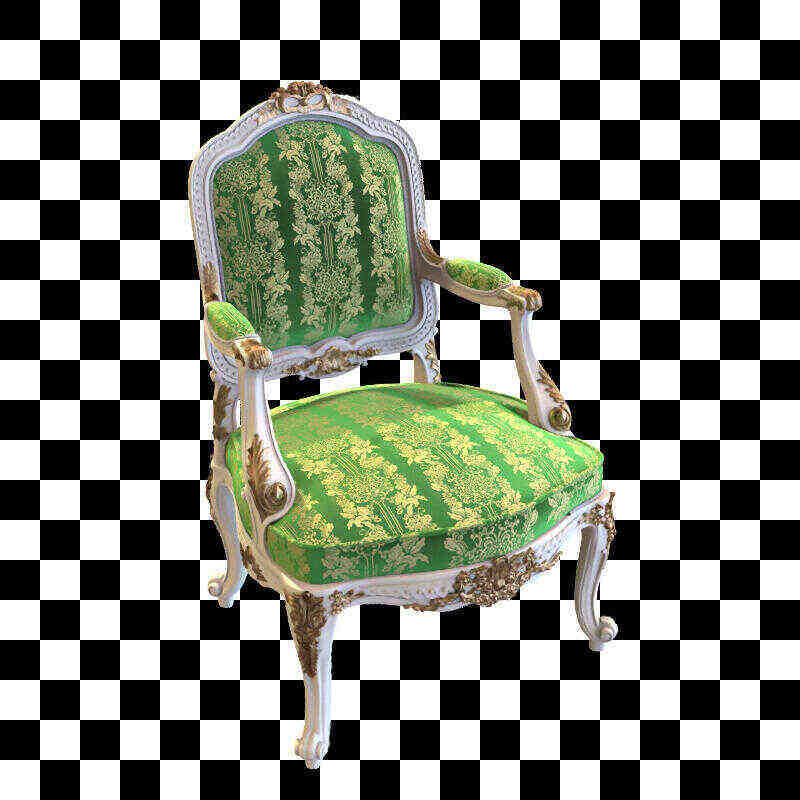}{\MipNeRF{}, 1m it.}
            \labelImage{\textwidth}{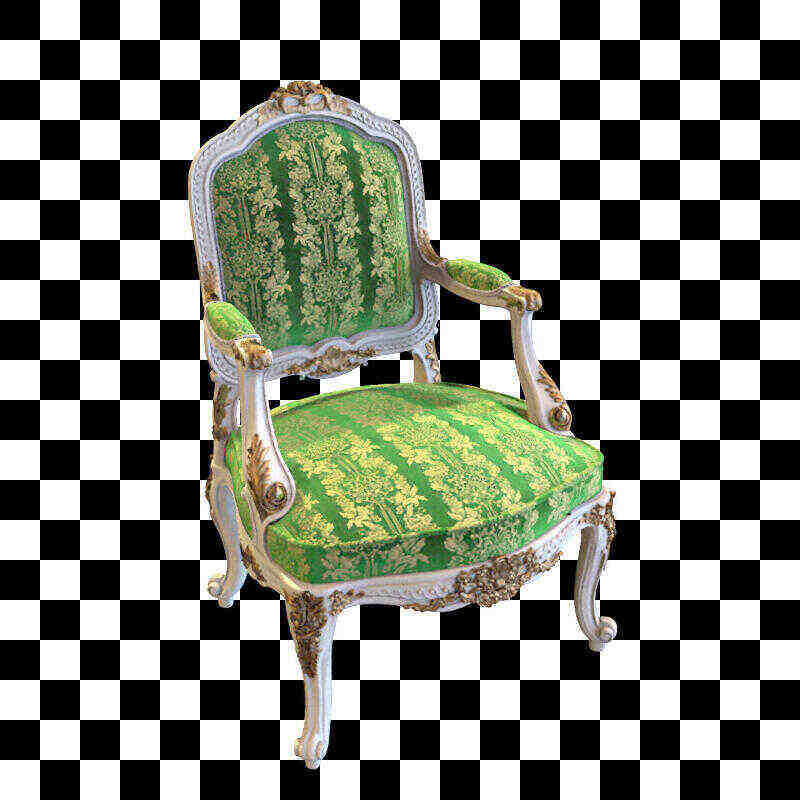}{\ours{}, 115k it.}
            \label{fig:qualitativeEvaluation:chairPhoto.jpg}
            \imageGap{}%
        \end{subfigure}%
        \begin{subfigure}[t]{0.19\textwidth}
            \centering%
            \labelImage{\textwidth}{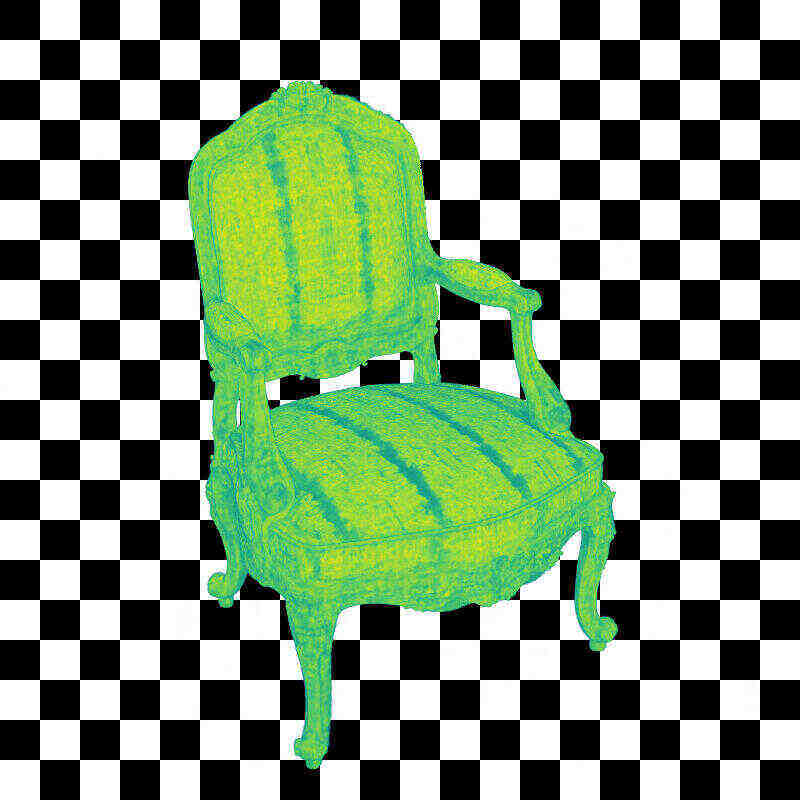}{\MipNeRF{}, 1m it.}
            \labelImage{\textwidth}{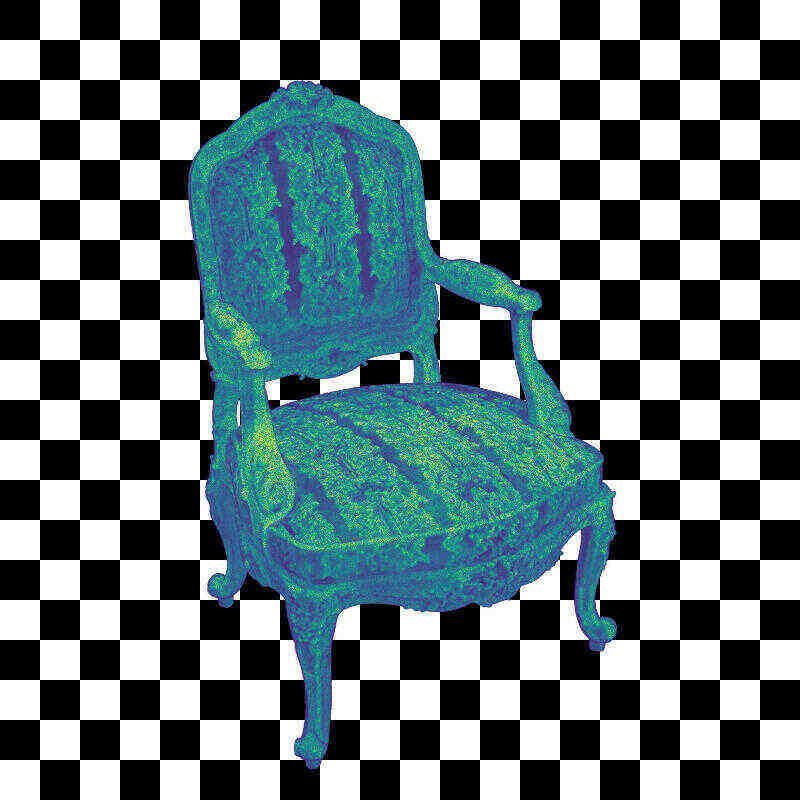}{\ours{}, 115k it.}%
            \label{fig:qualitativeEvaluation:chairDensity.jpg}
            \imageGap{}%
        \end{subfigure}%
        \begin{subfigure}[t]{0.19\textwidth}
            \centering%
            \normalMap{\textwidth}{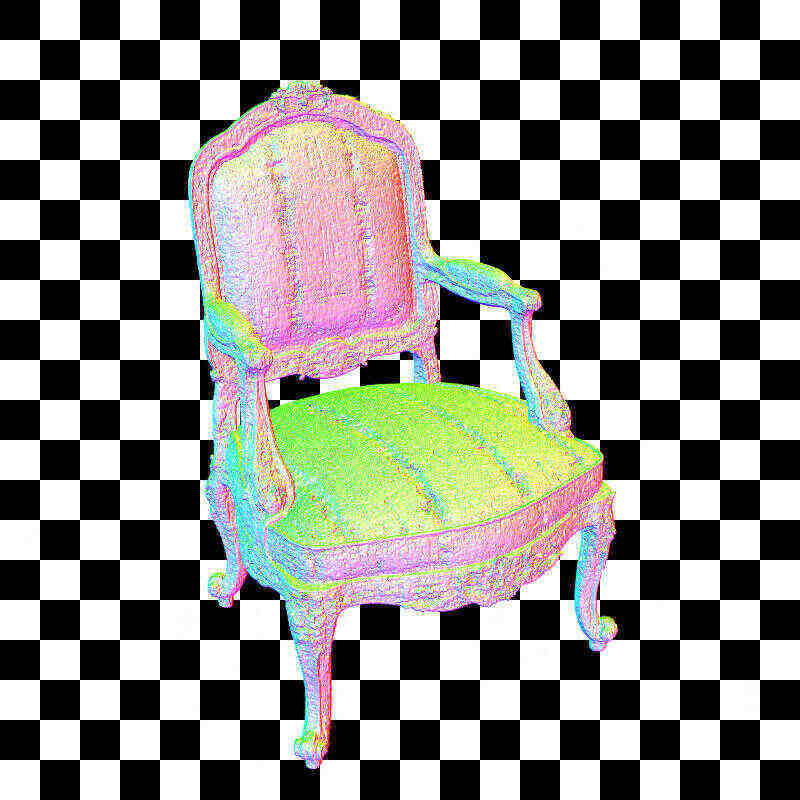}{\MipNeRF{}, 1m it.}
            \normalMap{\textwidth}{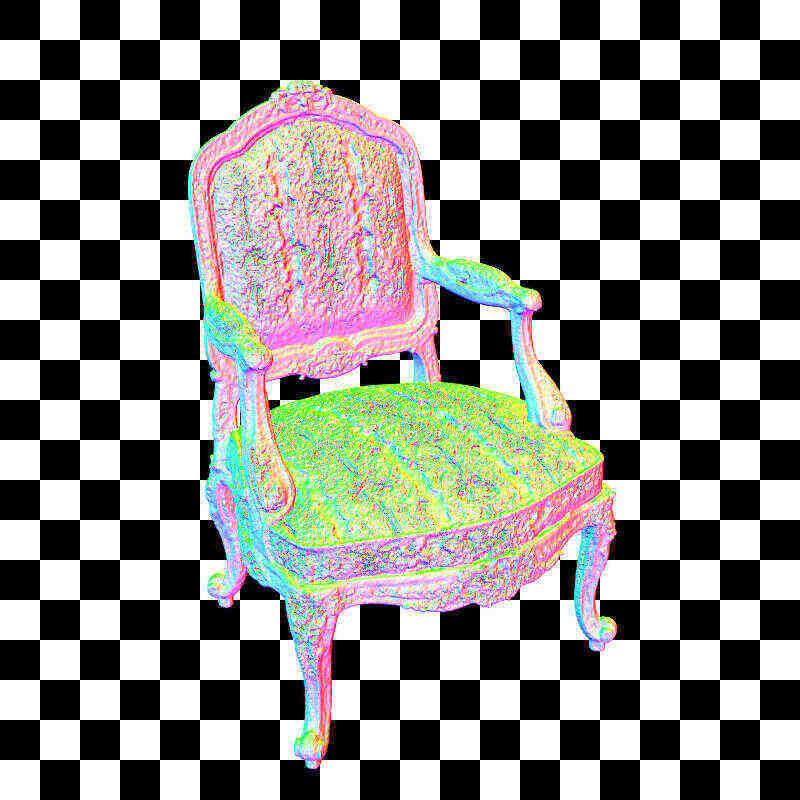}{\ours{}}%
            \label{fig:qualitativeEvaluation:chairNormals.jpg}
            \imageGap{}%
        \end{subfigure}%
        \begin{subfigure}[t]{0.19\textwidth}
            \centering%
            \labelImage{\textwidth}{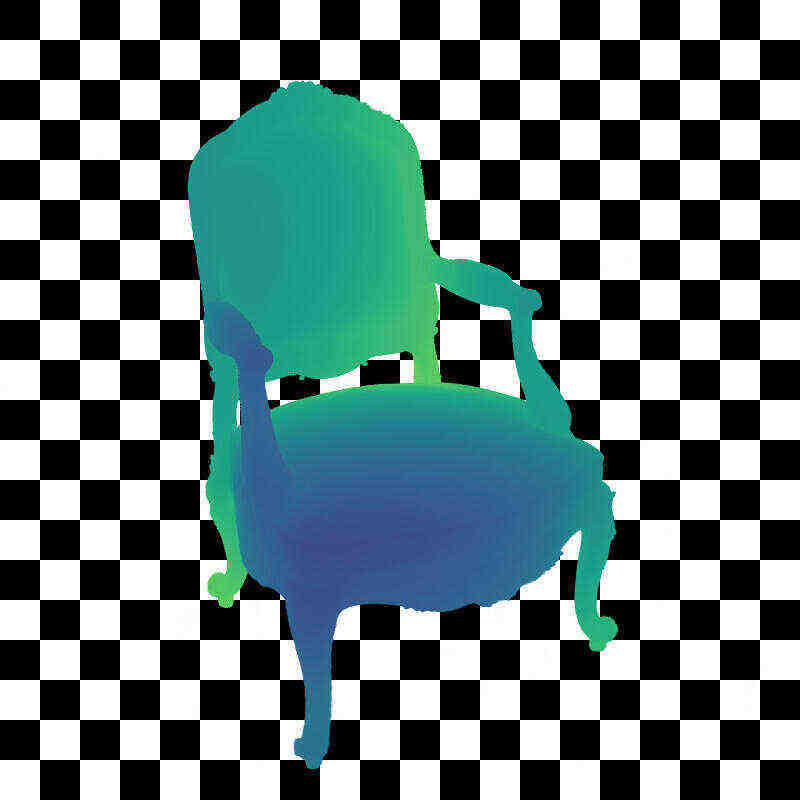}{\MipNeRF{}, 1m it.}
            \labelImage{\textwidth}{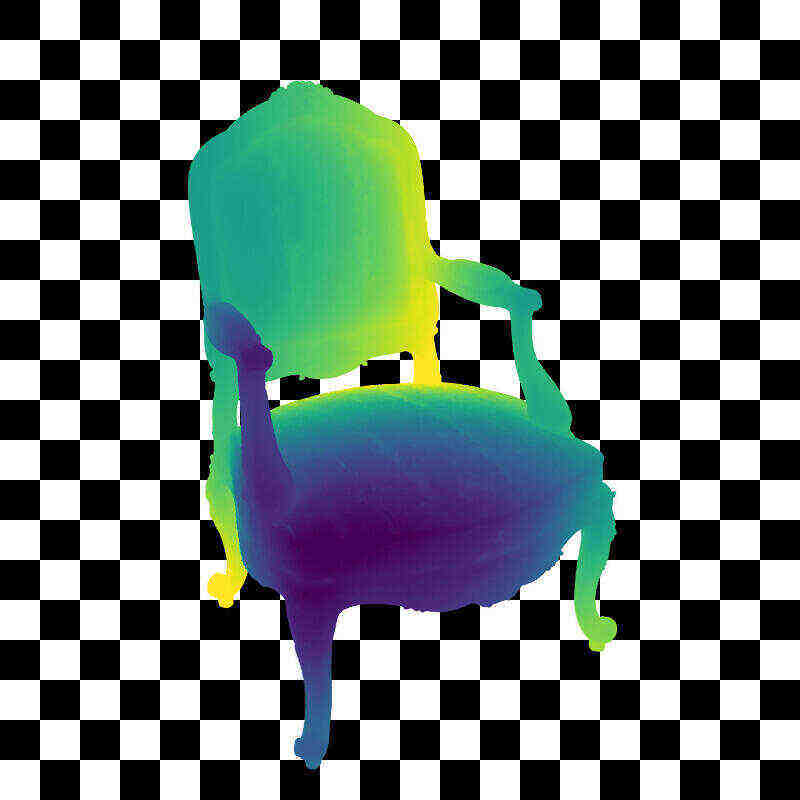}{\ours{}}
            \label{fig:qualitativeEvaluation:chairDepth.jpg}
            \imageGap{}%
        \end{subfigure} \\
        %
        %
        \raisebox{-0.5\height}[0pt][0pt]
        {
            \begin{subfigure}[t]{0.19\textwidth}
                \centering%
                \labelImage{\textwidth}{Results/QualitativeEvaluation/ShipView305/ShipView305L0Original}{gt}
                \label{fig:qualitativeEvaluation:shipOriginal}
                \imageGap{}%
            \end{subfigure}
        } &%
        \begin{subfigure}[t]{0.19\textwidth}
            \centering%
            \labelImage{\textwidth}{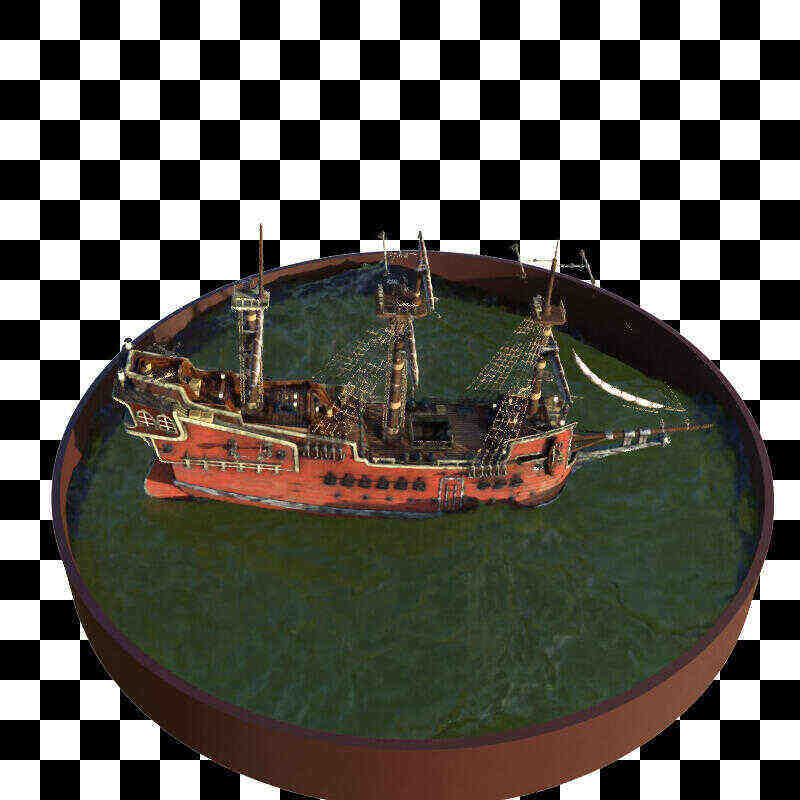}{\MipNeRF{}, 1m it.}
            \labelImage{\textwidth}{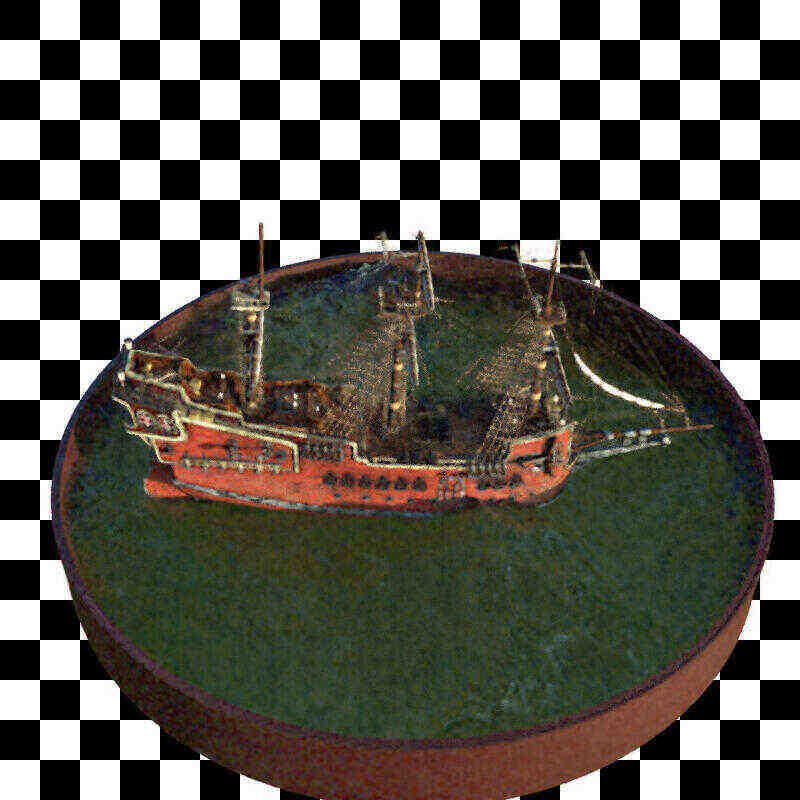}{\ours{}, 120k it.}
            \label{fig:qualitativeEvaluation:shipPhoto.jpg}
            \imageGap{}%
        \end{subfigure}%
        \begin{subfigure}[t]{0.19\textwidth}
            \centering%
            \labelImage{\textwidth}{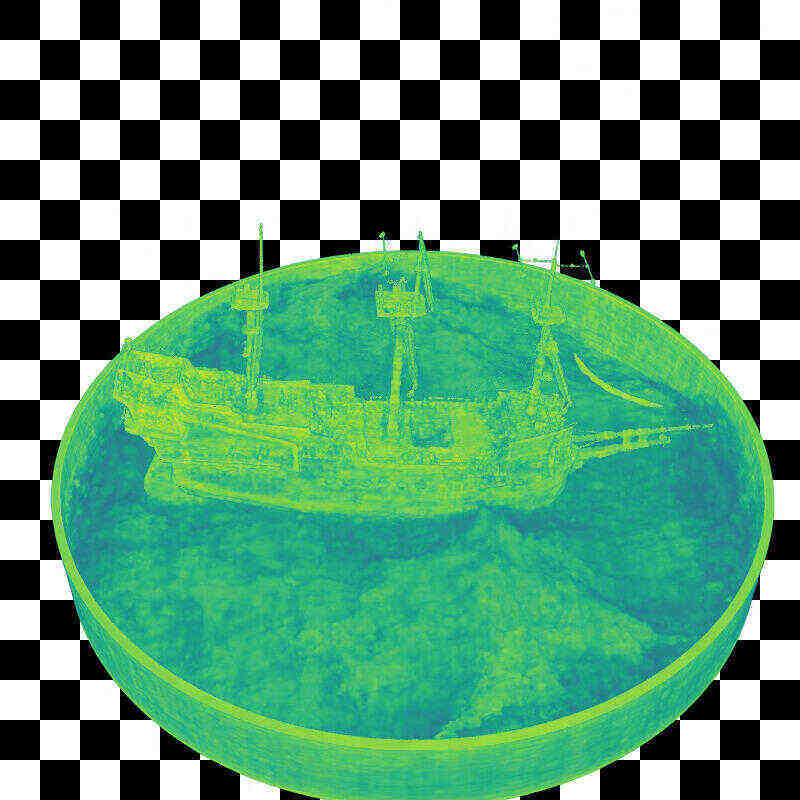}{\MipNeRF{}, 1m it.}
            \labelImage{\textwidth}{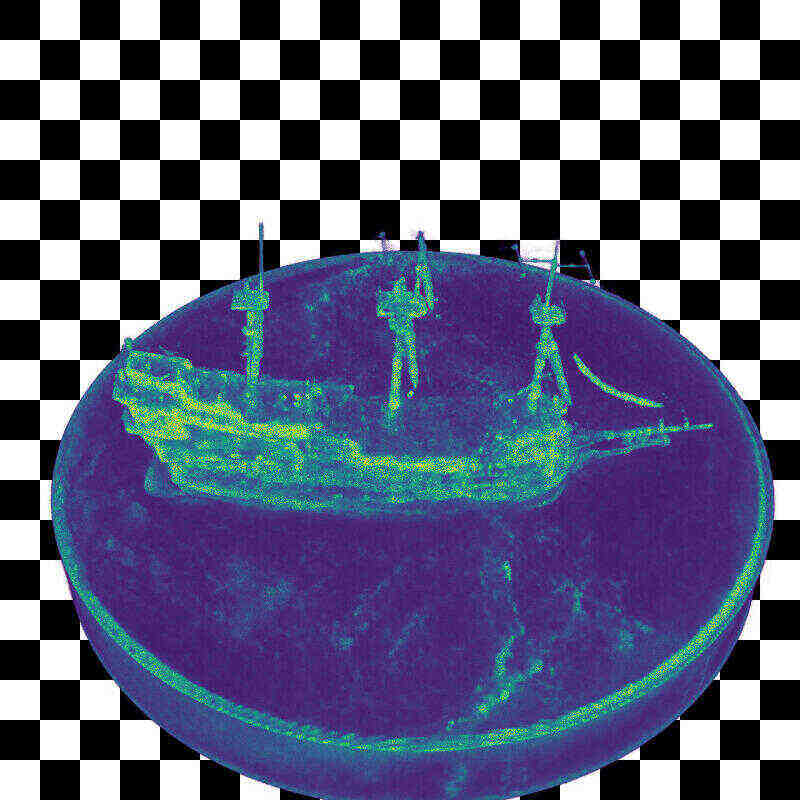}{\ours{}, 120k it.}
            \label{fig:qualitativeEvaluation:shipDensity.jpg}
            \imageGap{}%
        \end{subfigure}%
        \begin{subfigure}[t]{0.19\textwidth}
            \centering%
            \normalMap{\textwidth}{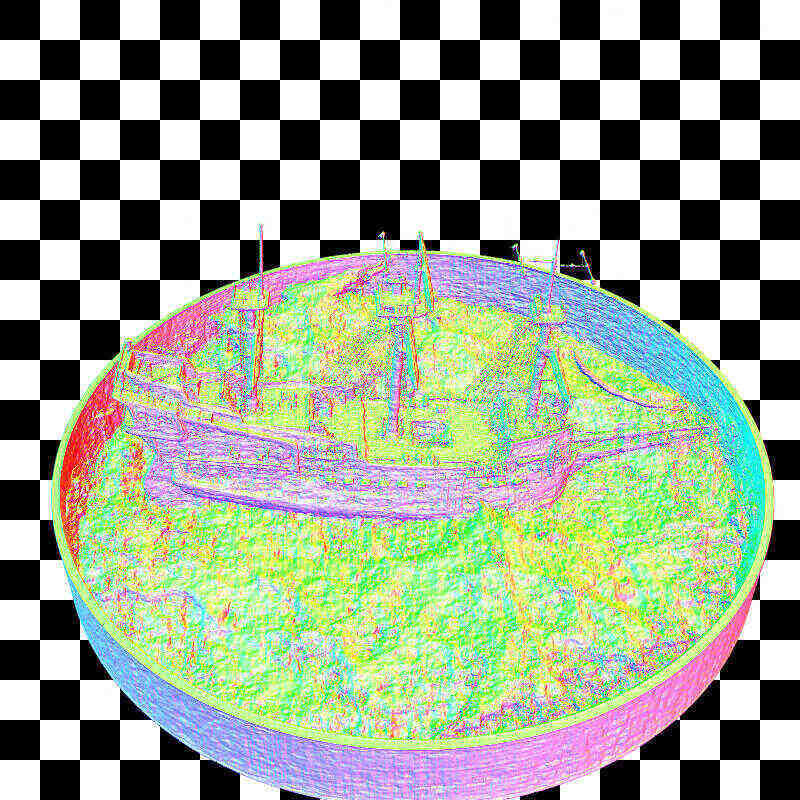}{\MipNeRF{}, 1m it.}
            \normalMap{\textwidth}{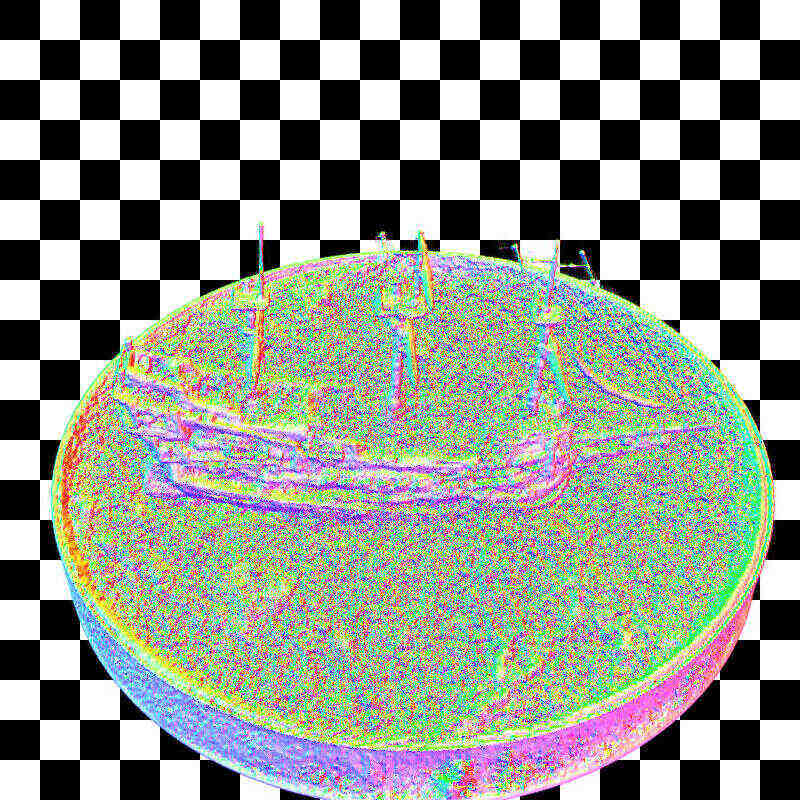}{\ours{}, 120k it.}
            \label{fig:qualitativeEvaluation:shipNormals.jpg}
            \imageGap{}%
        \end{subfigure}%
        \begin{subfigure}[t]{0.19\textwidth}
            \centering%
            \labelImage{\textwidth}{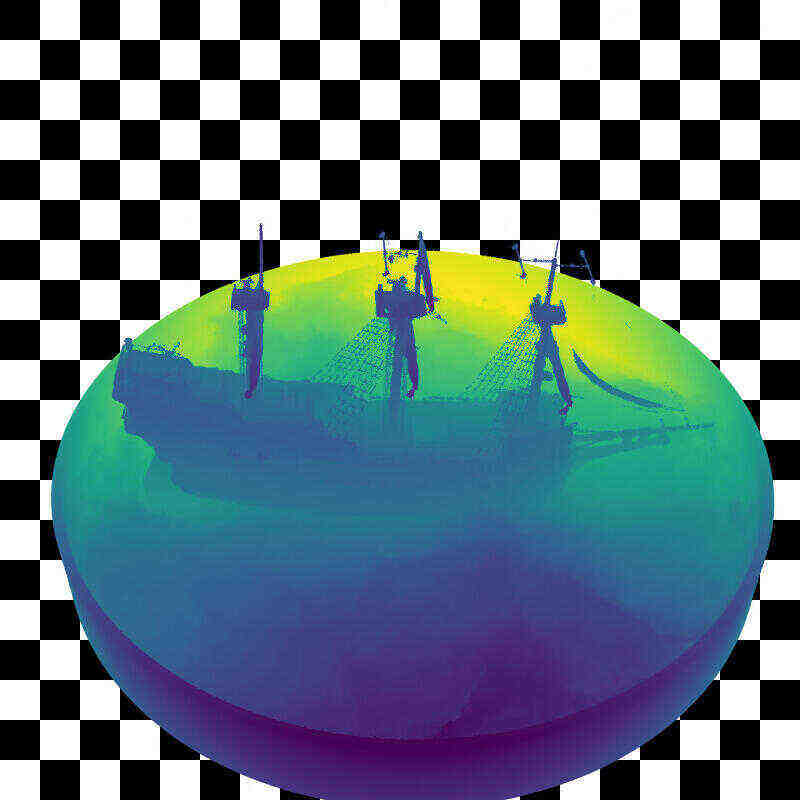}{\MipNeRF{}, 1m it.}
            \labelImage{\textwidth}{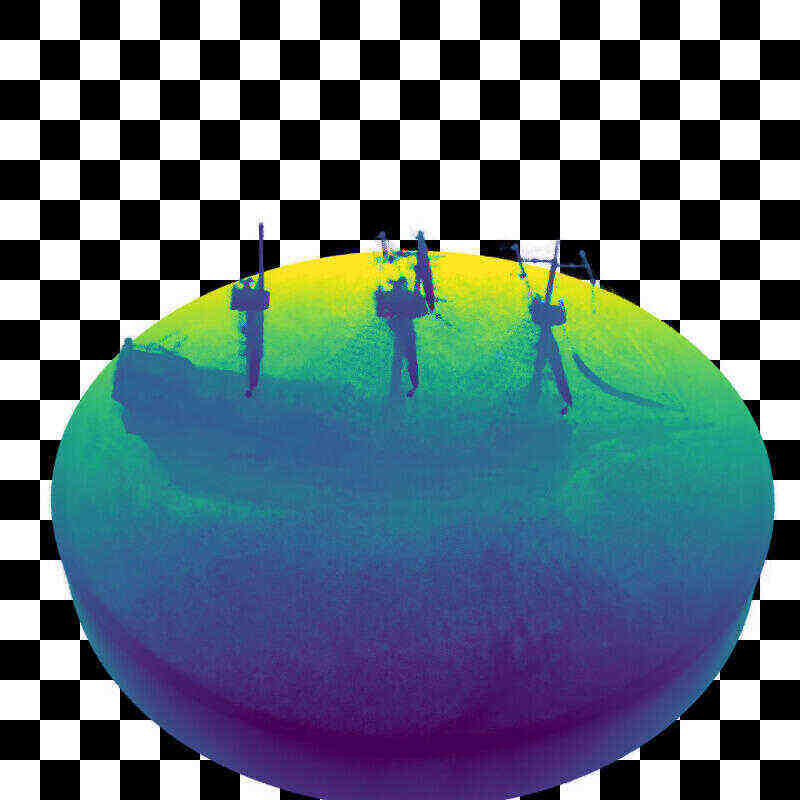}{\ours{}, 120k it.}
            \label{fig:qualitativeEvaluation:shipDepth.jpg}
            \imageGap{}%
        \end{subfigure}
    \end{tabular}
    %
    %
    \caption%
    {%
        Qualitative evaluation using \MipNeRF{} and ours
        on the synthetic chair and ship scene \cite{mildenhall2020NeRF}
        with (from left to right):
        original hold-out ground truth image;
        photo-realistic reconstruction;
        density (\MipNeRF{}) or opacity (\ours{});
        surface normals
        and depth map visualization.
    }%
    \label{fig:qualitativeEvaluationNeRFSynthetic}
\end{figure*}

\begin{figure*}%
    \centering%
    \begin{tabular}{ll}
        %
        %
        \begin{subfigure}[t]{0.16\textwidth}
            \centering%
            \labelImage{\textwidth}{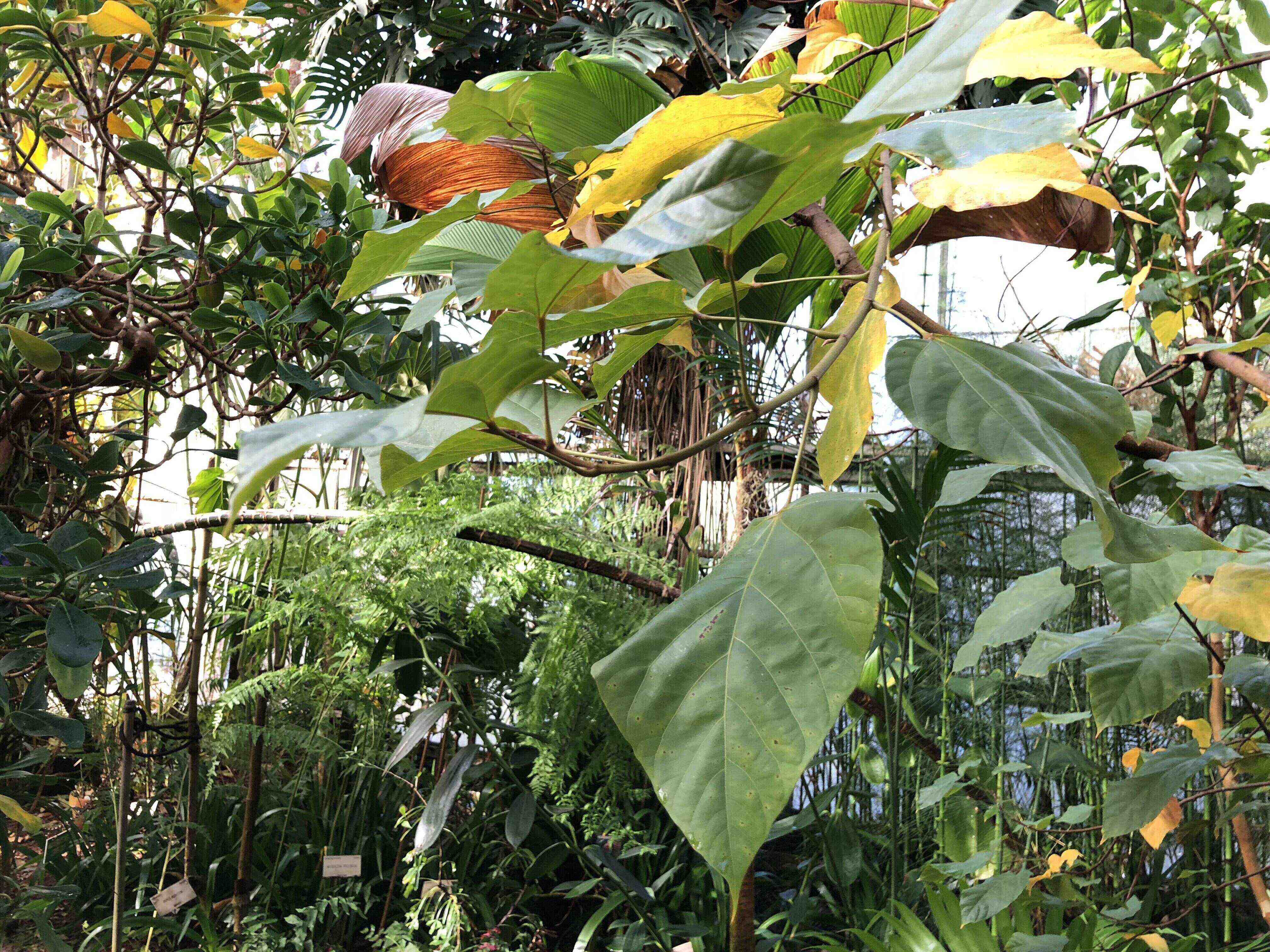}{gt}
            \label{fig:qualitativeEvaluation:leavesOriginal}
            \imageGap{}%
        \end{subfigure}
        &%
        \begin{subfigure}[t]{0.16\textwidth}
            \centering%
            \labelImage{\textwidth}{Results/QualitativeEvaluation/LeavesView008/JaxNeRF/View08}{\JaxNeRF{}}
            \label{fig:qualitativeEvaluation:leavesPhotoNeRF}
            \imageGap{}%
        \end{subfigure}%
        \begin{subfigure}[t]{0.16\textwidth}
            \centering%
                \labelImage{\textwidth}{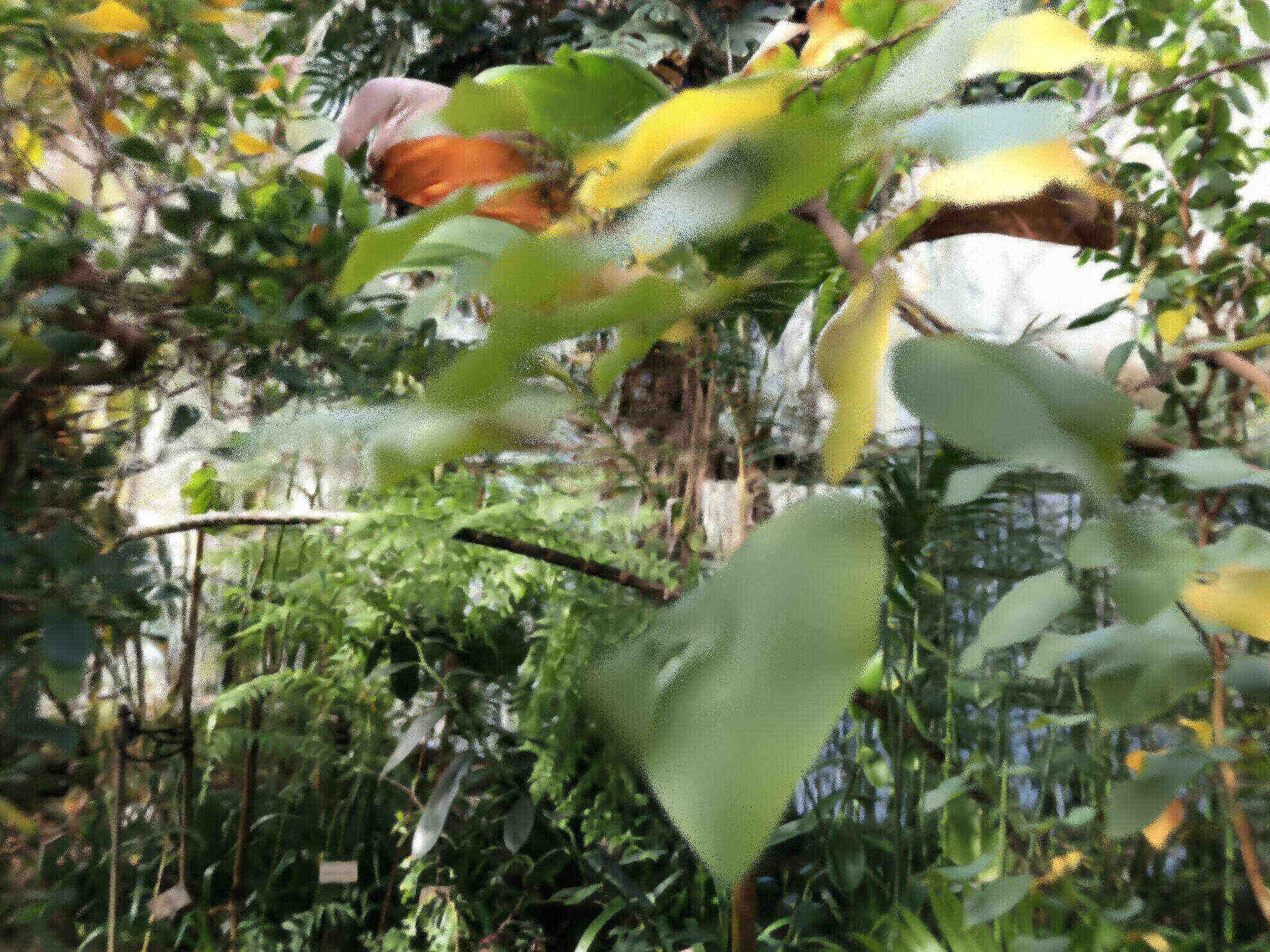}{\ours{}, 60k it.}
            \label{fig:qualitativeEvaluation:leavesPhotoOurs}
            \imageGap{}%
        \end{subfigure}%
        \begin{subfigure}[t]{0.16\textwidth}
            \centering%
            \labelImage{\textwidth}{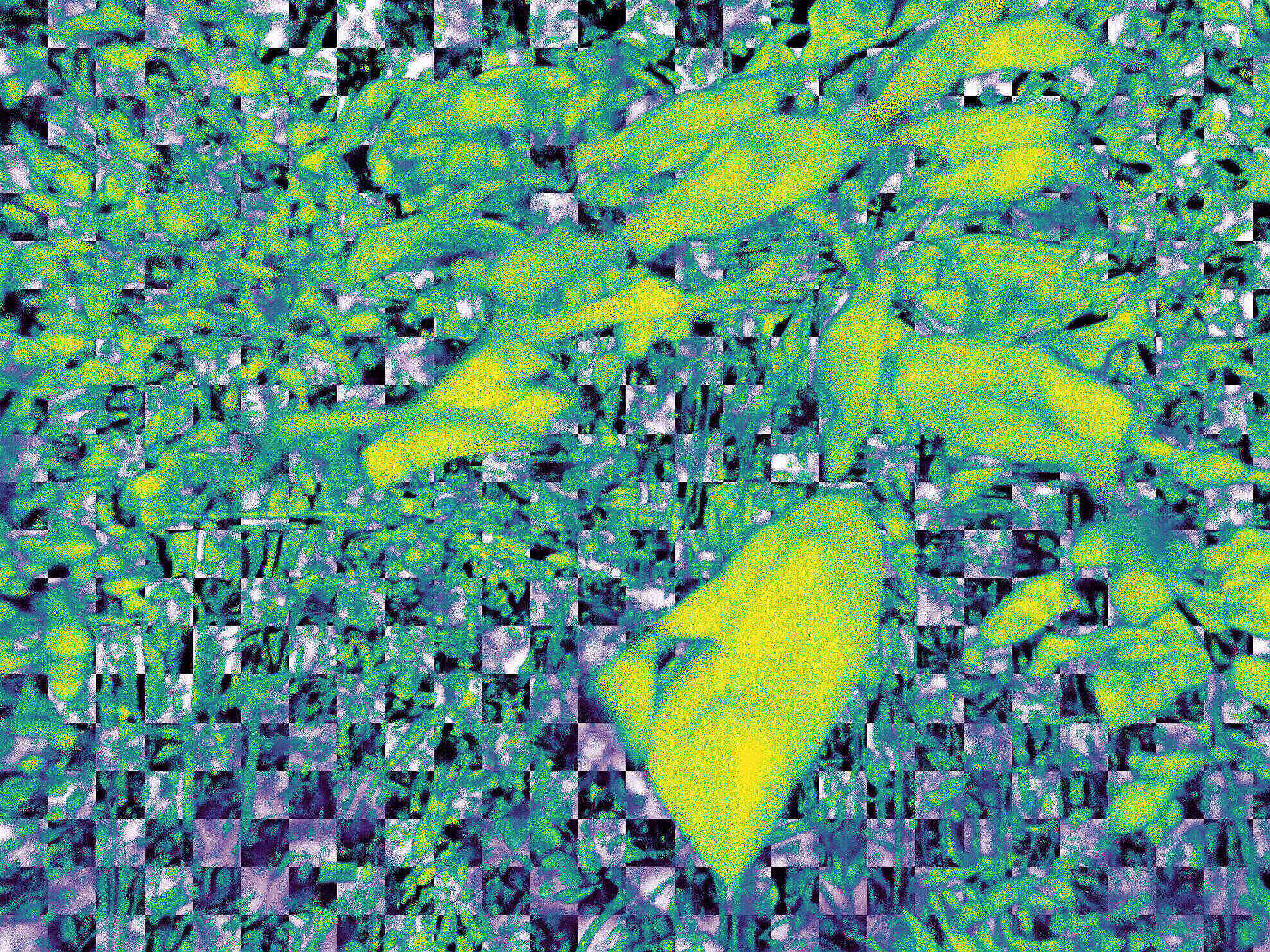}{\ours{}}%
            \label{fig:qualitativeEvaluation:leavesDensity.jpg}
            \imageGap{}%
        \end{subfigure}%
        \begin{subfigure}[t]{0.16\textwidth}
            \centering%
            \normalMap{\textwidth}{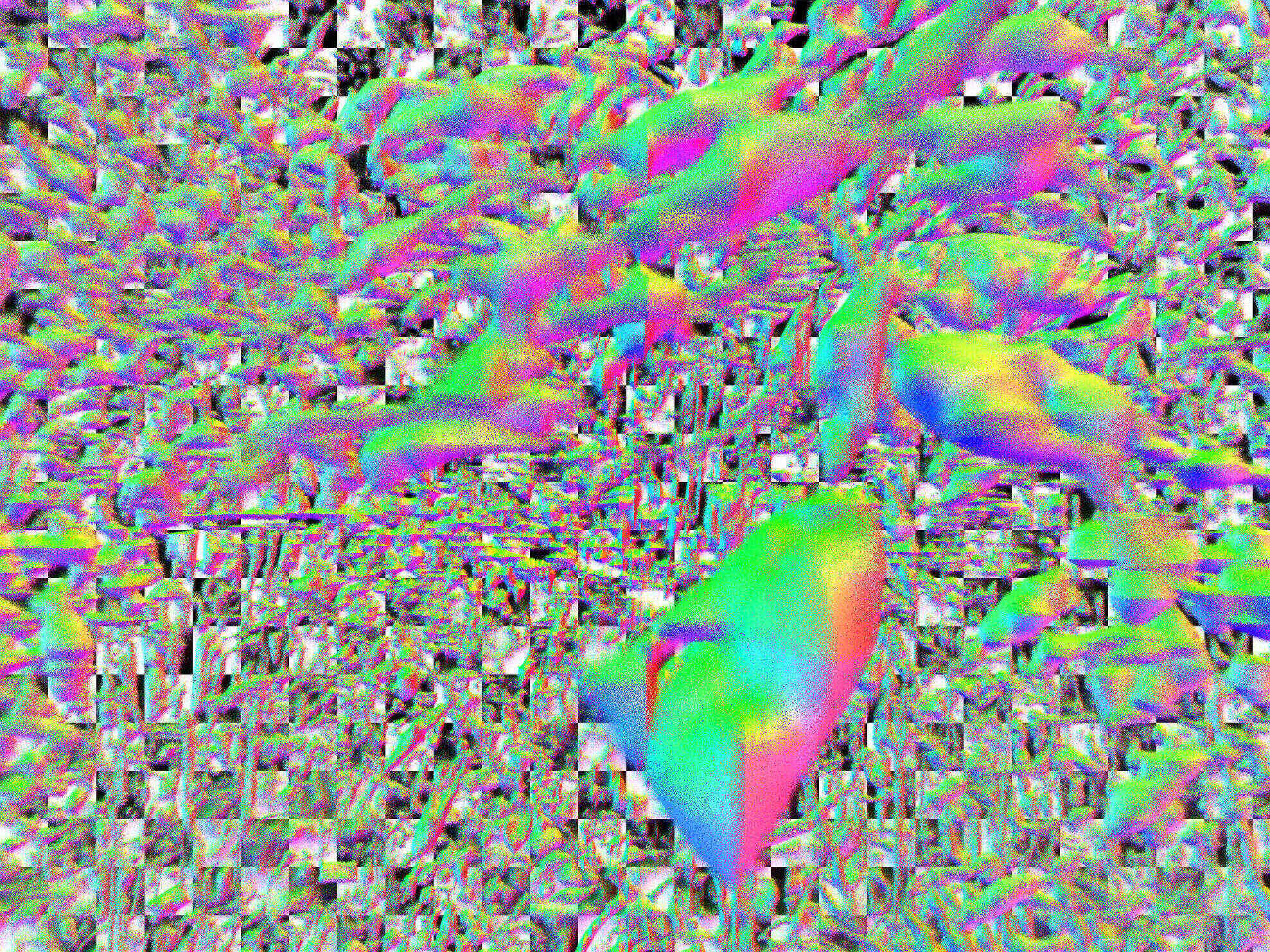}{\ours{}}
            \label{fig:qualitativeEvaluation:leavesNormals.jpg}
            \imageGap{}%
        \end{subfigure}%
        \begin{subfigure}[t]{0.16\textwidth}
            \centering%
            \labelImage{\textwidth}{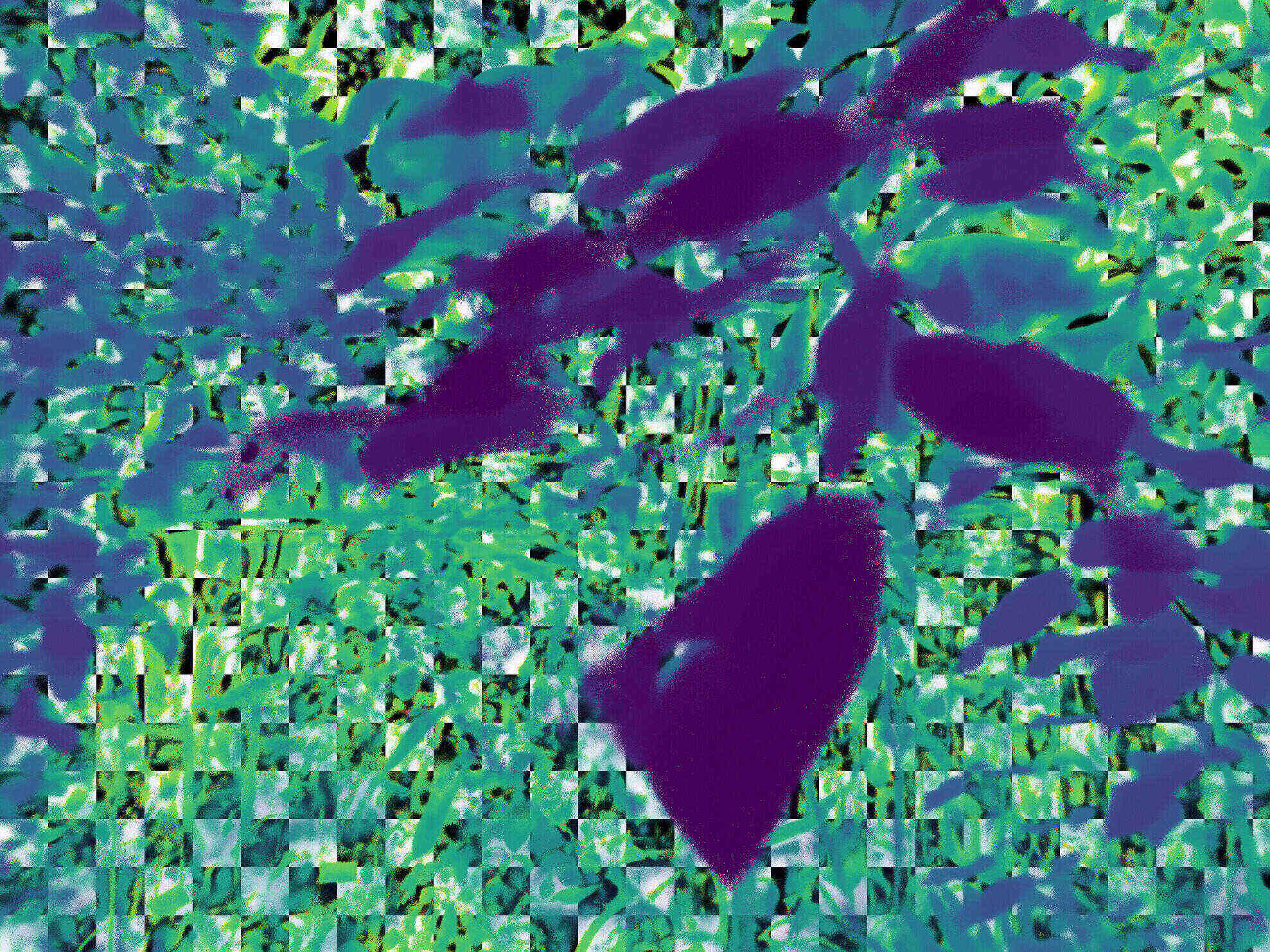}{\ours{}}%
            \label{fig:qualitativeEvaluation:leavesDepth.jpg}
            \imageGap{}%
        \end{subfigure} \\
        %
        %
        \begin{subfigure}[t]{0.16\textwidth}
            \centering%
            \labelImage{\textwidth}{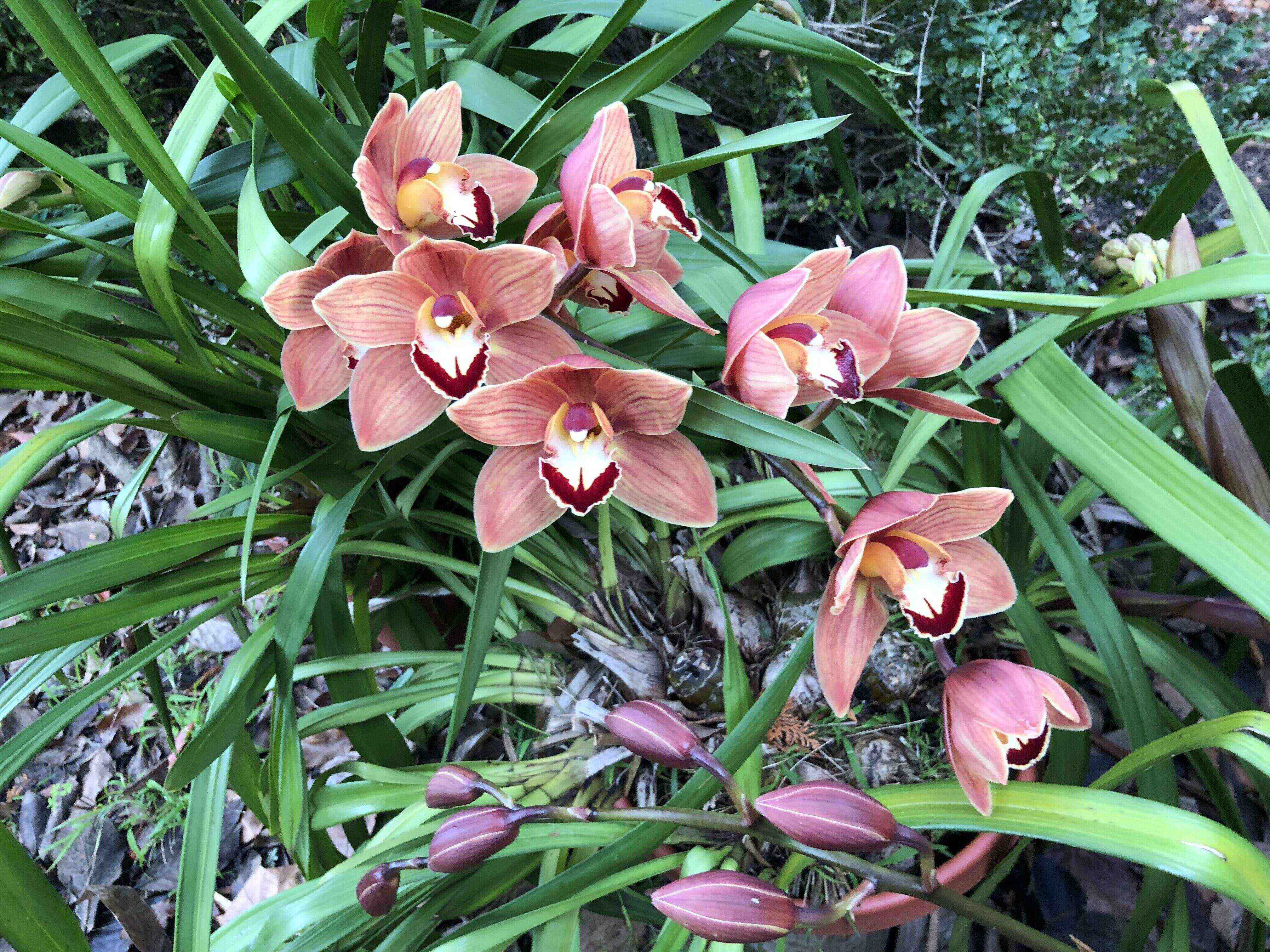}{gt}
            \label{fig:qualitativeEvaluation:orchidsOriginal}
            \imageGap{}%
        \end{subfigure}
        &%
        \begin{subfigure}[t]{0.16\textwidth}
            \centering%
            \labelImage{\textwidth}{Results/QualitativeEvaluation/OrchidsView08/JaxNeRF/View08}{\JaxNeRF{}}
            \label{fig:qualitativeEvaluation:orchidsPhotoNeRF}
            \imageGap{}%
        \end{subfigure}%
        \begin{subfigure}[t]{0.16\textwidth}
            \centering%
            \labelImage{\textwidth}{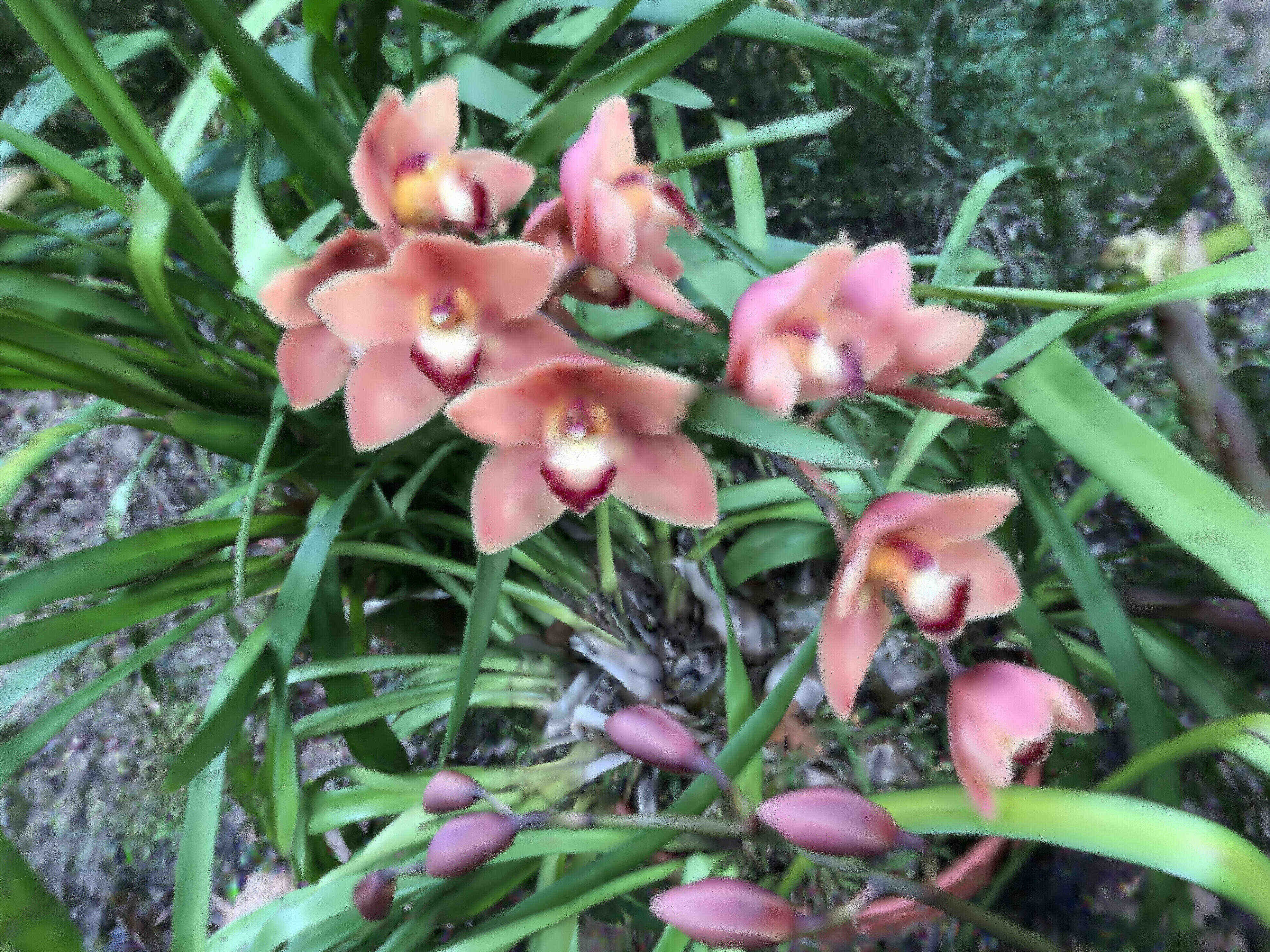}{\ours{}, 40k it.}
            \label{fig:qualitativeEvaluation:orchidsPhotoOurs}
            \imageGap{}%
        \end{subfigure}%
        \begin{subfigure}[t]{0.16\textwidth}
            \centering%
            \labelImage{\textwidth}{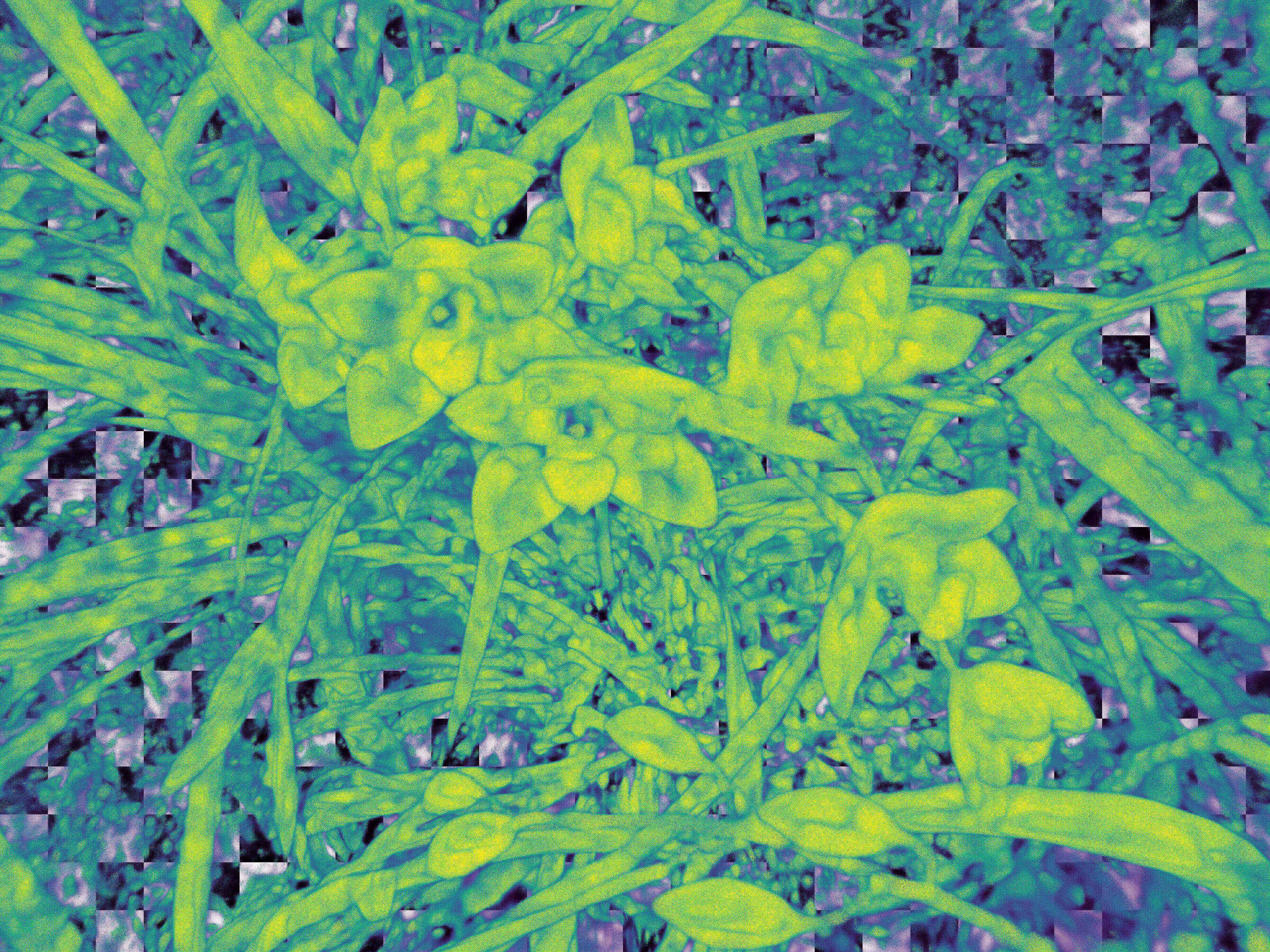}{\ours{}}%
            \label{fig:qualitativeEvaluation:orchidsDensity.jpg}
            \imageGap{}%
        \end{subfigure}%
        \begin{subfigure}[t]{0.16\textwidth}
            \centering%
            \normalMap{\textwidth}{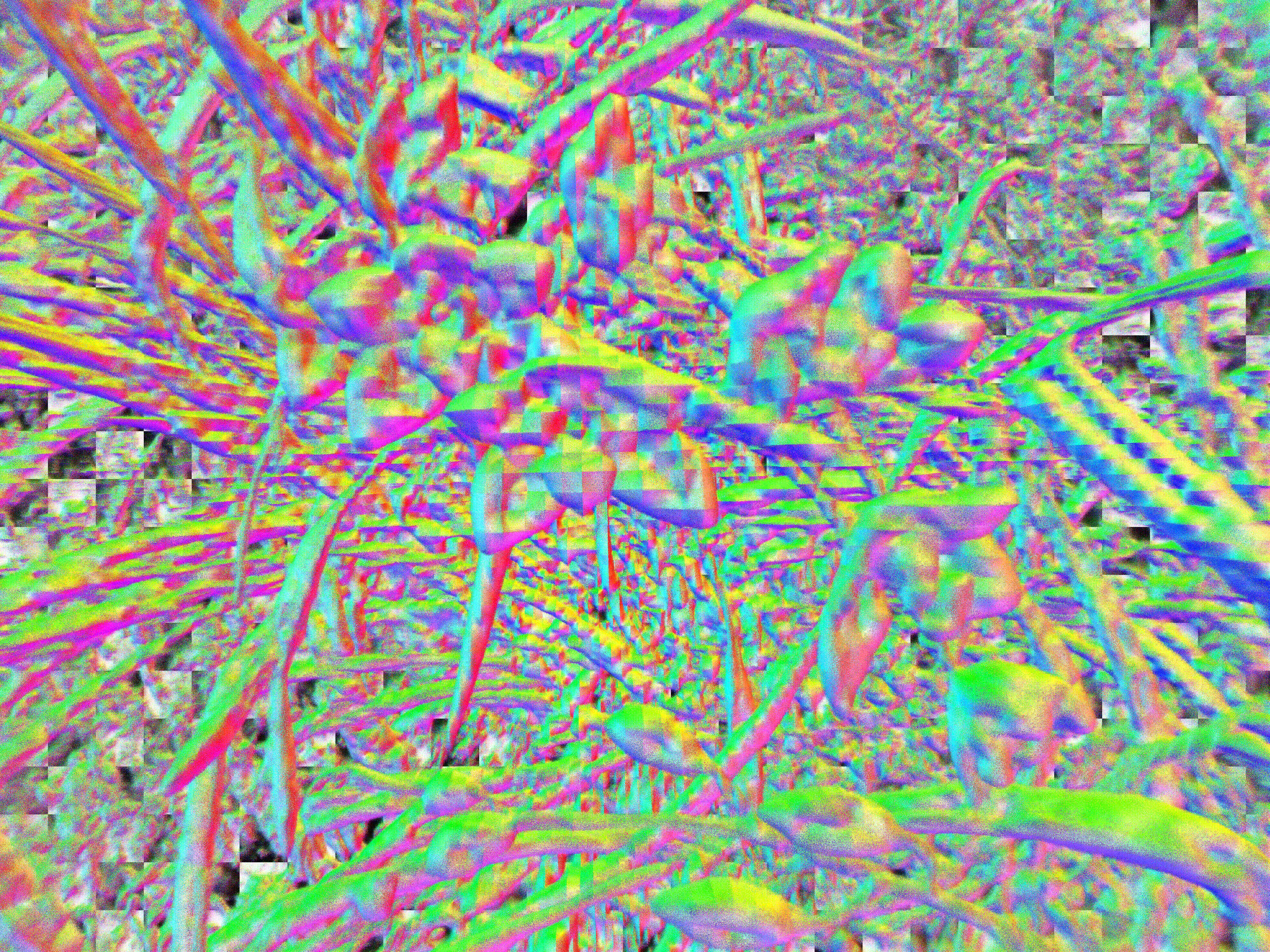}{\ours{}}%
            \label{fig:qualitativeEvaluation:orchidsNormals.jpg}
            \imageGap{}%
        \end{subfigure}%
        \begin{subfigure}[t]{0.16\textwidth}
            \centering%
            \labelImage{\textwidth}{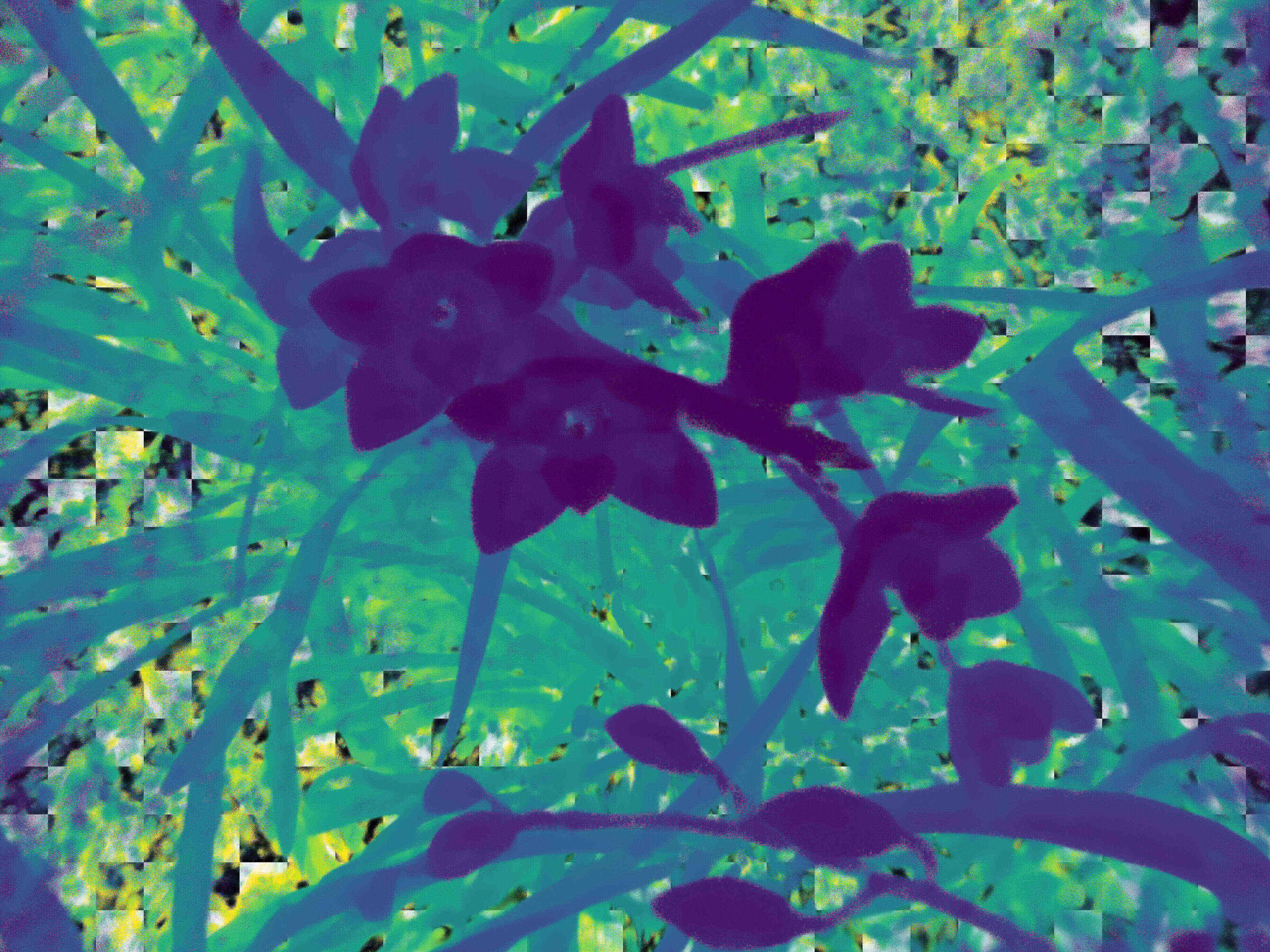}{\ours{}}
            \label{fig:qualitativeEvaluation:orchidsDepth.jpg}
            \imageGap{}%
        \end{subfigure}
    \end{tabular}
    %
    %
    \caption%
    {%
        Qualitative evaluation using \JaxNeRF{} and ours
        on the leaves and orchid scene \cite{mildenhall2019LLFF}
        with (from left to right):
        original hold-out ground truth images;
        photo-realistic reconstructions (\JaxNeRF{} and \ours{});
        renderings of opacity, surface normals and depths (\ours{} only).
    }%
    \label{fig:qualitativeEvaluationLLFF}
\end{figure*}

\Fig{\ref{fig:qualitativeEvaluationNeRFSynthetic}} shows reconstruction for exemplary synthetic scenes
from \NeRF{} \cite{mildenhall2020NeRF}
where our method performs similar to \MipNeRF{}.
Note that
our algorithm is able to reconstruct many of the fine and thin structures like the ship ropes
despite that it starts with a coarse grid only.
\Fig{\ref{fig:qualitativeEvaluationLLFF}} demonstrates that our approach can also reconstruct
large scale outdoor scenes with complex scene geometry.
Though,
our models exhibit fewer details
than \JaxNeRF{} (improved original \NeRF{}).

\paragraph{Quantitative evaluation}

\begin{figure}
    \begin{subfigure}[t]{0.19\textwidth}
        \centering%
        \labelImage{\textwidth}{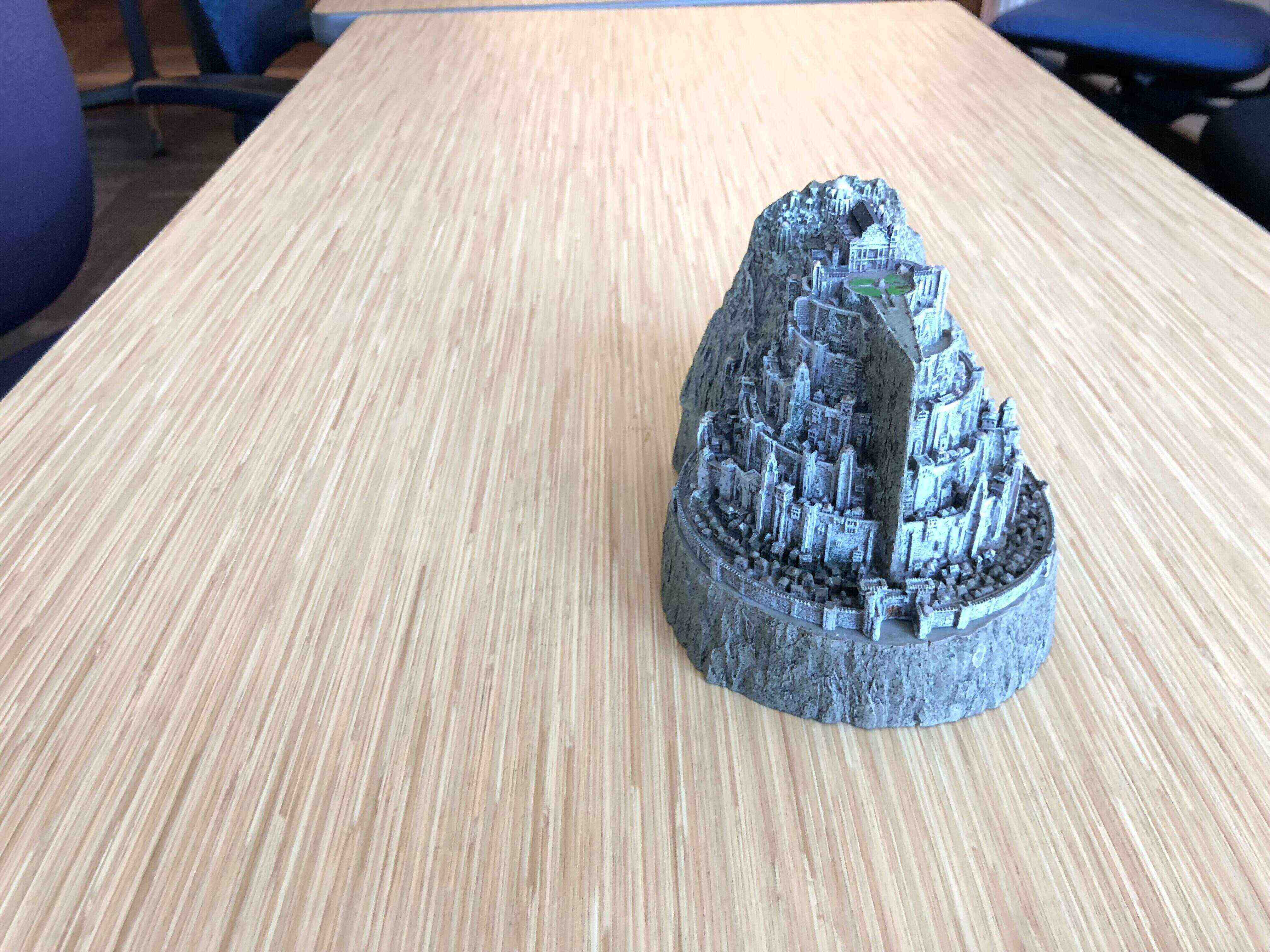}{gt}
        \imageGap{}%
    \end{subfigure}
    \begin{subfigure}[t]{0.19\textwidth}
        \centering%
        \labelImage{\textwidth}{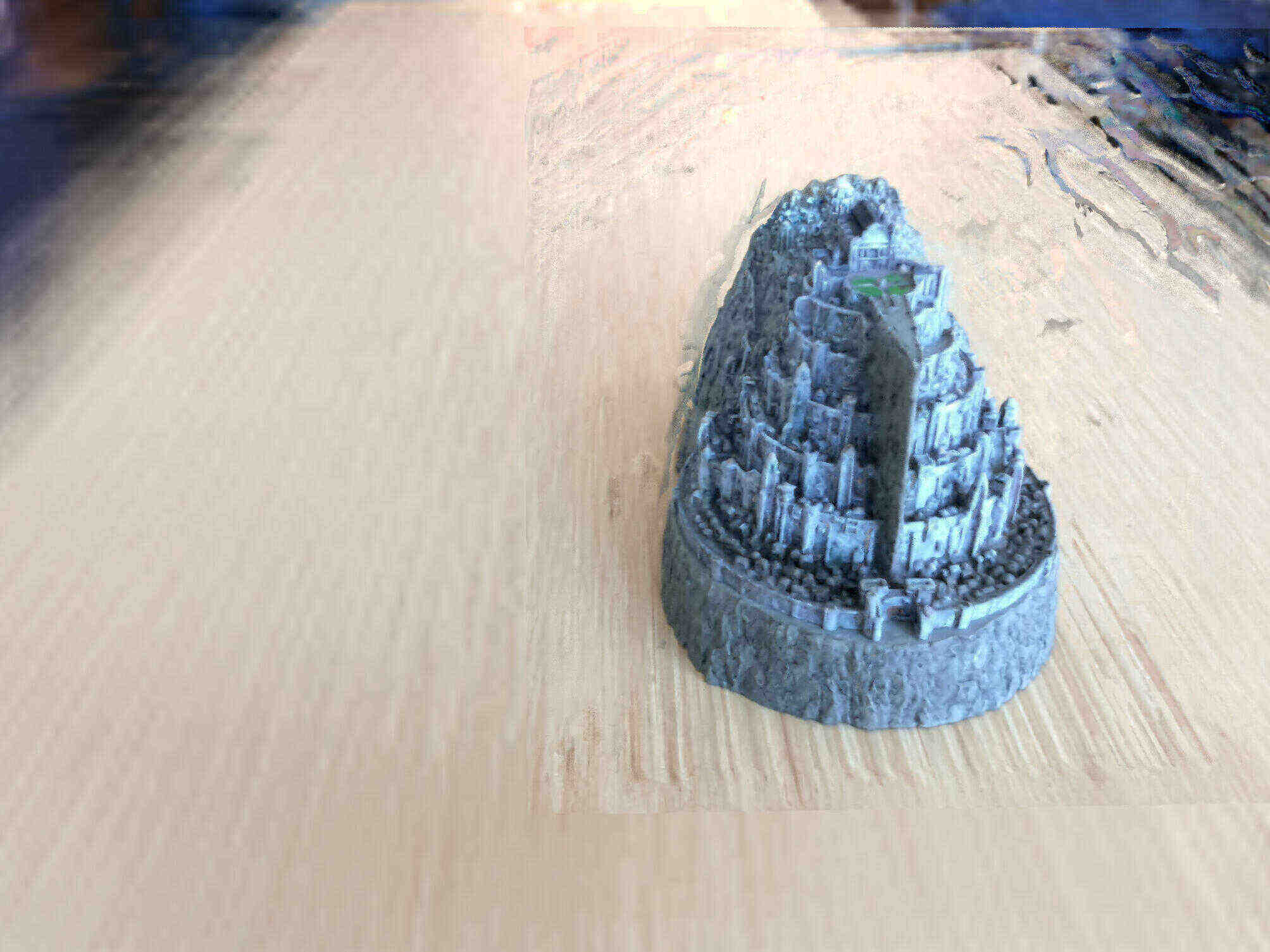}{\ours{}}
        \imageGap{}%
    \end{subfigure} \\
    \begin{subfigure}[t]{0.19\textwidth}
        \centering%
        \labelImage{\textwidth}{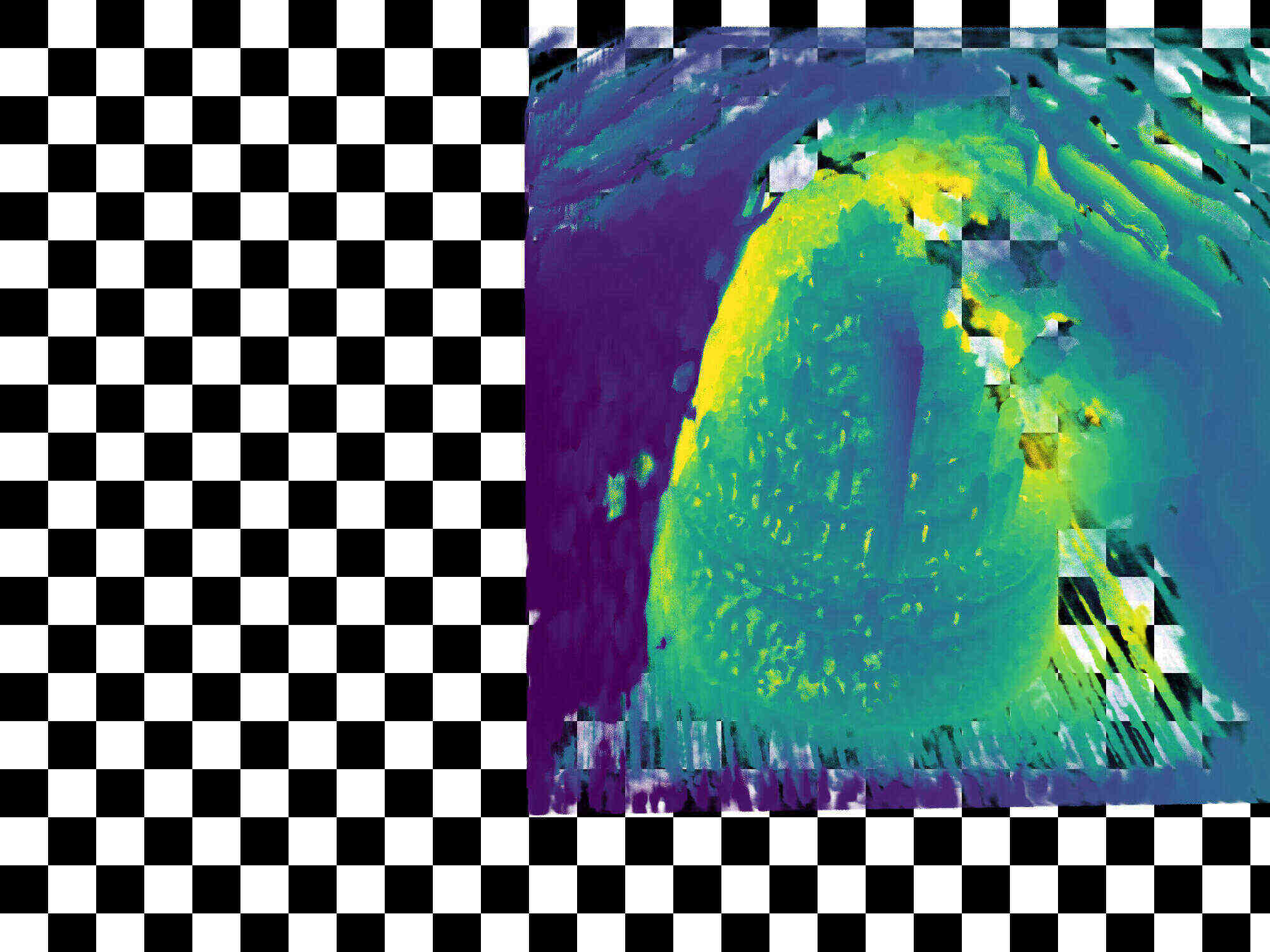}{depth}
        \imageGap{}%
    \end{subfigure}%
    \begin{subfigure}[t]{0.19\textwidth}
        \centering%
        \normalMap{\textwidth}{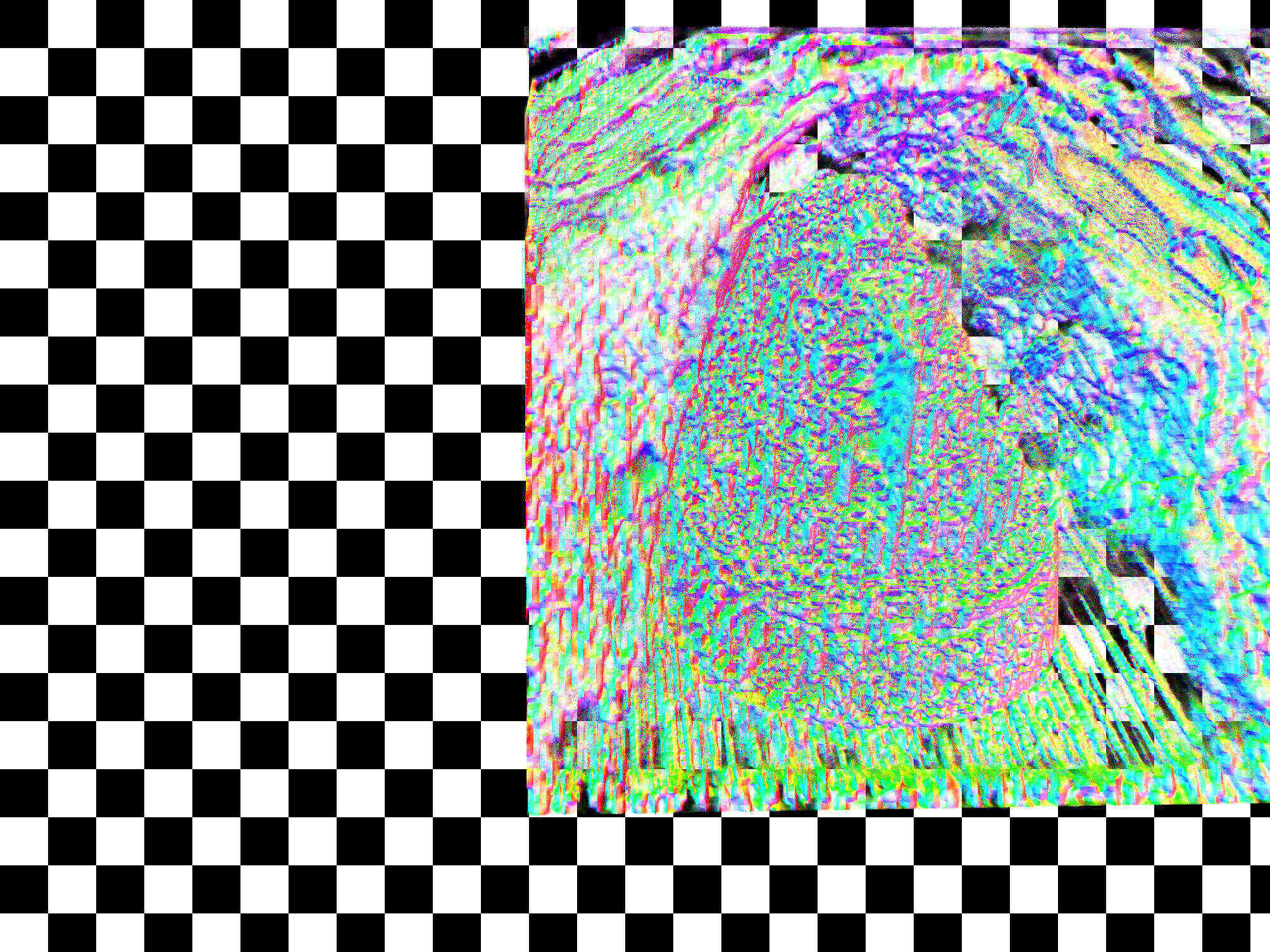}{normals}
        \imageGap{}%
    \end{subfigure}%
    %
    %
    \caption%
    {%
        Fortress scene \cite{mildenhall2019LLFF} reconstruction
        after 170k iterations
        based on a manually defined tight scene \AABB{}.
    }%
    \label{fig:fortressLimitingAABB}
\end{figure}

%
%
\Tab{\ref{tab:quantitativeEvaluationSyntheticNeRF}} shows
that our method reconstructs high quality models
that are similar to the state-of-the-art implicit \MipNeRF{} approach.
Our models perform slightly worse due to view ray sampling noise and
due to the fact that they cannot capture high-frequency reflections well.
We omit a qualitative comparison against \JaxNeRF{} on the \LLFF{} scenes.
Our models consistently perform worse
mainly due to the limiting initial \SVO{} creation
and due to the fact that we do not dynamically adapt
the user-defined scene \AABB{}.
\Fig{\ref{fig:fortressLimitingAABB}} shows that our method
reconstructs the object of interest inside the scene \AABB{}.
However,
it is surrounded by clutter
that the optimizer added
to account for the table outside the \AABB{}
which cannot be represented well by the environment map.

%
%

%
%
\begin{table}
    \begin{tabular}{|l||c|c||c|c||}
    \hline
     & \multicolumn{2}{c||}{PSNR$\uparrow$} & \multicolumn{2}{c|}{SSIM$\uparrow$} \\
    \hline
     & \MipNeRF{} & ours & \MipNeRF{} & ours \\
    \hline
     Chair     & 35.19 & 28.73 & 0.9891 & 0.9887 \\
     Drums     & 26.16 & 24.22 & 0.9597 & 0.9657 \\
     Ficus     & 32.34 & 27.36 & 0.9861 & 0.9864 \\
     Hot dog   & 37.18 & 31.71 & 0.9921 & 0.9883 \\
     Bulldozer & 35.76 & 28.38 & 0.9903 & 0.9817 \\
     Materials & 31.50 & 26.29 & 0.9808 & 0.9662 \\
     Mic       & 36.22 & 30.23 & 0.9941 & 0.9888 \\
     Ship      & 29.33 & 25.38 & 0.9297 & 0.9409 \\
    \hline
     Average   & 32.96 & 27.79 & 0.9777 & 0.9758 \\
    \hline
    \end{tabular}
    \caption%
    {%
        Quantitative novel view synthesis comparison of
        the proposed explicit models (ours)
        with
        \MipNeRF{} \cite{barron2021Mip-NeRF}
        on the synthetic \NeRF{} scenes \cite{mildenhall2020NeRF}.
    }%
    \label{tab:quantitativeEvaluationSyntheticNeRF}
\end{table}

\section{Conclusion}
\label{sec:conclusion}

\paragraph{Limitations and future work}
First, 
the biggest limitation of our method is the simple initial \SVO{} creation.
If the assumption is broken that the geometry outside the \AABB{} is infinitely far away,
the optimizer creates "clutter"\ in free space.
If the \AABB{} is initially very big,
either the voxels are too coarse or the dense initial grid consumes too much memory for fine voxels.
A solution would be creating the dense \SVO{} according to the camera distribution and
dynamically adding new root nodes during optimization if necessary.
Second,
using only \glspl{SH} to store the outgoing radiance of scene surfaces
limits them to low frequency radiance distributions.
Hence, our representation fails to capture sharp highlights for shiny materials.
E.g., the water reflections of the \NeRF{} ship scene are such a failure case.
Investigating different ways to represent hemispherical radiance distributions
within our models, e.g., using kernel-based methods, is an interesting future direction.
Third,
directly optimizing the $\SLF{}$ "bakes the scene light"\ into the surfaces.
This reduces model flexibility and for example possibilities for further editing.
E.g., in case of the bulldozer scene of \Fig{\ref{fig:editingExample}},
it is desirable to not only remove the shovel, but also the shadow it casts onto the vehicle.
However,
decomposing \glspl{SLF} into incoming radiance and surface materials seems possible
since their convolution is also differentiable and can be optimized using \SGD{}.
Though, this is a very complex and ambiguous problem, 
which is out of scope of this work, but also promising for future research.

\paragraph{Summary}
In this work, we presented a novel representation which explicitly represents a 3D scene using 
a sparse multi-resolution grid storing
surface geometry statistically via an opacity field and
the radiance outgoing from this surface geometry via a volumetric \SH{} \SLF{}.
Further,
we showed how to initialize and optimize such a scene model in a coarse-to-fine manner
using \SGD{} and inverse differentiable rendering for multiple views.
Finally,
we provided reconstruction results comparable to state-of-the-art approaches regarding novel view synthesis
and demonstrated the advantages of our novel approach regarding
its versatility for capturing as well as
its suitability for post processing use cases,
such as interactive editing.



\bibliographystyle{ACM-Reference-Format}
\bibliography{Bibliography}

\appendix
\newpage
\section{Supplementary material}
In the following,
we show additional results for 
specific parts of our method
like changing model parameters, such as the \SH{} band count,
as well as more quantitative and qualitative comparisons of our method
against state-of-the-art implicit alternatives.
Also we show how simple our \LiLU{} constraints are
using their full \PyTorch{} \cite{pytorch2019} code:

\begin{verbatim}  
import torch 

class LiLU(torch.autograd.Function):
    @staticmethod
    def forward(
        ctx, x: torch.Tensor
    ) -> torch.Tensor:
        ctx.border = x.le(0.0)
        y = x.masked_fill(ctx.border, 0.0)
        return y

    @staticmethod
    def backward(
        ctx, g_y: torch.Tensor
    ) -> torch.Tensor:
        if not ctx.needs_input_grad[0]:
            return (None,)
            
            
            
        # shrinking?
        # (assuming loss minimization)
        s = g_y > 0.0
        # and at border?
        b = ctx.border
        # not okay
        no = torch.logical_and(b, s)
        g_in = g_y.masked_fill(no, 0.0)
        return (g_in,)
        
\end{verbatim}        


%
%
\Tab{\ref{tab:quantitativeEvaluationSyntheticNeRFLPIPS}} compares
our approach regarding \LPIPS{}
on the 
\NeRF{} scenes
where \MipNeRF{} performs better.

%
%
\begin{table}[b]
    \centering
    \begin{tabular}{|l||c|c|}
    \hline
    \LPIPS{}$\downarrow$ & \MipNeRF{} & ours \\
    \hline
     Chair     & 0.013 & 0.036 \\
     Drums     & 0.064 & 0.072 \\
     Ficus     & 0.021 & 0.045 \\
     Hot dog   & 0.020 & 0.061 \\
     Bulldozer & 0.015 & 0.043 \\
     Materials & 0.027 & 0.078 \\
     Mic       & 0.006 & 0.027 \\
     Ship      & 0.128 & 0.183 \\
    \hline
     Average   & 0.0367 & 0.0679 \\
    \hline
    \end{tabular}
    \caption%
    {%
        \LPIPS{} comparison
        using the proposed explicit models (ours)
        with
        \MipNeRF{} \cite{barron2021Mip-NeRF}
        on the scenes from
        \cite{mildenhall2020NeRF}.
    }%
    \label{tab:quantitativeEvaluationSyntheticNeRFLPIPS}
\end{table}

\begin{figure*}%
    \centering%
    \begin{tabular}{l}
        %
        %
        \begin{subfigure}[t]{0.195\textwidth}
            \centering%
            \labelImage{\textwidth}{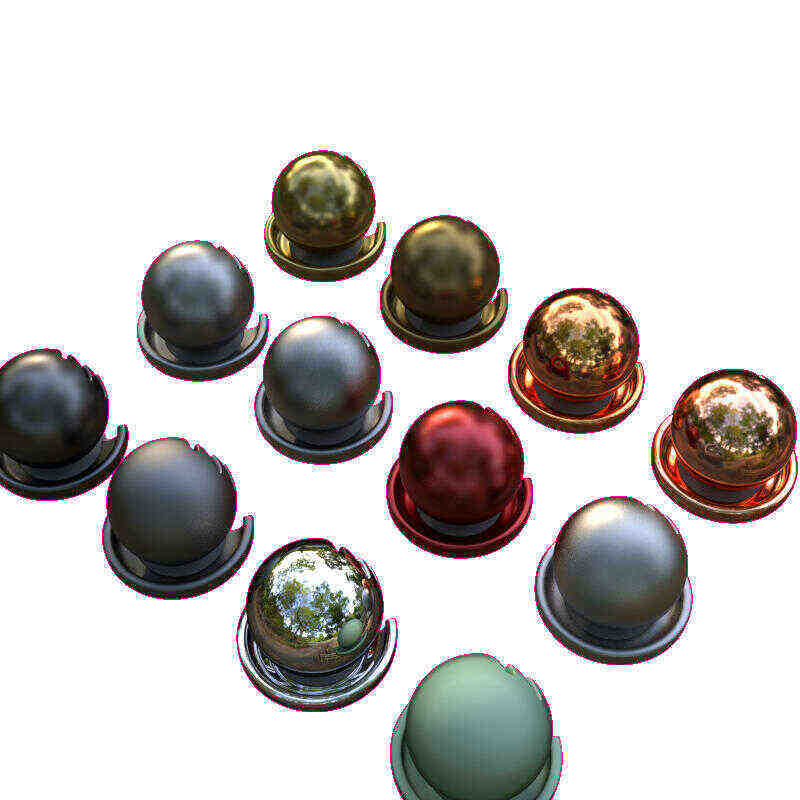}{gt}
            \label{fig:qualitativeEvaluation:materialsSHBandCount1GroundTruth}
            \imageGap{}%
        \end{subfigure}%
        \begin{subfigure}[t]{0.195\textwidth}
            \centering%
            \labelImage{\textwidth}{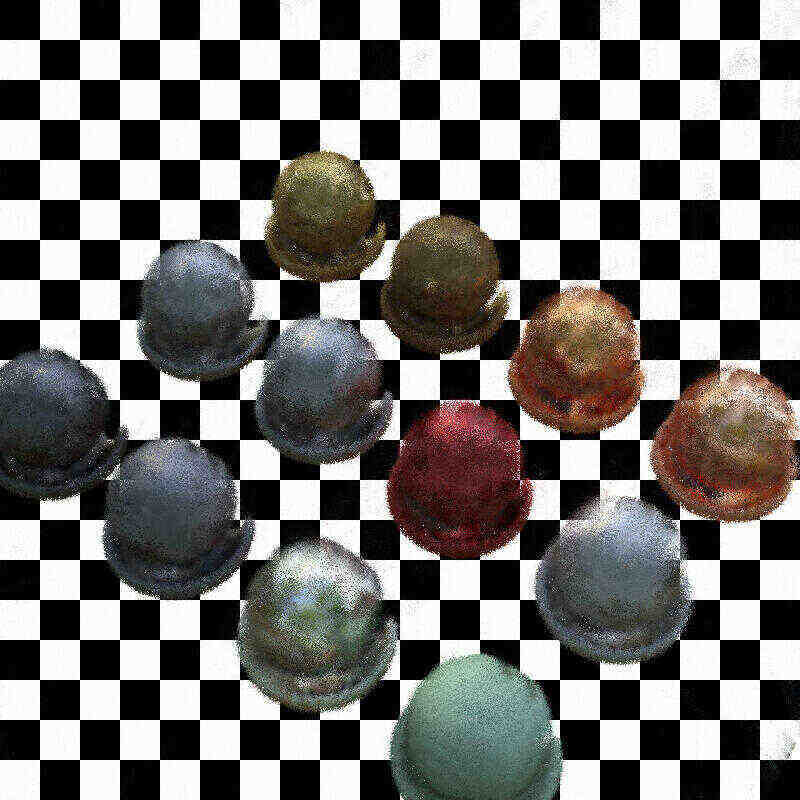}{b=1}
            \label{fig:qualitativeEvaluation:materialsSHBandCount1Photo.jpg}
            \imageGap{}%
        \end{subfigure}%
        \begin{subfigure}[t]{0.195\textwidth}
            \centering%
            \labelImage{\textwidth}{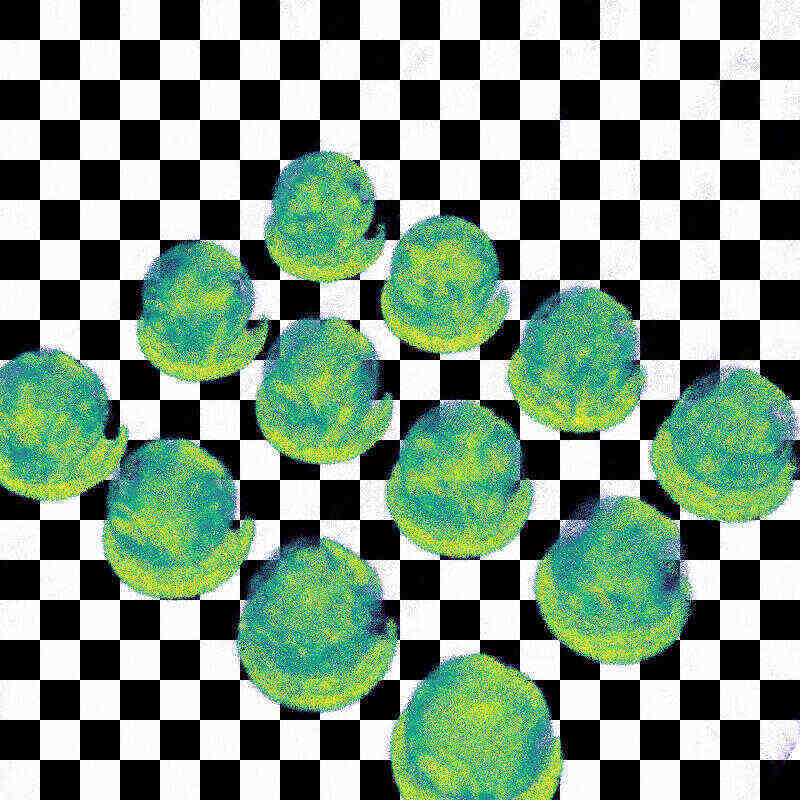}{b=1}
            \label{fig:qualitativeEvaluation:materialsSHBandCount1Density.jpg}
            \imageGap{}%
        \end{subfigure}%
        \begin{subfigure}[t]{0.195\textwidth}
            \centering%
            \normalMap{\textwidth}{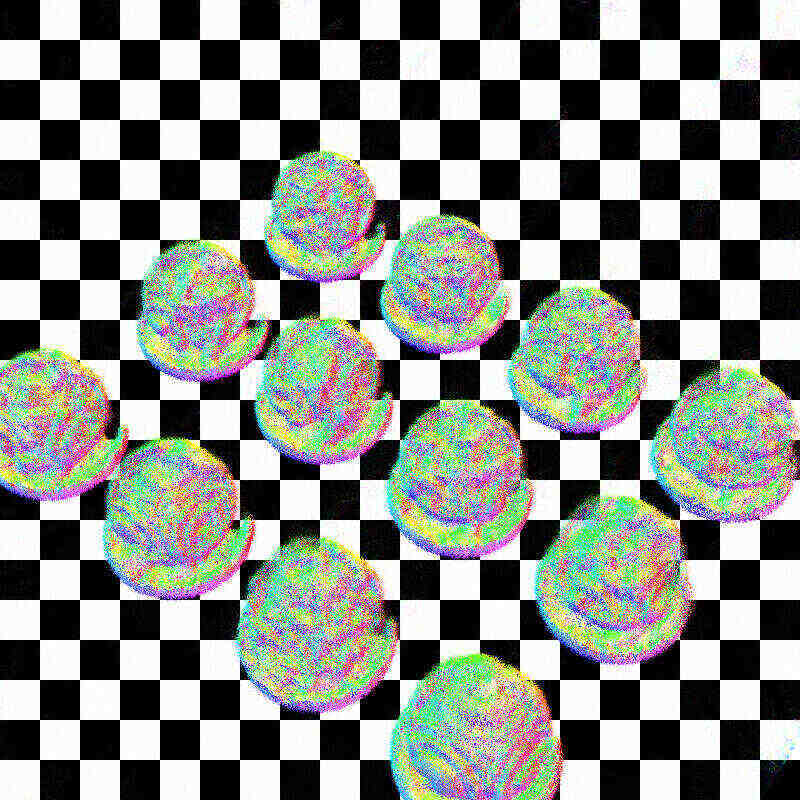}{b=1}
            \label{fig:qualitativeEvaluation:materialsSHBandCount1Normals.jpg}
            \imageGap{}%
        \end{subfigure}%
        \begin{subfigure}[t]{0.195\textwidth}
            \centering%
            \labelImage{\textwidth}{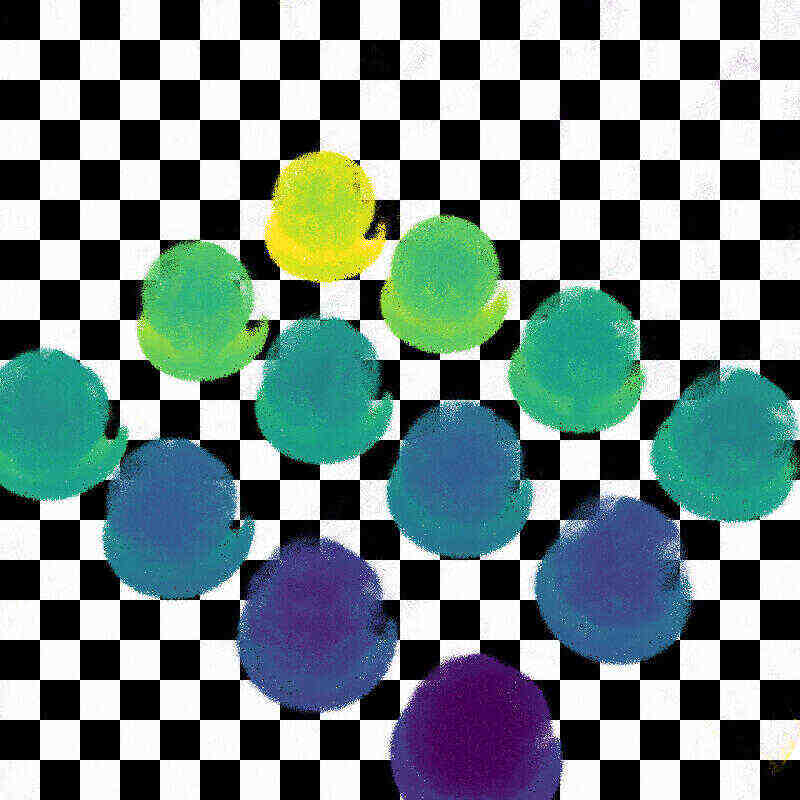}{b=1}
            \label{fig:qualitativeEvaluation:materialsSHBandCount1Depth.jpg}
            \imageGap{}%
        \end{subfigure} \\
        %
        %
        \begin{subfigure}[t]{0.195\textwidth}
            \centering%
            \labelImage{\textwidth}{Results/AblationStudy/SHBandCount/GroundTruth/MaterialsView300/Photo.jpg}{gt}
            \label{fig:qualitativeEvaluation:materialsSHBandCount2GroundTruth}
            \imageGap{}%
        \end{subfigure}%
        \begin{subfigure}[t]{0.195\textwidth}
            \centering%
            \labelImage{\textwidth}{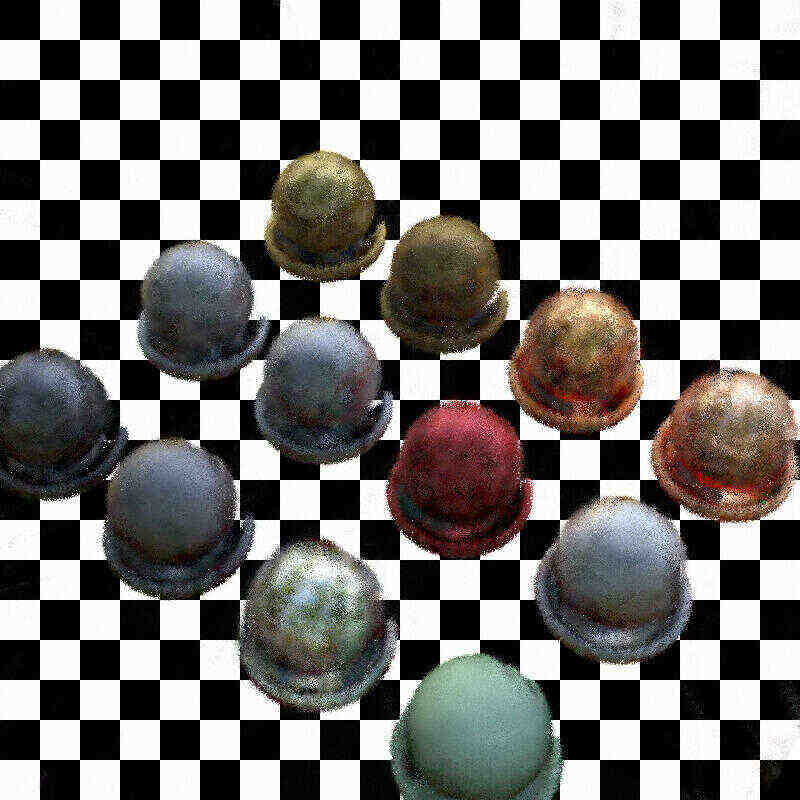}{b=2}
            \label{fig:qualitativeEvaluation:materialsSHBandCount2Photo.jpg}
            \imageGap{}%
        \end{subfigure}%
        \begin{subfigure}[t]{0.195\textwidth}
            \centering%
            \labelImage{\textwidth}{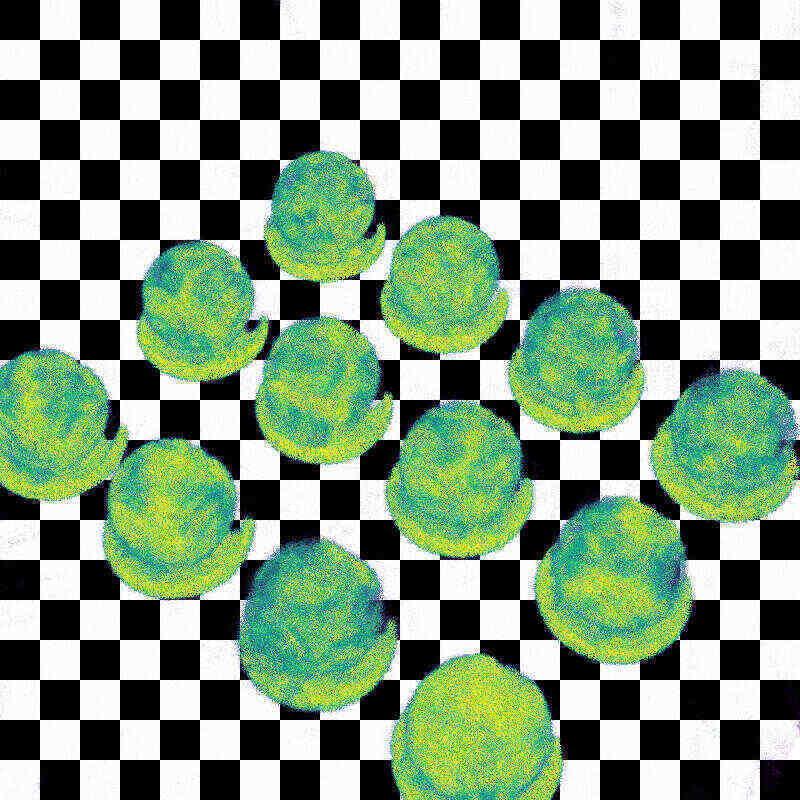}{b=2}
            \label{fig:qualitativeEvaluation:materialsSHBandCount2Density.jpg}
            \imageGap{}%
        \end{subfigure}%
        \begin{subfigure}[t]{0.195\textwidth}
            \centering%
            \normalMap{\textwidth}{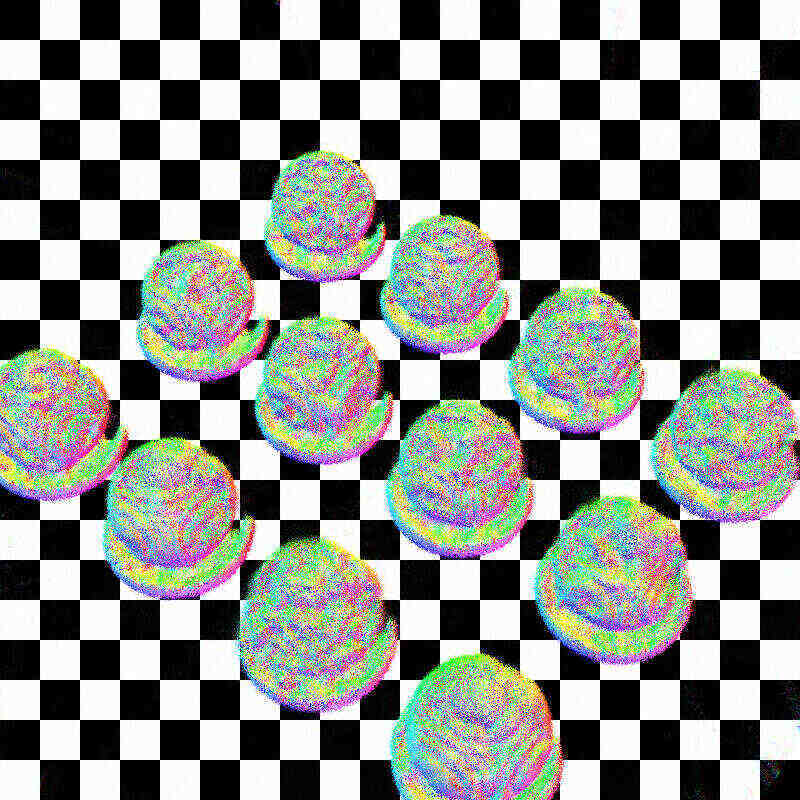}{b=2}
            \label{fig:qualitativeEvaluation:materialsSHBandCount2Normals.jpg}
            \imageGap{}%
        \end{subfigure}%
        \begin{subfigure}[t]{0.195\textwidth}
            \centering%
            \labelImage{\textwidth}{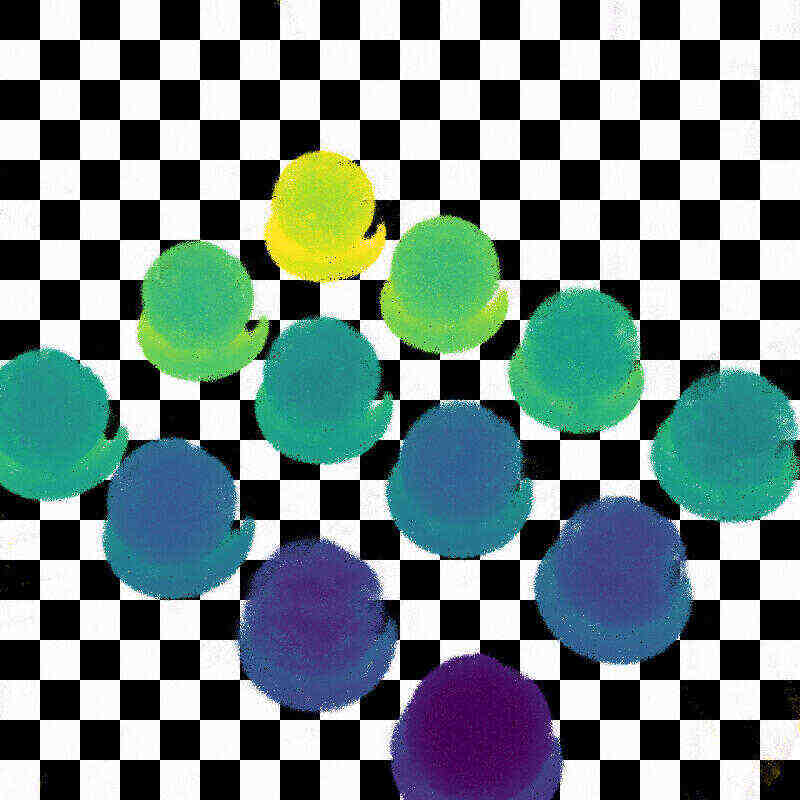}{b=2}
            \label{fig:qualitativeEvaluation:materialsSHBandCount2Depth.jpg}
            \imageGap{}%
        \end{subfigure} \\
        %
        %
        \begin{subfigure}[t]{0.195\textwidth}
            \centering%
            \labelImage{\textwidth}{Results/AblationStudy/SHBandCount/GroundTruth/MaterialsView300/Photo.jpg}{gt}
            \label{fig:qualitativeEvaluation:materialsSHBandCount3GroundTruth}
            \imageGap{}%
        \end{subfigure}%
        \begin{subfigure}[t]{0.195\textwidth}
            \centering%
            \labelImage{\textwidth}{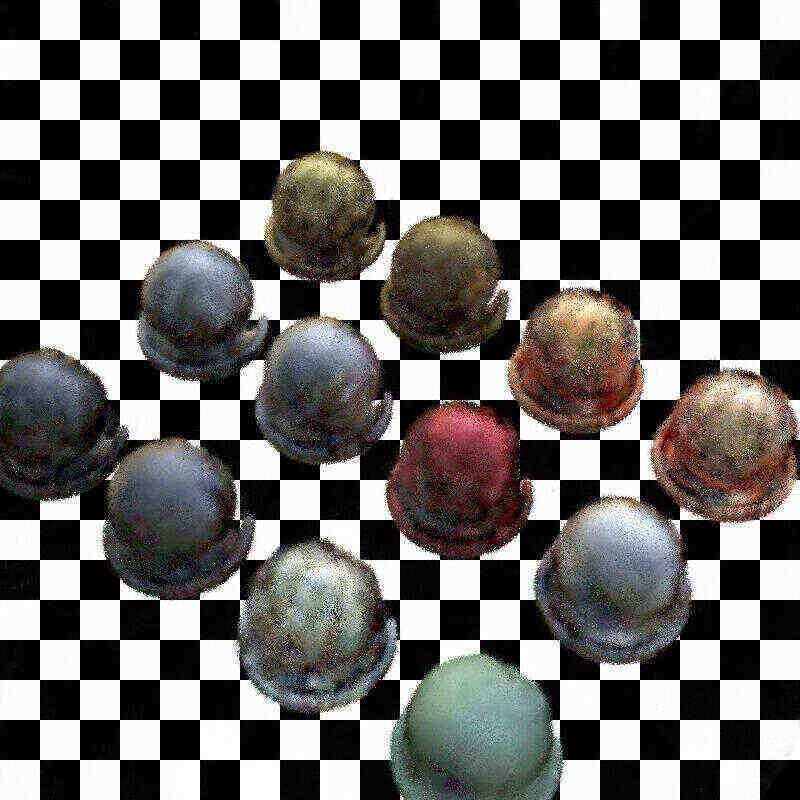}{b=3}
            \label{fig:qualitativeEvaluation:materialsSHBandCount3Photo.jpg}
            \imageGap{}%
        \end{subfigure}%
        \begin{subfigure}[t]{0.195\textwidth}
            \centering%
            \labelImage{\textwidth}{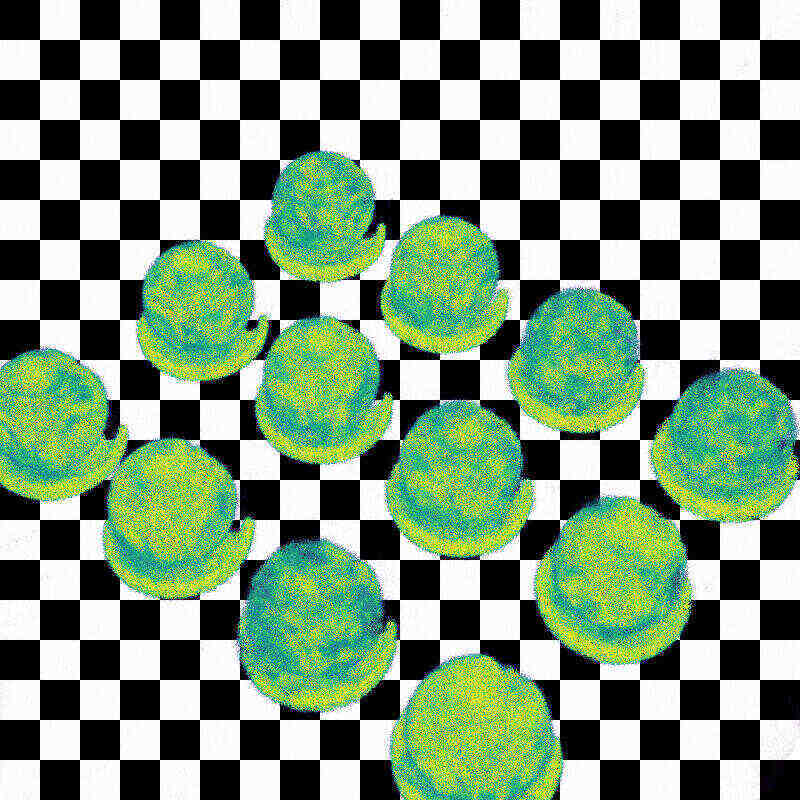}{b=3}
            \label{fig:qualitativeEvaluation:materialsSHBandCount3Density.jpg}
            \imageGap{}%
        \end{subfigure}%
        \begin{subfigure}[t]{0.195\textwidth}
            \centering%
            \normalMap{\textwidth}{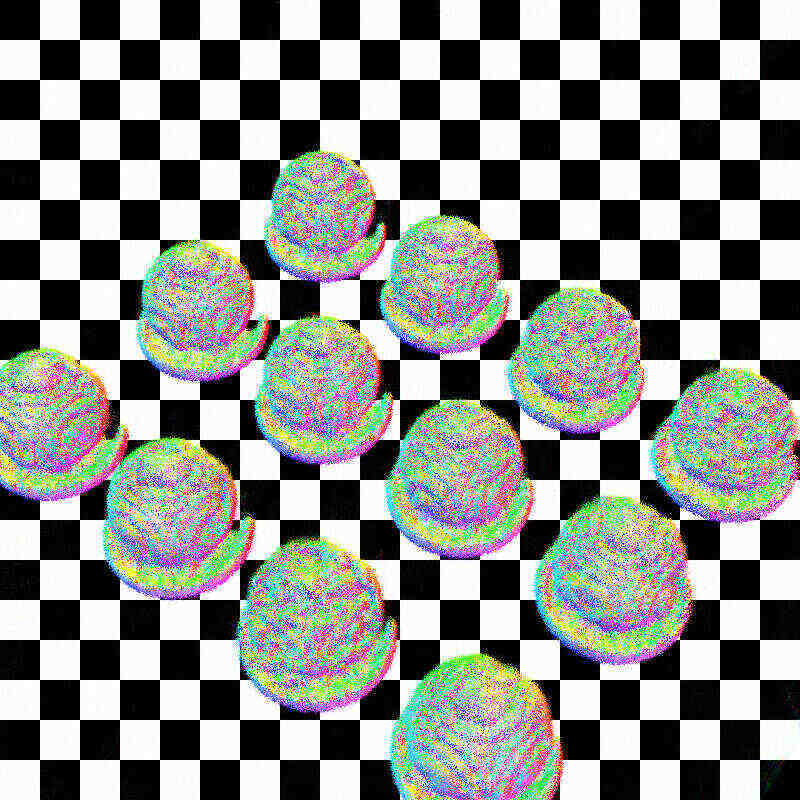}{b=3}
            \label{fig:qualitativeEvaluation:materialsSHBandCount3Normals.jpg}
            \imageGap{}%
        \end{subfigure}%
        \begin{subfigure}[t]{0.195\textwidth}
            \centering%
            \labelImage{\textwidth}{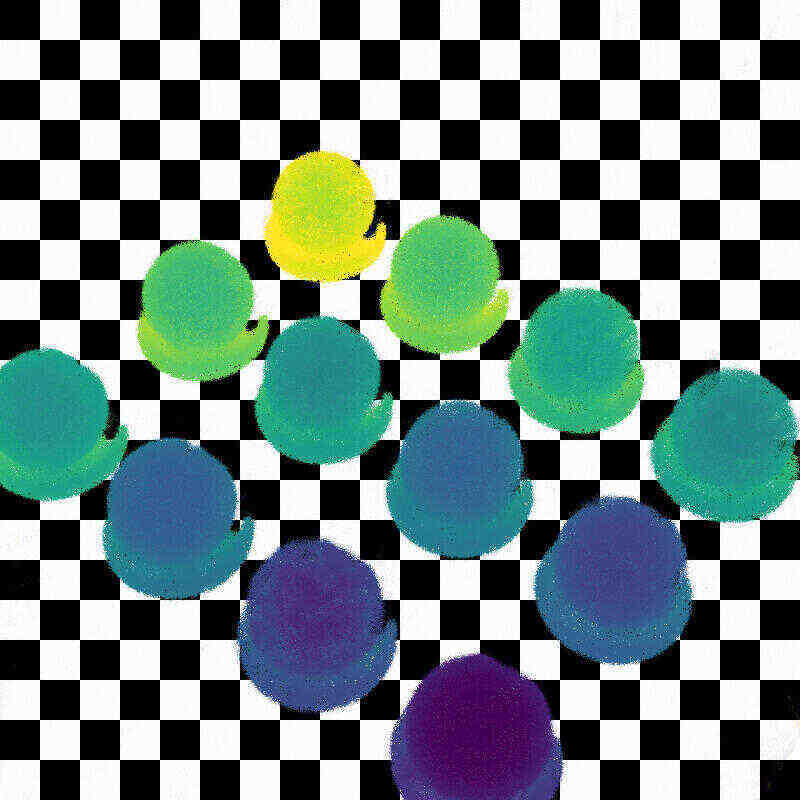}{b=3}
            \label{fig:qualitativeEvaluation:materialsSHBandCount3Depth.jpg}
            \imageGap{}%
        \end{subfigure} \\
        %
        %
        \begin{subfigure}[t]{0.195\textwidth}
            \centering%
            \labelImage{\textwidth}{Results/AblationStudy/SHBandCount/GroundTruth/MaterialsView300/Photo.jpg}{gt}
            \label{fig:qualitativeEvaluation:materialsSHBandCount4GroundTruth}
            \imageGap{}%
        \end{subfigure}%
        \begin{subfigure}[t]{0.195\textwidth}
            \centering%
            \labelImage{\textwidth}{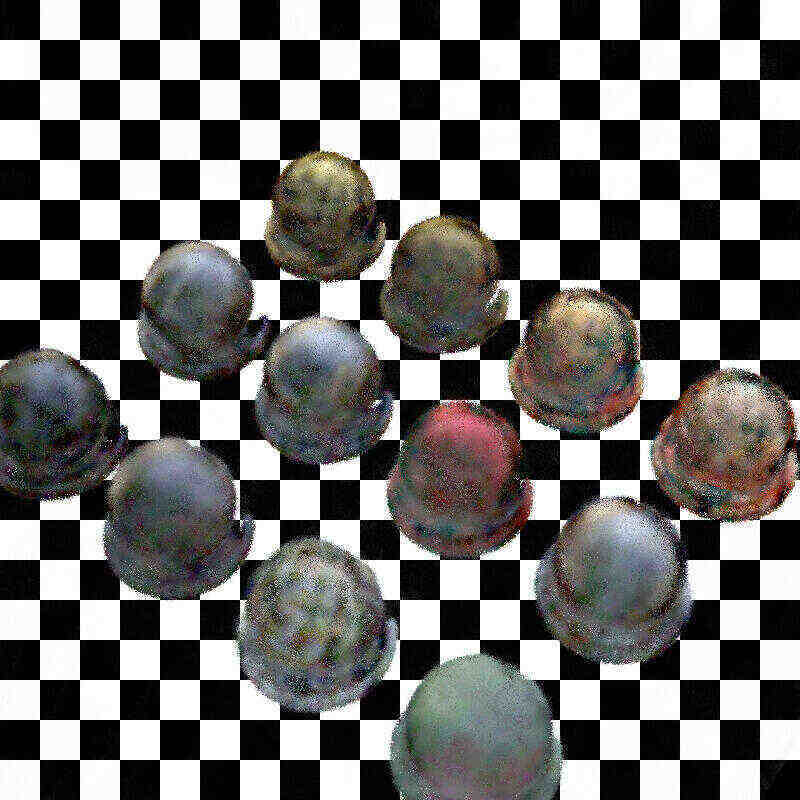}{b=4}
            \label{fig:qualitativeEvaluation:materialsSHBandCount4Photo.jpg}
            \imageGap{}%
        \end{subfigure}%
        \begin{subfigure}[t]{0.195\textwidth}
            \centering%
            \labelImage{\textwidth}{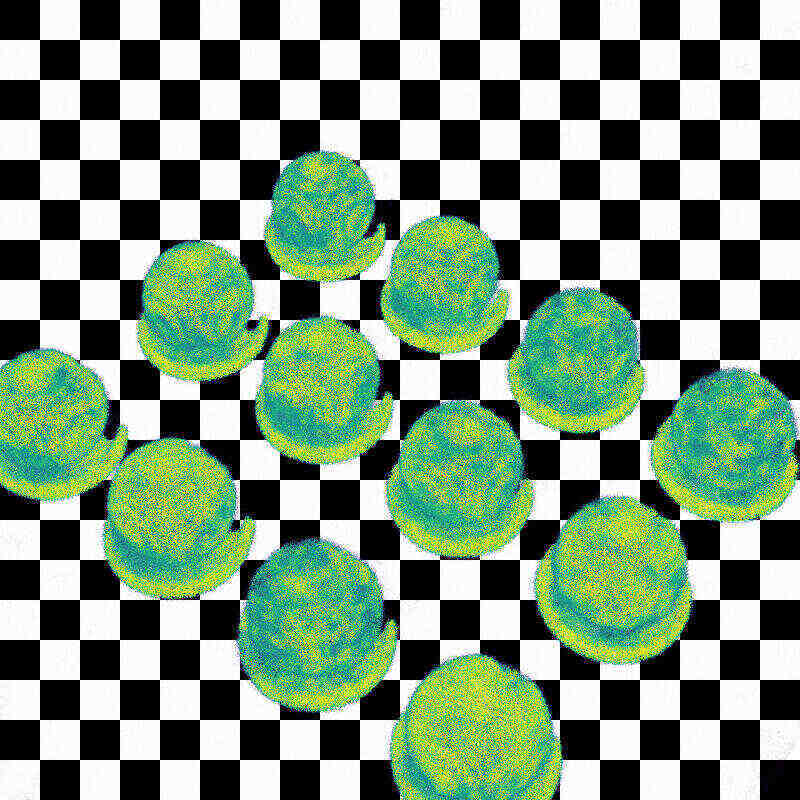}{b=4}
            \label{fig:qualitativeEvaluation:materialsSHBandCount4Density.jpg}
            \imageGap{}%
        \end{subfigure}%
        \begin{subfigure}[t]{0.195\textwidth}
            \centering%
            \normalMap{\textwidth}{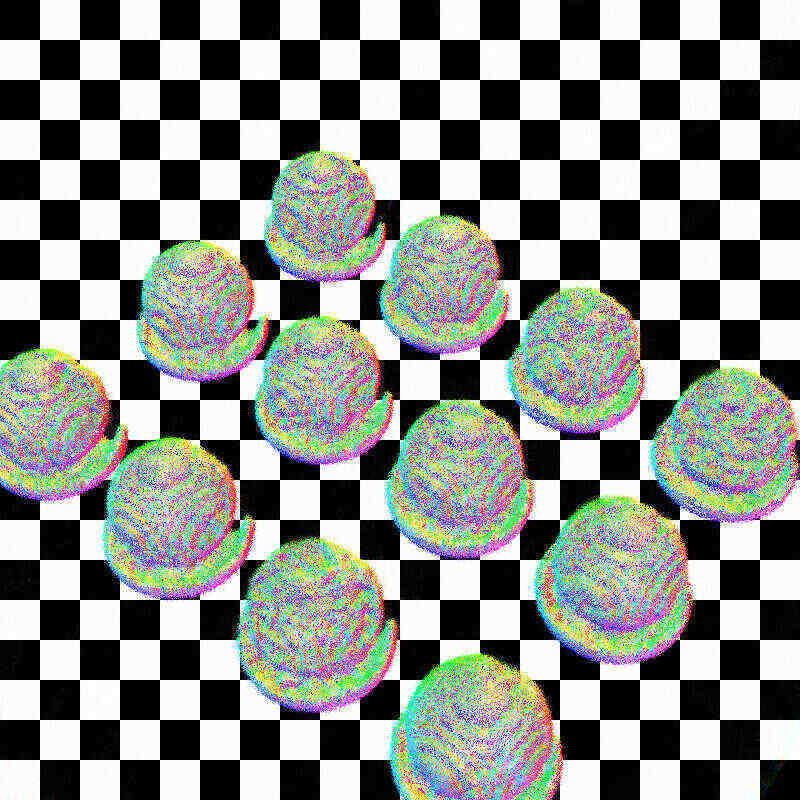}{b=4}
            \label{fig:qualitativeEvaluation:materialsSHBandCount4Normals.jpg}
            \imageGap{}%
        \end{subfigure}%
        \begin{subfigure}[t]{0.195\textwidth}
            \centering%
            \labelImage{\textwidth}{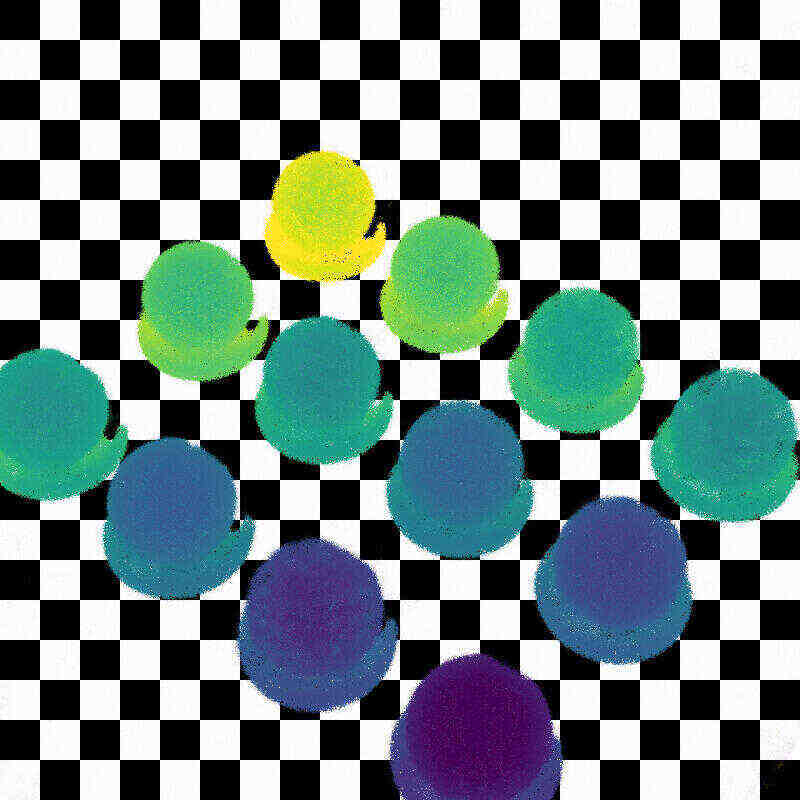}{b=4}
            \label{fig:qualitativeEvaluation:materialsSHBandCount4Depth.jpg}
            \imageGap{}%
        \end{subfigure}
    \end{tabular}
    %
    %
    \caption%
    {%
        \SH{} band count comparison
        with reconstructed models after 30k mini batch iterations
        on the synthetic materials scene \cite{mildenhall2020NeRF}
        (from left to right):
        hold-out ground truth image (gt),
        photo-realistic reconstruction,
        opacity,
        surface normals and
        depth map.
    }%
    \label{fig:SHBandCountComparison}
\end{figure*}

%
%
\Fig{\ref{fig:SHBandCountComparison}} shows differences in reconstruction quality
for varying numbers of \SH{} bands
to represent outgoing surface radiance.
Increasing the \SH{} band count first increases reconstruction quality.
However, the quality decreases again starting with 4 bands.
We presume that optimization convergence decreases
due to the higher degree of the underlying polynomials
which are not localized in angular space.
The frequency-based design of the \glspl{SH}
presumably compounds optimization
as each drawn radiance sample gradient influences all \SH{} coefficients.

%
%

\begin{figure*}%
    \centering%
    \begin{tabular}{l}
        %
        %
        \begin{subfigure}[t]{0.245\textwidth}
            \centering%
            \labelImage{\textwidth}{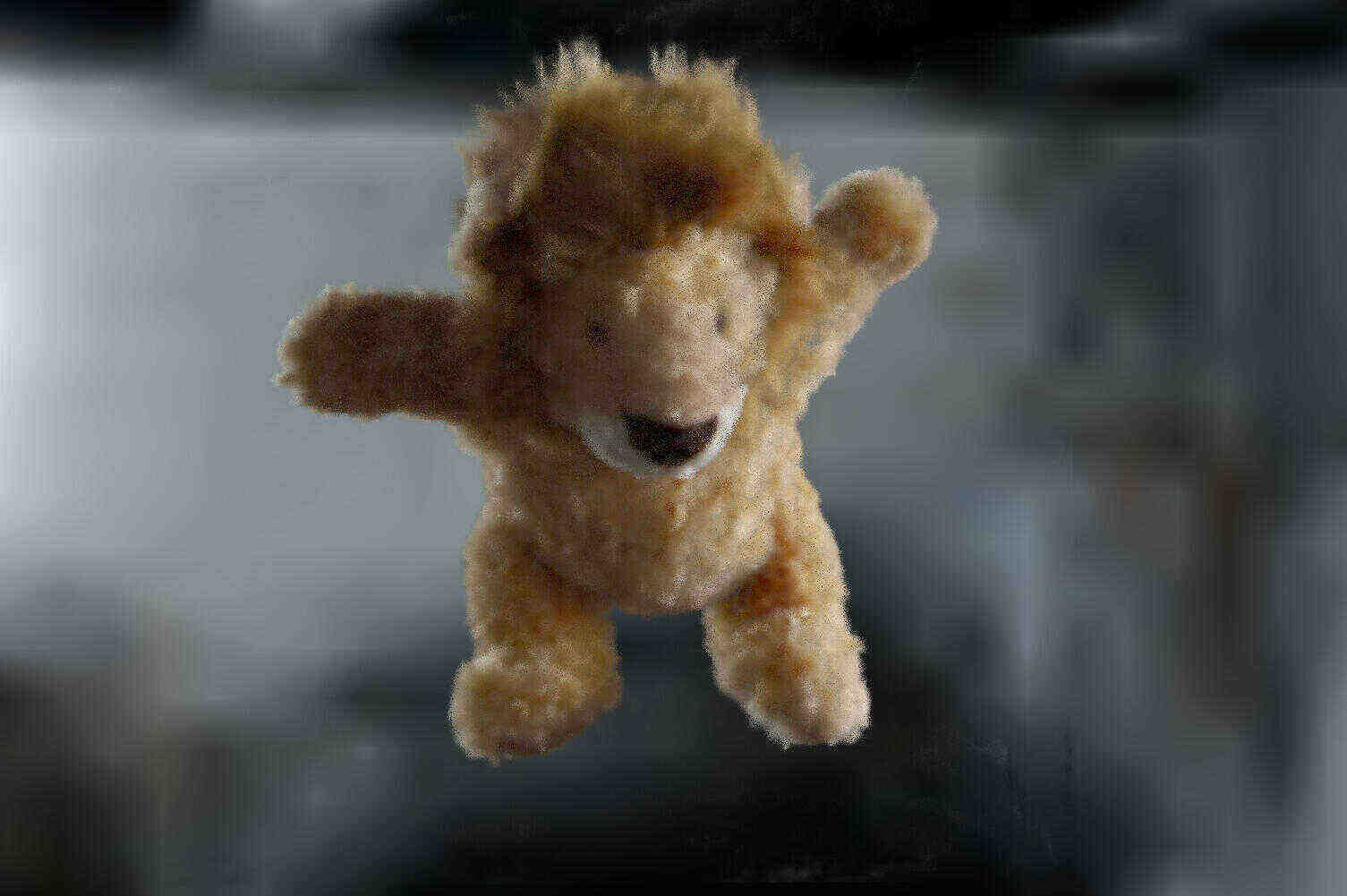}{$\lambda=0.1$}
            \imageGap{}%
        \end{subfigure}%
        \begin{subfigure}[t]{0.245\textwidth}
            \centering%
            \labelImage{\textwidth}{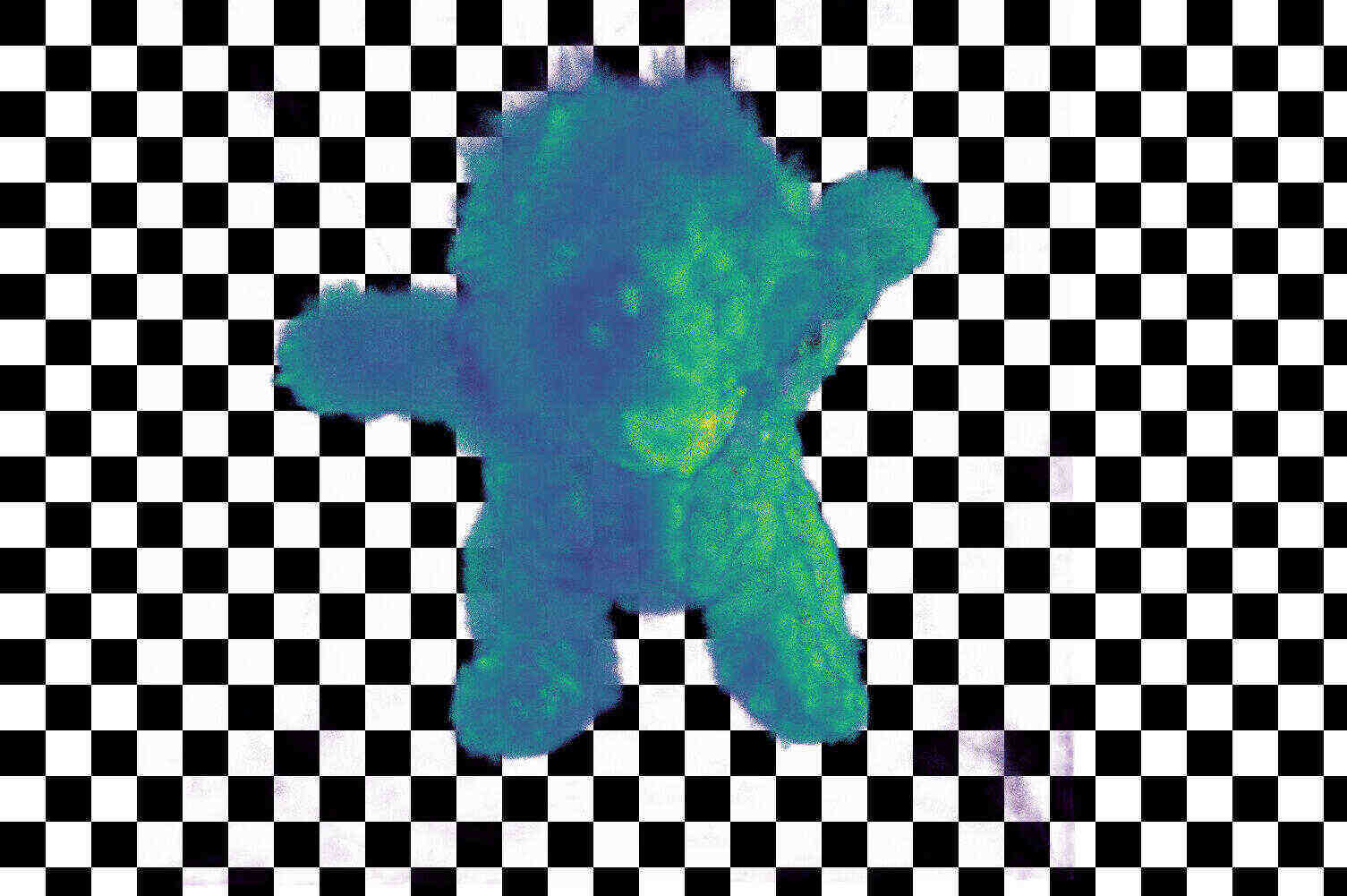}{$\lambda=0.1$}
            \imageGap{}%
        \end{subfigure}%
        \begin{subfigure}[t]{0.245\textwidth}
            \centering%
            \normalMap{\textwidth}{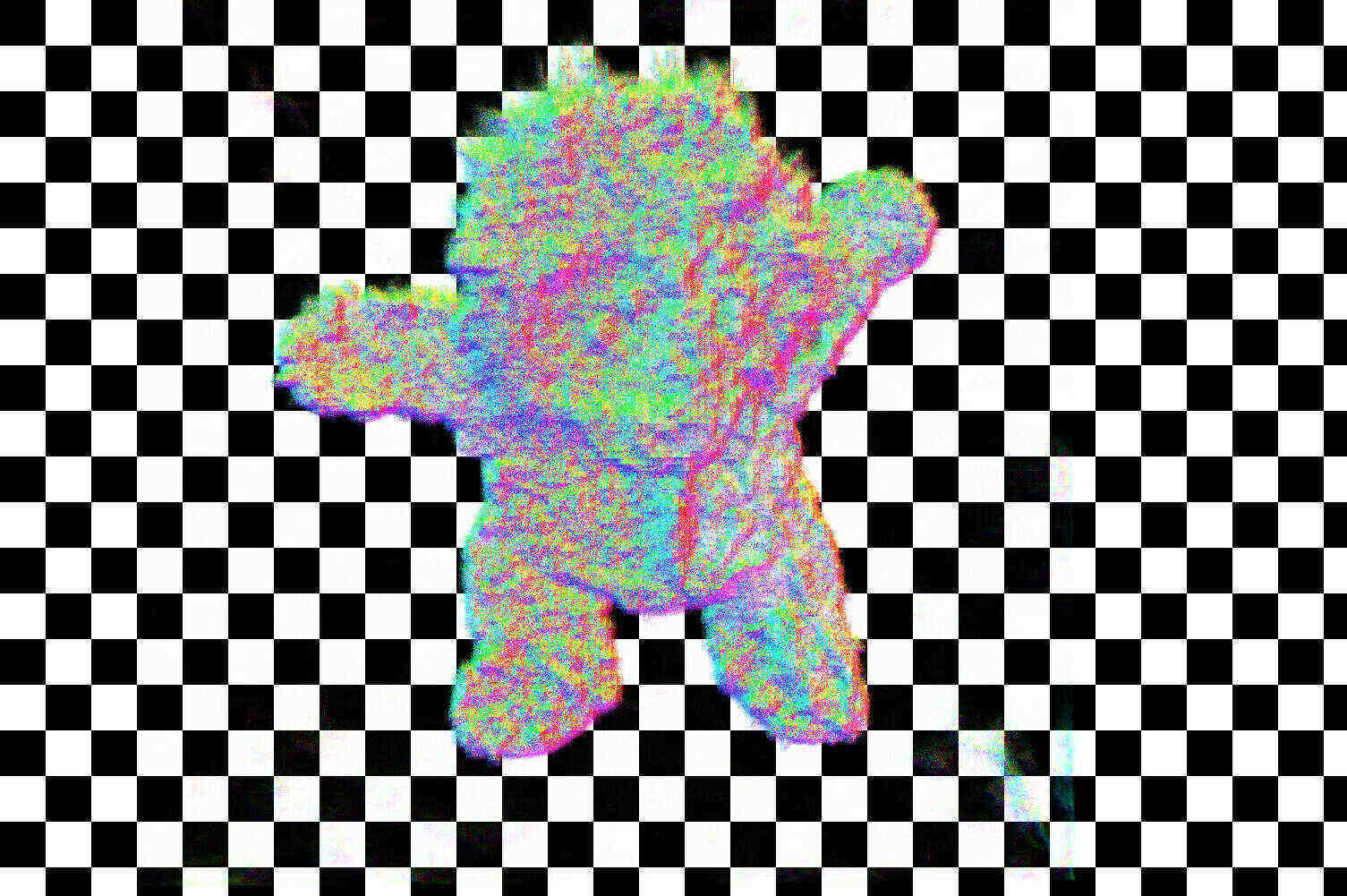}{$\lambda=0.1$}
            \imageGap{}%
        \end{subfigure}%
        \begin{subfigure}[t]{0.245\textwidth}
            \centering%
            \labelImage{\textwidth}{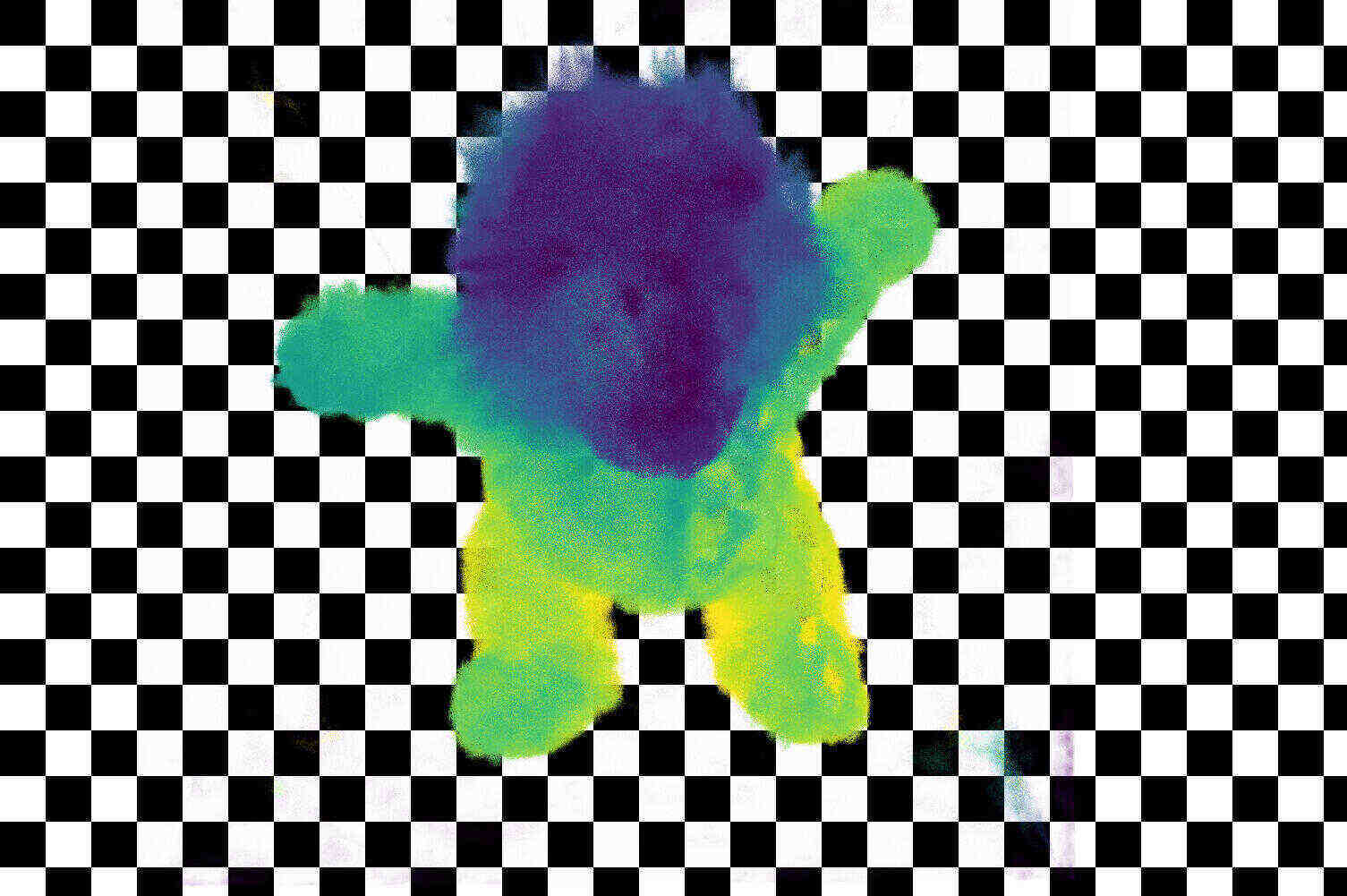}{$\lambda=0.1$}
            \imageGap{}%
        \end{subfigure} \\
        %
        %
        \begin{subfigure}[t]{0.245\textwidth}
            \centering%
            \labelImage{\textwidth}{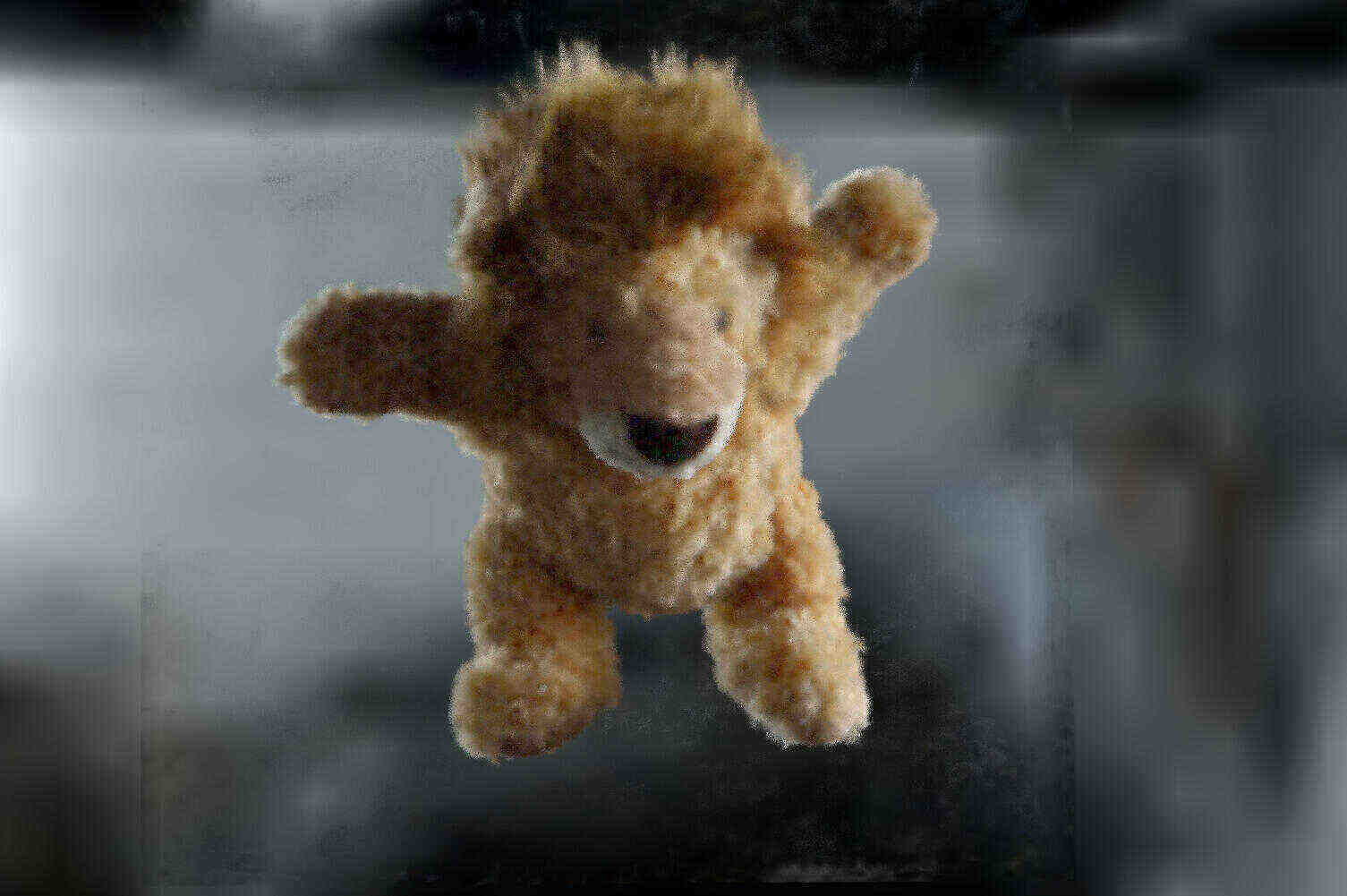}{$\lambda=0.01$}
            \imageGap{}%
        \end{subfigure}%
        \begin{subfigure}[t]{0.245\textwidth}
            \centering%
            \labelImage{\textwidth}{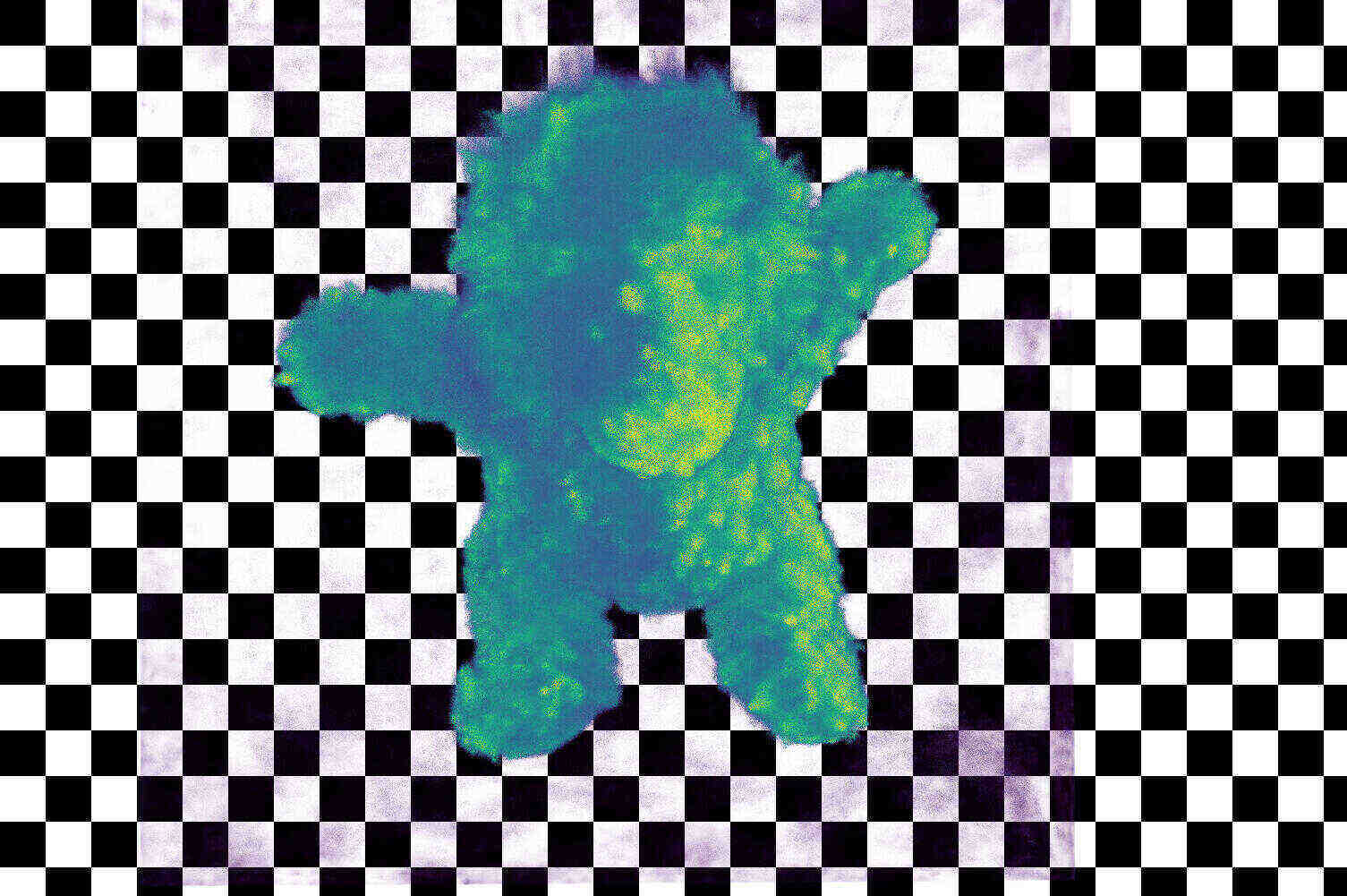}{$\lambda=0.01$}
            \imageGap{}%
        \end{subfigure}%
        \begin{subfigure}[t]{0.245\textwidth}
            \centering%
            \normalMap{\textwidth}{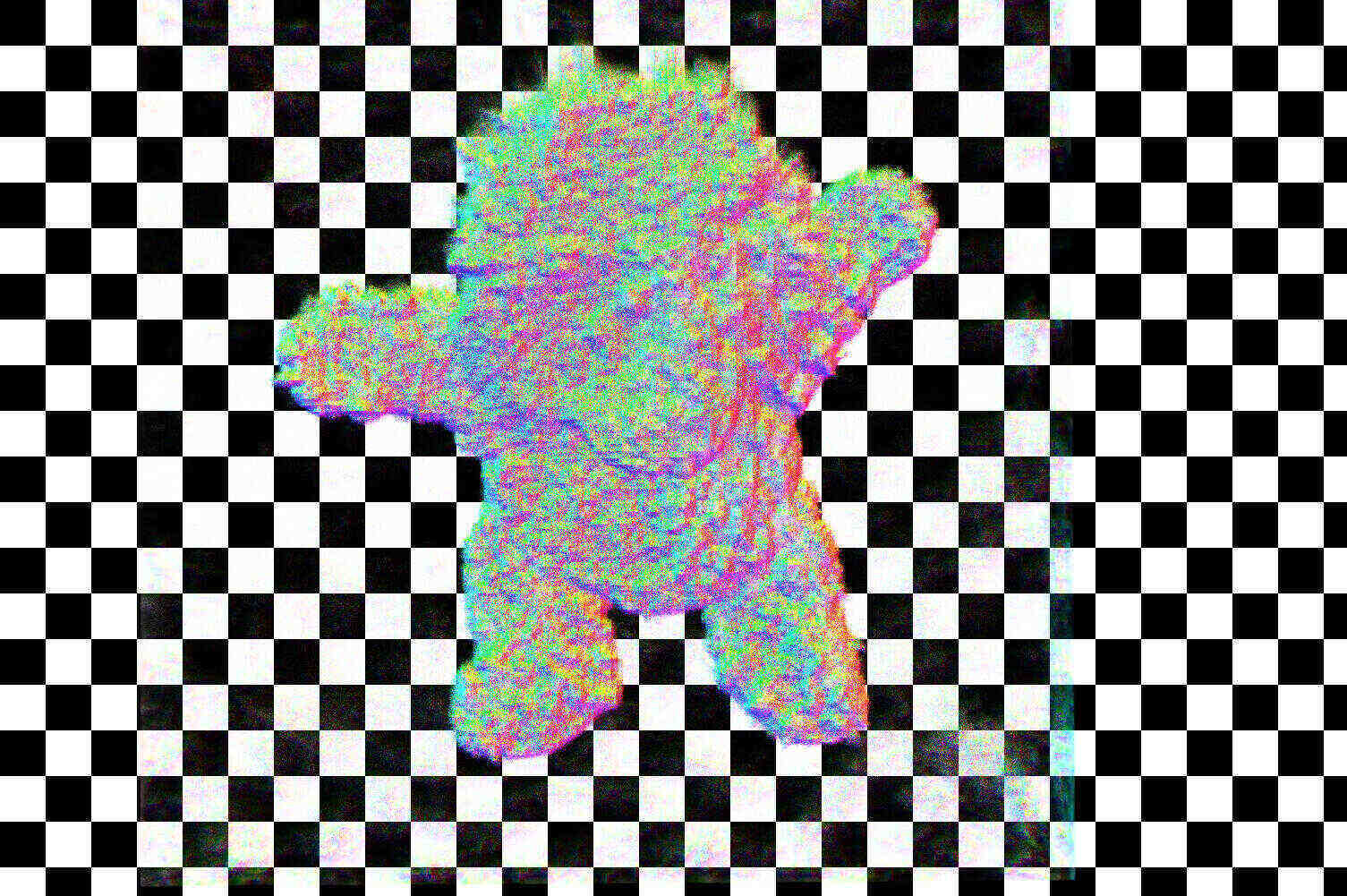}{$\lambda=0.01$}
            \imageGap{}%
        \end{subfigure}%
        \begin{subfigure}[t]{0.245\textwidth}
            \centering%
            \labelImage{\textwidth}{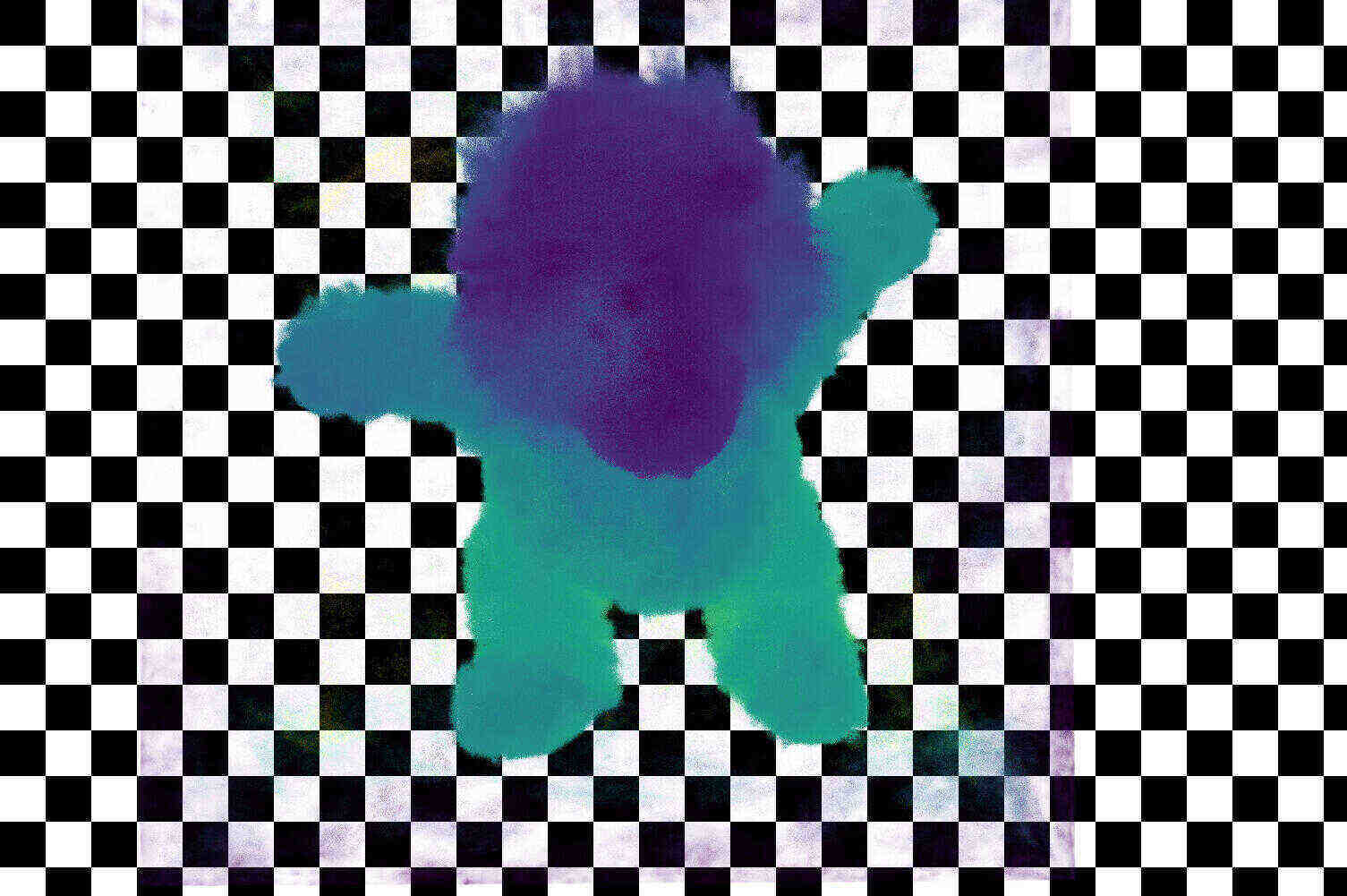}{$\lambda=0.01$}
            \imageGap{}%
        \end{subfigure} \\
        %
        %
        \begin{subfigure}[t]{0.245\textwidth}
            \centering%
            \labelImage{\textwidth}{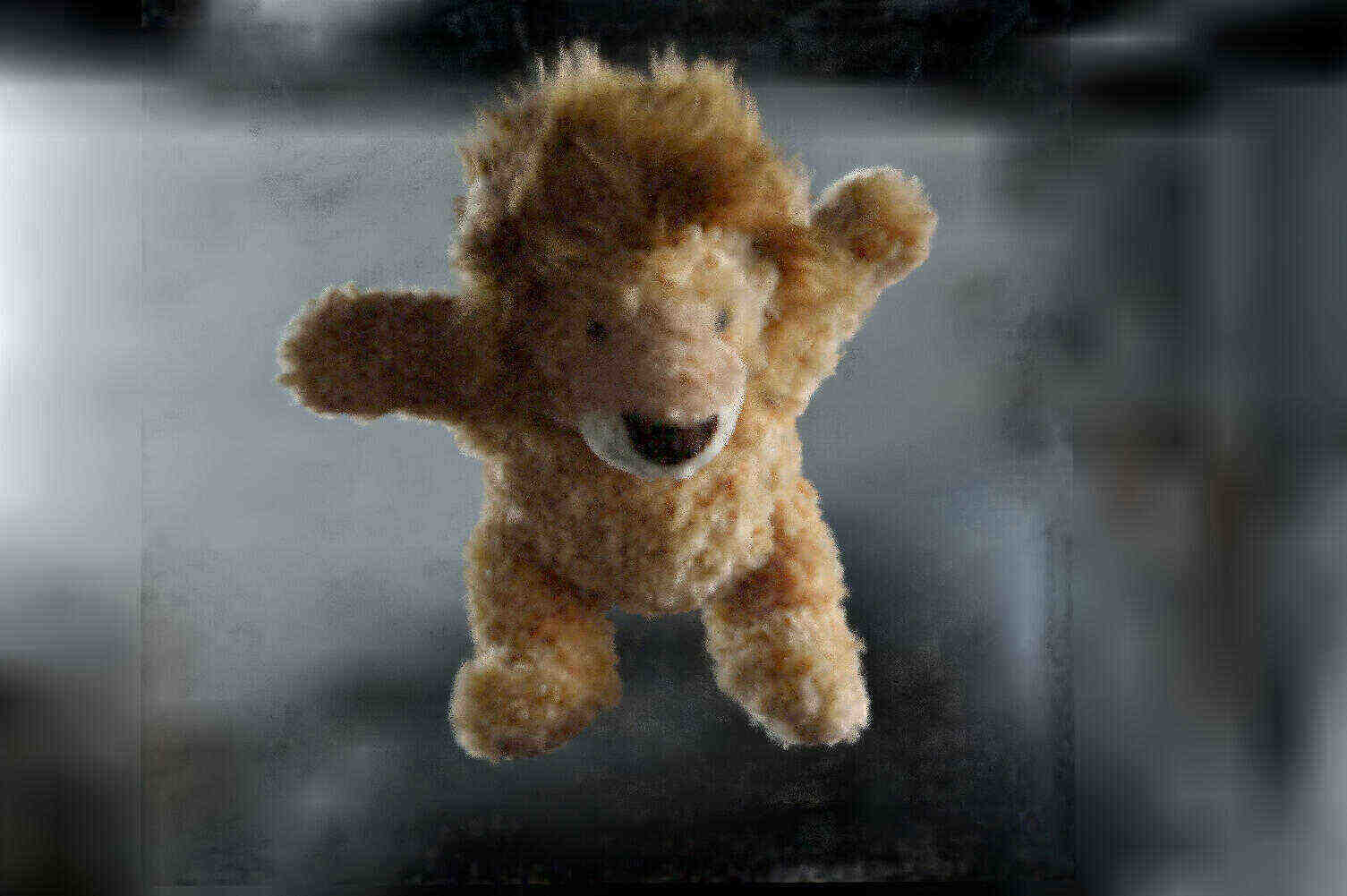}{$\lambda=1e-3$}
            \imageGap{}%
        \end{subfigure}%
        \begin{subfigure}[t]{0.245\textwidth}
            \centering%
            \labelImage{\textwidth}{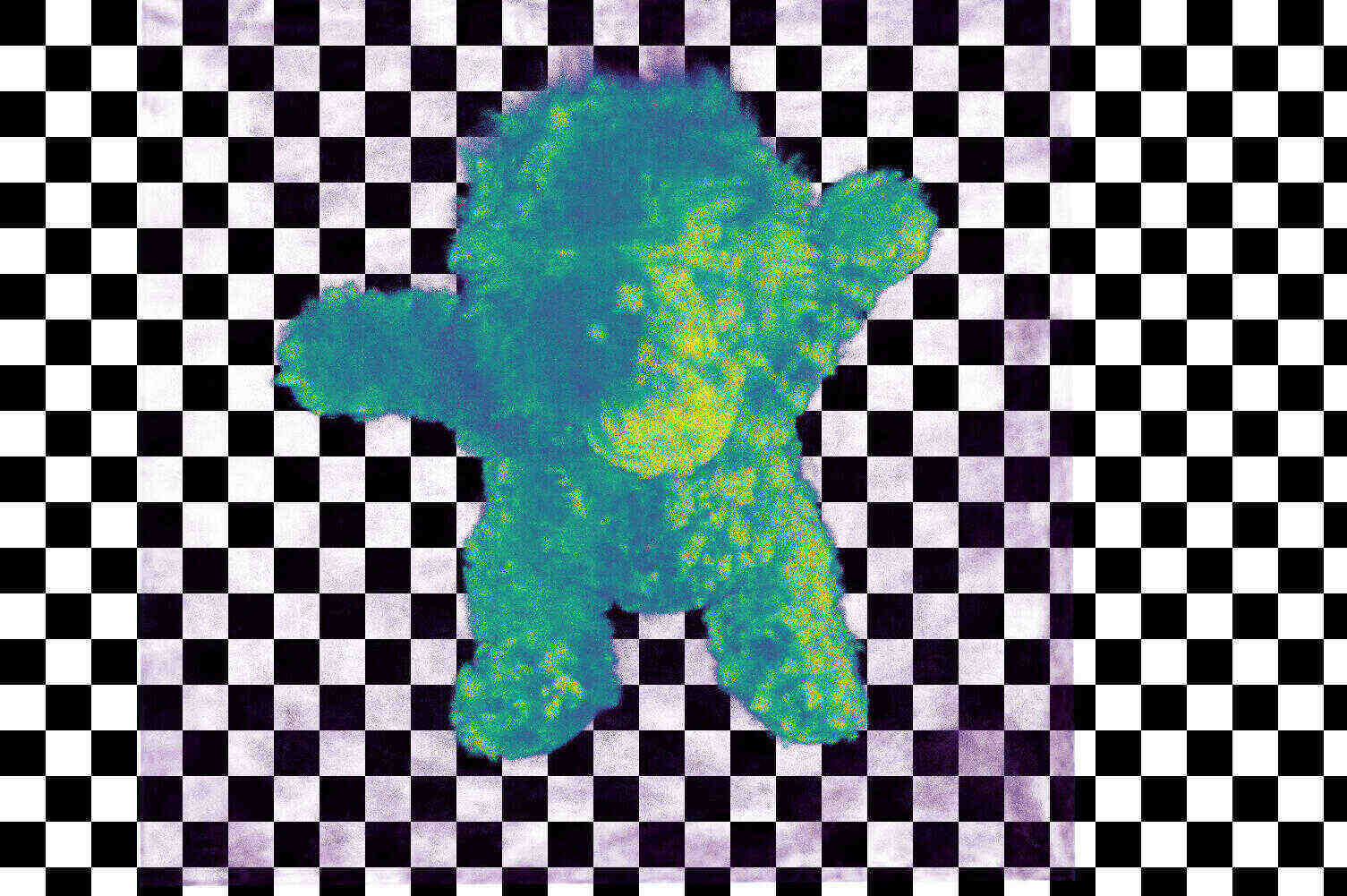}{$\lambda=1e-3$}
            \imageGap{}%
        \end{subfigure}%
        \begin{subfigure}[t]{0.245\textwidth}
            \centering%
            \normalMap{\textwidth}{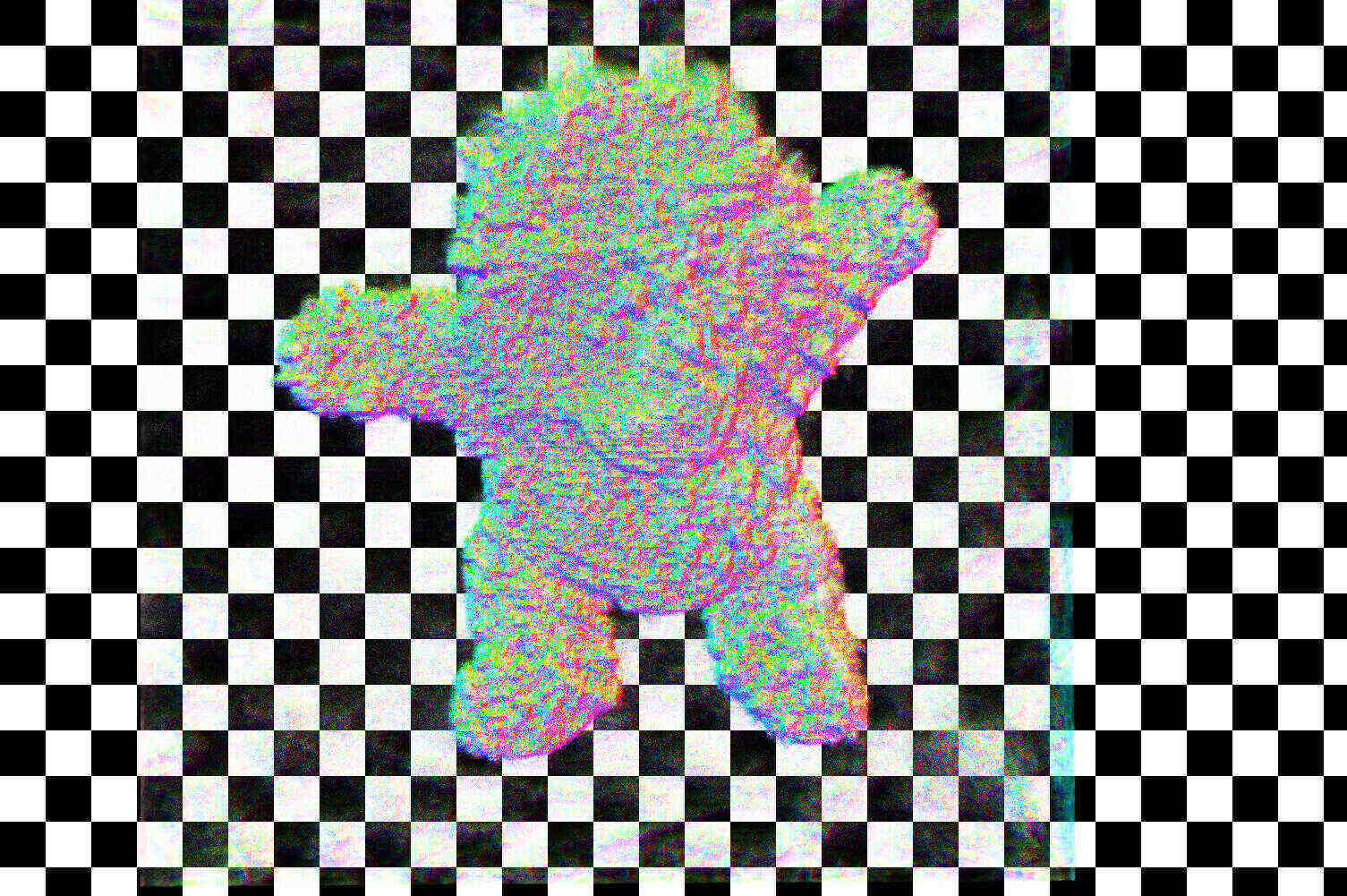}{$\lambda=1e-3$}
            \imageGap{}%
        \end{subfigure}%
        \begin{subfigure}[t]{0.245\textwidth}
            \centering%
            \labelImage{\textwidth}{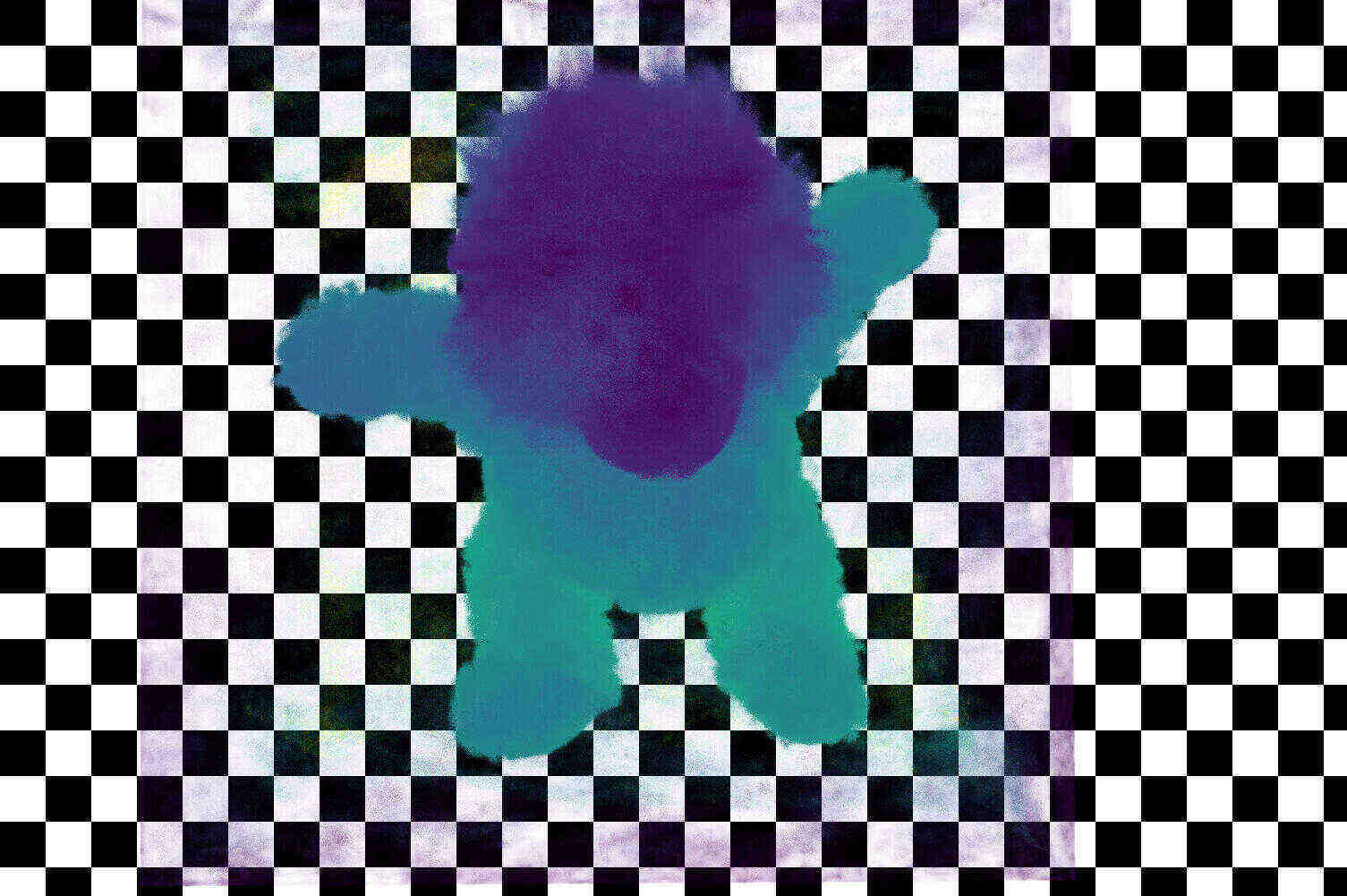}{$\lambda=1e-3$}
            \imageGap{}%
        \end{subfigure} \\
        %
        %
        \begin{subfigure}[t]{0.245\textwidth}
            \centering%
            \labelImage{\textwidth}{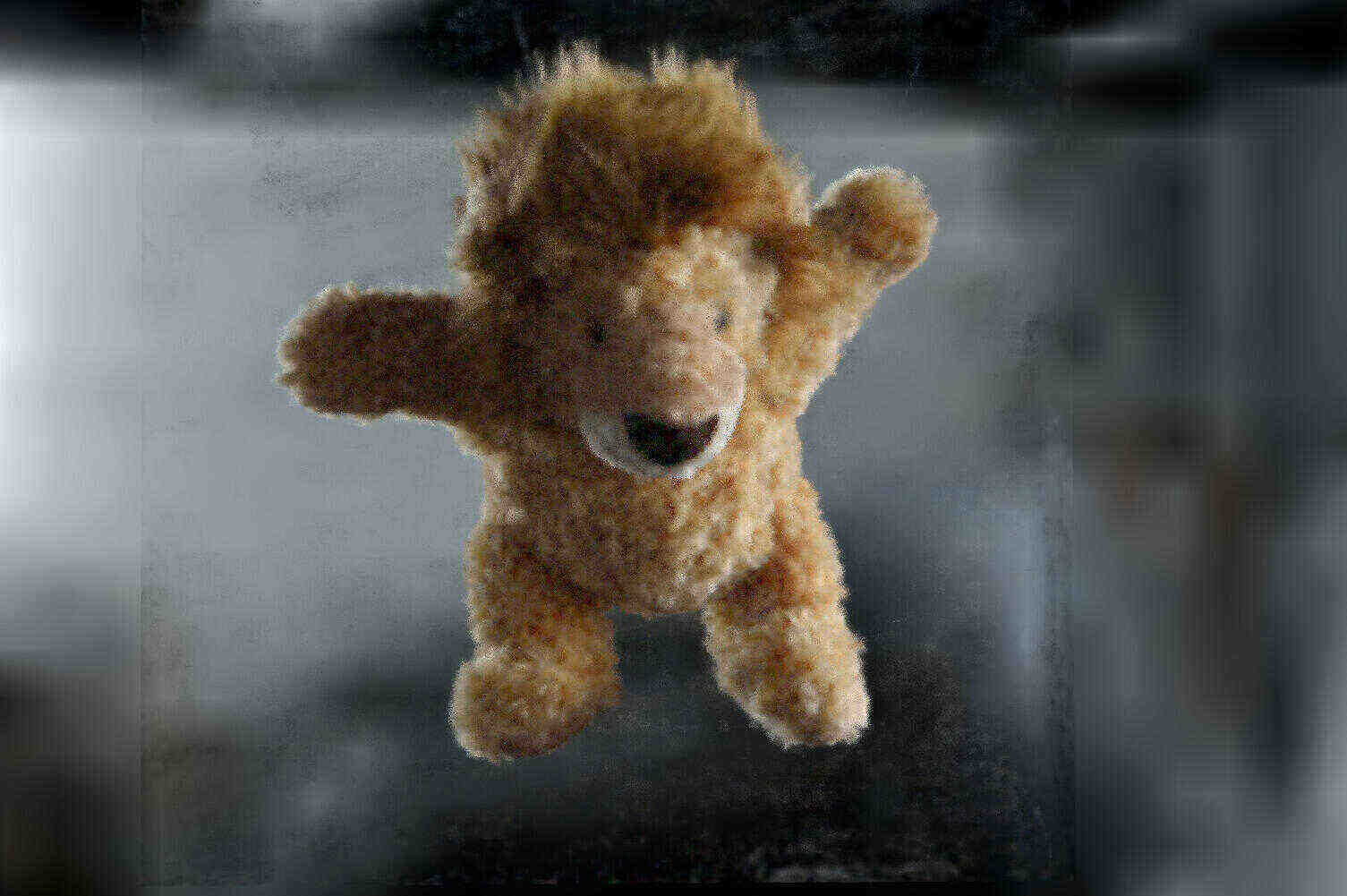}{$\lambda=1e-4$}
            \imageGap{}%
        \end{subfigure}%
        \begin{subfigure}[t]{0.245\textwidth}
            \centering%
            \labelImage{\textwidth}{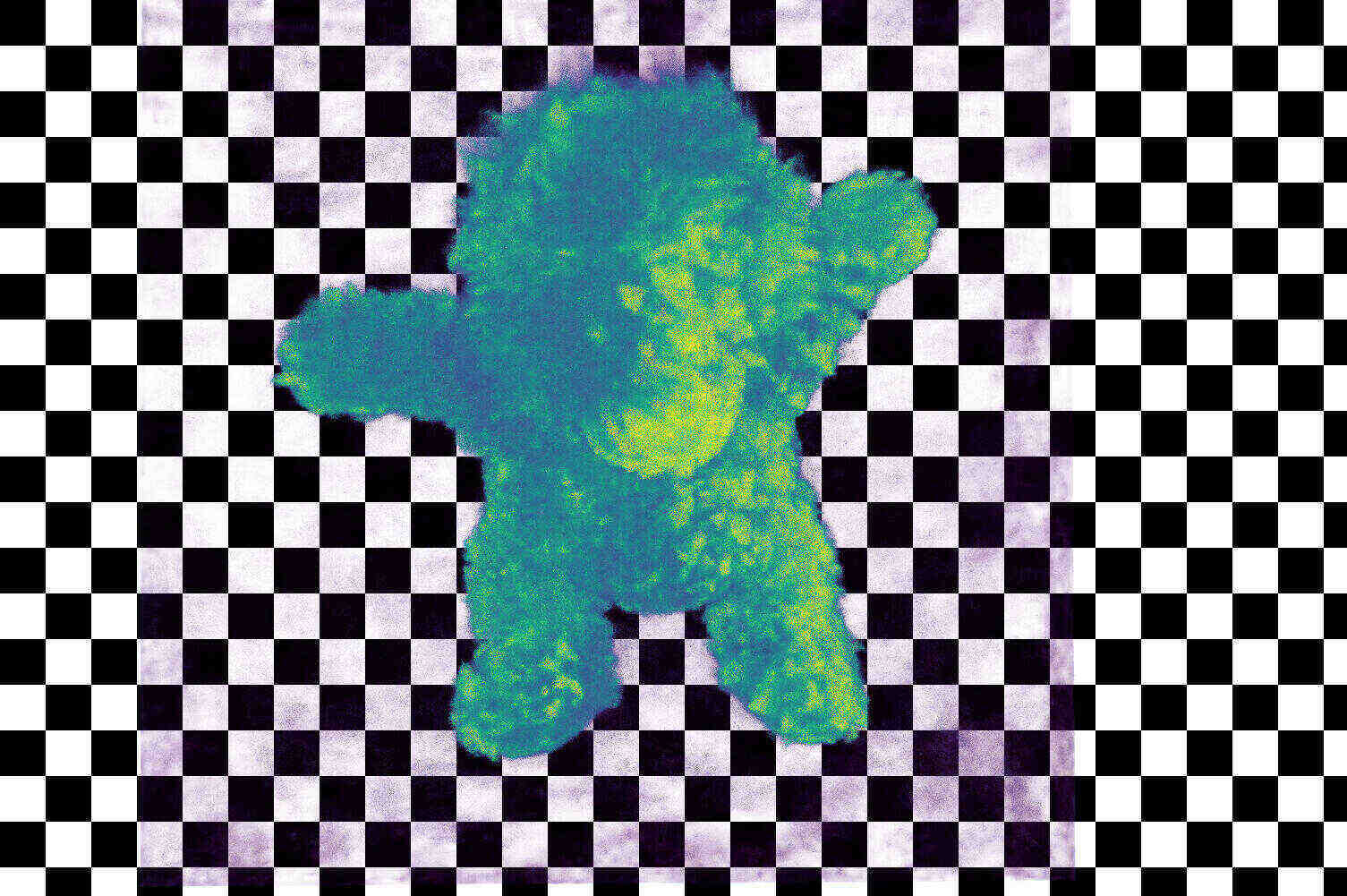}{$\lambda=1e-4$}
            \imageGap{}%
        \end{subfigure}%
        \begin{subfigure}[t]{0.245\textwidth}
            \centering%
            \normalMap{\textwidth}{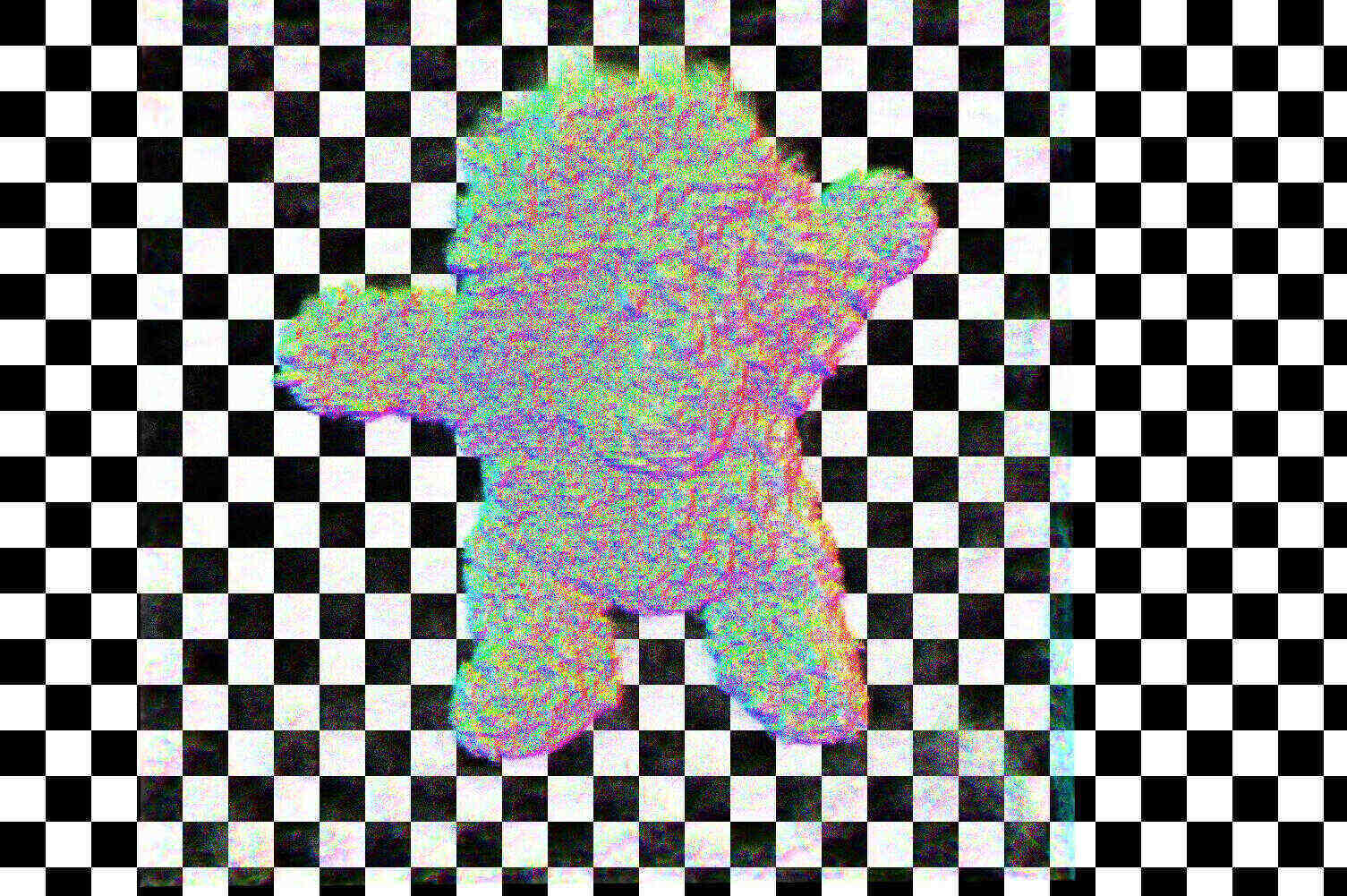}{$\lambda=1e-4$}
            \imageGap{}%
        \end{subfigure}%
        \begin{subfigure}[t]{0.245\textwidth}
            \centering%
            \labelImage{\textwidth}{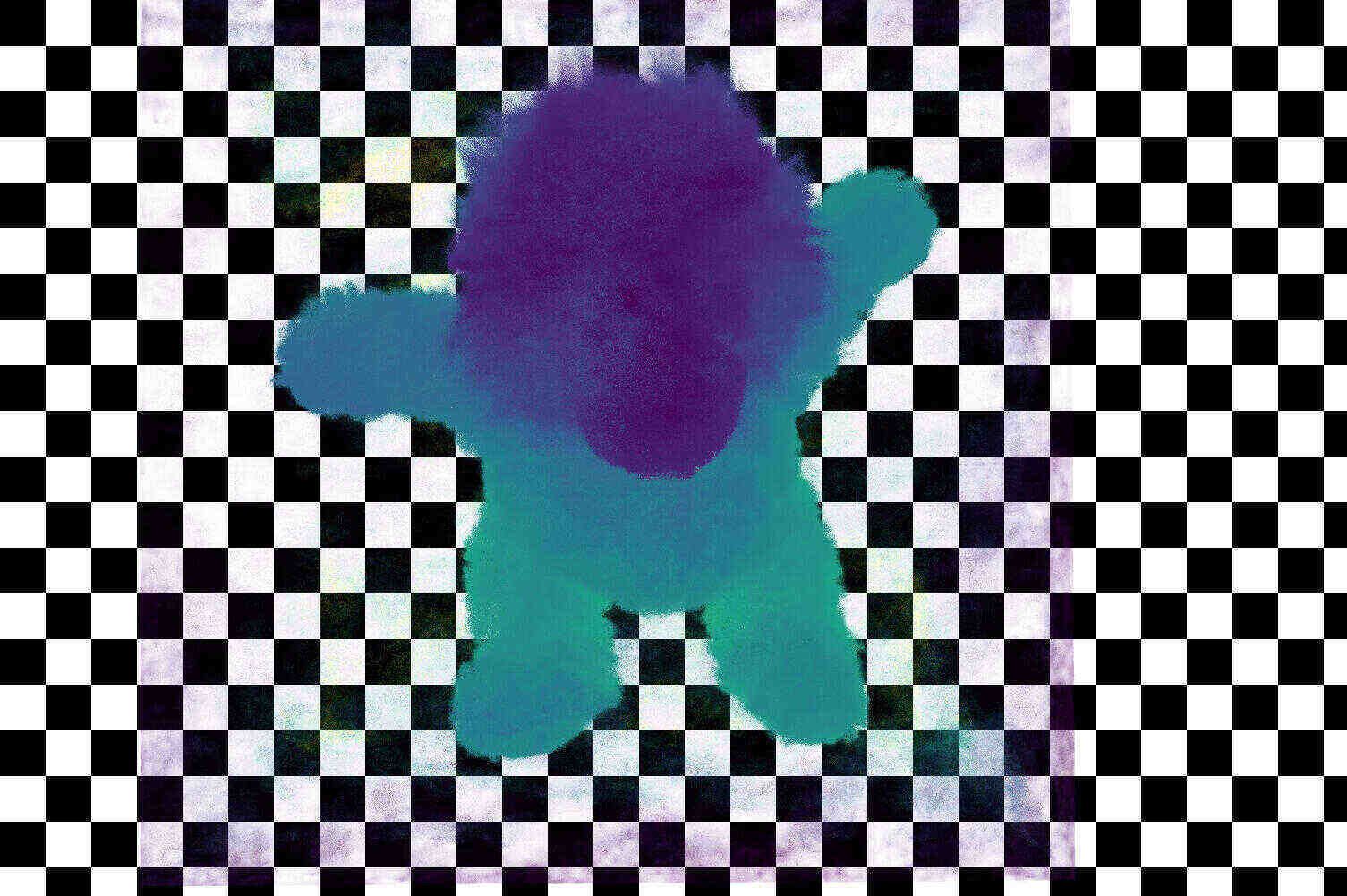}{$\lambda=1e-4$}
            \imageGap{}%
        \end{subfigure}%
    \end{tabular}
    %
    %
    \caption%
    {%
        Varying prior strengths experiment showing Lion reconstructions
        after 50k iterations with decreasing $lambda$ factors and
        thus also decreasing model smoothness and increasing free space clutter
        from top to bottom.
    }%
    \label{fig:priorStrengths}
\end{figure*}

%
%
\Fig{\ref{fig:priorStrengths}} compares reconstruction results for varying prior strengths.
They induce overly smooth results for $\lambda=0.1$ and $\lambda=0.01$
and have almost no impact for $\lambda=1e-4$.
Note that
the zero opacity prior also reduces the initial clutter in free space more
for higher $\lambda$ values.
Thus, we generally set $\lambda=1e-3$ for our experiments.

%
%

\begin{figure*}%
    \centering%
    \begin{tabular}{l}
        %
        %
        \begin{subfigure}[t]{0.245\textwidth}
            \centering%
            \labelImage{\textwidth}{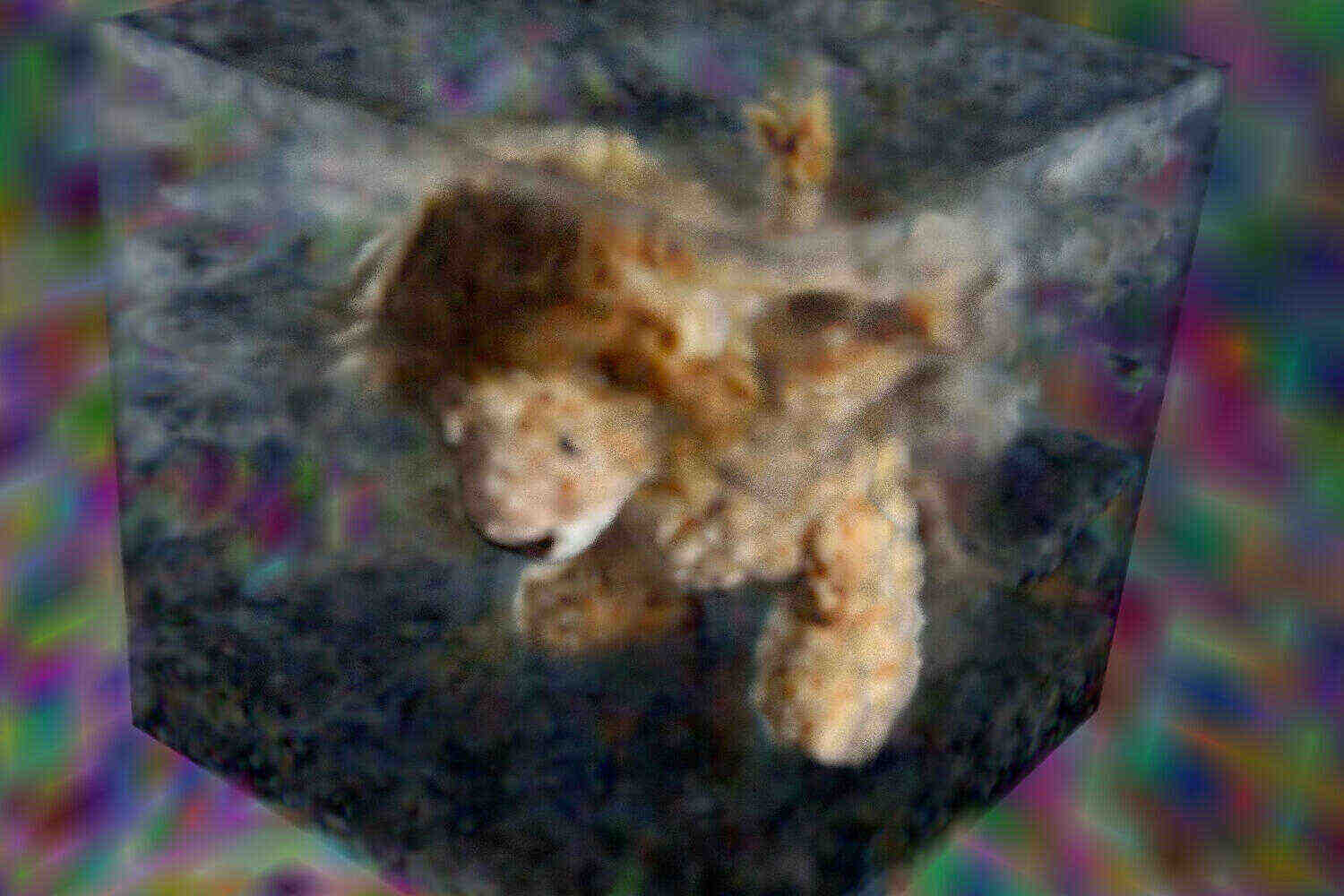}{2.5k it., 16/512/128}%
            \imageGap{}%
        \end{subfigure}%
        \begin{subfigure}[t]{0.245\textwidth}
            \centering%
            \labelImage{\textwidth}{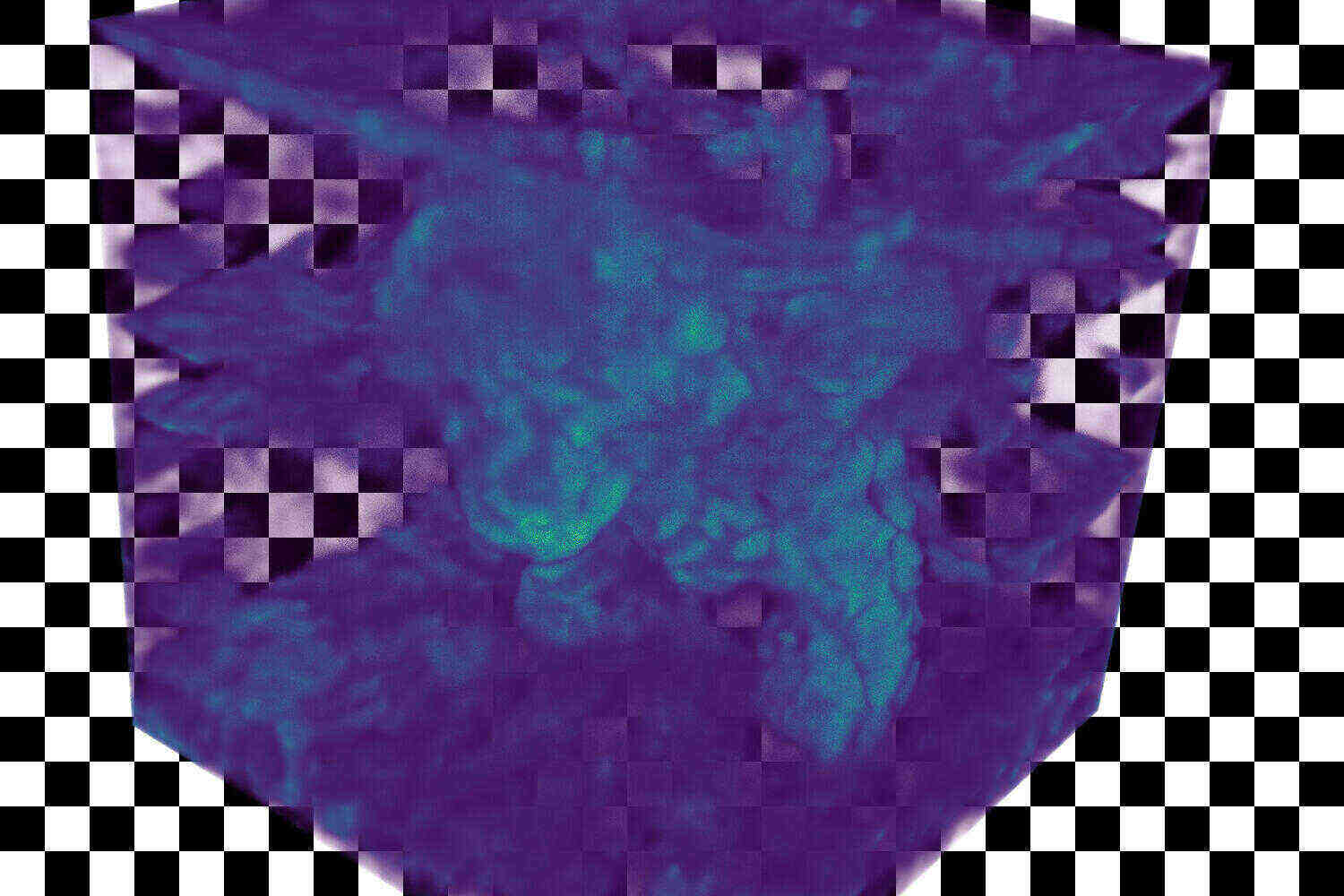}{2.5k it., 16/512/128}%
            \imageGap{}%
        \end{subfigure}%
        \begin{subfigure}[t]{0.245\textwidth}
            \centering%
            \normalMap{\textwidth}{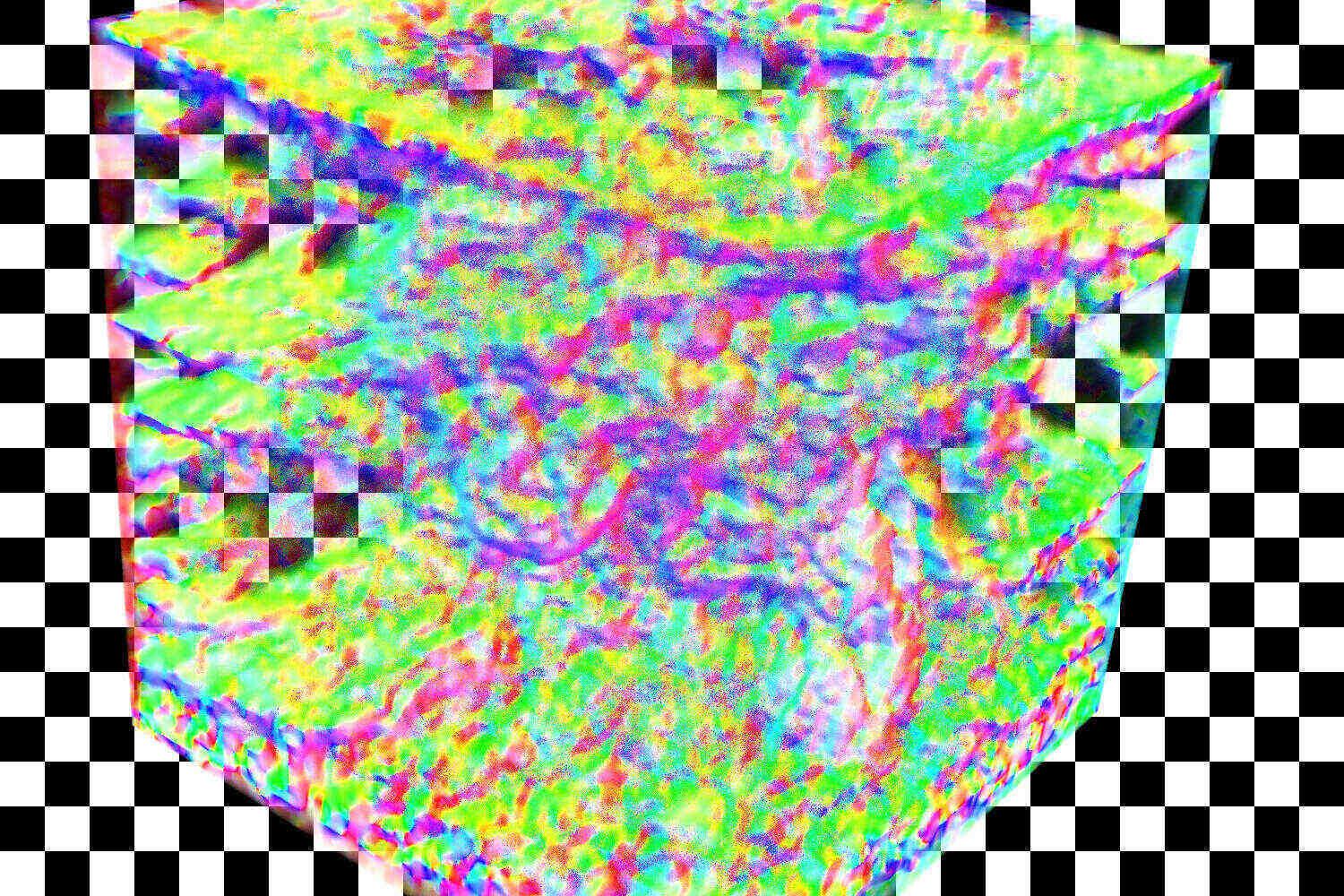}{2.5k it., 16/512/128}%
            \imageGap{}%
        \end{subfigure}%
        \begin{subfigure}[t]{0.245\textwidth}
            \centering%
            \labelImage{\textwidth}{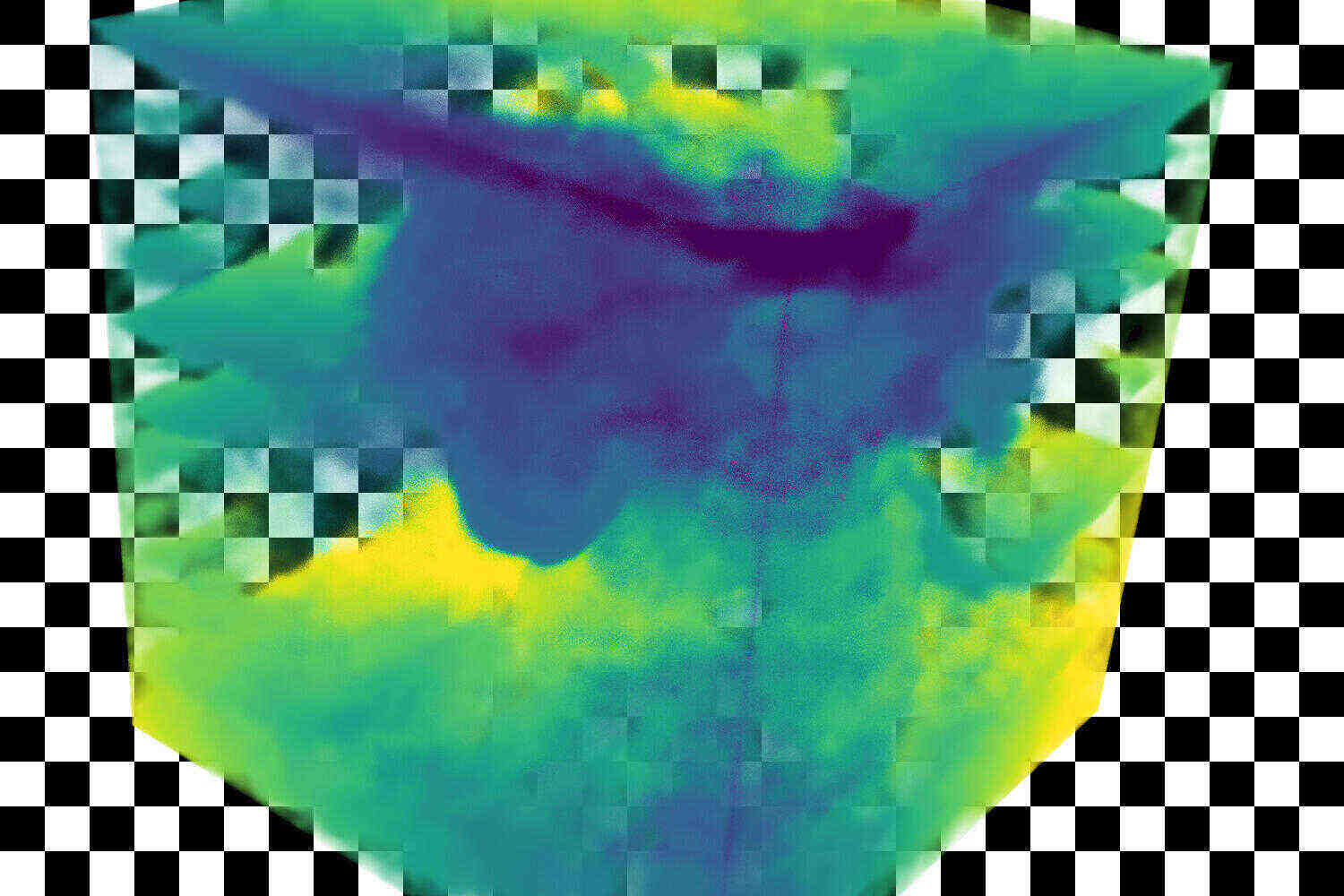}{2.5k it., 16/512/128}%
            \imageGap{}%
        \end{subfigure} \\
        %
        %
        \begin{subfigure}[t]{0.245\textwidth}
            \centering%
            \labelImage{\textwidth}{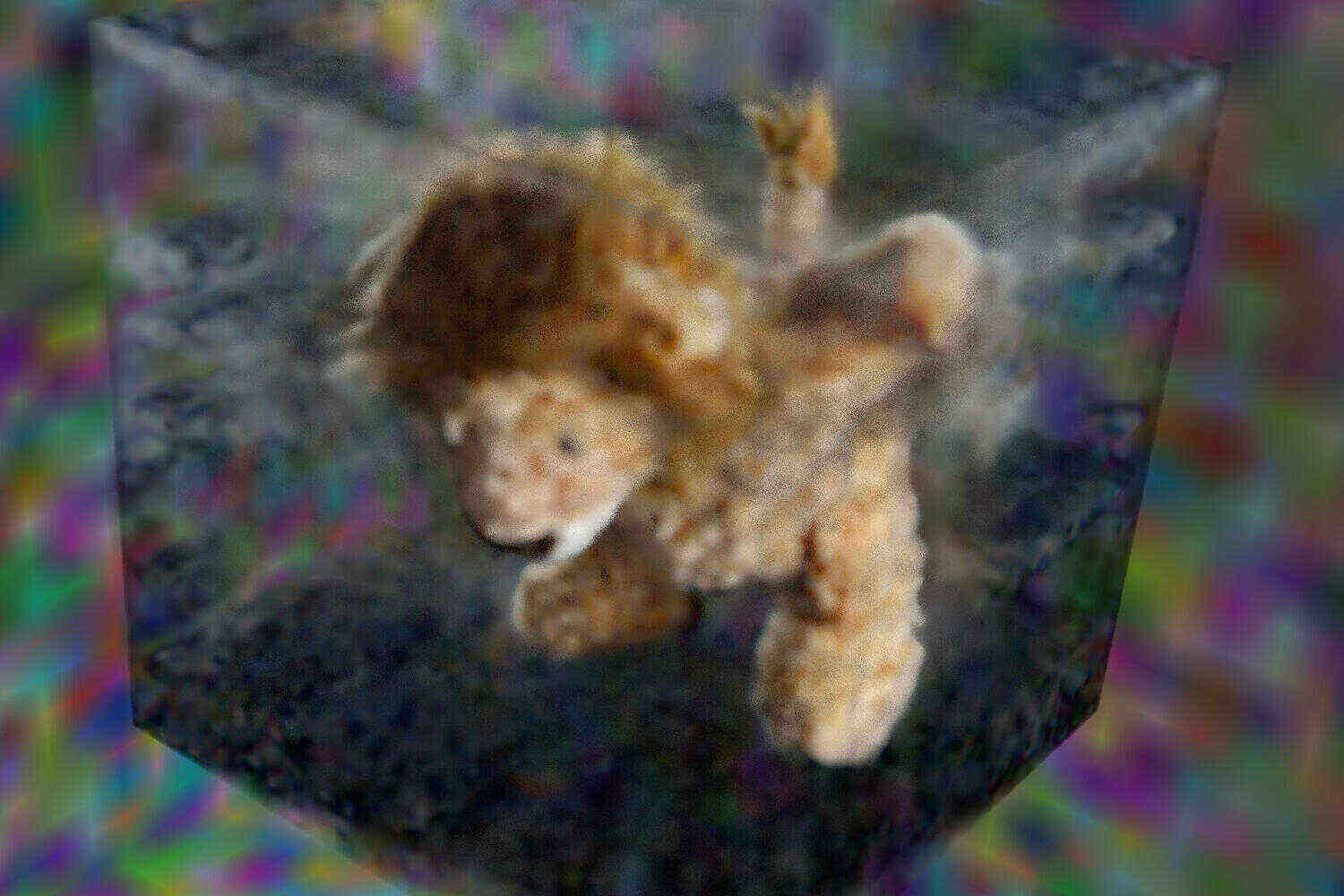}{2.5k it., 8/256/64}%
            \imageGap{}%
        \end{subfigure}%
        \begin{subfigure}[t]{0.245\textwidth}
            \centering%
            \labelImage{\textwidth}{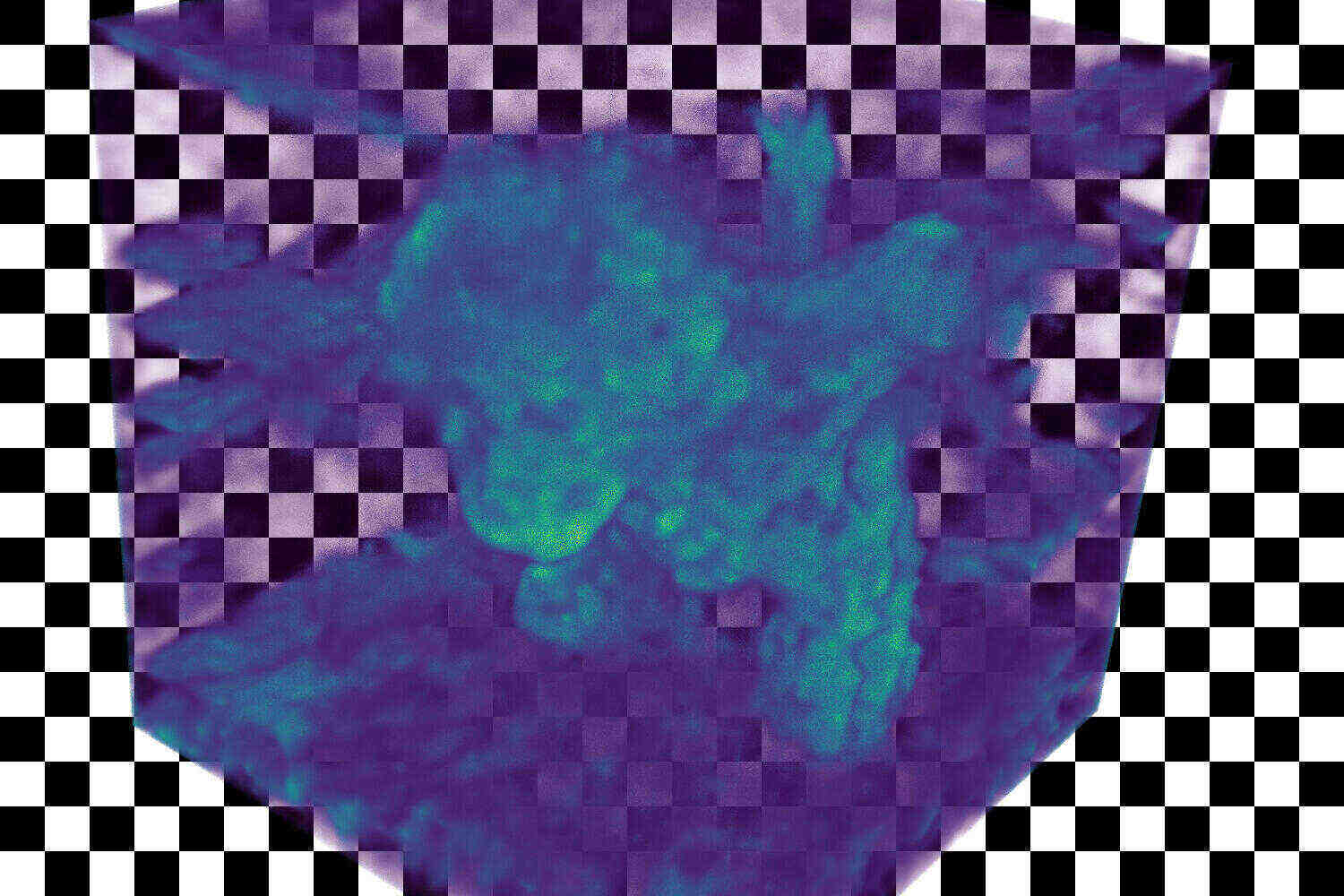}{2.5k it., 8/256/64}%
            \imageGap{}%
        \end{subfigure}%
        \begin{subfigure}[t]{0.245\textwidth}
            \centering%
            \normalMap{\textwidth}{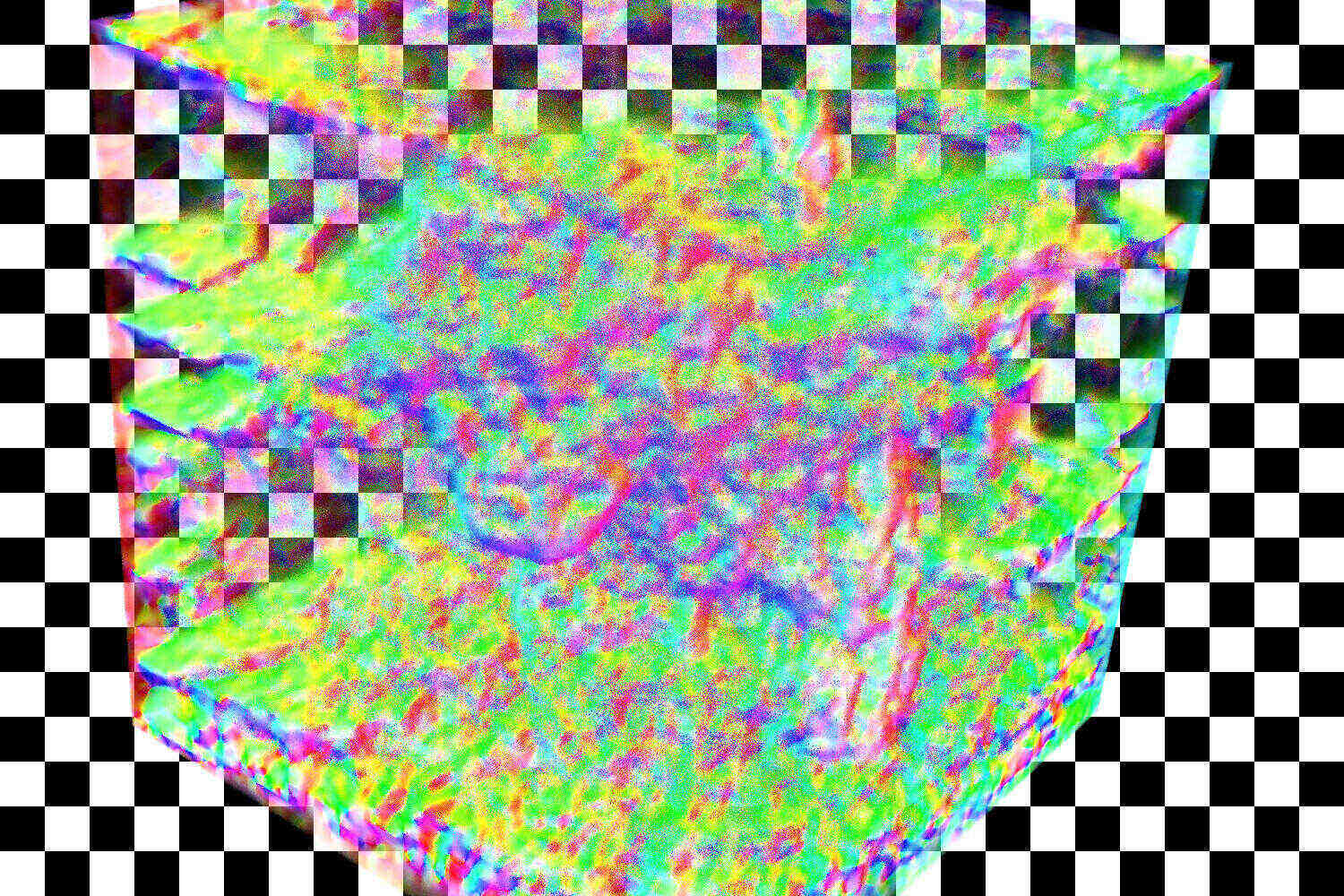}{2.5k it., 8/256/64}%
            \imageGap{}%
        \end{subfigure}%
        \begin{subfigure}[t]{0.245\textwidth}
            \centering%
            \labelImage{\textwidth}{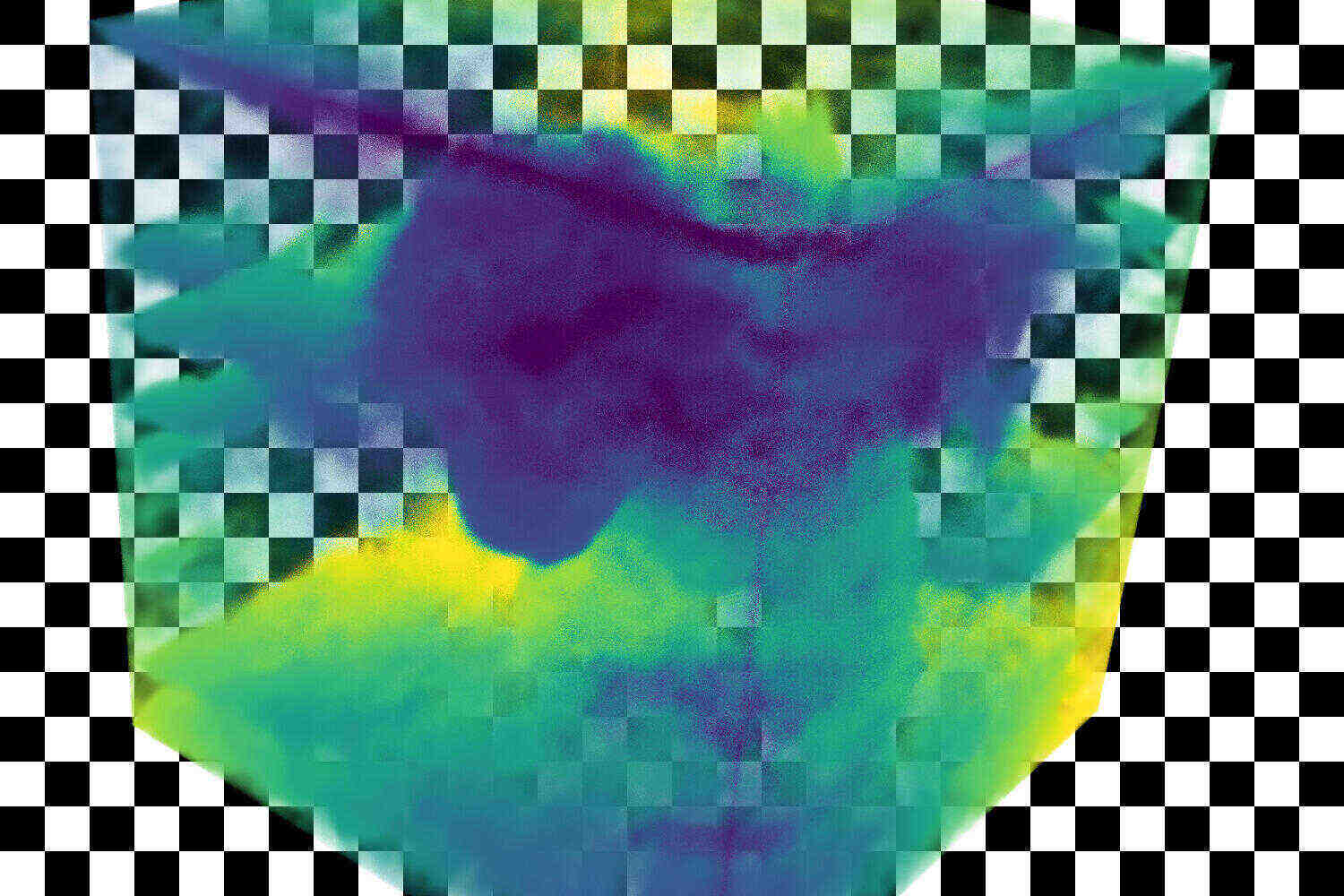}{2.5k it., 8/256/64}%
            \imageGap{}%
        \end{subfigure} \\
        %
        %
        \begin{subfigure}[t]{0.245\textwidth}
            \centering%
            \labelImage{\textwidth}{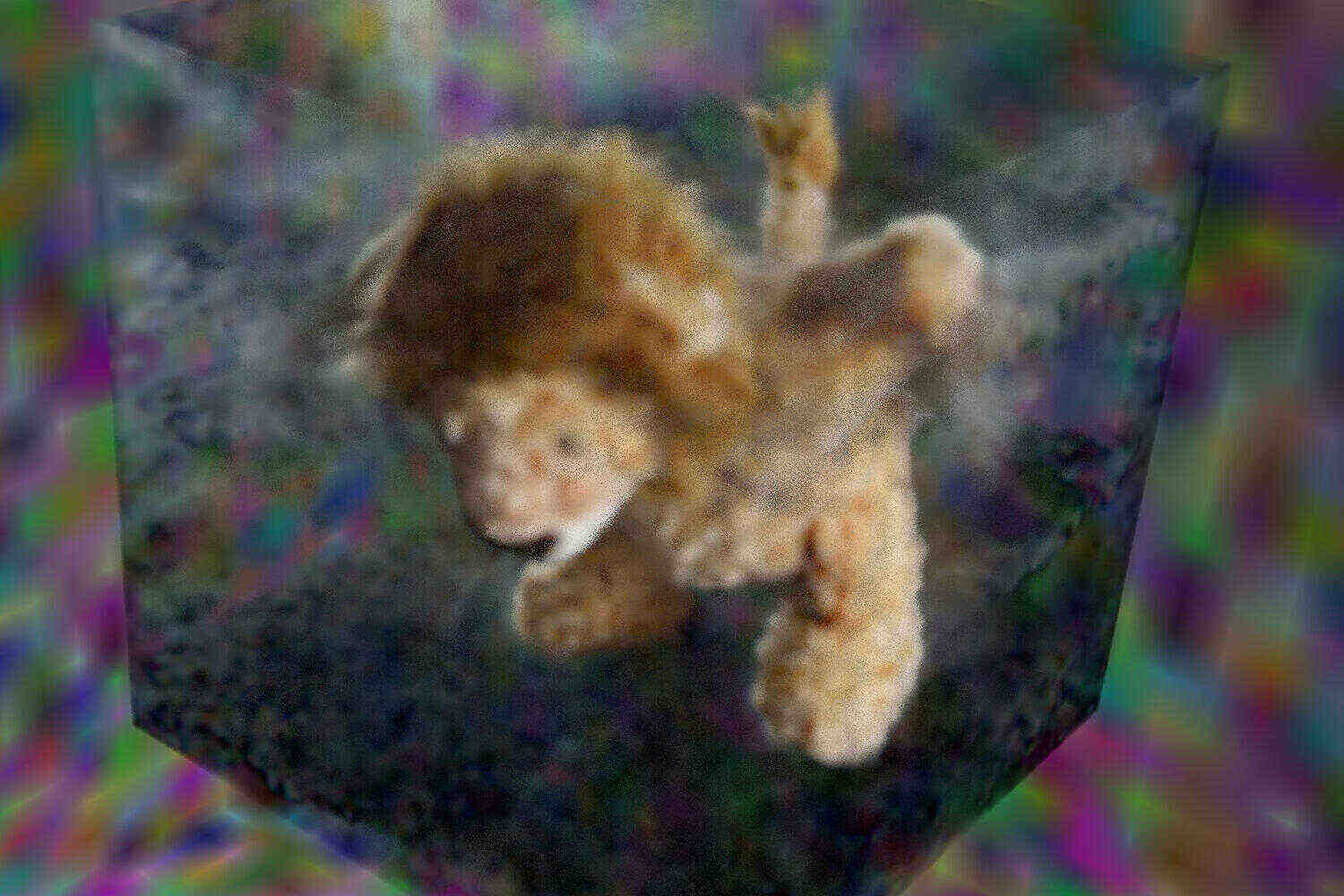}{2.5k it., 8/256/32}%
            \imageGap{}%
        \end{subfigure}%
        \begin{subfigure}[t]{0.245\textwidth}
            \centering%
            \labelImage{\textwidth}{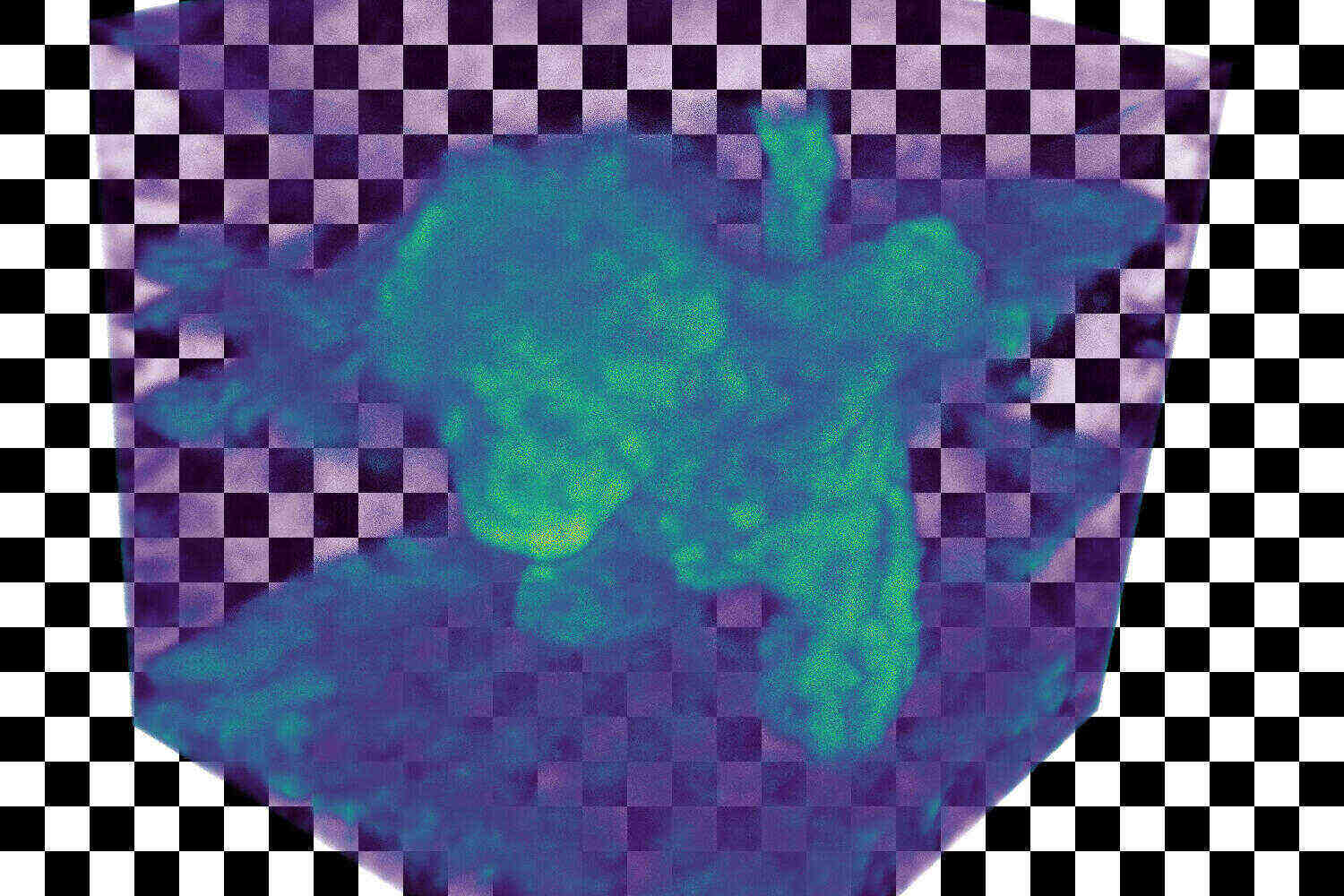}{2.5k it., 8/256/32}%
            \imageGap{}%
        \end{subfigure}%
        \begin{subfigure}[t]{0.245\textwidth}
            \centering%
            \normalMap{\textwidth}{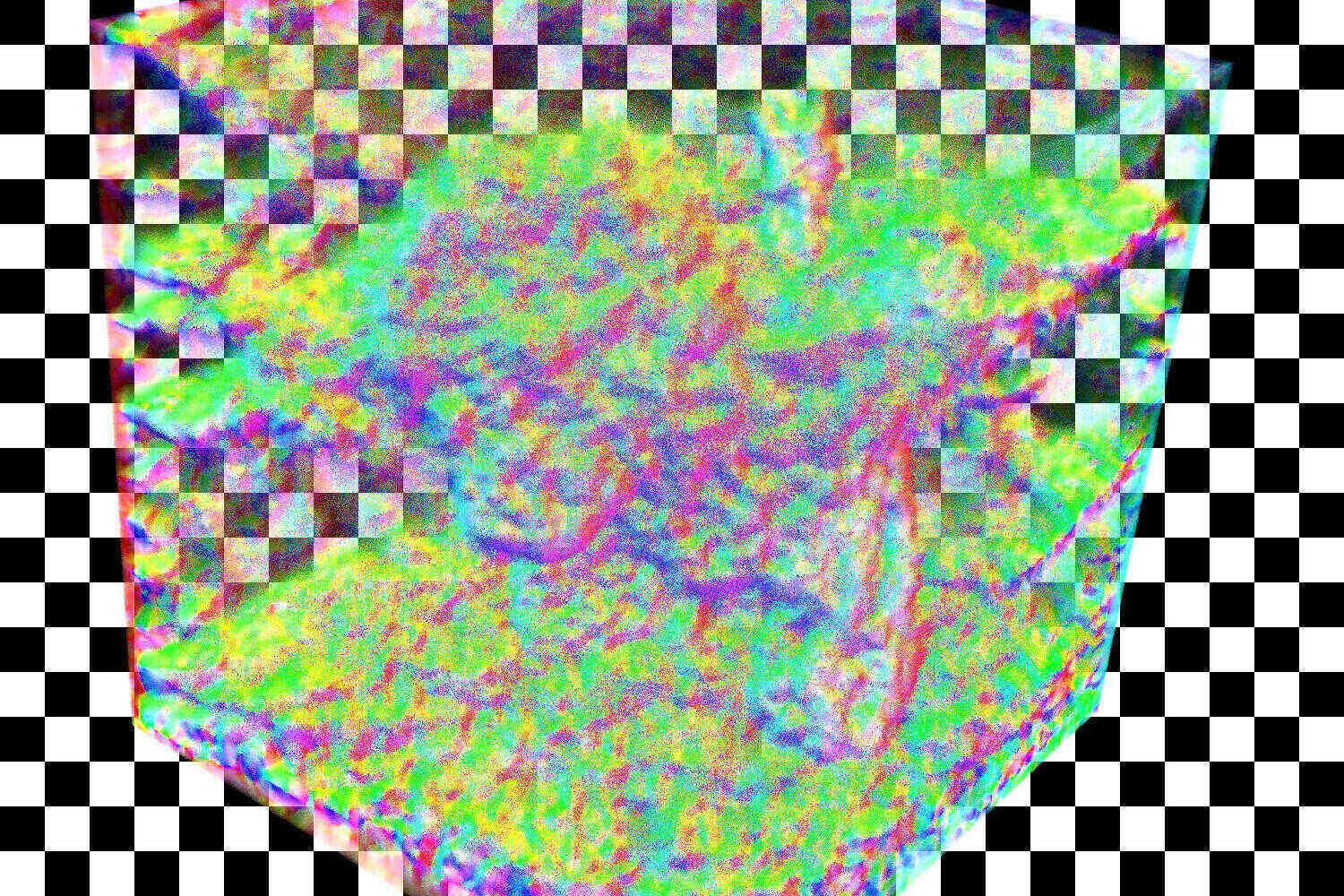}{2.5k it., 8/256/32}%
            \imageGap{}%
        \end{subfigure}%
        \begin{subfigure}[t]{0.245\textwidth}
            \centering%
            \labelImage{\textwidth}{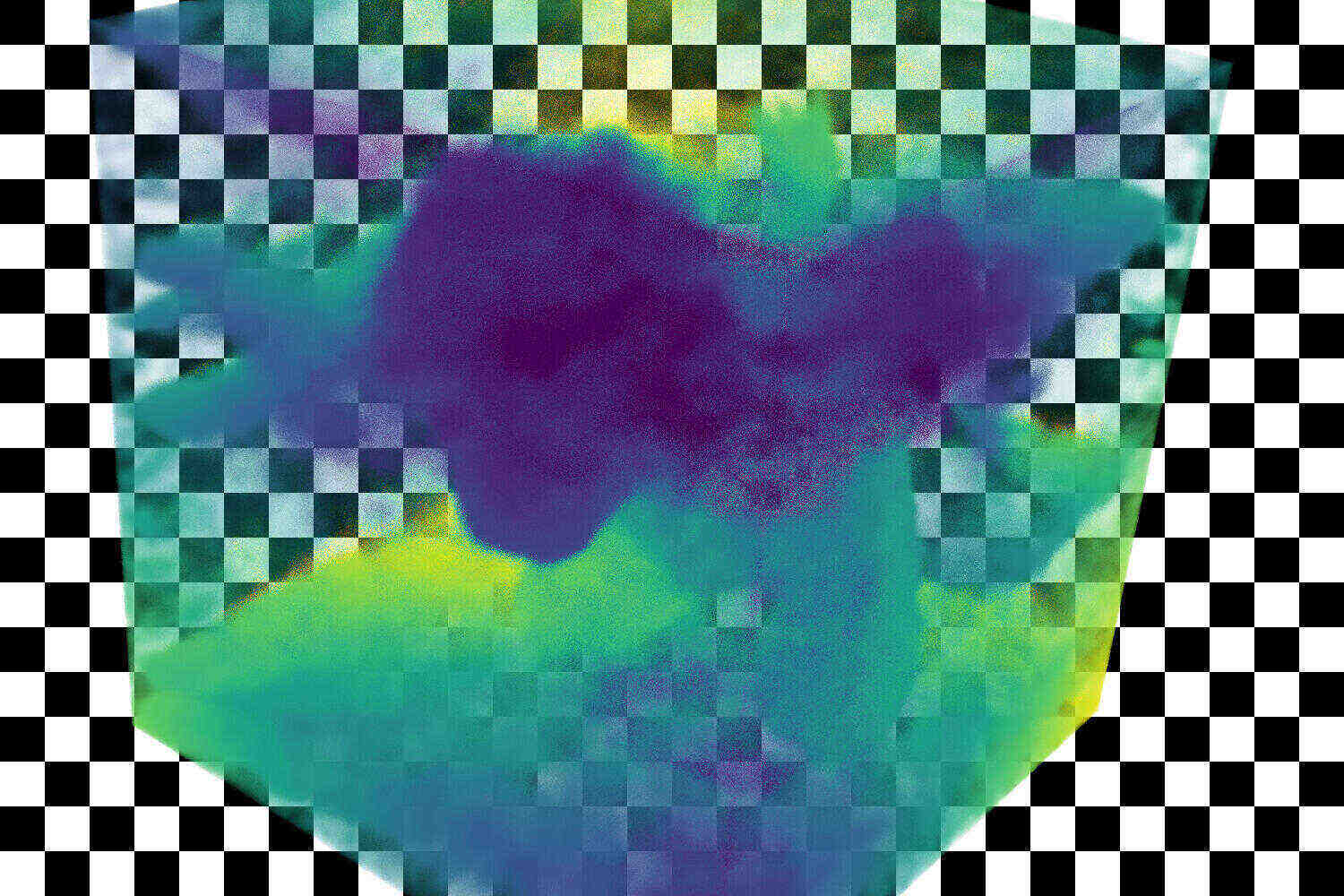}{2.5k it., 8/256/32}%
            \imageGap{}%
        \end{subfigure} \\
        %
        %
        \begin{subfigure}[t]{0.245\textwidth}
            \centering%
            \labelImage{\textwidth}{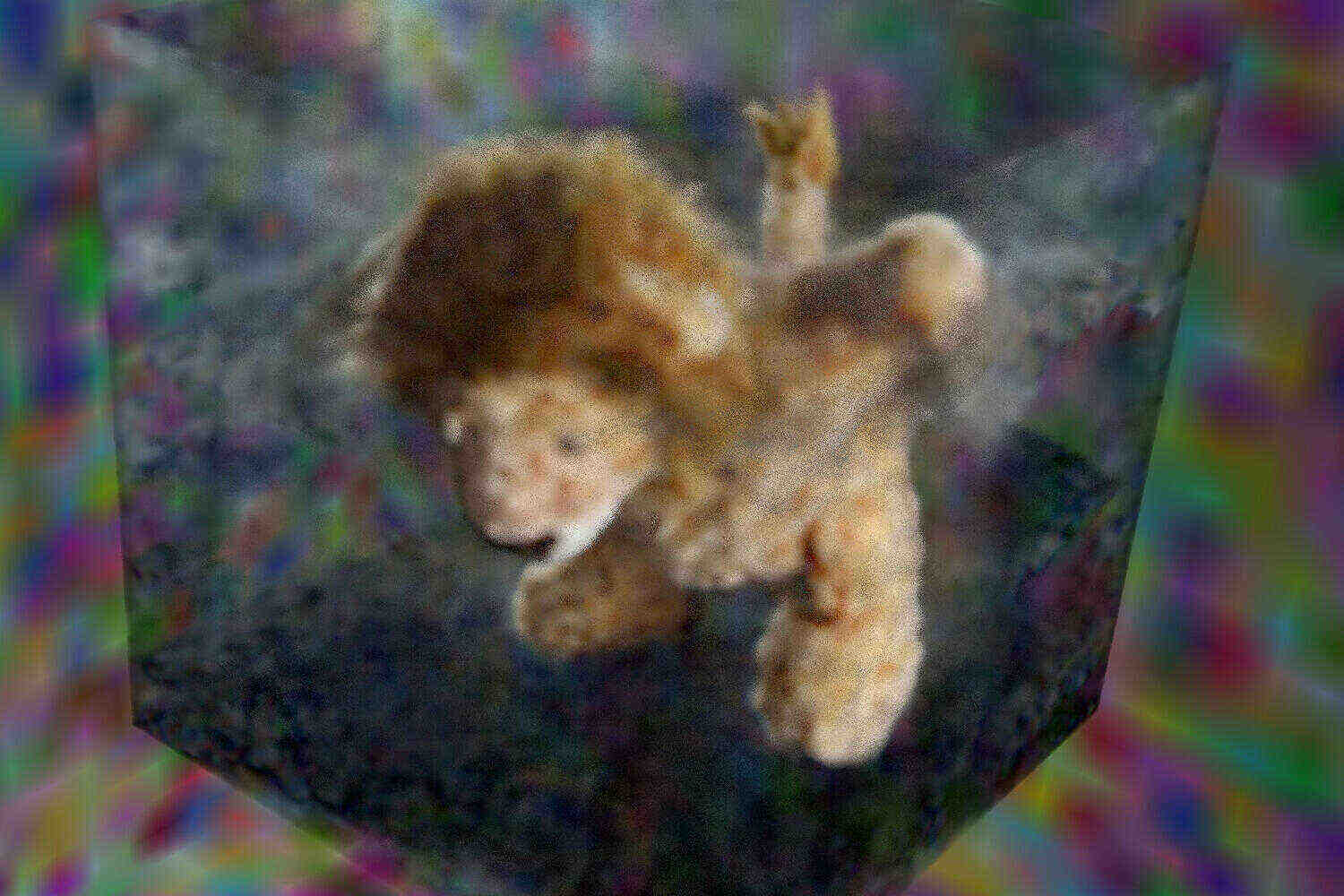}{2.5k it., 8/256/32, uniform}%
            \imageGap{}%
        \end{subfigure}%
        \begin{subfigure}[t]{0.245\textwidth}
            \centering%
            \labelImage{\textwidth}{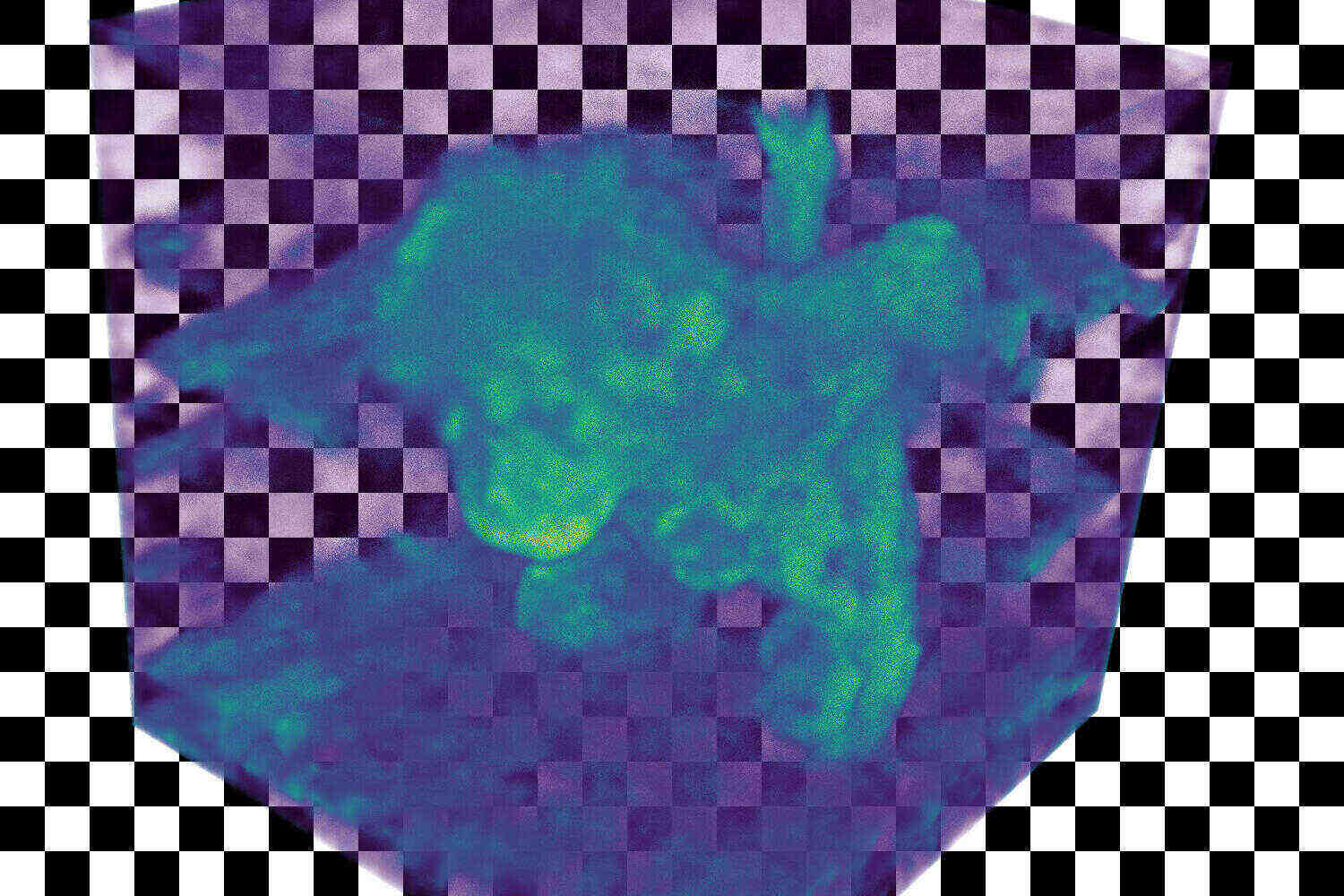}{2.5k it., 8/256/32, uniform}%
            \imageGap{}%
        \end{subfigure}%
        \begin{subfigure}[t]{0.245\textwidth}
            \centering%
            \normalMap{\textwidth}{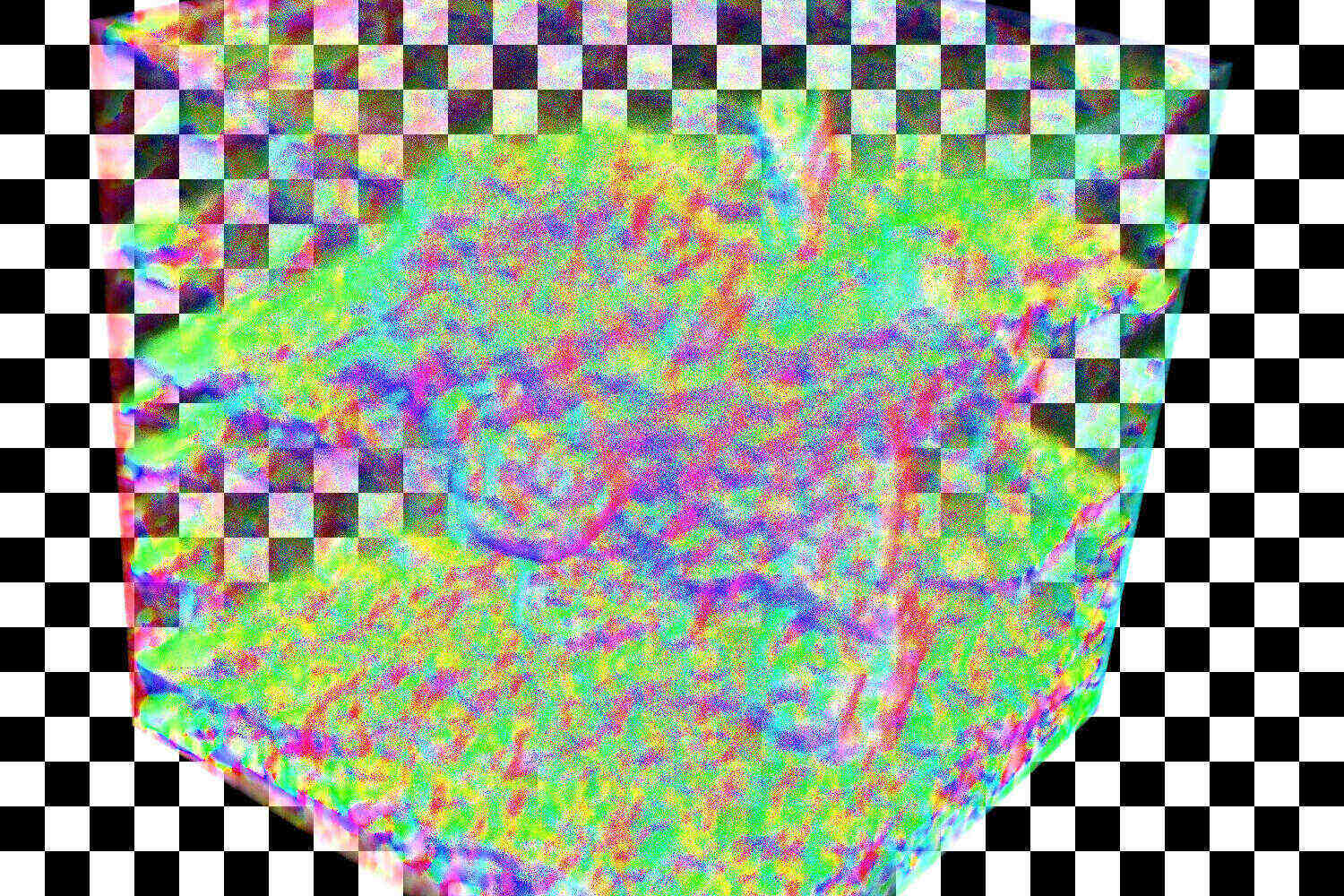}{2.5k it., 8/256/32, uniform}%
            \imageGap{}%
        \end{subfigure}%
        \begin{subfigure}[t]{0.245\textwidth}
            \centering%
            \labelImage{\textwidth}{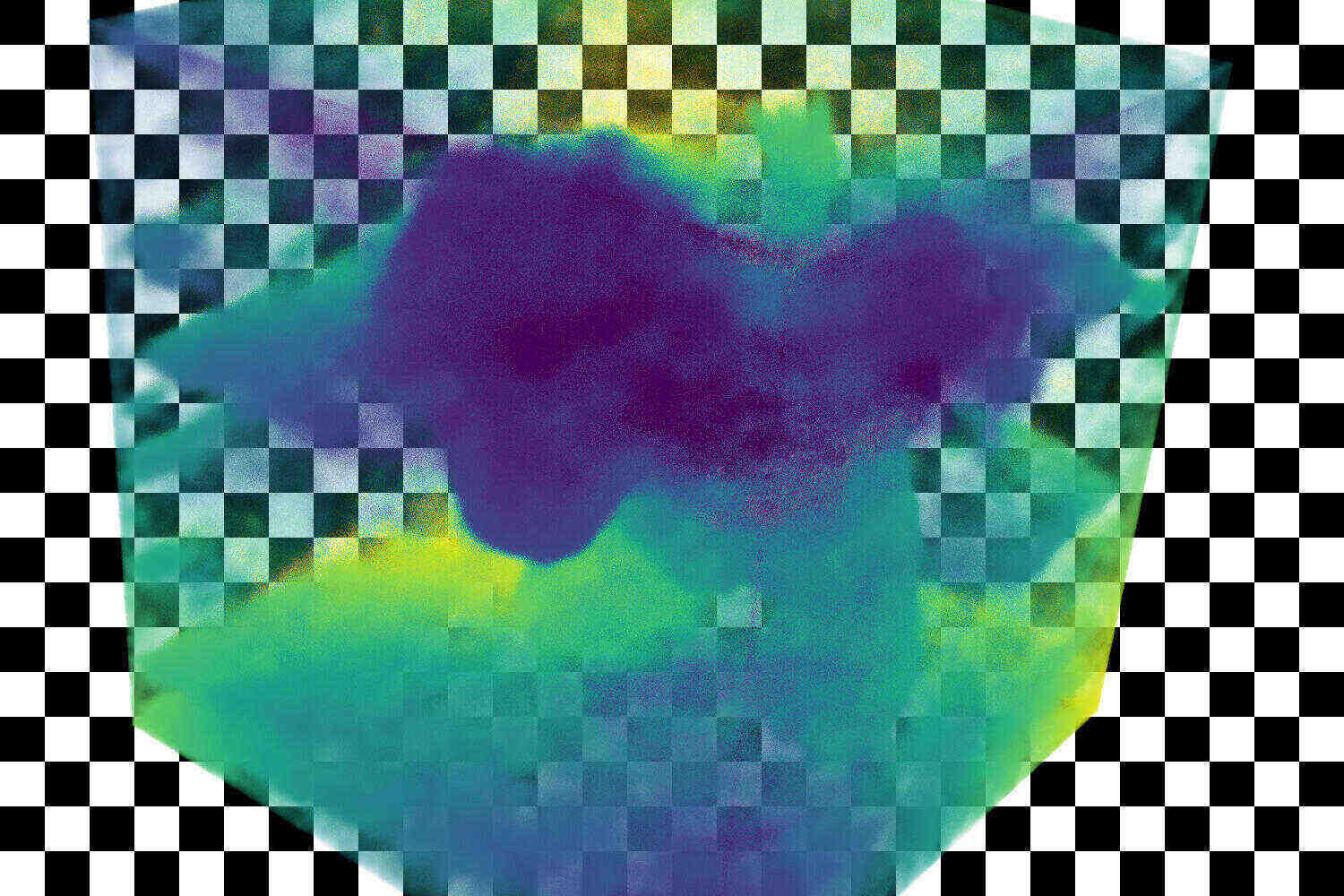}{2.5k it., 8/256/32, uniform}%
            \imageGap{}%
        \end{subfigure}\\%
        %
        %
        \begin{subfigure}[t]{0.245\textwidth}
            \centering%
            \labelImage{\textwidth}{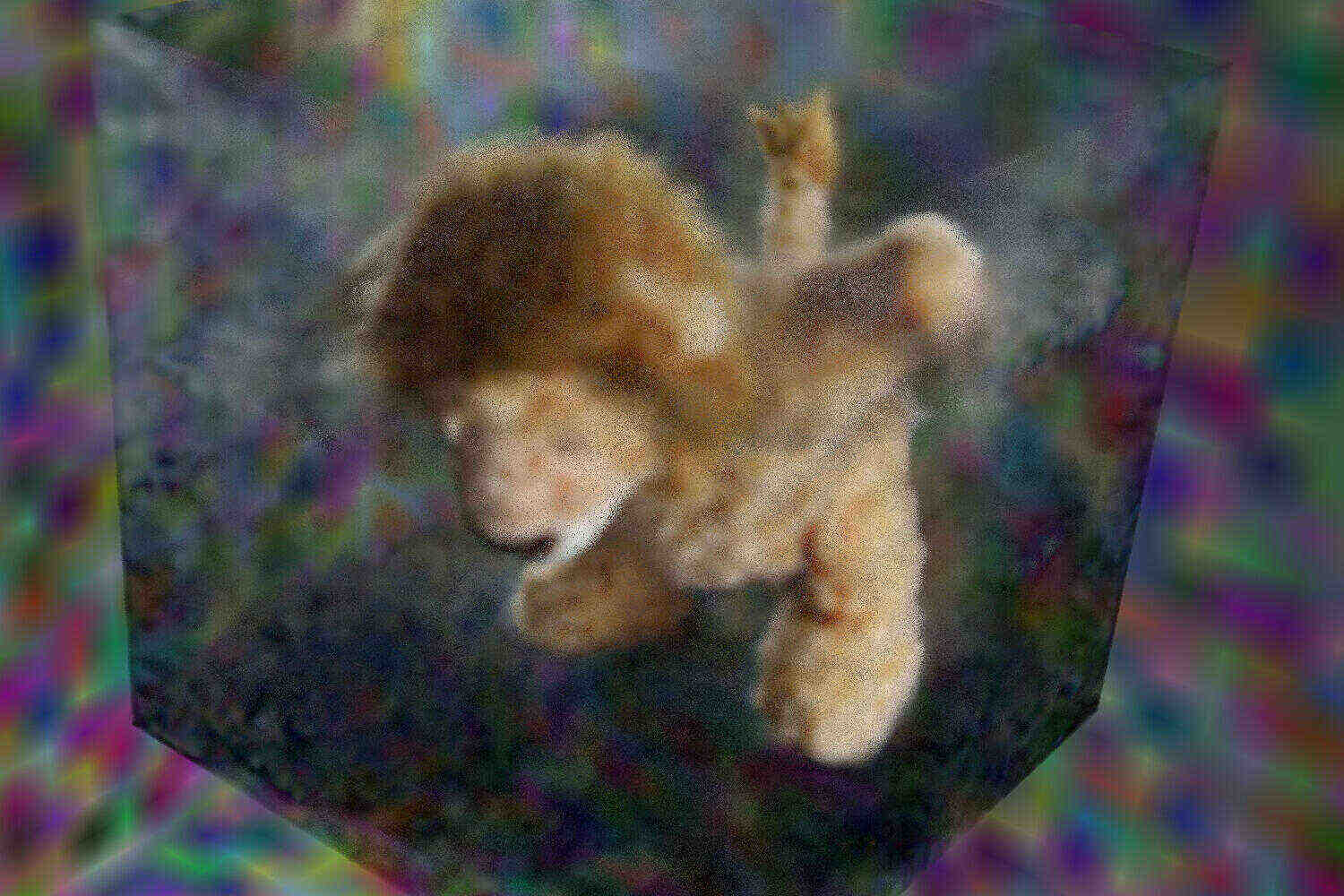}{2.5k it., 8/256/16}%
            \imageGap{}%
        \end{subfigure}%
        \begin{subfigure}[t]{0.245\textwidth}
            \centering%
            \labelImage{\textwidth}{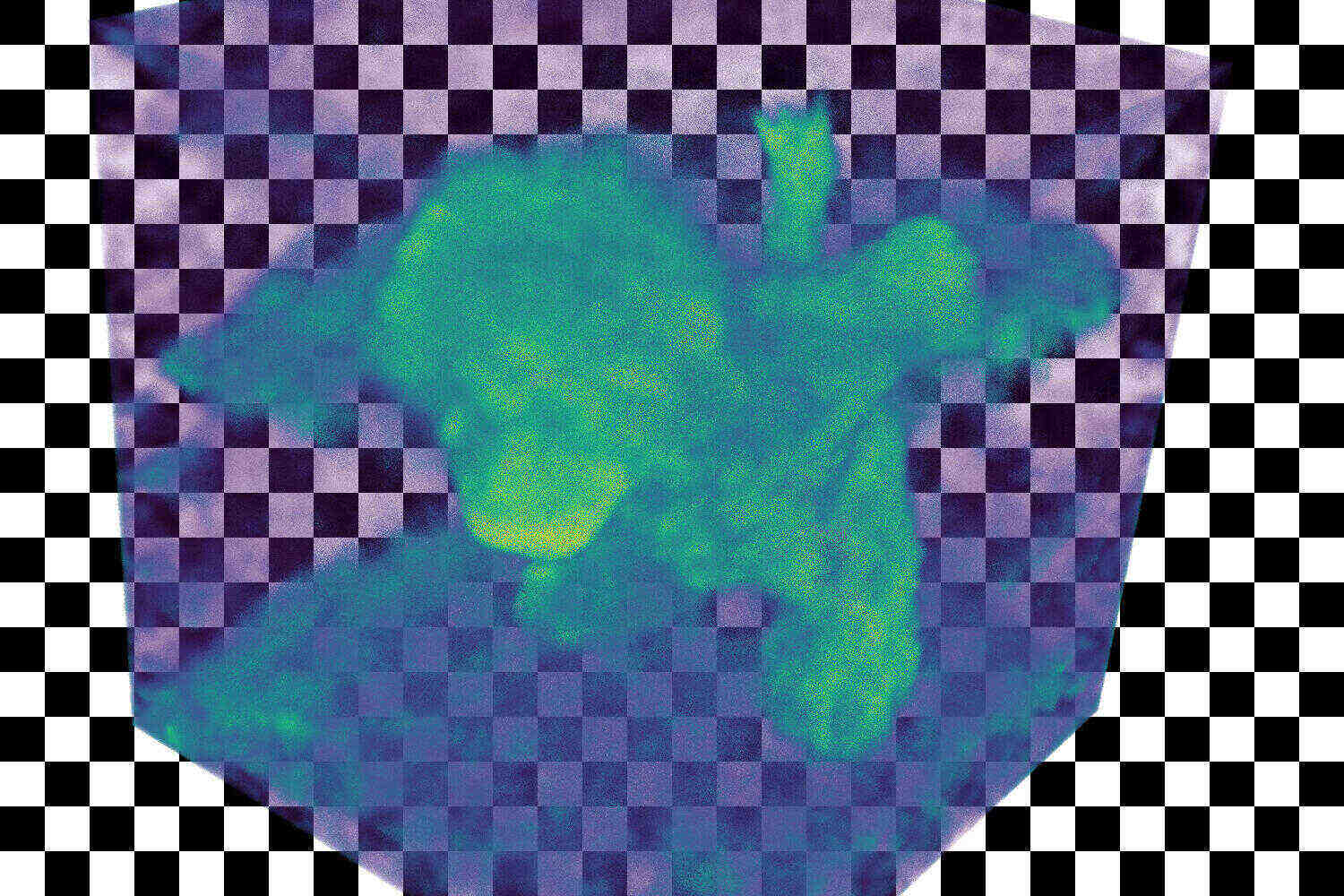}{2.5k it., 8/256/16}%
            \imageGap{}%
        \end{subfigure}%
        \begin{subfigure}[t]{0.245\textwidth}
            \centering%
            \normalMap{\textwidth}{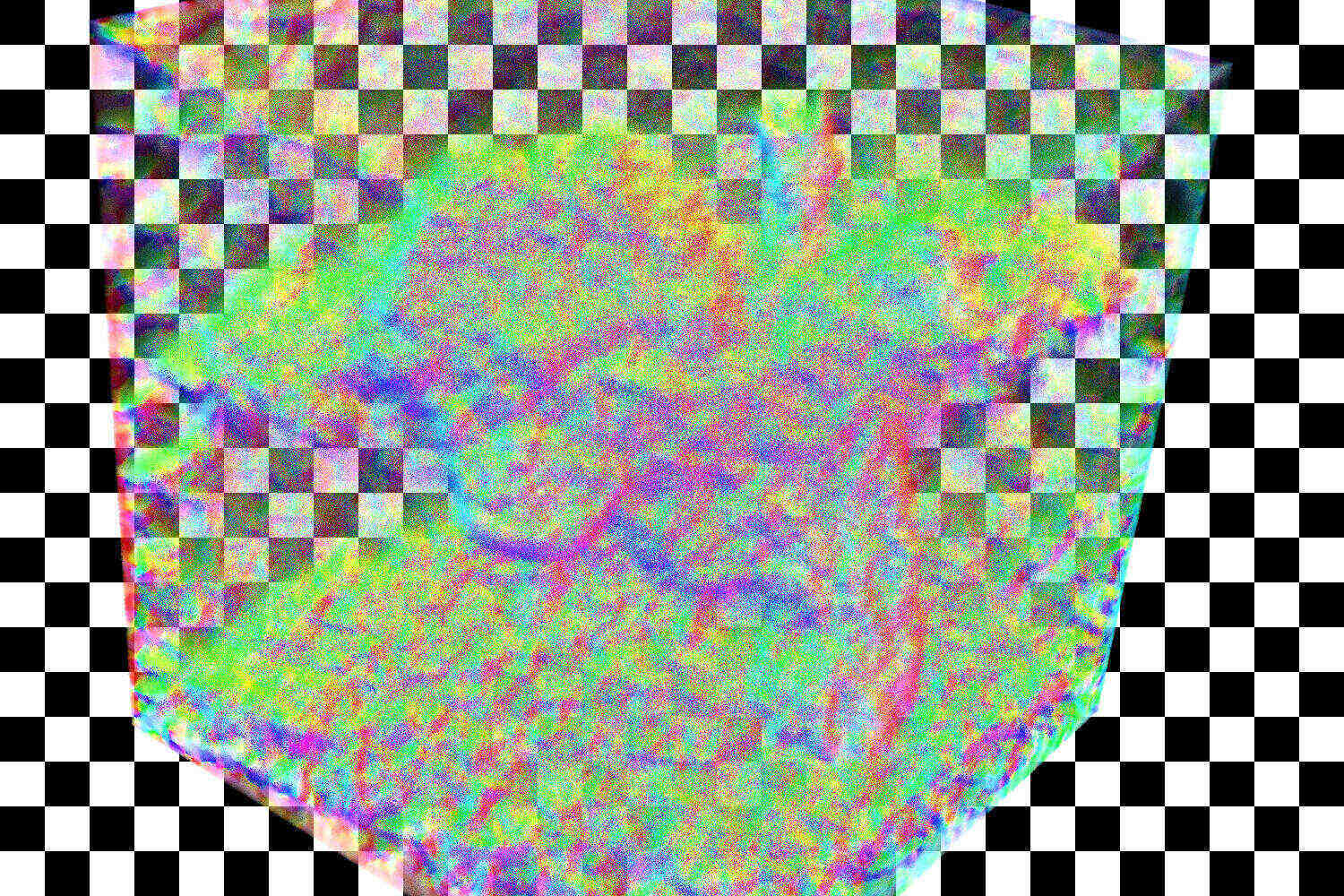}{2.5k it., 8/256/16}%
            \imageGap{}%
        \end{subfigure}%
        \begin{subfigure}[t]{0.245\textwidth}
            \centering%
            \labelImage{\textwidth}{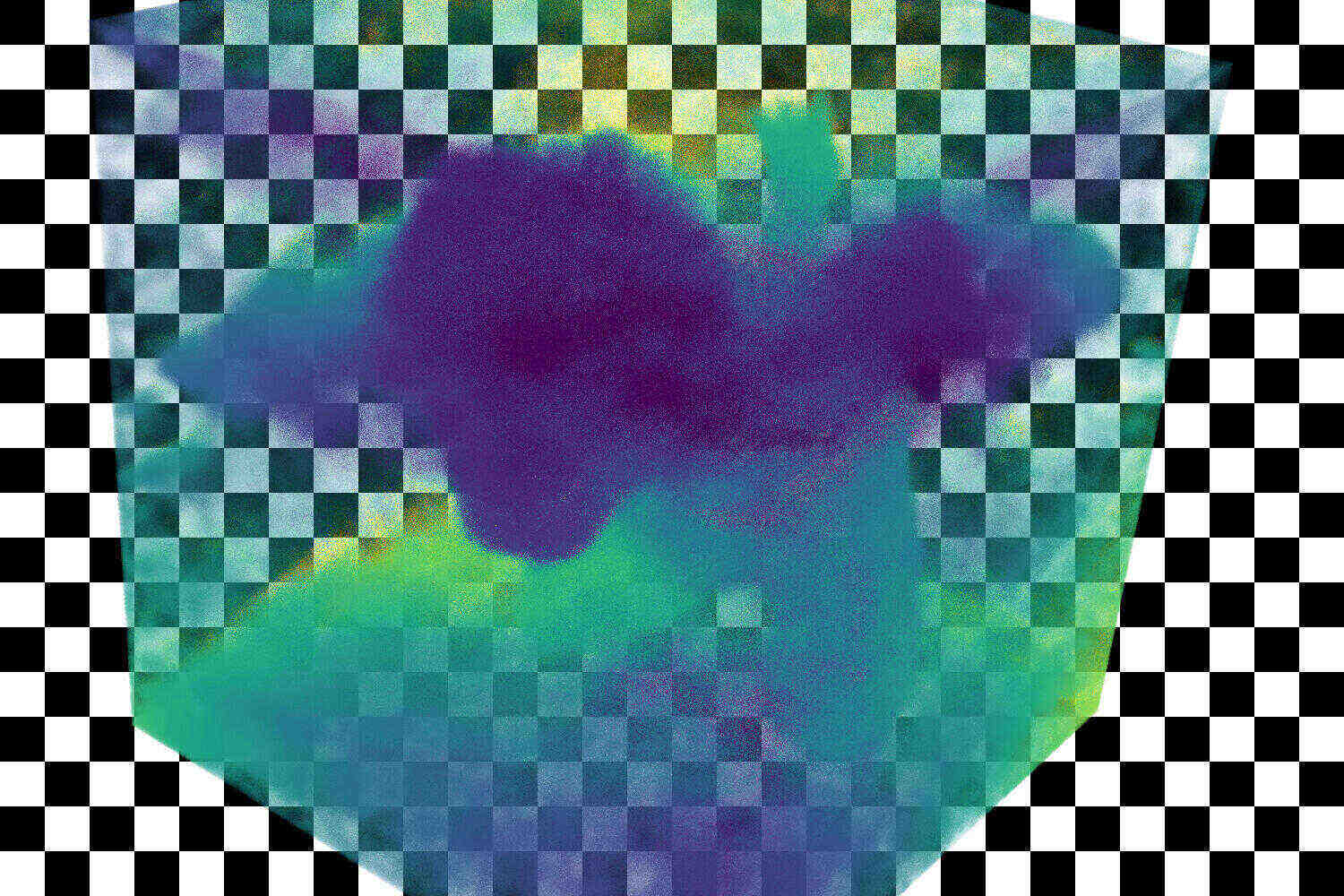}{2.5k it., 8/256/16}%
            \imageGap{}%
        \end{subfigure} \\
    \end{tabular}
    %
    %
    \caption%
    {%
        Scene sampling influence on early results of the Lion scene with varying
        sampling budgets $N/\maxSamplesPerRayUninformed/\maxSamplesPerRayInformed$
        which respectively are the
        samples per node edge length;
        maximum number of samples per ray after uninformed
        and after opacity-based filtering.
    }%
    \label{fig:sceneSamplingBudgets2.5k}
\end{figure*}

\begin{figure*}%
    \centering%
    \begin{tabular}{l}
        %
        %
        \begin{subfigure}[t]{0.245\textwidth}
            \centering%
            \labelImage{\textwidth}{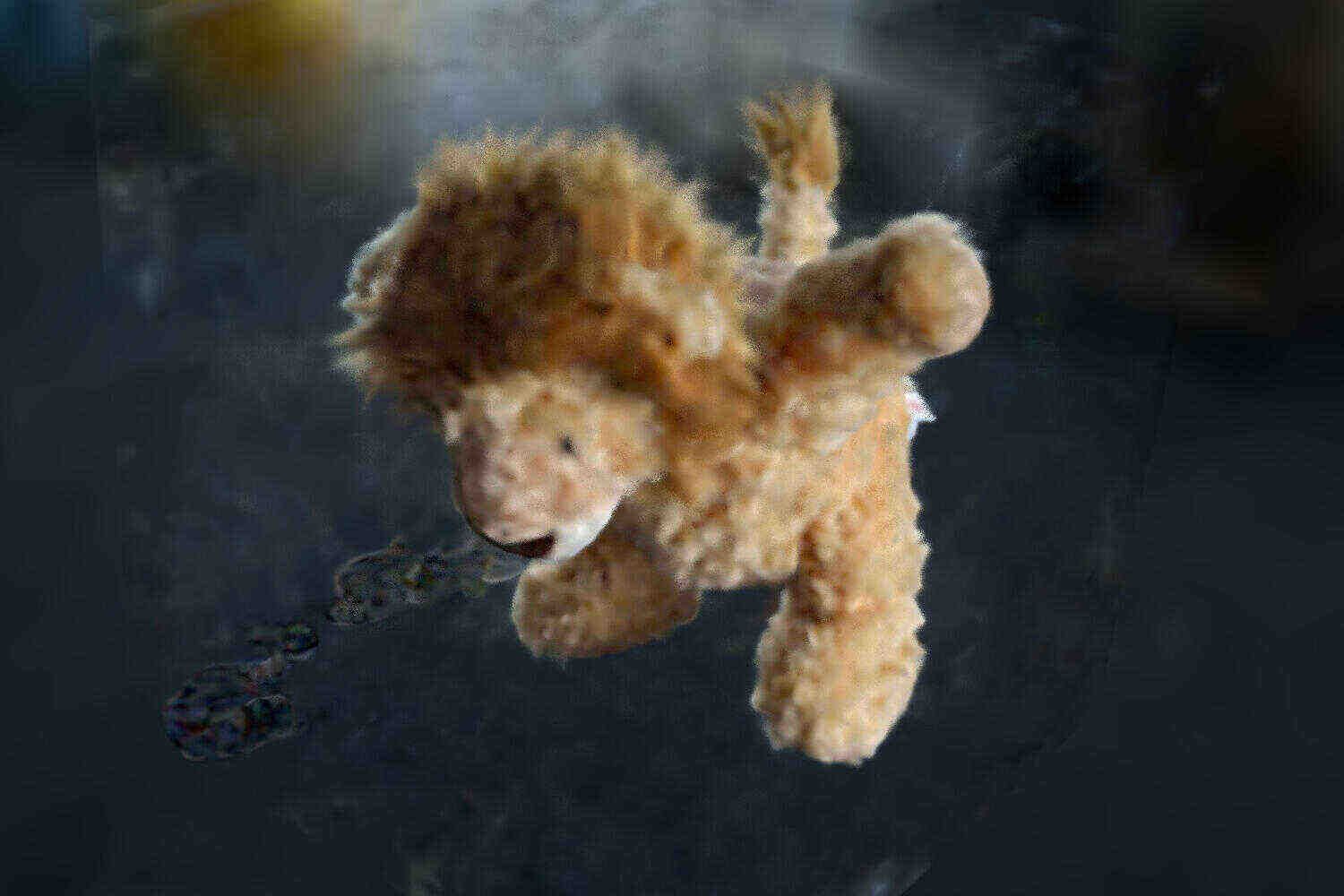}{40k it., 16/512/128}%
            \imageGap{}%
        \end{subfigure}%
        \begin{subfigure}[t]{0.245\textwidth}
            \centering%
            \labelImage{\textwidth}{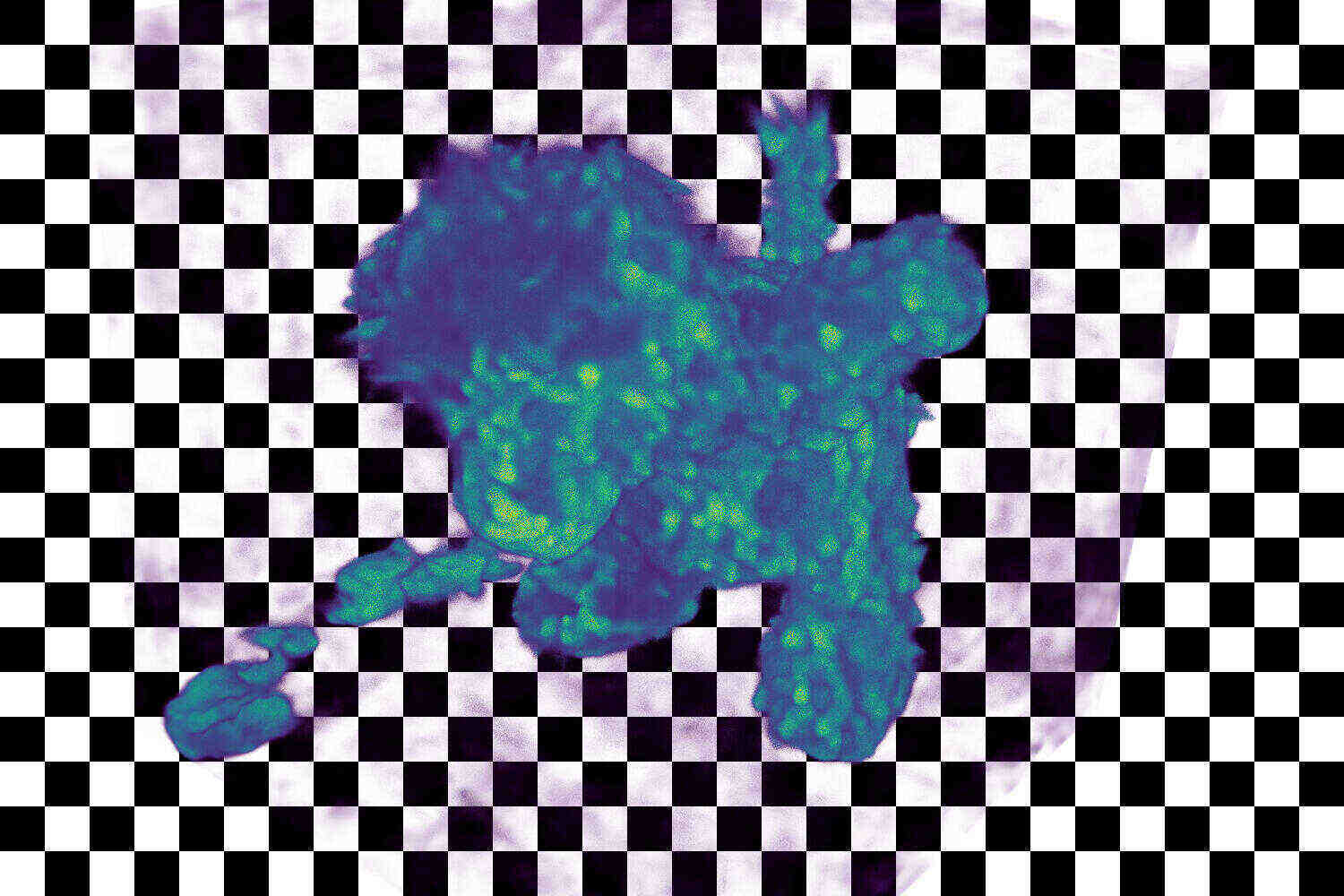}{40k it., 16/512/128}%
            \imageGap{}%
        \end{subfigure}%
        \begin{subfigure}[t]{0.245\textwidth}
            \centering%
            \normalMap{\textwidth}{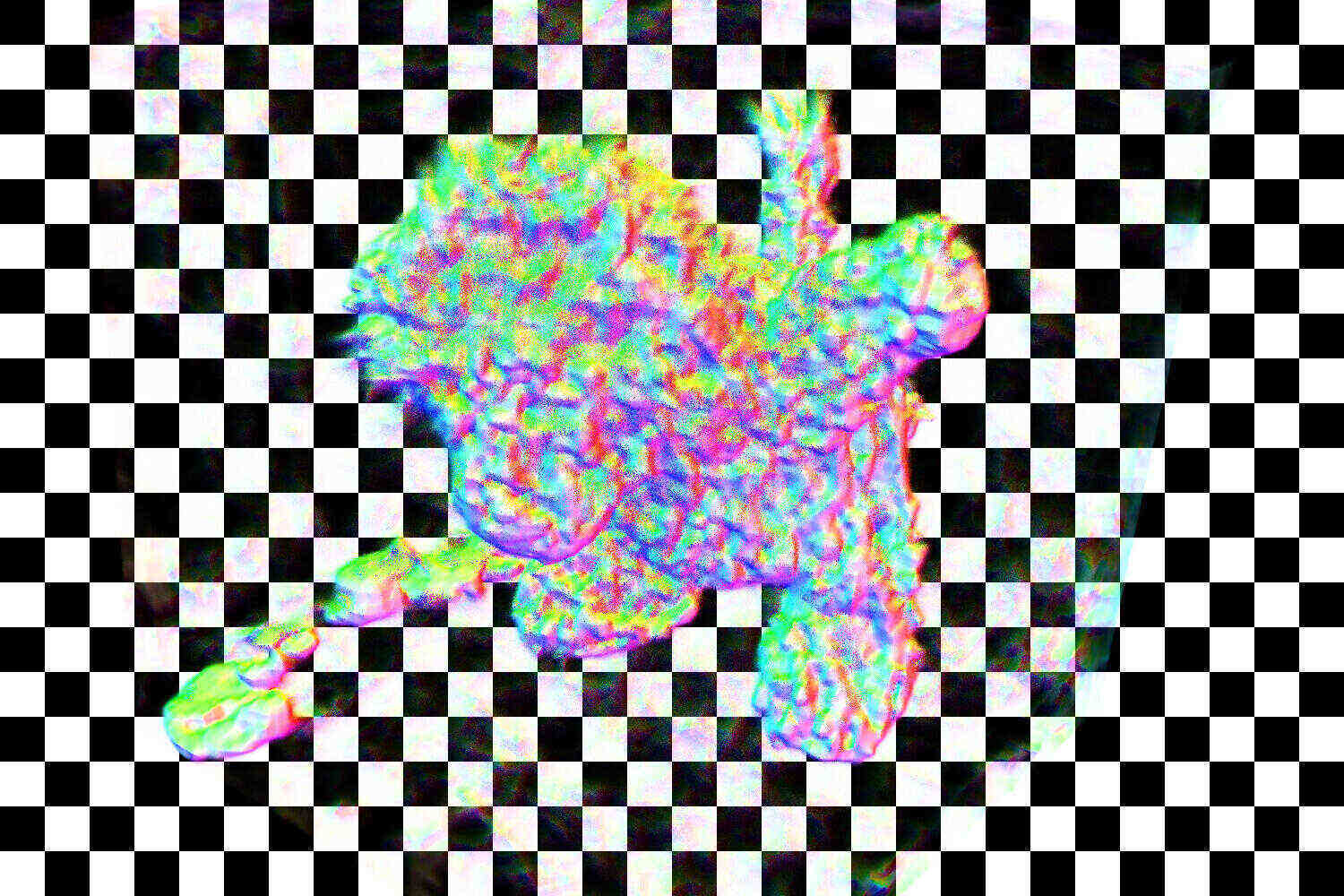}{40k it., 16/512/128}%
            \imageGap{}%
        \end{subfigure}%
        \begin{subfigure}[t]{0.245\textwidth}
            \centering%
            \labelImage{\textwidth}{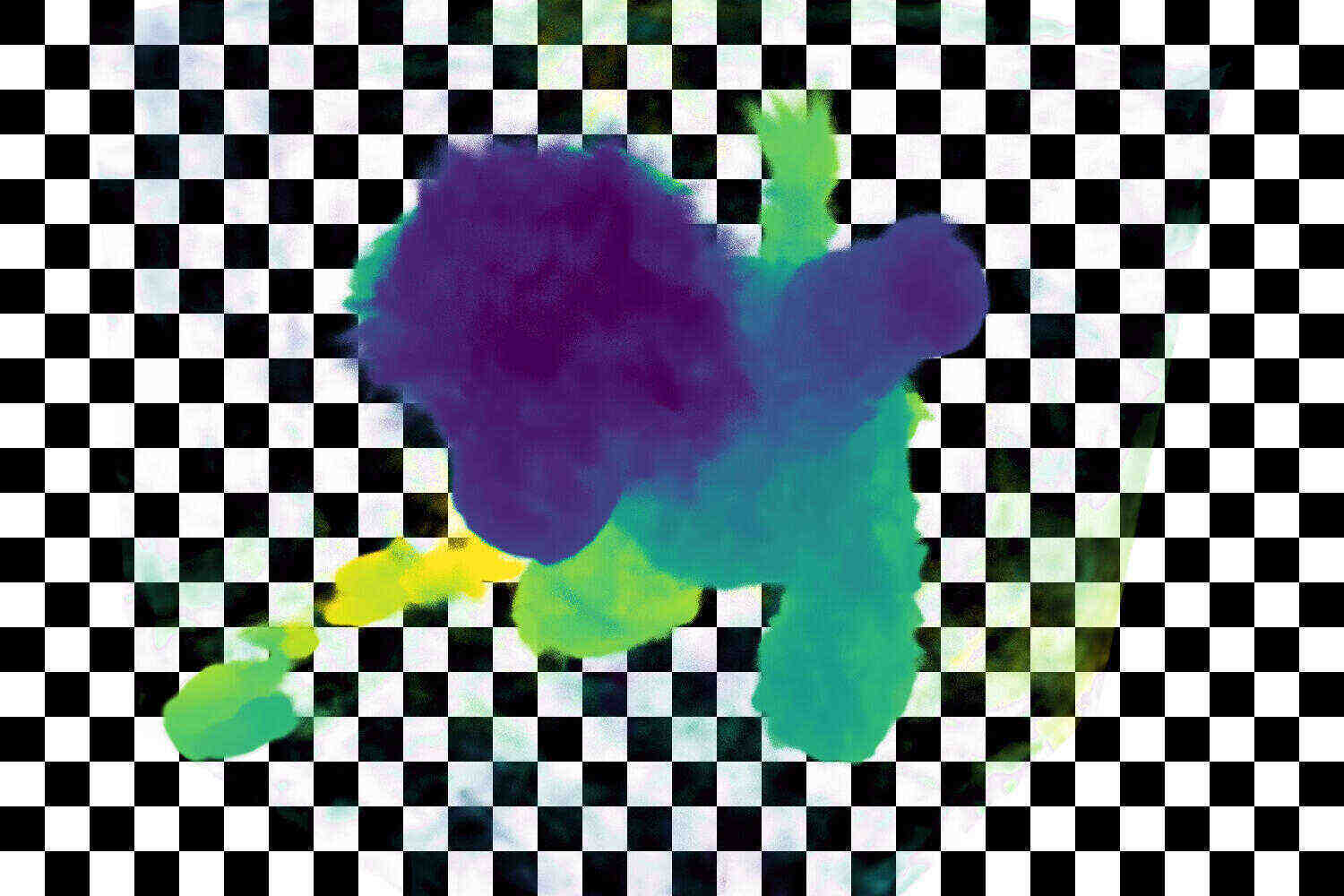}{40k it., 16/512/128}%
            \imageGap{}%
        \end{subfigure} \\
        %
        %
        \begin{subfigure}[t]{0.245\textwidth}
            \centering%
            \labelImage{\textwidth}{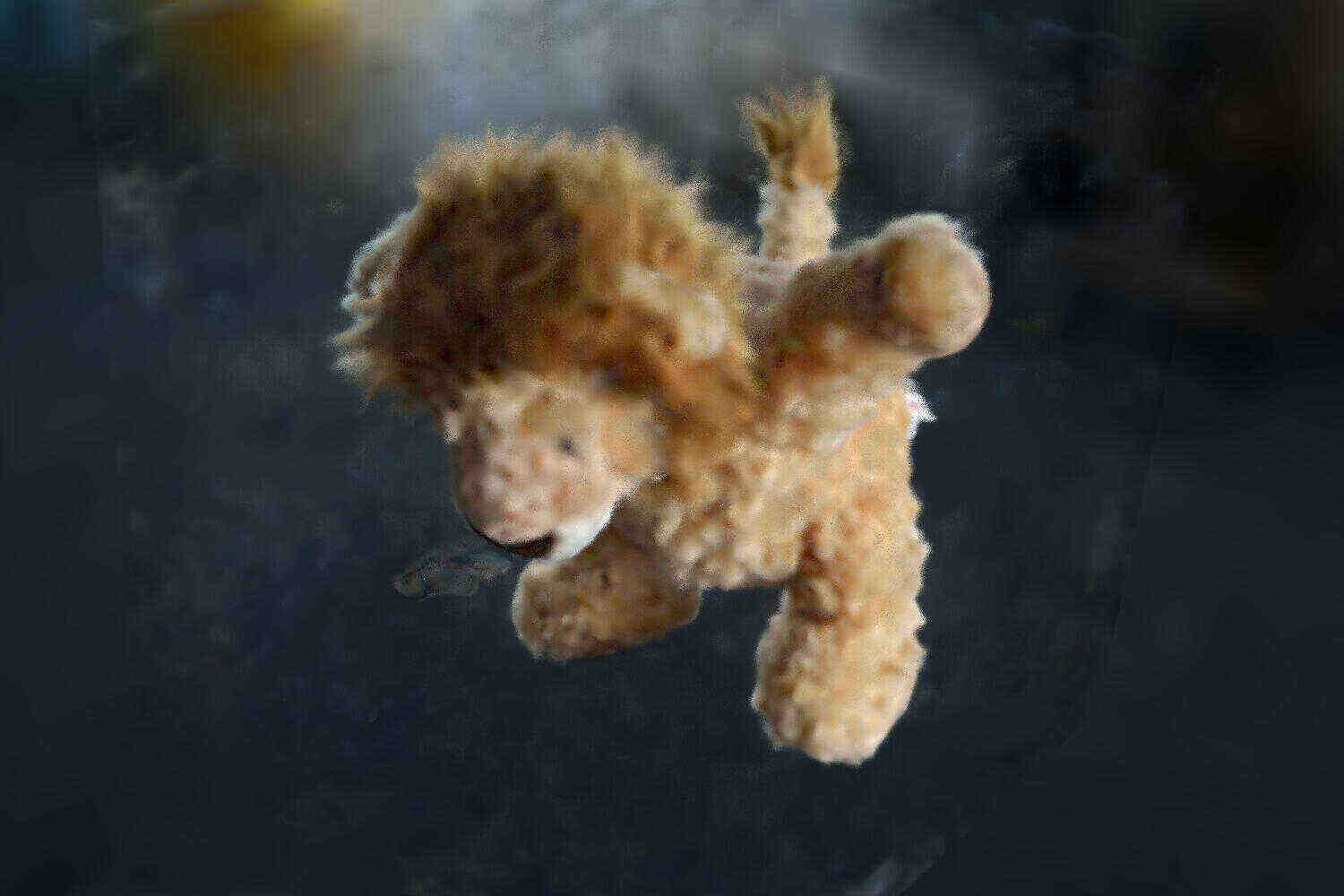}{40k it., 8/256/64}%
            \imageGap{}%
        \end{subfigure}%
        \begin{subfigure}[t]{0.245\textwidth}
            \centering%
            \labelImage{\textwidth}{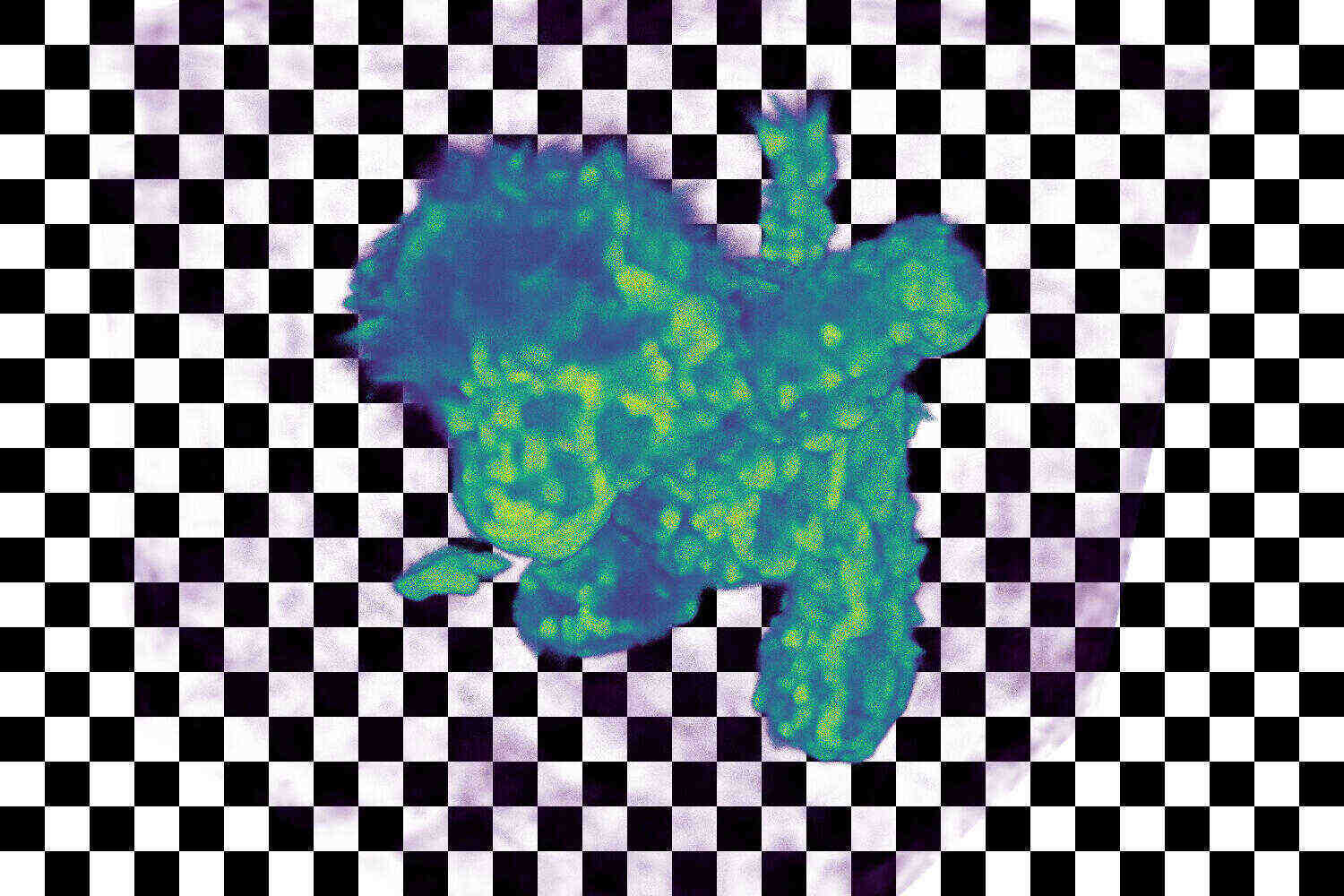}{40k it., 8/256/64}%
            \imageGap{}%
        \end{subfigure}%
        \begin{subfigure}[t]{0.245\textwidth}
            \centering%
            \normalMap{\textwidth}{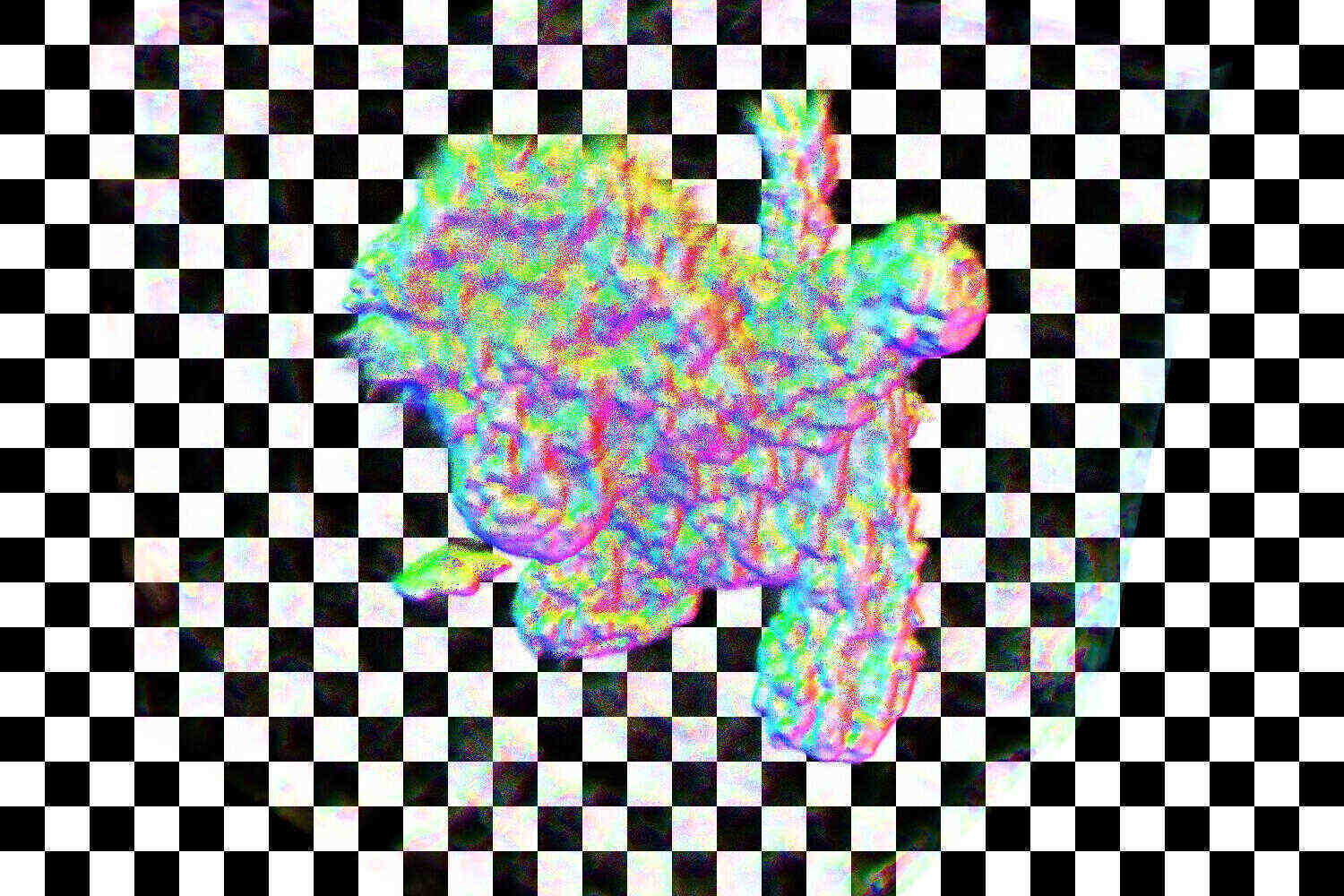}{40k it., 8/256/64}%
            \imageGap{}%
        \end{subfigure}%
        \begin{subfigure}[t]{0.245\textwidth}
            \centering%
            \labelImage{\textwidth}{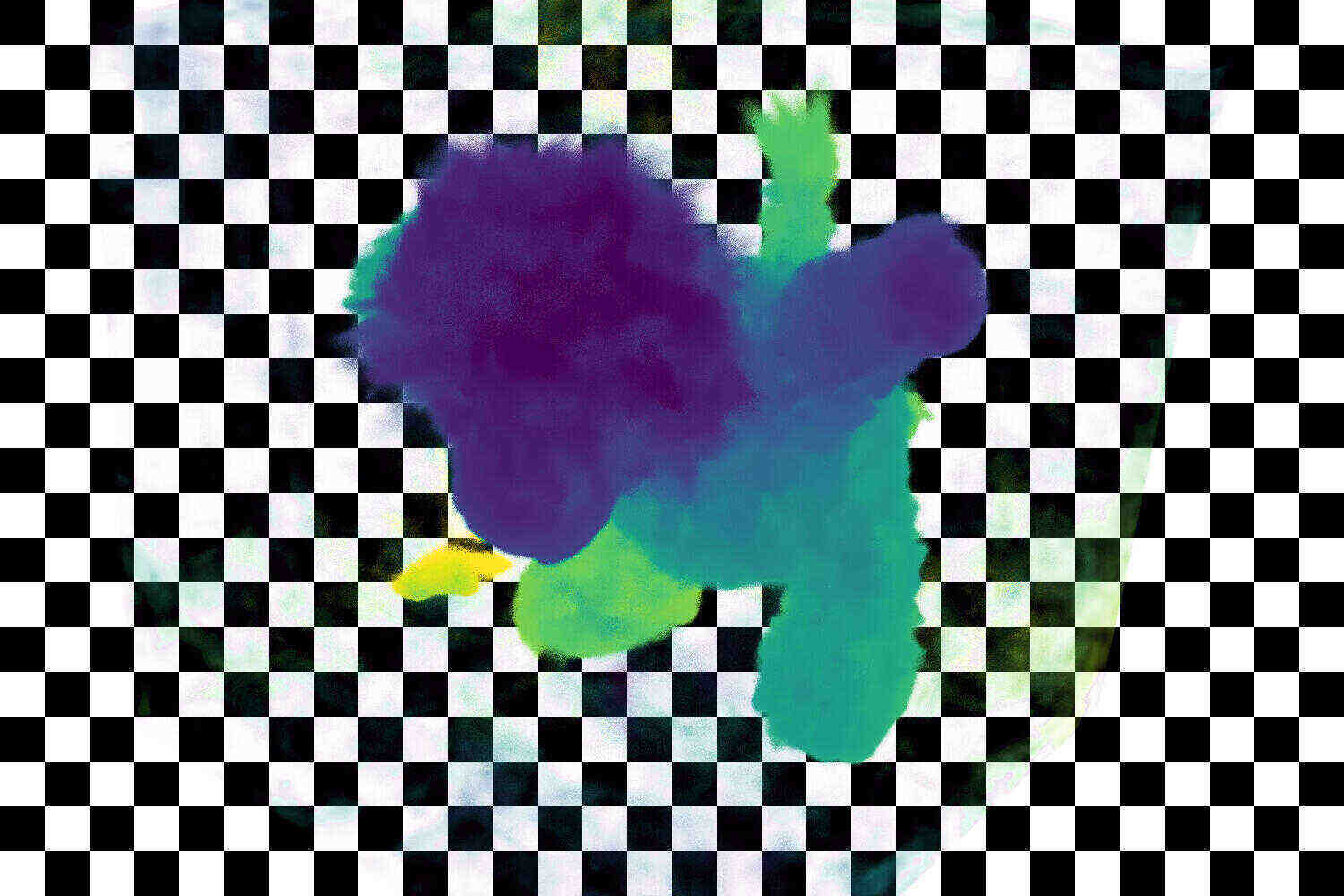}{40k it., 8/256/64}%
            \imageGap{}%
        \end{subfigure} \\
        %
        %
        \begin{subfigure}[t]{0.245\textwidth}
            \centering%
            \labelImage{\textwidth}{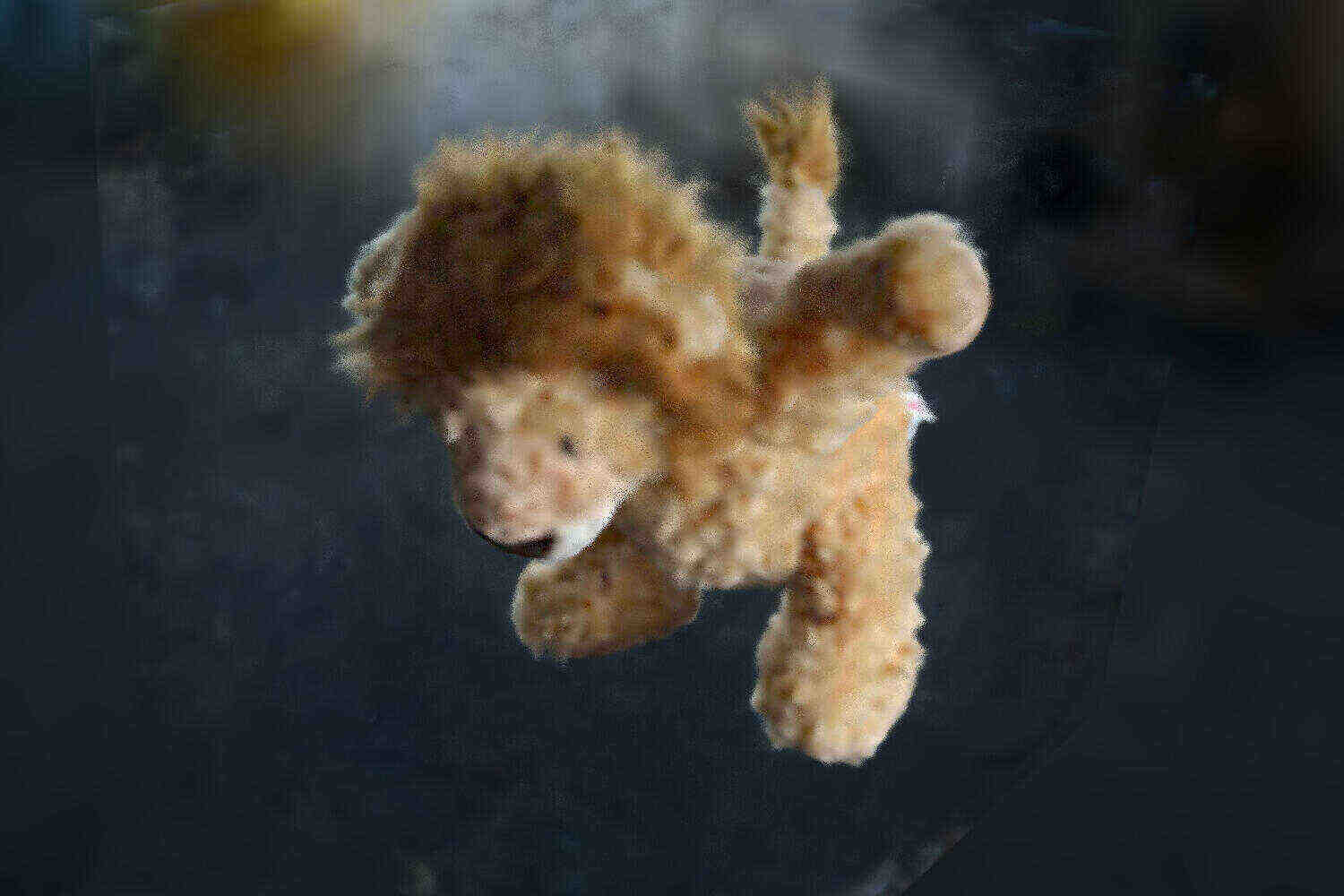}{40k it., 8/256/32}%
            \imageGap{}%
        \end{subfigure}%
        \begin{subfigure}[t]{0.245\textwidth}
            \centering%
            \labelImage{\textwidth}{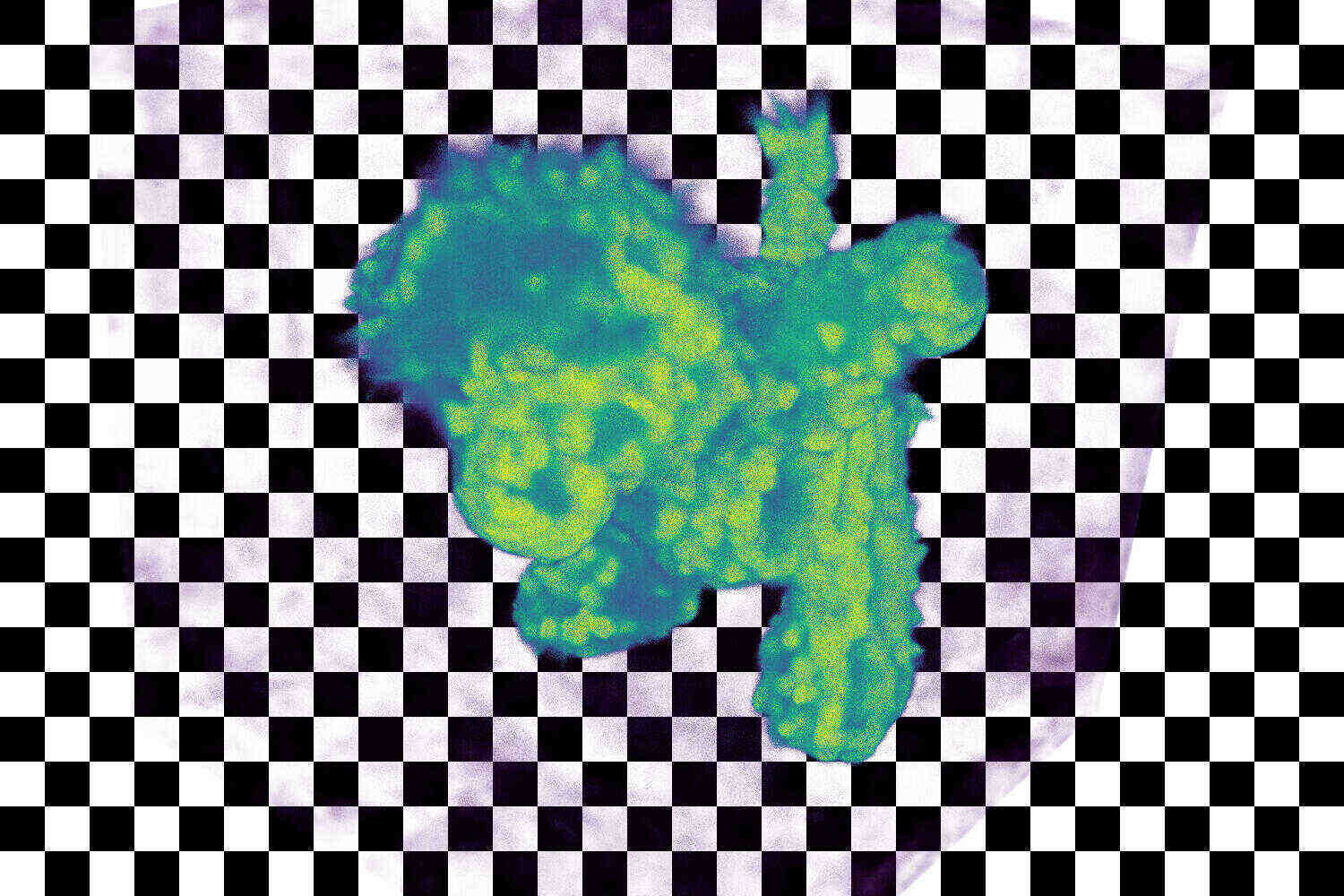}{40k it., 8/256/32}%
            \imageGap{}%
        \end{subfigure}%
        \begin{subfigure}[t]{0.245\textwidth}
            \centering%
            \normalMap{\textwidth}{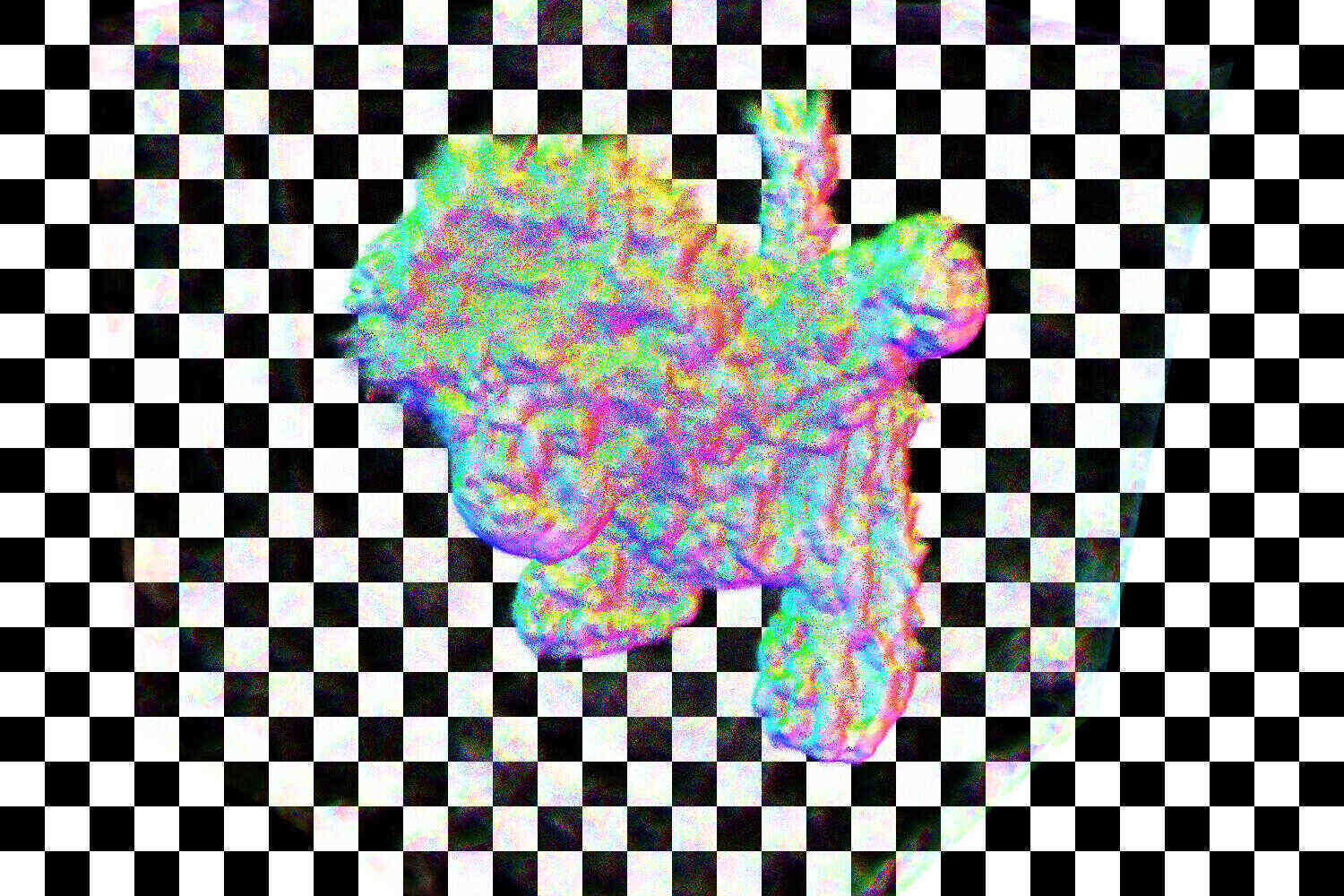}{40k it., 8/256/32}%
            \imageGap{}%
        \end{subfigure}%
        \begin{subfigure}[t]{0.245\textwidth}
            \centering%
            \labelImage{\textwidth}{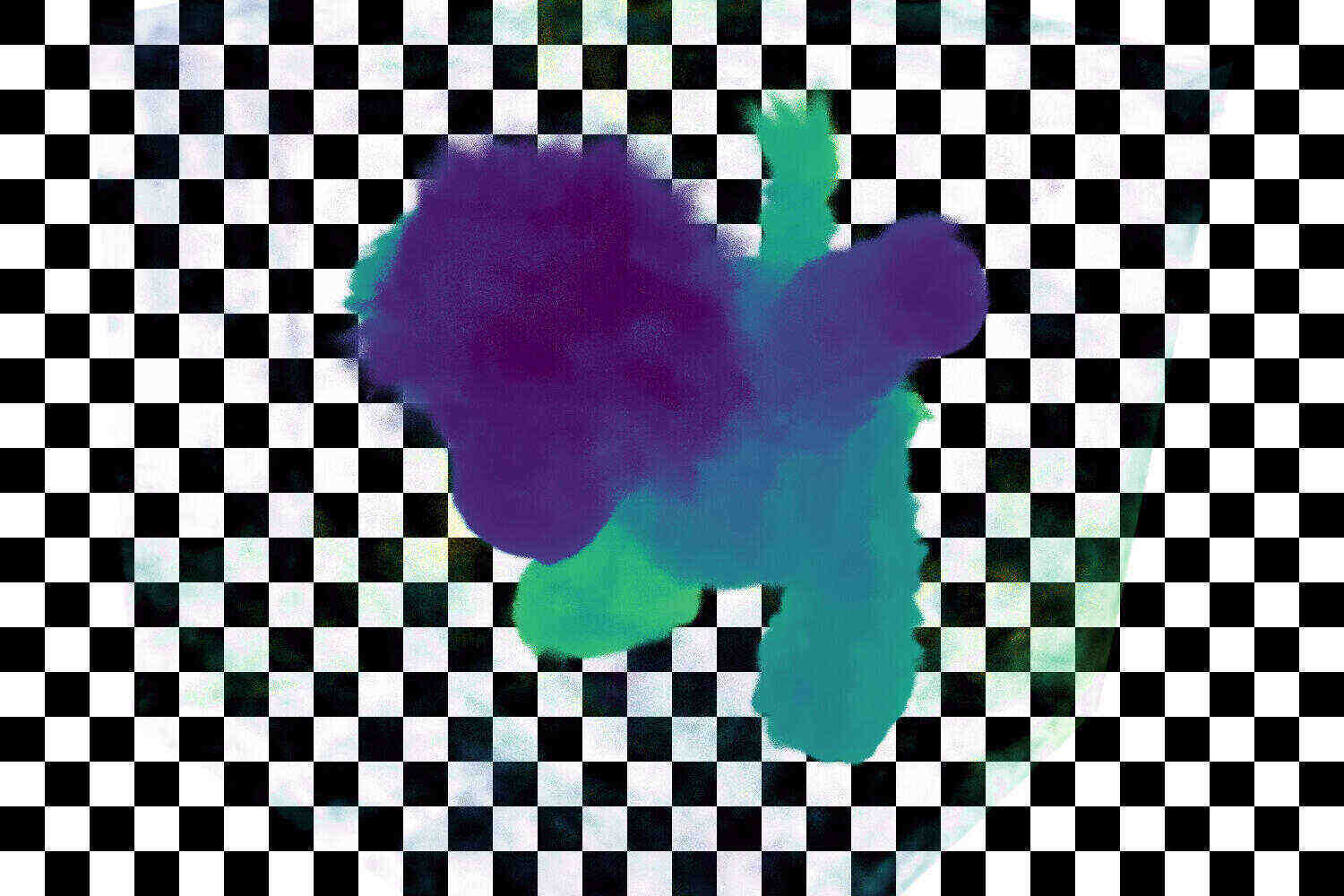}{40k it., 8/256/32}%
            \imageGap{}%
        \end{subfigure} \\
        %
        %
        \begin{subfigure}[t]{0.245\textwidth}
            \centering%
            \labelImage{\textwidth}{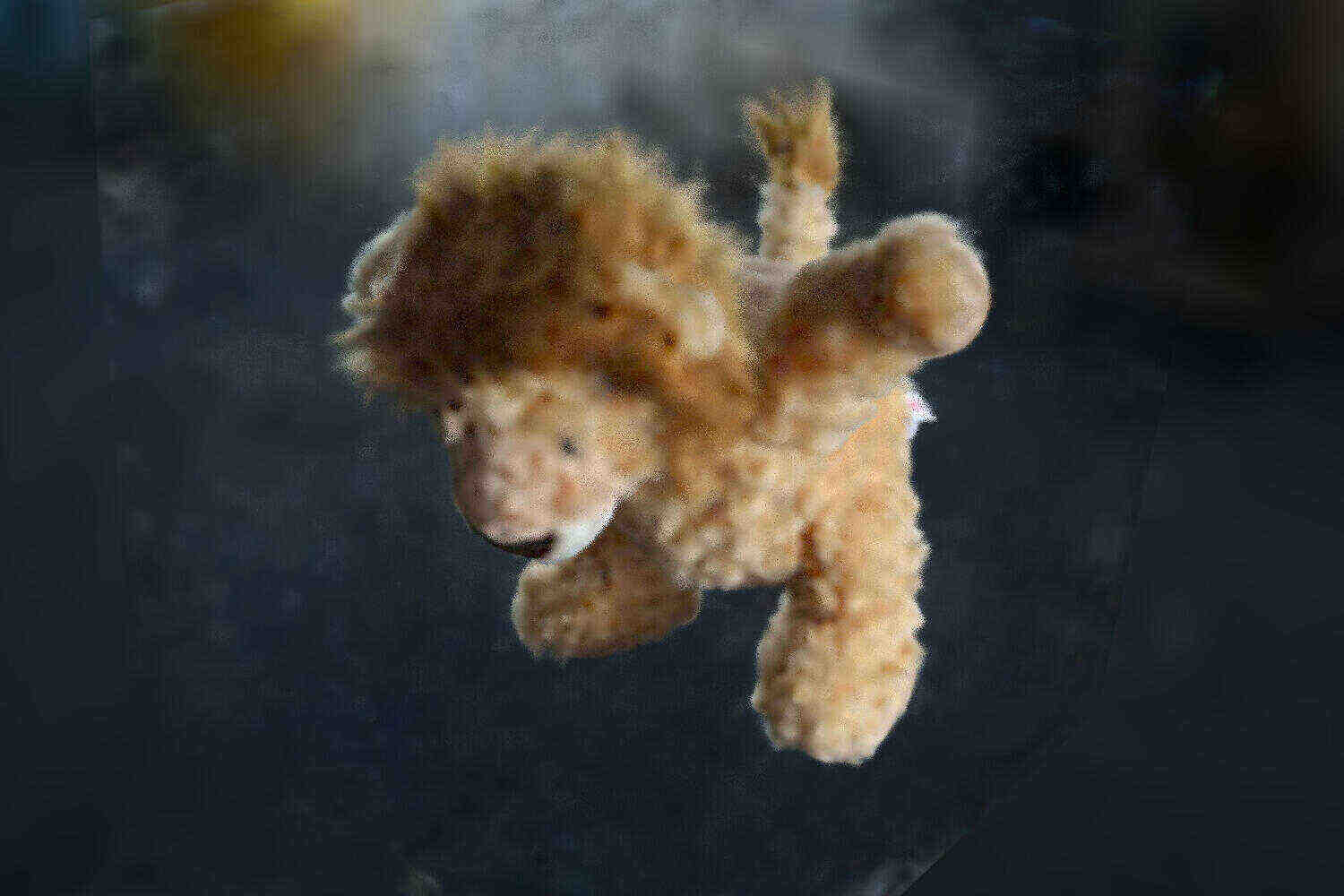}{40k it., 8/256/32, uniform}%
            \imageGap{}%
        \end{subfigure}%
        \begin{subfigure}[t]{0.245\textwidth}
            \centering%
            \labelImage{\textwidth}{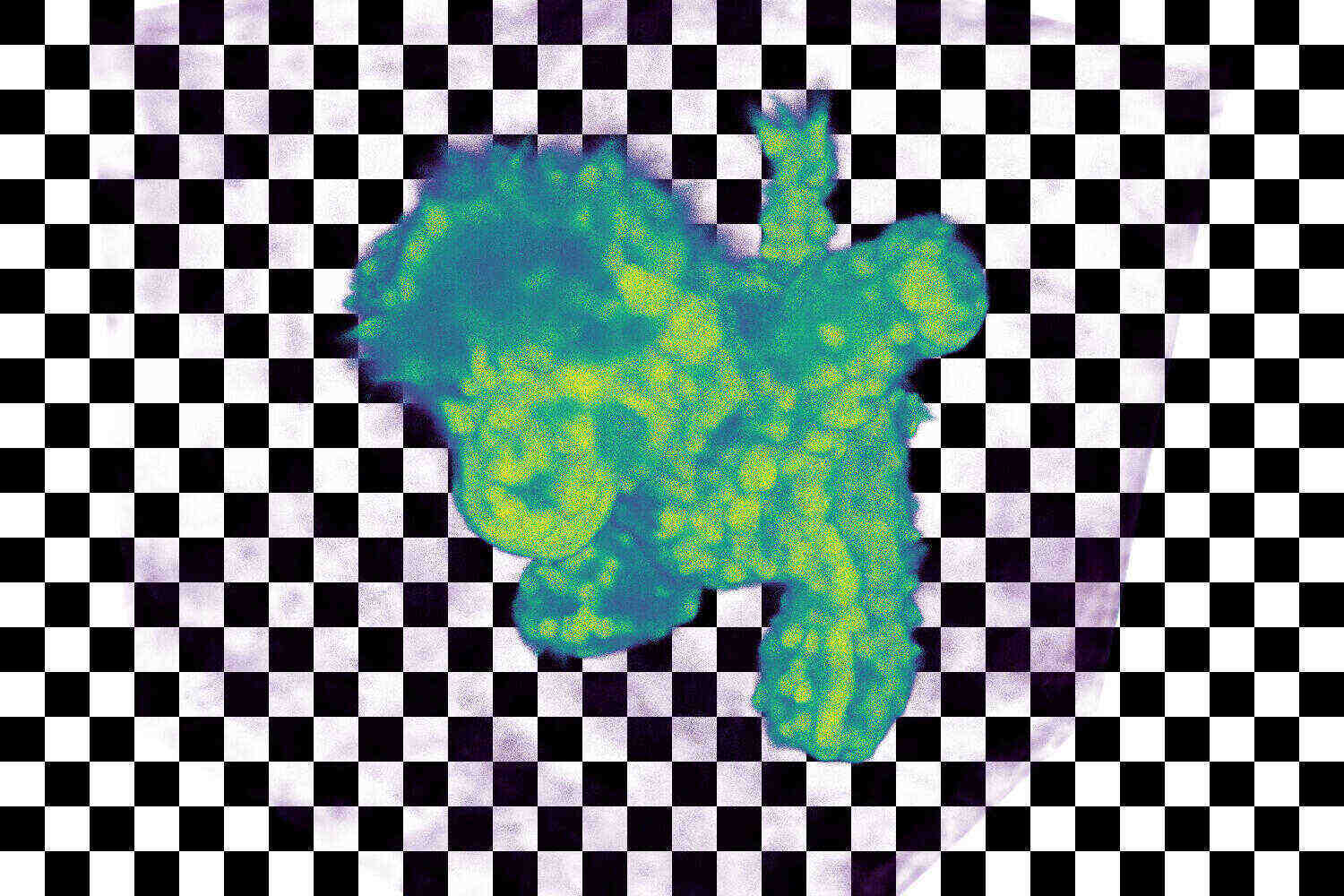}{40k it., 8/256/32, uniform}%
            \imageGap{}%
        \end{subfigure}%
        \begin{subfigure}[t]{0.245\textwidth}
            \centering%
            \normalMap{\textwidth}{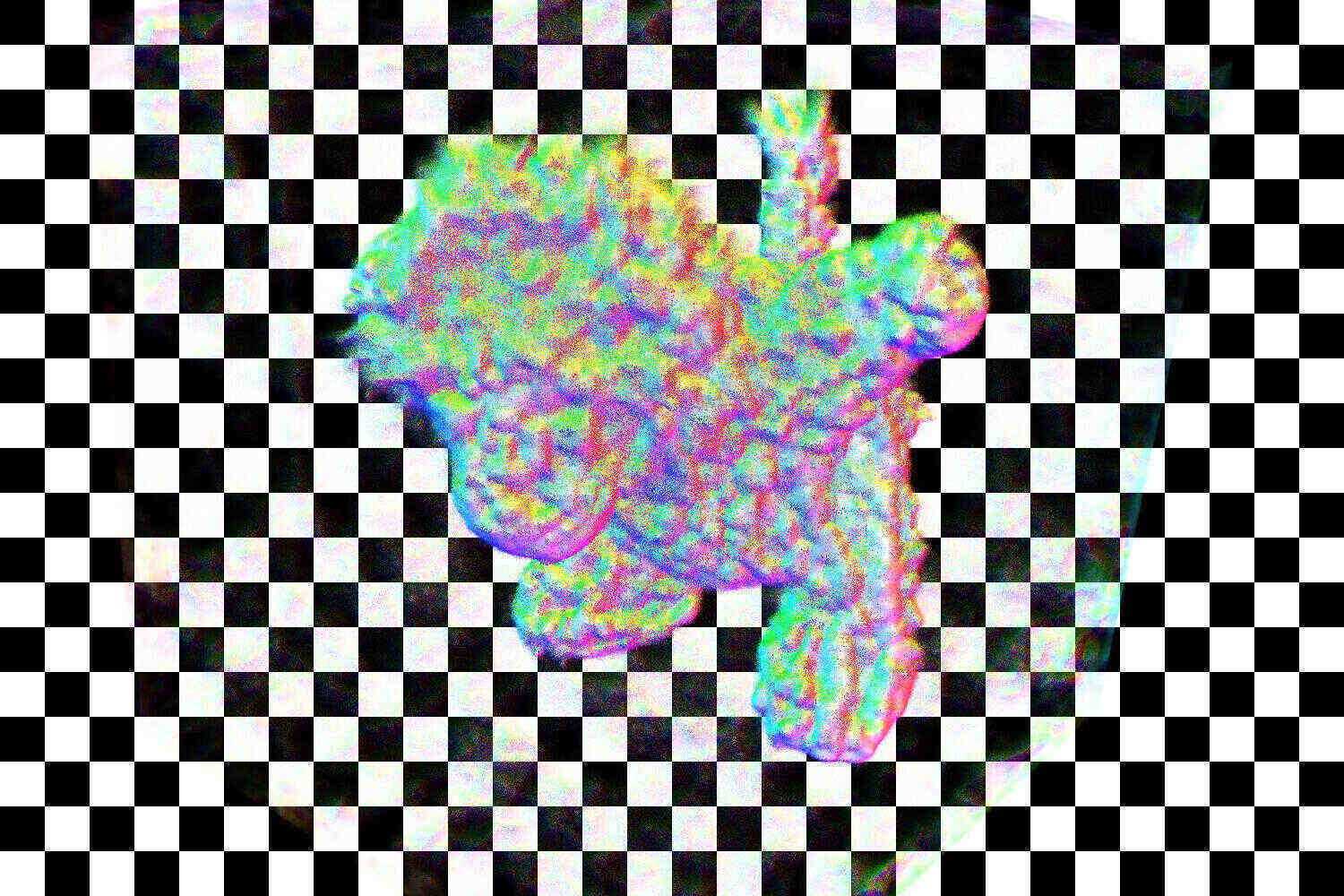}{40k it., 8/256/32, uniform}%
            \imageGap{}%
        \end{subfigure}%
        \begin{subfigure}[t]{0.245\textwidth}
            \centering%
            \labelImage{\textwidth}{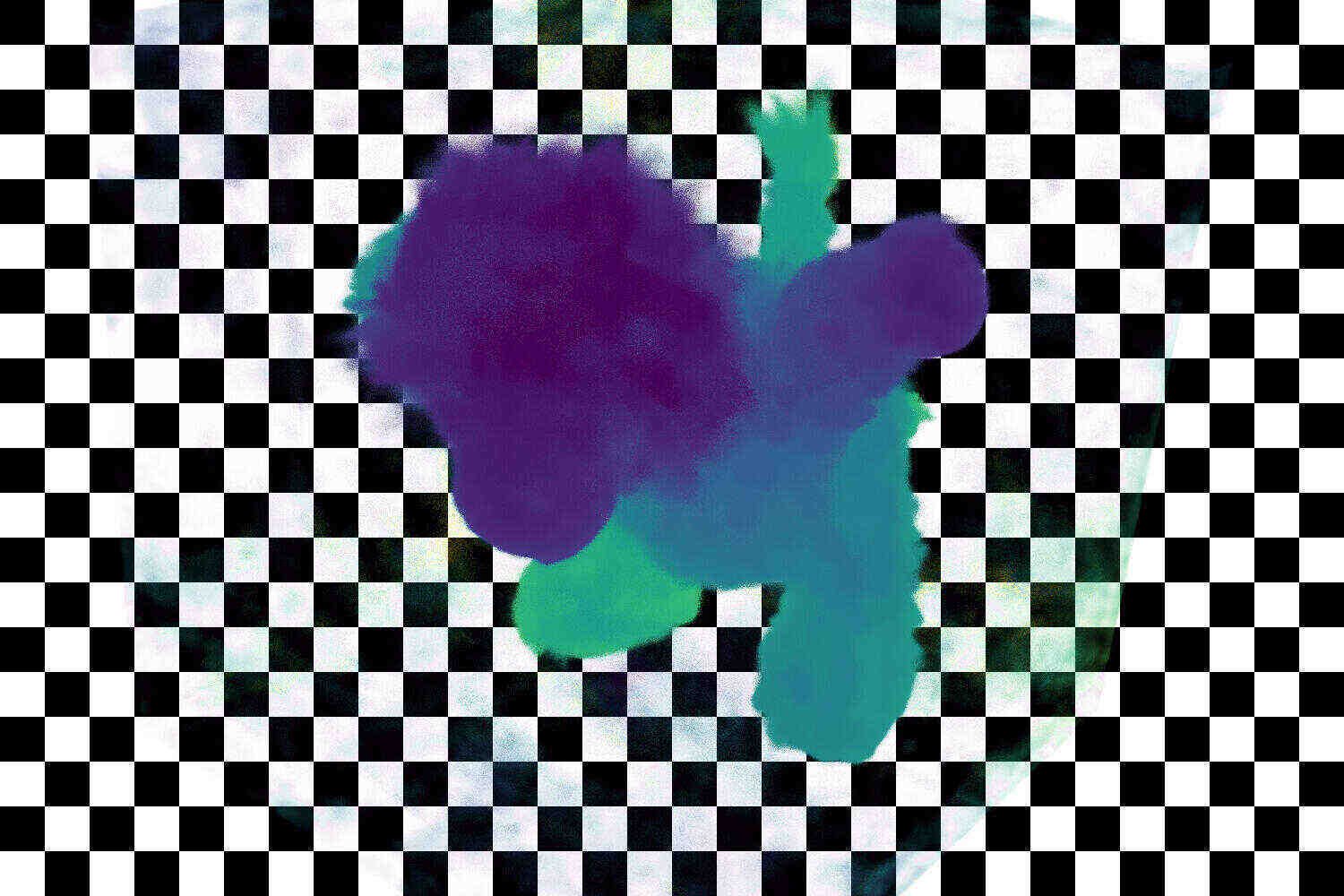}{40k it., 8/256/32, uniform}%
            \imageGap{}%
        \end{subfigure} \\%
        %
        %
        \begin{subfigure}[t]{0.245\textwidth}
            \centering%
            \labelImage{\textwidth}{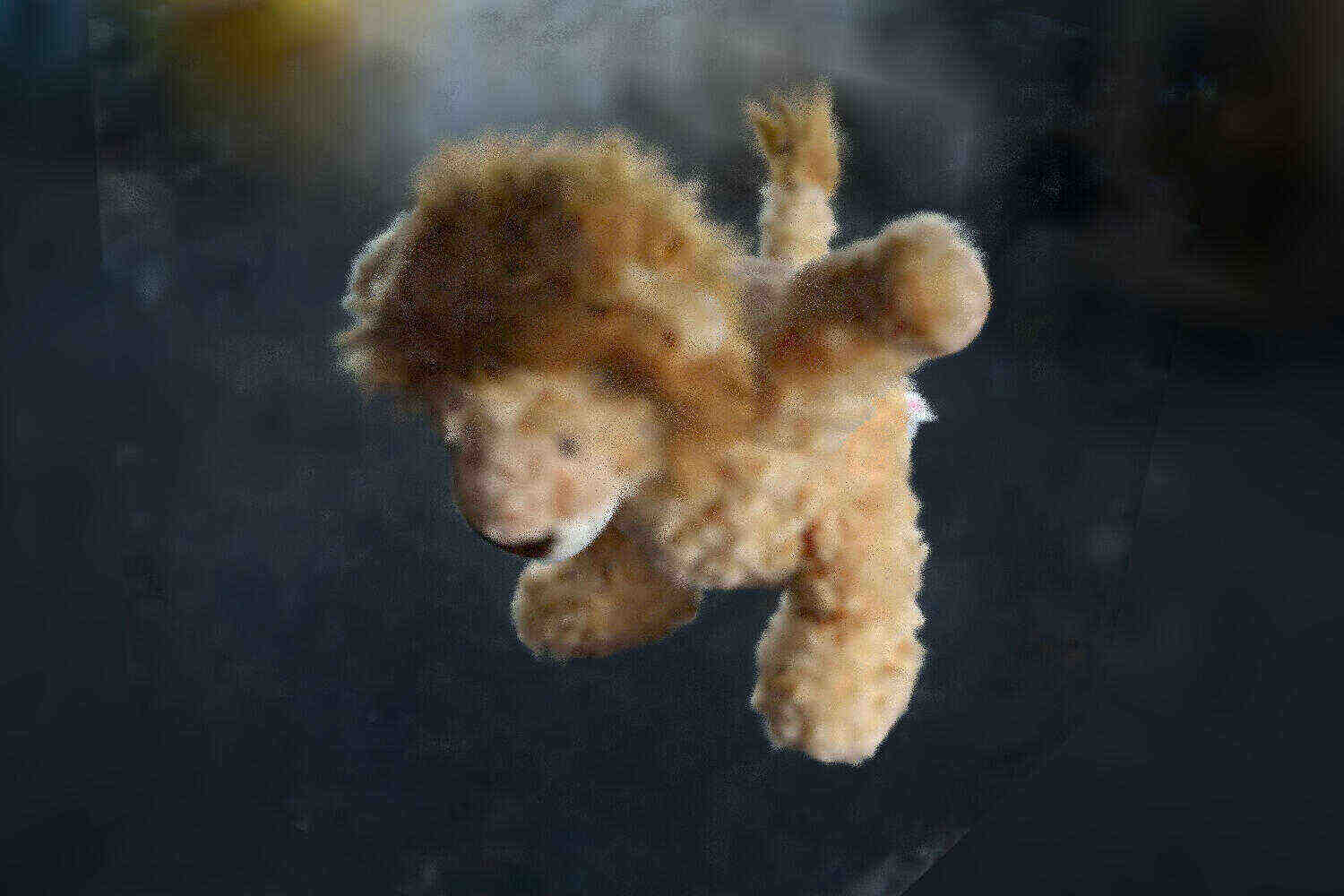}{40k it., 8/256/16}%
            \imageGap{}%
        \end{subfigure}%
        \begin{subfigure}[t]{0.245\textwidth}
            \centering%
            \labelImage{\textwidth}{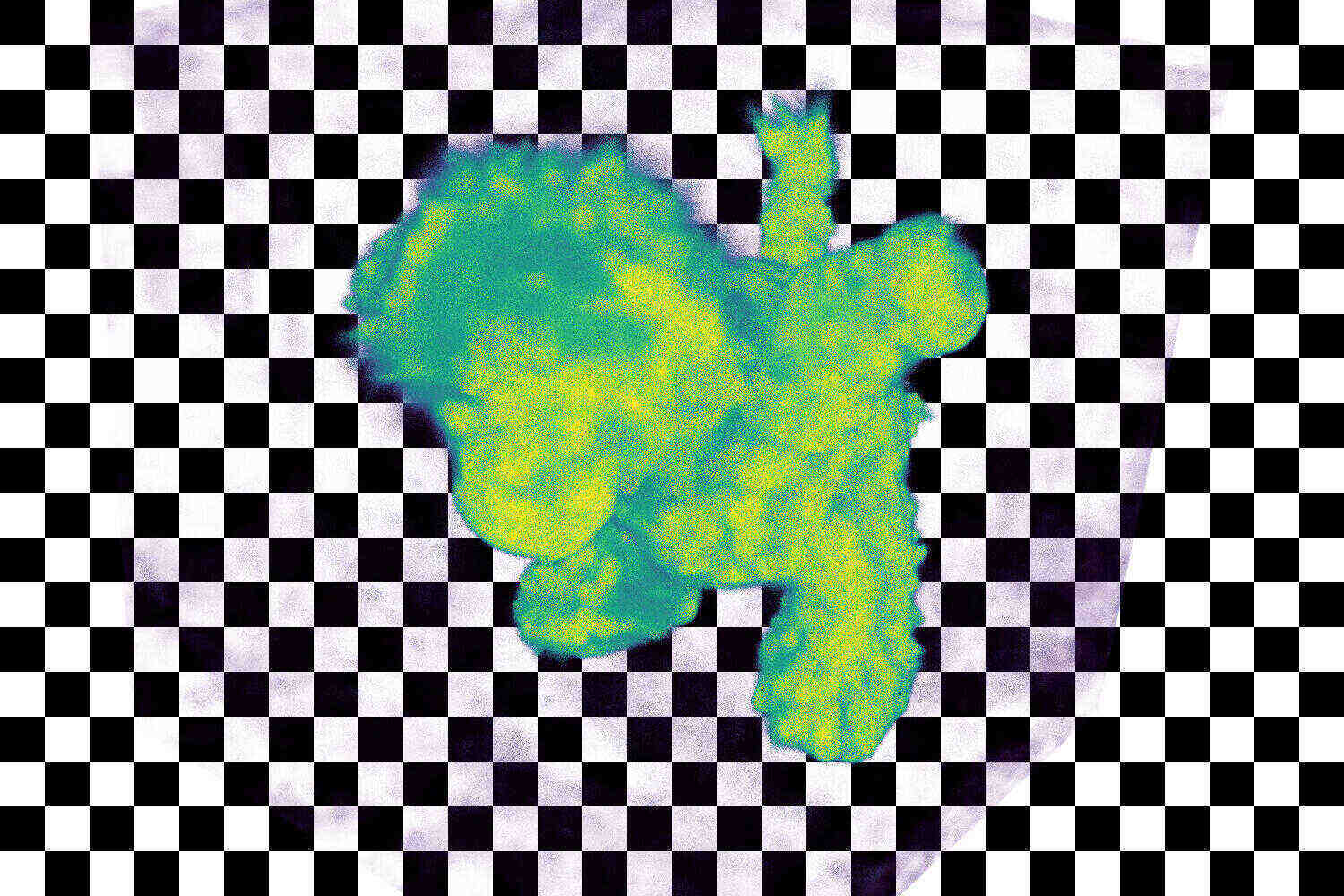}{40k it., 8/256/16}%
            \imageGap{}%
        \end{subfigure}%
        \begin{subfigure}[t]{0.245\textwidth}
            \centering%
            \normalMap{\textwidth}{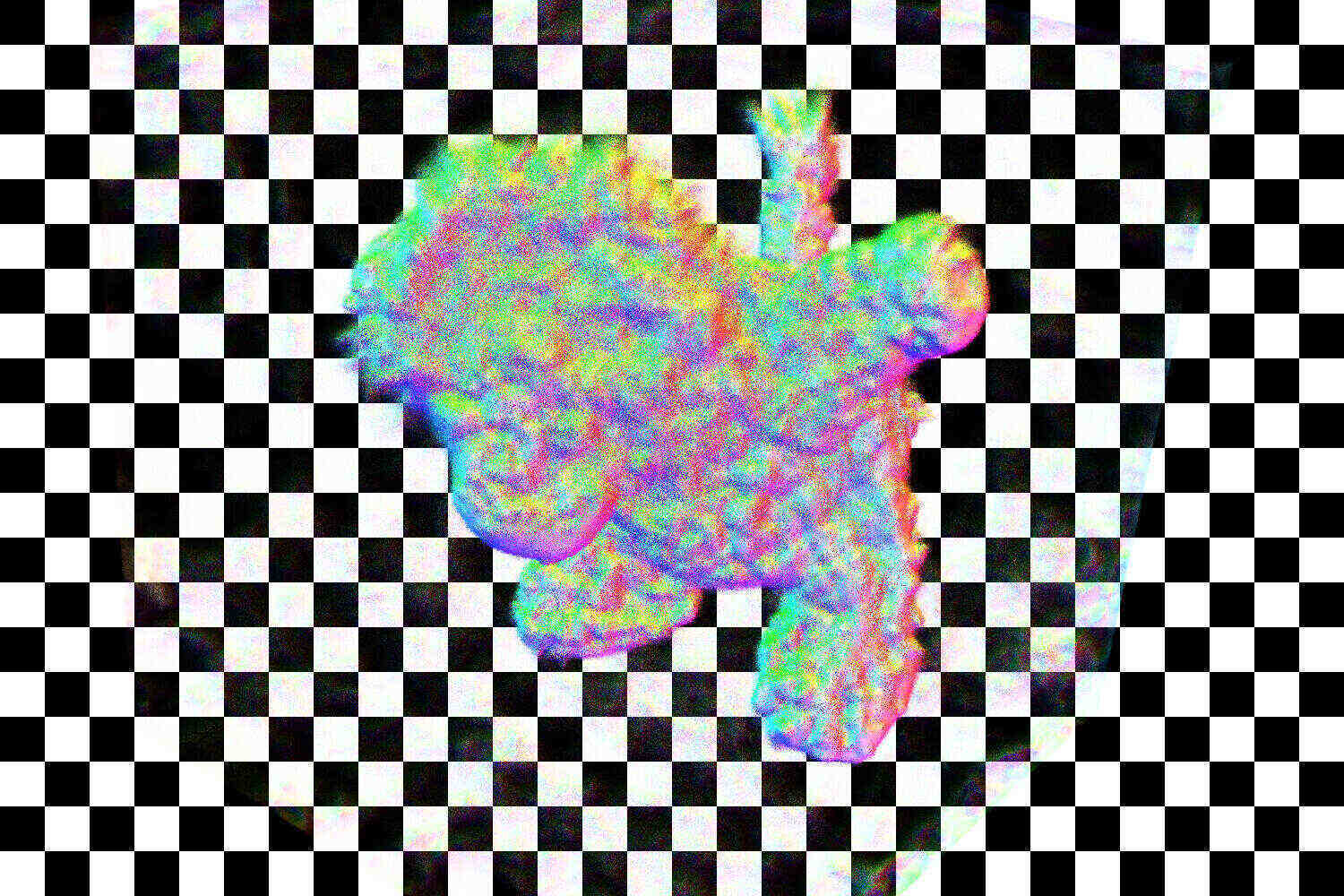}{40k it., 8/256/16}%
            \imageGap{}%
        \end{subfigure}%
        \begin{subfigure}[t]{0.245\textwidth}
            \centering%
            \labelImage{\textwidth}{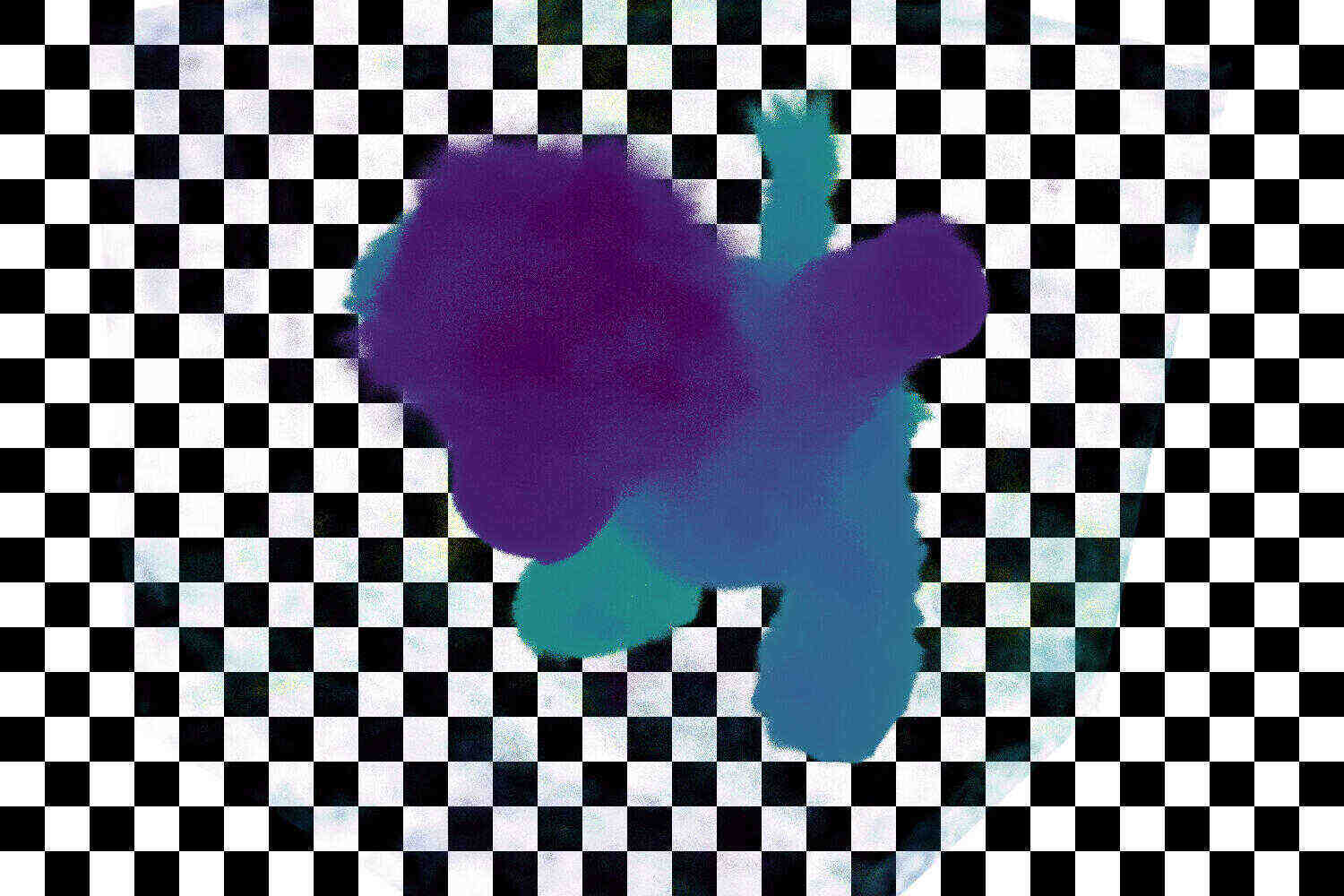}{40k it., 8/256/16}%
            \imageGap{}%
        \end{subfigure}%
    \end{tabular}
    %
    %
    \caption%
    {%
        Scene sampling influence on results of the Lion scene after 40k iterations
        and with varying sampling budgets $N/\maxSamplesPerRayUninformed/\maxSamplesPerRayInformed$.
    }%
    \label{fig:sceneSamplingBudgets40k}
\end{figure*}

%
%
We investigated the influence of different sampling budgets during optimization.
\Fig{\ref{fig:sceneSamplingBudgets2.5k}} and \Fig{\ref{fig:sceneSamplingBudgets40k}} demonstrate
that \Adam{} is very robust against the noise introduced by only small sampling budgets.
Interestingly,
the reconstructions are actually better
(less false clutter in free space)
when only employing smaller and cheaper sampling budgets.
We presume that
small sampling budgets increase the probability of sampling around false occluders
which mitigates the lack of required gradients they induce on occluded voxels.
Note that smaller sampling budgets partially blur the fields
and result in voxels within objects being occupied
despite our zero opacity prior.
Uniform instead of stratified sampling within a ray-intersected node (4th row only)
reduces reconstruction quality only little in this case.

%
%

%
%

\begin{figure*}%
    \centering%
    \begin{subfigure}[t]{0.245\textwidth}
        \centering%
        \labelImage{\textwidth}{Results/QualitativeEvaluation/SyntheticNeRFScenesOverview/00ChairView363Original}{gt}
        \labelImage{\textwidth}{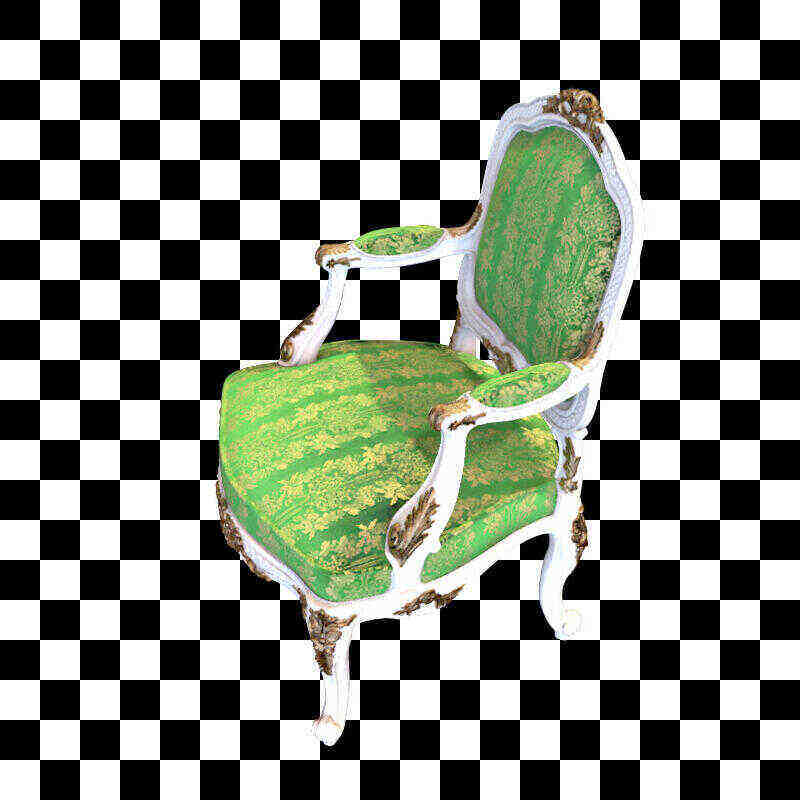}{115k it.}
        \label{fig:qualitativeEvaluationOverviewChair}
        \imageGap{}%
    \end{subfigure}%
    \begin{subfigure}[t]{0.245\textwidth}
        \centering%
        \labelImage{\textwidth}{Results/QualitativeEvaluation/SyntheticNeRFScenesOverview/01DrumsView340Original}{gt}
        \labelImage{\textwidth}{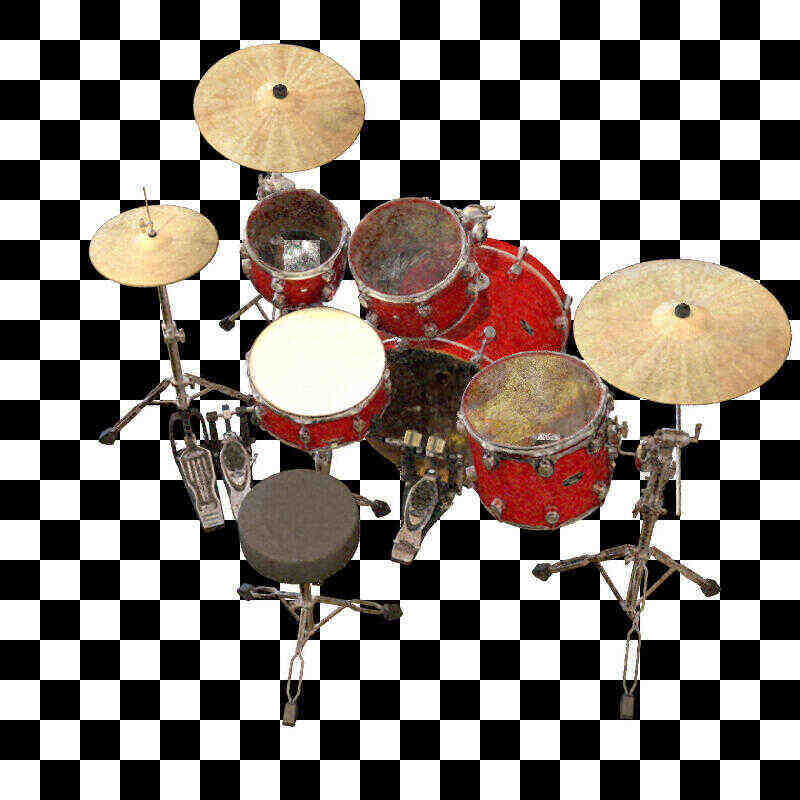}{115k it.}
        \label{fig:qualitativeEvaluationOverviewDrums}
        \imageGap{}%
    \end{subfigure}%
    \begin{subfigure}[t]{0.245\textwidth}
        \centering%
        \labelImage{\textwidth}{Results/QualitativeEvaluation/SyntheticNeRFScenesOverview/02FicusView372Original}{gt}
        \labelImage{\textwidth}{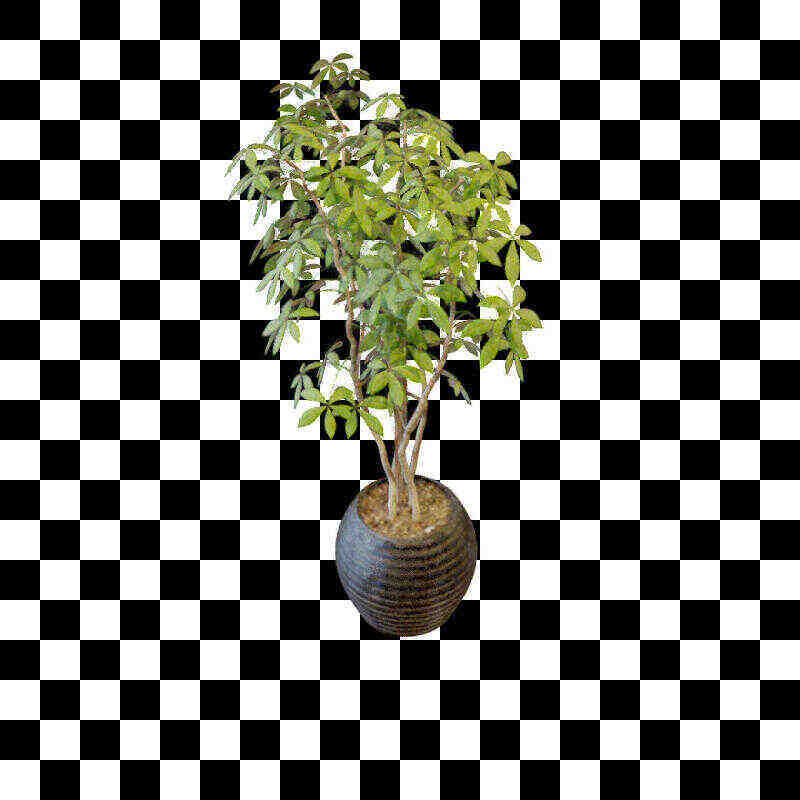}{115k it.}
        \label{fig:qualitativeEvaluationOverviewFicus}
        \imageGap{}%
    \end{subfigure}%
    \begin{subfigure}[t]{0.245\textwidth}
        \centering%
        \labelImage{\textwidth}{Results/QualitativeEvaluation/SyntheticNeRFScenesOverview/03HotDogView310Original}{gt}
        \labelImage{\textwidth}{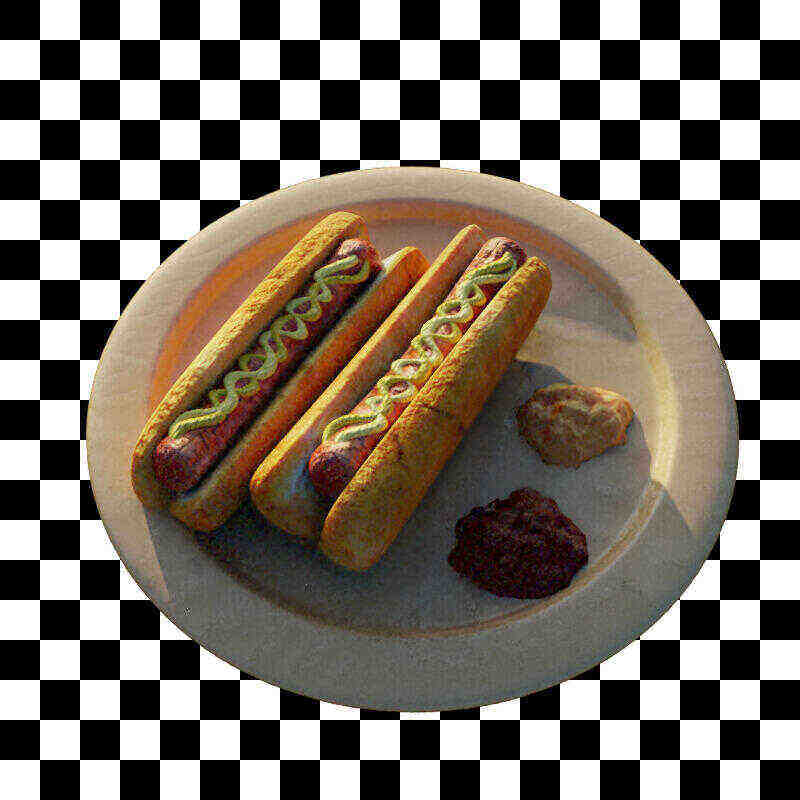}{115k it.}
        \label{fig:qualitativeEvaluationOverviewHotDog}
        \imageGap{}%
    \end{subfigure} \\
    \begin{subfigure}[t]{0.245\textwidth}
        \centering%
        \labelImage{\textwidth}{Results/QualitativeEvaluation/SyntheticNeRFScenesOverview/04LegoView320Original}{gt}
        \labelImage{\textwidth}{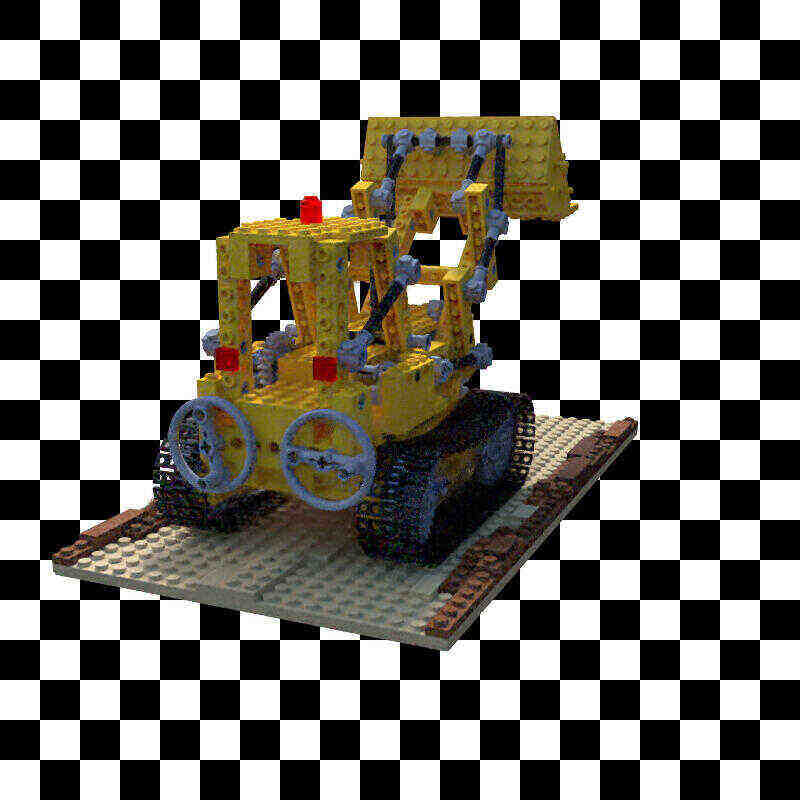}{115k it.}
        \label{fig:qualitativeEvaluationOverviewLego}
        \imageGap{}%
    \end{subfigure}%
    \begin{subfigure}[t]{0.245\textwidth}
        \centering%
        \labelImage{\textwidth}{Results/QualitativeEvaluation/SyntheticNeRFScenesOverview/05MaterialsView397Original}{gt}
        \labelImage{\textwidth}{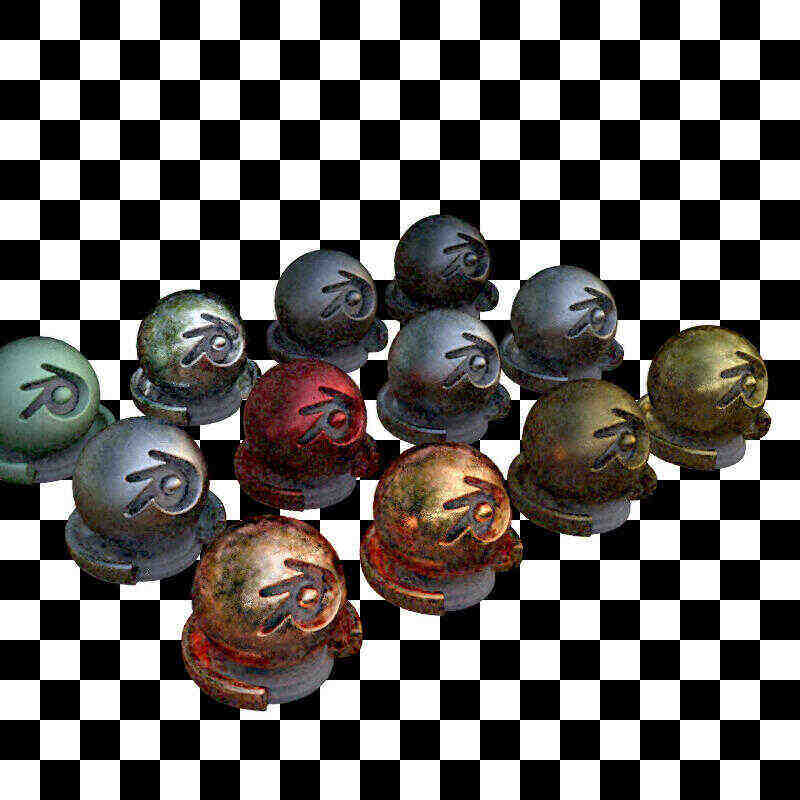}{115k it.}
        \label{fig:qualitativeEvaluationOverviewMaterials}
        \imageGap{}%
    \end{subfigure}%
    \begin{subfigure}[t]{0.245\textwidth}
        \centering%
        \labelImage{\textwidth}{Results/QualitativeEvaluation/SyntheticNeRFScenesOverview/06MicView357Original}{gt}
        \labelImage{\textwidth}{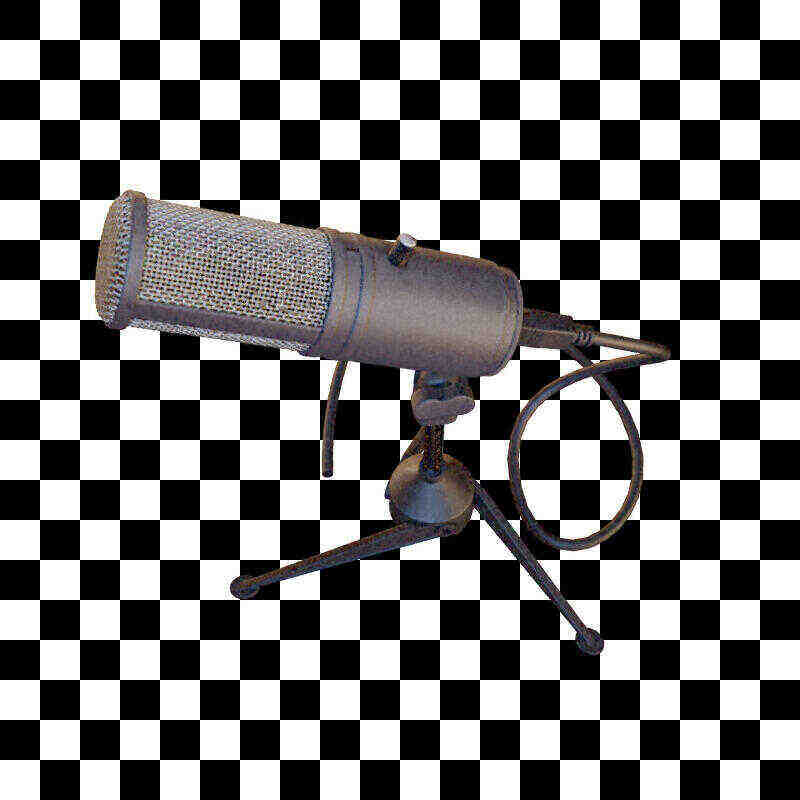}{115k it.}
        \label{fig:qualitativeEvaluationOverviewMic}
        \imageGap{}%
    \end{subfigure}%
    \begin{subfigure}[t]{0.245\textwidth}
        \centering%
        \labelImage{\textwidth}{Results/QualitativeEvaluation/SyntheticNeRFScenesOverview/07ShipView372Original}{gt}
        \labelImage{\textwidth}{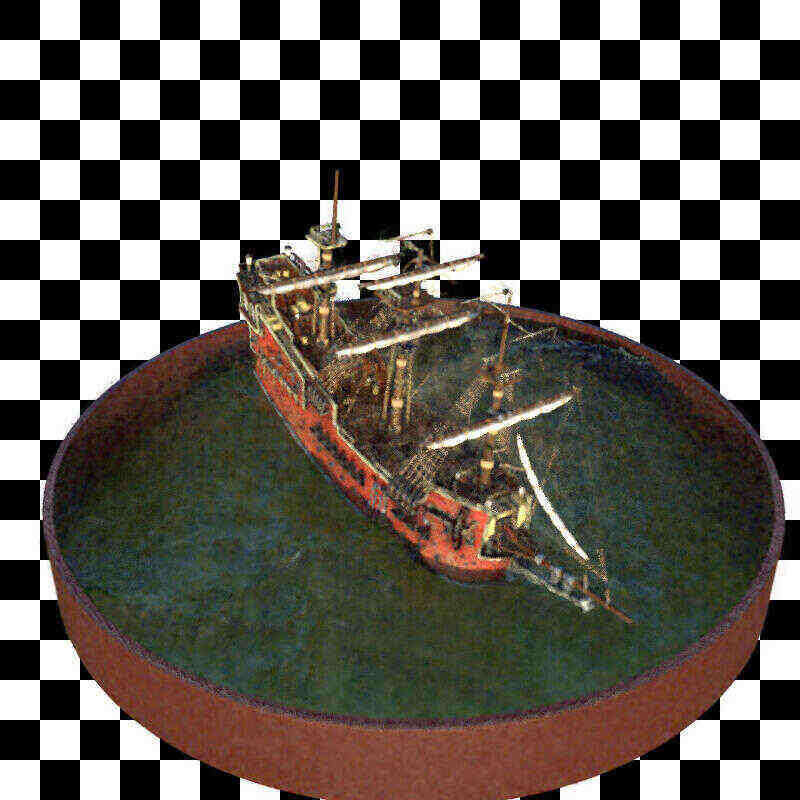}{120k it.}
        \label{fig:qualitativeEvaluationOverviewShip}
        \imageGap{}%
    \end{subfigure}%
    %
    %
    \caption%
    {%
        Qualitative evaluation using an overview of our results
        for all of the synthetic \NeRF{} scenes \cite{mildenhall2020NeRF}.
    }%
    \label{fig:qualitativeEvaluationOverviewSynthNeRFScenes}
\end{figure*}

%
%
The overview of \Fig{\ref{fig:qualitativeEvaluationOverviewSynthNeRFScenes}} demonstrates the
novel view synthesis performance of our method
for all of the synthetic \NeRF{} scenes \cite{mildenhall2020NeRF}.
Our method reconstructs the scenes
within comparably few optimization iterations and
with high quality that is similar to state-of-the-art implicit approaches like \MipNeRF{}.
However,
the surface light field based on low frequency \glspl{SH}
is clearly not able to represent sharp reflections
as for example the drums scene comparison shows.

%
%

\begin{figure*}%
    \centering%
    \begin{subfigure}[t]{0.19\textwidth}
        \centering%
        \labelImage{\textwidth}{Results/CheckeredCube/CubeVariation04_2022_01_26_18_44_49_f6_It135K/225/GroundTruth}{gt}%
    \end{subfigure}%
    \begin{subfigure}[t]{0.19\textwidth}
        \labelImage{\textwidth}{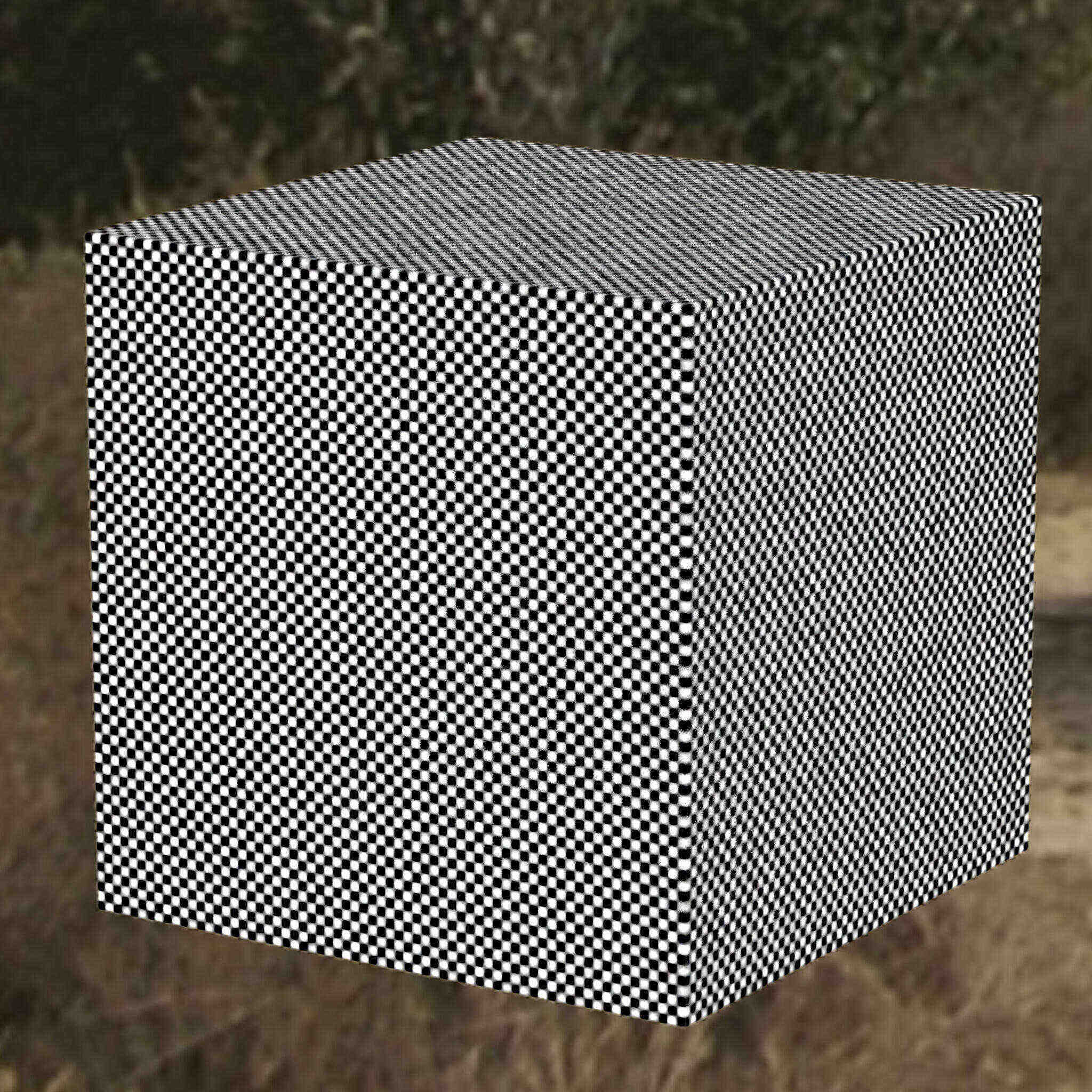}{115k it.}%
    \end{subfigure}%
    \begin{subfigure}[t]{0.19\textwidth}
        \labelImage{\textwidth}{Results/CheckeredCube/CubeVariation04_2022_01_26_18_44_49_f6_It135K/225/Errors}{}%
    \end{subfigure}%
    \begin{subfigure}[t]{0.19\textwidth}
        \normalMap{\textwidth}{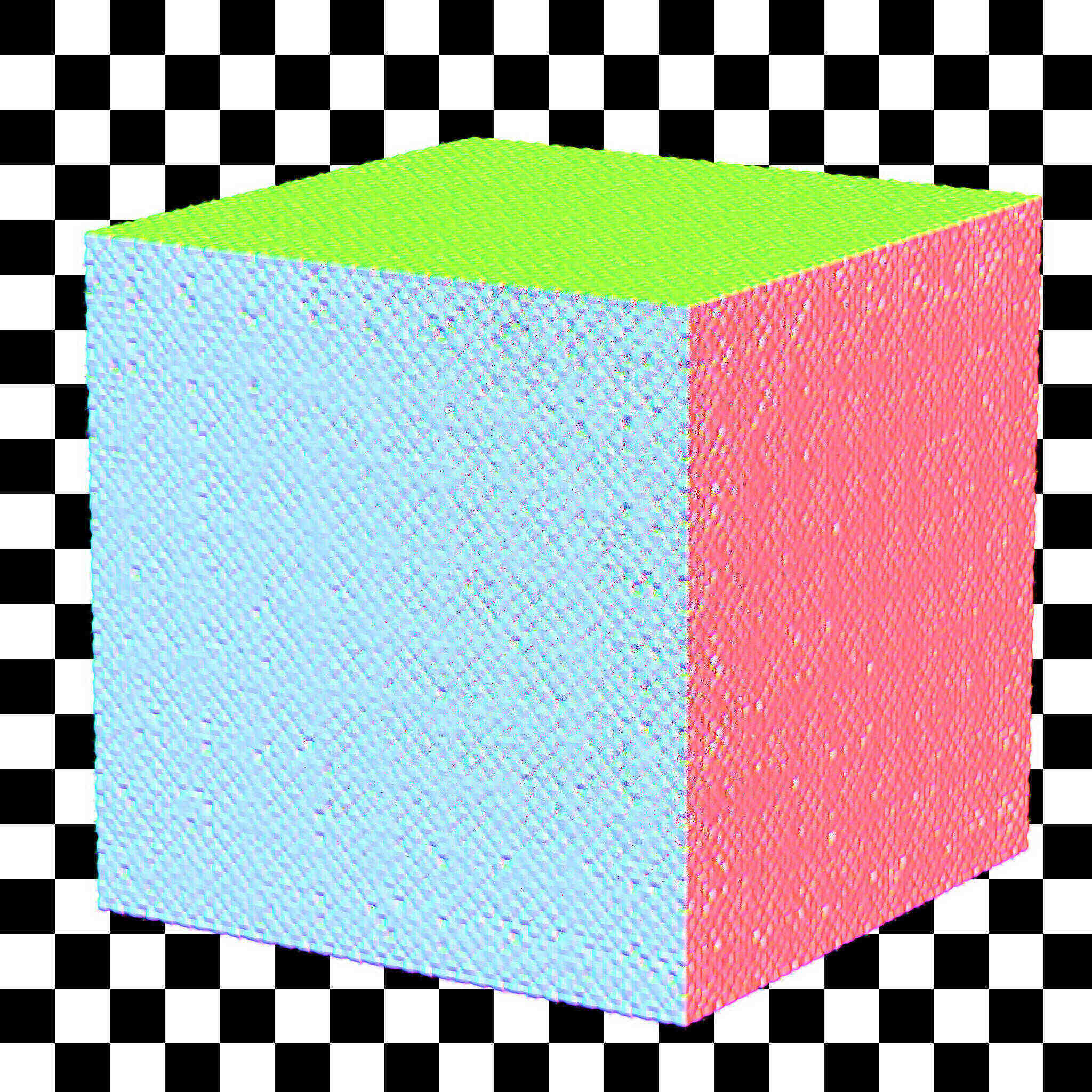}{}%
    \end{subfigure}%
    \begin{subfigure}[t]{0.19\textwidth}
        \labelImage{\textwidth}{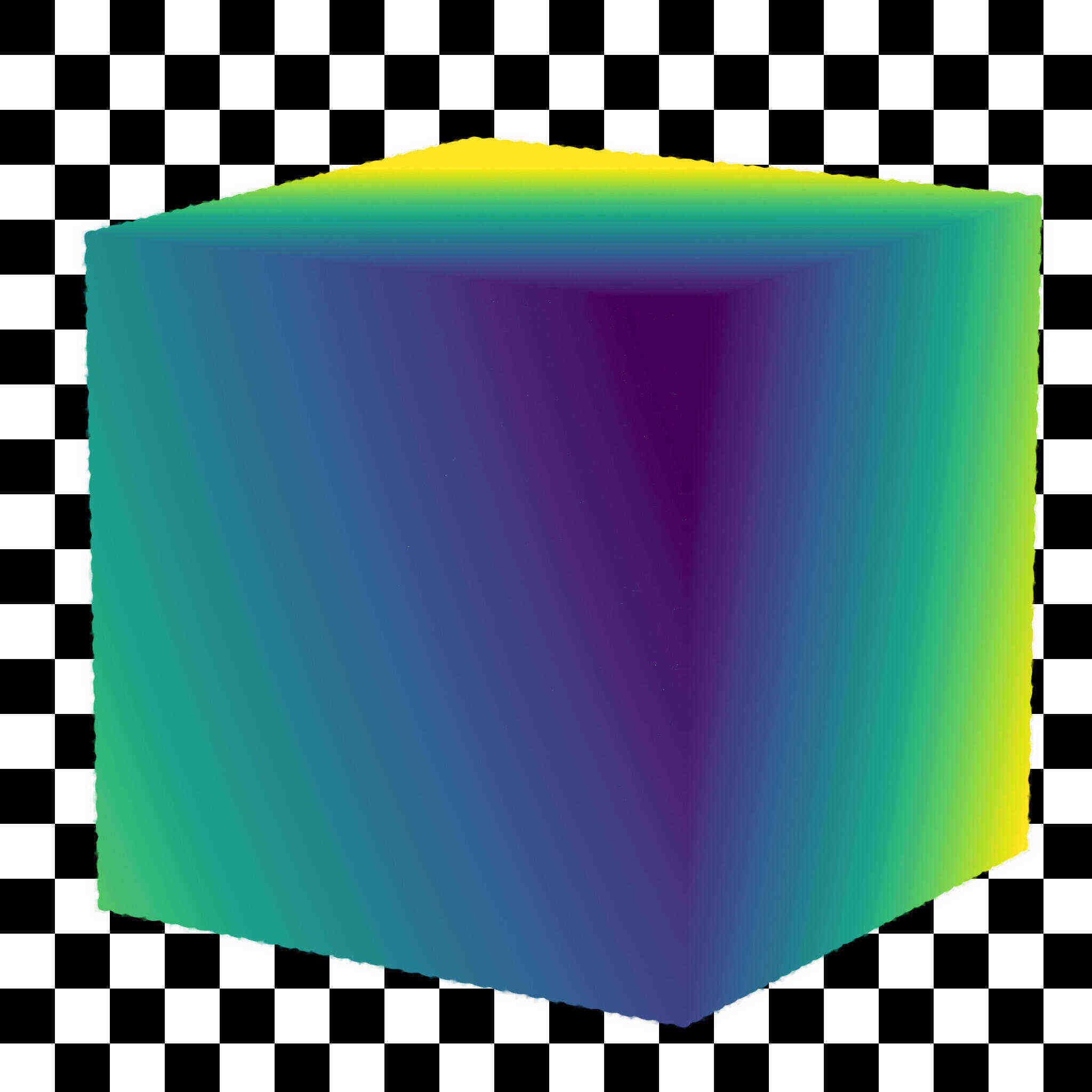}{}%
    \end{subfigure}\\%
    \begin{subfigure}[t]{0.19\textwidth}
        \centering%
        \labelImage{\textwidth}{Results/CheckeredCube/CubeVariation04_2022_01_26_18_44_49_f6_It135K/226/GroundTruth}{gt}%
    \end{subfigure}%
    \begin{subfigure}[t]{0.19\textwidth}
        \labelImage{\textwidth}{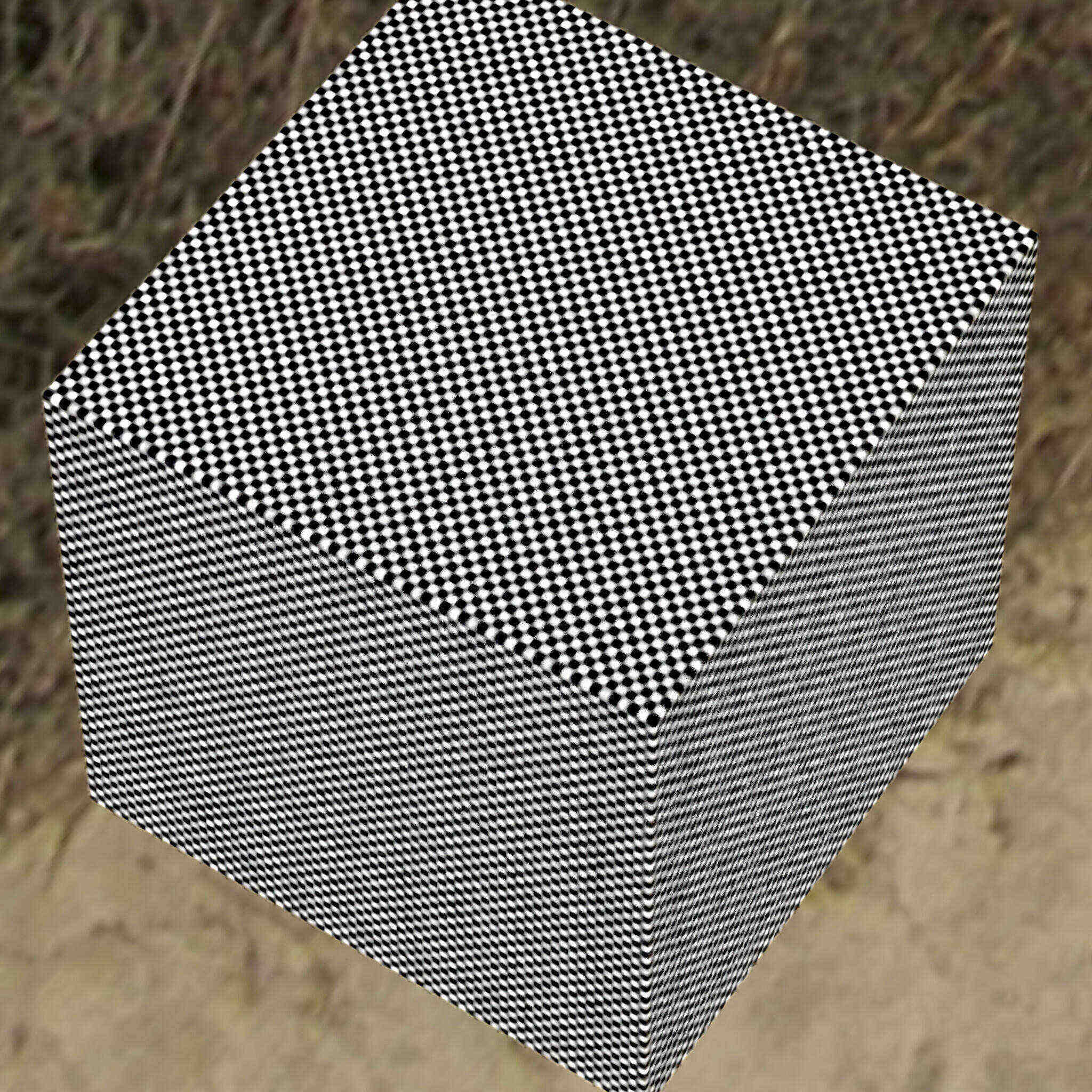}{115k it.}%
    \end{subfigure}%
    \begin{subfigure}[t]{0.19\textwidth}
        \labelImage{\textwidth}{Results/CheckeredCube/CubeVariation04_2022_01_26_18_44_49_f6_It135K/226/Errors}{}%
    \end{subfigure}%
    \begin{subfigure}[t]{0.19\textwidth}
        \normalMap{\textwidth}{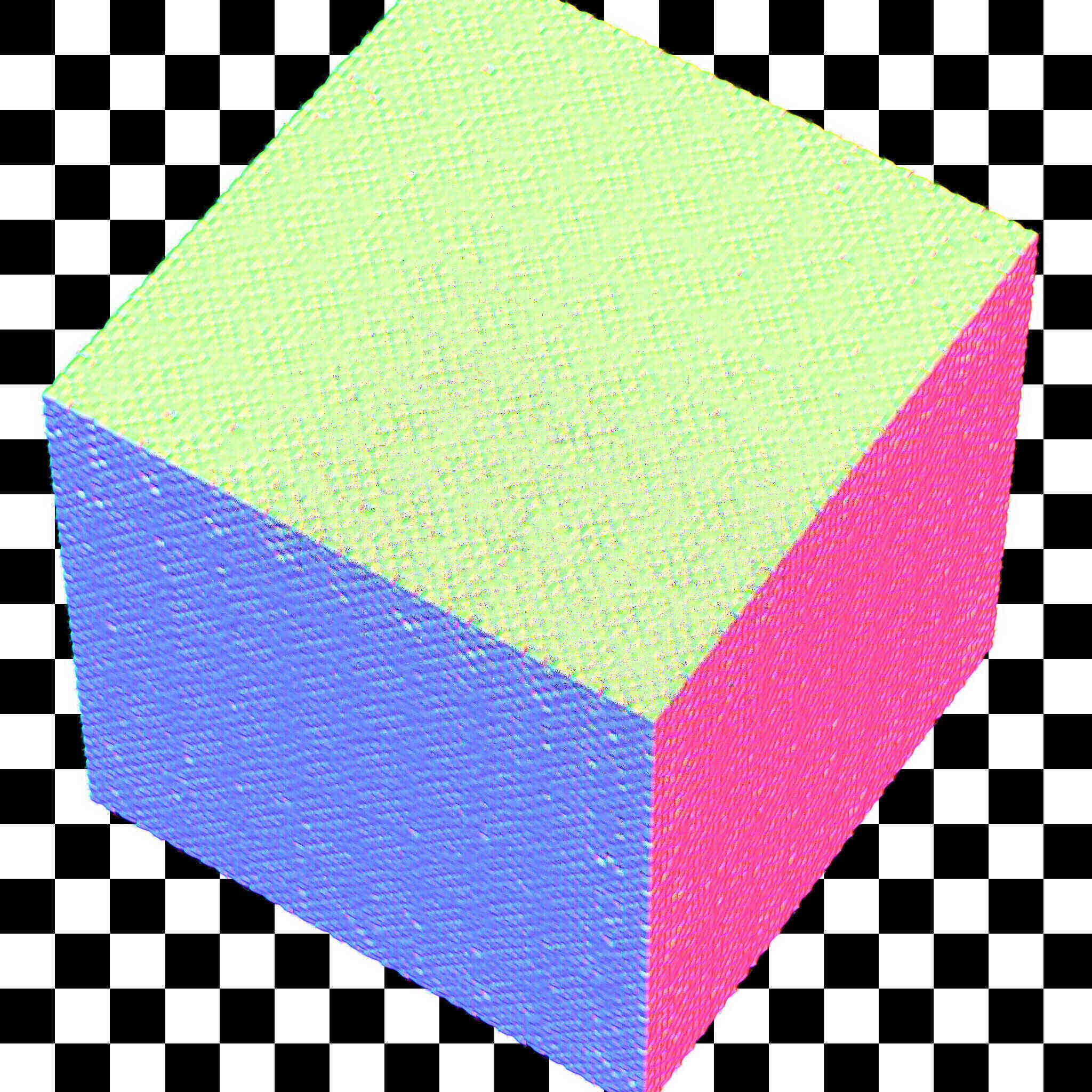}{}%
    \end{subfigure}%
    \begin{subfigure}[t]{0.19\textwidth}
        \labelImage{\textwidth}{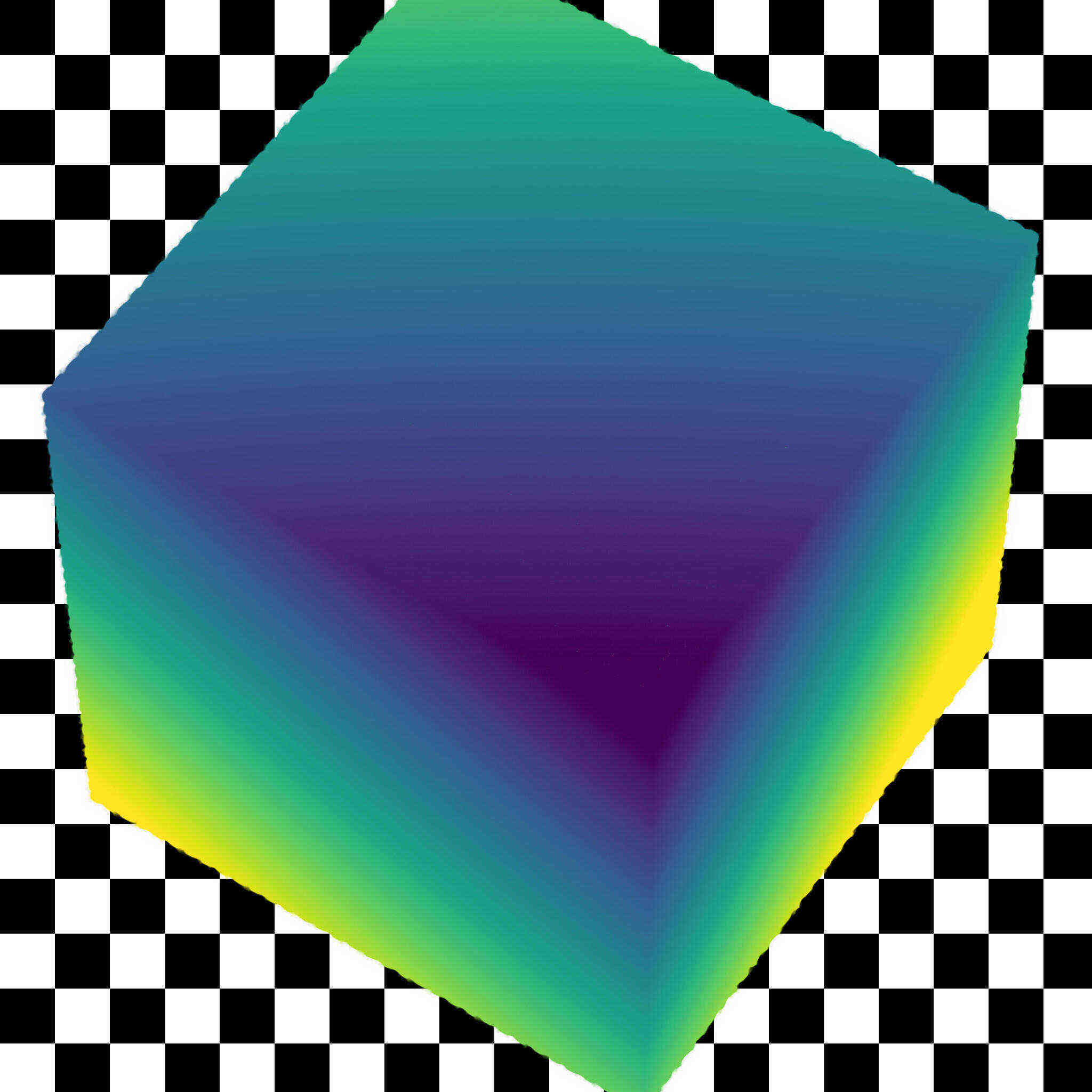}{}%
    \end{subfigure}\\%
    %
    %
    \caption%
    {%
        Robustness experiment on a synthetic scene
        consisting of an axis-aligned checkered cube
        flying inside a beach environment map
        showing (left to right):
        hold out ground truth;
        photo realistic reconstruction;
        photo consistency errors;
        surface normals and depths.
    }%
    \label{fig:cubeExperiment}
\end{figure*}

We tested the robustness of our method against fine details with opposing surface radiance
(black and white checker pattern) using the synthetic checkered cube scene
shown in \Fig{\ref{fig:cubeExperiment}}.
Our method is able to reconstruct the checker pattern and does not fail
with a average gray surface estimate despite the \SGD{}-based optimization.

%
%

\end{document}